\documentclass{article}

\usepackage{microtype}
\usepackage{graphicx}
\usepackage{subfigure}
\usepackage{booktabs} % for professional tables

\usepackage{hyperref}
\usepackage{multirow}
\usepackage{tabularray}
\usepackage{array}
\usepackage{dashrule}

 \usepackage[accepted]{icml2025}

\usepackage{amssymb}
\usepackage{mathtools}
\usepackage{amsthm}

\usepackage[capitalize,noabbrev]{cleveref}

\theoremstyle{plain}

\theoremstyle{definition}

\theoremstyle{remark}

\usepackage[textsize=tiny]{todonotes}

\usepackage{mathtools}
\usepackage{enumitem}
\usepackage{mathtools}
\newcommand{\defeq}{\vcentcolon=}

\usepackage{booktabs}
\usepackage{amsthm}

\usepackage{graphicx}
\usepackage{subcaption}
\usepackage{bm}

\usepackage{xcolor}
\definecolor{maroon}{cmyk}{0, 0.87, 0.68, 0.32}
\definecolor{halfgray}{gray}{0.55}

\definecolor{lred}{RGB}{252, 224, 225}
\definecolor{lorange}{RGB}{255, 226, 187}
\definecolor{lyellow}{RGB}{253, 249, 192}
\definecolor{lblue}{RGB}{194, 232, 247}
\definecolor{lgreen}{RGB}{204, 231, 207}
\definecolor{lgrey}{RGB}{230, 230, 230}
\definecolor{llgrey}{RGB}{245, 245, 245}
\definecolor{lpurple}{RGB}{197, 190, 223}
\definecolor{lmagenta}{RGB}{243, 200, 220}
\definecolor{laqua}{RGB}{104, 165, 179}

\definecolor{rev}{RGB}{0,0,0}

\definecolor{mydarkblue}{rgb}{0,0.08,0.45}
\hypersetup{ %
    pdftitle={},
    pdfauthor={},
    pdfsubject={},
    pdfkeywords={},
    pdfborder=0 0 0,
    pdfpagemode=UseNone,
    colorlinks=true,
    linkcolor=mydarkblue,
    citecolor=mydarkblue,
    filecolor=mydarkblue,
    urlcolor=mydarkblue,
    pdfview=FitH
}

\newtheorem{prop}{Proposition}

\newcommand{\vx}{\bm{x}}
\newcommand{\vz}{\bm{z}}
\newcommand{\vu}{\bm{u}}
\newcommand{\vt}{\bm{t}}
\icmltitlerunning{Can Transformers Learn Full Bayesian Inference In Context?}

\begin{document}

\twocolumn[
\icmltitle{Can Transformers Learn Full Bayesian Inference In Context?}



\icmlsetsymbol{equal}{*}

\begin{icmlauthorlist}
\icmlauthor{Arik Reuter}{lmu}
\icmlauthor{Tim G. J. Rudner}{nyu}
\icmlauthor{Vincent Fortuin}{tum,hel,mcml}
\icmlauthor{David R\"ugamer}{lmu,mcml}
\end{icmlauthorlist}

\icmlaffiliation{lmu}{Department of Statistics, LMU Munich,
Munich, Germany}
\icmlaffiliation{mcml}{Munich Center for Machine Learning (MCML), Munich, Germany}
\icmlaffiliation{nyu}{Center for Data Science, New York University, New York, USA}
\icmlaffiliation{tum}{Department of Computer Science, Technical University of
Munich, Munich, Germany}
\icmlaffiliation{hel}{Helmholtz AI, Munich, Germany.}

\icmlcorrespondingauthor{Arik Reuter}{arik.reuter@campus.lmu.de}
%\icmlcorrespondingauthor{Firstname2 Lastname2}{first2.last2@www.uk}

%\author{Arik Reuter\textsuperscript{1}, Tim G. J. Rudner\textsuperscript{2}, Vincent Fortuin\textsuperscript{3,4,5} \& David Rügamer\textsuperscript{1,5}\\
%\textsuperscript{1}LMU Munich, \textsuperscript{2}New York University, 
%\textsuperscript{3}Technical University of Munich, \textsuperscript{4}Helmholtz AI, \\ \textsuperscript{5}Munich Center for Machine Learning \\

\icmlkeywords{In-Context Learning, Prior-Data Fitted Networks, Bayesian Inference}

\vskip 0.3in
]



\printAffiliationsAndNotice{} % otherwise use the standard text.

\begin{abstract}
Transformers have emerged as the dominant architecture in the field of deep learning, with a broad range of applications and remarkable in-context learning (ICL) capabilities. While not yet fully understood, ICL has already proved to be an intriguing phenomenon, allowing transformers to learn in context---without requiring further training. In this paper, we further advance the understanding of ICL by demonstrating that transformers can perform full Bayesian inference for commonly used statistical models in context. More specifically, we introduce a general framework that builds on ideas from prior fitted networks and continuous normalizing flows and enables us to infer complex posterior distributions for models such as generalized linear models and latent factor models. Extensive experiments on real-world datasets demonstrate that our ICL approach yields posterior samples that are similar in quality to state-of-the-art MCMC or variational inference methods that do not operate in context. The source code for this paper is available at \url{https://github.com/ArikReuter/ICL_for_Full_Bayesian_Inference}
\end{abstract}

\section{Introduction}

In-context learning (ICL) has become a fundamental principle in natural language processing (NLP) with large language models (LLMs) as ubiquitous in-context learners.
The core principle of ICL is that a system adapts to a given task based on information provided in its context. This enables the system to address complex problems, such as question answering or text summarization, using a fixed model without requiring any gradient-based fine-tuning, simply by referencing the context. Thereby, ICL enables the generation of real-time solutions through a localized understanding of data without explicit re-training \citep{dong2022survey, garg2022can}. 
%A simple example of utilizing the ICL capabilities of LLMs is text summarization: After inputting a text together with a suitable prompt (the \textit{context}), the LLM first computes an internal representation of the context. Then, a summary is generated through autoregressive sampling while referring to the context using self-attention. The model thus learns a new (summarized) representation of the original text in-context. 

A fundamental benefit of ICL with LLMs is its versatility. Almost every NLP task involving small data can be solved in context using LLMs, while the performance often surpasses existing baselines \citep{touvron2023llama, achiam2023gpt, team2023gemini}. Additionally, achieving this performance can be very straightforward, requiring only suitably formulated prompts in natural language. Excellent results across a broad variety of tasks, combined with fast inference times and ease of usability, have made in-context learning a machine learning tool employed by millions of people \citep{eloundou2023gpts}.

Furthermore, ICL has recently shown remarkable promise for regression and classification tasks involving tabular data, with tabular prior-data fitted networks (TabPFNs) dominating benchmarks alongside minimal prediction time \citep{hollmann2022tabpfn, hollmann2025accurate, hootabular, robertson2024fairpfn}. While the internet serves as a suitable source for the massive data needed to train in-context learners on text, TabPFNs demonstrate that training on purely synthetic data facilitates the development of in-context learners for tabular data.

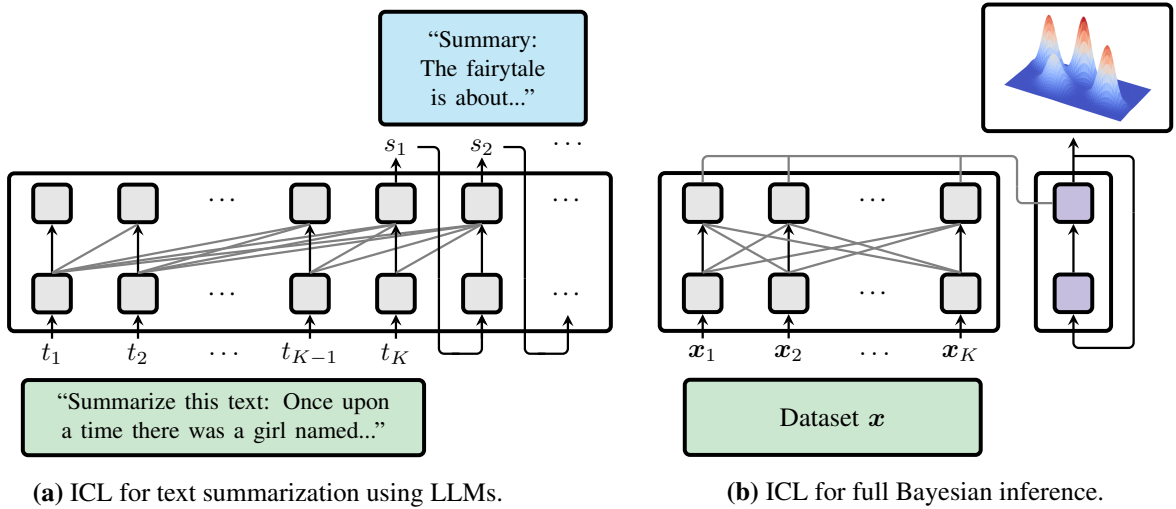
\begin{figure*}
    \centering
    % First figure
    \vspace*{5pt}
    \hspace*{10pt}
    \scalebox{1.0}{
    \begin{minipage}{0.4\textwidth}
        \centering
        \begin{tikzpicture}[x=1.0cm,y=1.0cm,scale=1,baseline={(0,0)}]

            \tikzstyle{arrow} = [->,>=stealth, line width = 0.1mm, rounded corners=2pt]
            \tikzstyle{arrow_thick} = [->,>=stealth, line width = 0.3mm, rounded corners=2pt]
            \tikzstyle{arrow_back} = [<-,>=stealth, line width = 0.1mm, rounded corners=2pt]
            \tikzstyle{arrow_double} = [<->,>=stealth, line width = 0.1mm, rounded corners=2pt]
            \tikzstyle{line} = [-,>=stealth, line width = 0.1mm, rounded corners=2pt]
            \tikzstyle{line_thick} = [-,>=stealth, line width = 0.3mm, rounded corners=2pt]
            \tikzstyle{line_gray} = [-,>=stealth, line width = 0.3mm, color = gray, rounded corners=2pt]
            
            \node (background) [rectangle,rounded corners = 2pt, text centered, draw=black, line width = 1.5pt, minimum width = 8.0cm, minimum height = 2.1cm, fill = white, xshift = 0cm, yshift = -0.65cm] {};

            \node (context) [rectangle,rounded corners = 2pt, text centered, draw=black, line width = 1.5pt, minimum width = 5.1cm, minimum height = 1cm, fill = lgreen, xshift = -1.15cm, yshift = -2.85cm, text width = 5.1cm] {\small ``Summarize this text: Once upon a time there was a girl named...'' };

            \node (output) [rectangle,rounded corners = 2pt, text centered, draw=black, line width = 1.5pt, minimum width = 2.5cm, minimum height = 1.5cm, fill = lblue, xshift = 2.3cm, yshift = 1.8cm, text width = 2.5cm] {\small``Summary: The fairytale is about...'' };

            \node (i1) [xshift = -3.429cm, yshift = -2cm] {$t_1$};

            \node (i2) [xshift = -2.286cm, yshift = -2cm] {$t_2$};

            \node (i3) [xshift = -1.143cm, yshift = -2cm] {$\ldots$};

            \node (i4) [xshift = -0.0cm, yshift = -2cm] {$t_{K-1}$};

            \node (i5) [xshift = 1.143cm, yshift = -2cm] {$t_K$};

            \node (i6) [xshift = 2.286cm, yshift = -2cm] {};

            \node (i7) [xshift = 3.429cm, yshift = -2cm] {};

            \foreach \n in {1,2,4,5,6}{
                \node (j\n) [rectangle,rounded corners = 2pt, text centered, draw=black, line width = 1.5pt, minimum width = 0.5cm, minimum height = 0.5cm, fill = lgrey, above of = i\n, yshift = -0.2 cm] {} ;

                \node (k\n) [rectangle,rounded corners = 2pt, text centered, draw=black, line width = 1.5pt, minimum width = 0.5cm, minimum height = 0.5cm, fill = lgrey, above of = j\n, yshift = 0.2 cm] {} ;

                \draw[arrow_thick] ([yshift=2mm]i\n.center) -- (j\n);
            }

            \node (j3) [above of = i3, yshift = -0.2 cm] {\ldots};

            \node (k3) [above of = j3, yshift = 0.25 cm] {\ldots} ;

            \node (j7) [above of = i7, yshift = -0.2 cm] {\ldots};

            \node (k7) [above of = j7, yshift = 0.25 cm] {\ldots} ;

            \node (l5) [above of = k5, yshift = -0.25 cm] {$s_1$};

            \node (l6) [above of = k6, yshift = -0.25 cm] {$s_2$};

            \node (l7) [above of = k7, yshift = -0.25 cm] {$\ldots$};

%%%%%%%%%%%%%%%%%%%%%%%%%%%%%%%%%%%%%%%%%%%%%%%%%%%%%%%%%%%%%%%%%%%%%%%%%%%%%%%%%%%%%%%%%%%%%%%%%%%%%%%%%%%%%%%%%%%%%%%%%%%%%%%

            \draw [arrow_thick] (j1.north) -- (k1.south);
            \draw [line_gray] (j1.north) -- (k2.south);
            \draw [line_gray] (j1.north) -- (k4.south);
            \draw [line_gray] (j1.north) -- (k5.south);
            \draw [line_gray] (j1.north) -- (k6.south);

            \draw [arrow_thick] (j2.north) -- (k2.south);
            \draw [line_gray] (j2.north) -- (k4.south);
            \draw [line_gray] (j2.north) -- (k5.south);
            \draw [line_gray] (j2.north) -- (k6.south);

            \draw [arrow_thick] (j4.north) -- (k4.south);
            \draw [line_gray] (j4.north) -- (k5.south);
            \draw [line_gray] (j4.north) -- (k6.south);

            \draw [arrow_thick] (j5.north) -- (k5.south);
            \draw [line_gray] (j5.north) -- (k6.south);
            \draw [arrow_thick] (j6.north) -- (k6.south);

            \draw [arrow_thick] (k5) -- ([yshift=0.9mm]l5.south);
            \draw [arrow_thick] (k6) -- ([yshift=0.9mm]l6.south);

            \draw [line_thick] (l5.east) -| (1.7144, -1);
            \draw [line_thick] (1.7144, -1) |- ([xshift=-3mm]i6.center);
            \draw [arrow_thick] ([xshift=-4mm]i6.center) -| (j6.south);

            \draw [line_thick] (l6.east) -| (2.857, -1);
            \draw [line_thick] (2.857, -1) |- ([xshift=-3mm]i7.center);
            \draw [arrow_thick] ([xshift=-4mm]i7.center) -| ([yshift=-1.4mm]j7.south);

        \end{tikzpicture}
    \\
    \vspace{3mm}
    \textbf{(a)} ICL for text summarization using LLMs.
    \end{minipage}
    }
    \hfill
    % Second figure
    \scalebox{1.0}{
    \begin{minipage}{0.4\textwidth}
        \vspace{-1mm}
        \centering
        \begin{tikzpicture}[x=1.0cm,y=1.0cm,scale=1,baseline={(0,0)}]

            \tikzstyle{arrow} = [->,>=stealth, line width = 0.1mm, rounded corners=2pt]
            \tikzstyle{arrow_thick} = [->,>=stealth, line width = 0.3mm, rounded corners=2pt]
            \tikzstyle{arrow_back} = [<-,>=stealth, line width = 0.1mm, rounded corners=2pt]
            \tikzstyle{arrow_double} = [<->,>=stealth, line width = 0.1mm, rounded corners=2pt]
            \tikzstyle{line} = [-,>=stealth, line width = 0.1mm, rounded corners=2pt]
            \tikzstyle{line_thick} = [-,>=stealth, line width = 0.3mm, rounded corners=2pt]
            \tikzstyle{line_gray} = [-,>=stealth, line width = 0.3mm, color = gray, rounded corners=2pt]
            \tikzstyle{line_gray_thick} = [-,>=stealth, line width = 0.3mm, color = gray, rounded corners=2pt]
            
            \node (background) [rectangle,rounded corners = 2pt, text centered, draw=black, line width = 1.5pt, minimum width = 4.5cm, minimum height = 2.1cm, fill = white, xshift = -1.75cm, yshift = -0.65cm] {};

            \node (background2) [rectangle,rounded corners = 2pt, text centered, draw=black, line width = 1.5pt, minimum width = 1cm, minimum height = 2.1cm, fill = white, xshift = 1.5cm, yshift = -0.65cm] {};

            \node (context) [rectangle,rounded corners = 2pt, text centered, draw=black, line width = 1.5pt, minimum width = 3.7cm, minimum height = 1cm, fill = lgreen, xshift = -1.70cm, yshift = -2.85cm, text width = 3.7cm] {Dataset $\vx$};

            \node (output) [rectangle,rounded corners = 2pt, text centered, draw=black, line width = 1.5pt, minimum width = 2.5cm, minimum height = 1.5cm, fill = white, xshift = 1.55cm, yshift = 1.8cm] 
            {{\includegraphics[height=1.45cm]{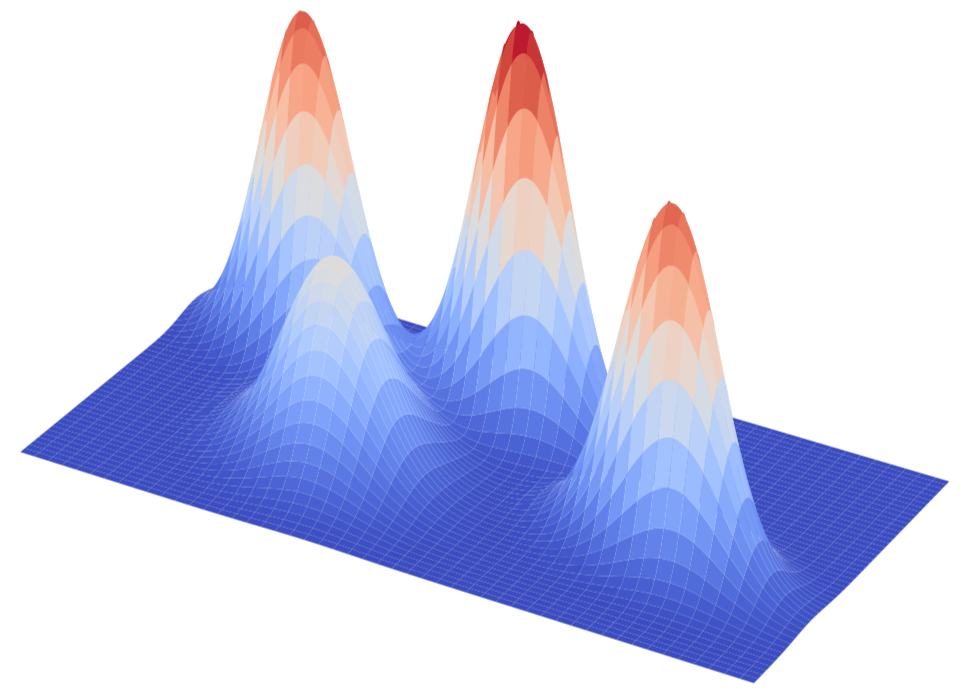}}};

            \node (i1) [xshift = -3.429cm, yshift = -2cm] {$\bm{x}_1$};

            \node (i2) [xshift = -2.286cm, yshift = -2cm] {$\bm{x}_2$};

            \node (i3) [xshift = -1.143cm, yshift = -2cm] {$\ldots$};

            \node (i4) [xshift = -0.0cm, yshift = -2cm] {$\bm{x}_K$};

            \node (i6) [xshift = 1.5cm, yshift = -2cm] {};

            \foreach \n in {1,2,4}{
                \node (j\n) [rectangle,rounded corners = 2pt, text centered, draw=black, line width = 1.5pt, minimum width = 0.5cm, minimum height = 0.5cm, fill = lgrey, above of = i\n, yshift = -0.2 cm] {} ;

                \node (k\n) [rectangle,rounded corners = 2pt, text centered, draw=black, line width = 1.5pt, minimum width = 0.5cm, minimum height = 0.5cm, fill = lgrey, above of = j\n, yshift = 0.2 cm] {} ;

                \draw[arrow_thick] ([yshift=2mm]i\n.center) -- (j\n);
            }

            \node (j3) [above of = i3, yshift = -0.2 cm] {\ldots};

            \node (k3) [above of = j3, yshift = 0.25 cm] {\ldots} ;

            \node (j6) [rectangle,rounded corners = 2pt, text centered, draw=black, line width = 1.5pt, minimum width = 0.5cm, minimum height = 0.5cm, fill = lpurple, above of = i6, yshift = -0.2 cm] {} ;

            \node (k6) [rectangle,rounded corners = 2pt, text centered, draw=black, line width = 1.5pt, minimum width = 0.5cm, minimum height = 0.5cm, fill = lpurple, above of = j6, yshift = 0.2 cm] {} ;

            \node (atts1) [yshift = 0.75 cm, xshift = -1.75cm] {} ;

            \node (l6) [yshift = 2.2cm, above of = i6] {
            %$\vz \sim P^{\vz|\vx}$
            };

%%%%%%%%%%%%%%%%%%%%%%%%%%%%%%%%%%%%%%%%%%%%%%%%%%%%%%%%%%%%%%%%%%%%%%%%%%%%%%%%%%%%%%%%%%%%%%%%%%%%%%%%%%%%%%%%%%%%%%%%%%%%%%%
        
            \draw [arrow_thick] (j1.north) -- (k1.south);
            \draw [line_gray] (j1.north) -- (k2.south);
            \draw [line_gray] (j1.north) -- (k4.south);
            
            \draw [arrow_thick] (j2.north) -- (k2.south);     
            \draw [line_gray] (j2.north) -- (k1.south);
            \draw [line_gray] (j2.north) -- (k4.south);

            \draw [arrow_thick] (j4.north) -- (k4.south);
            \draw [line_gray] (j4.north) -- (k1.south);
            \draw [line_gray] (j4.north) -- (k2.south);

            \draw [arrow_thick] (j6.north) -- (k6.south);

            \draw [line_thick] ([yshift=3.5mm]k6.north) -| ([yshift=-4mm, xshift=8mm]k6.north);
            \draw [line_thick] ([yshift=-4mm, xshift=8mm]k6.north) |- ([yshift=-4.5mm, xshift=3mm]j6.south);
            \draw[arrow_thick] ([yshift=-4.5mm, xshift=3mm]j6.south) -| (j6.south);

            \draw [line_gray] (k1.north) |- (atts1.south);
            \draw [line_gray] (k2.north) -- ([yshift=3.5mm]k2.north);
            \draw [line_gray] (k4.north) -- ([yshift=3.5mm]k4.north);

            \draw [line_gray_thick] (atts1.south) -| ([xshift = -0.75cm, yshift = 0.5cm]k6.south);
            \draw [line_gray_thick] ([xshift = -0.75cm, yshift = 0.5cm]k6.south) |- (k6.west); 

            \draw [arrow_thick] (k6.north) -- ([yshift = -2mm]l6.south);
        \end{tikzpicture}
    \\
    \vspace{3mm}
    \textbf{(b)} ICL for full Bayesian inference.
    \end{minipage}}
    \hspace*{10pt}
    \vspace*{5pt}
    \caption{\textbf{(a)} An LLM generates a summary $s_1, s_2, \ldots$ of a text $t_1, t_2, \ldots, t_K$ through autoregressive sampling while referring to the context using masked self-attention. 
\textbf{(b)} A dataset $\vx$ is processed with a transformer encoder. Subsequently, cross attention allows generating samples from the posterior conditioned on $\vx$ in context using a diffusion transformer (decoder). The samples are generated by solving a neural differential equation defining a continuous normalizing flow.}
    \label{fig:Overview}
\end{figure*}

While PFNs perform Bayesian inference, they target a univariate, typically discrete, posterior predictive distribution. In numerous applications, however, high-dimensional and continuous posteriors \smash{$P^{\vz|\vx}$} of (latent) variables $\vz$ given data $\vx$ play a key role.\footnote{We do not assume any specific form of $\vz$. That is, there can be a single $\vz_j$ associated with each data point $\vx_j$ in $\vx$, but the case where a single ``global'' $\vz$ governs the behavior of each $\vx_j$ in $\vx$ is equally included in this notation.} This includes areas such as healthcare \citep{kyrimi2021comprehensive, abdullah2022review, etzioni1995bayesian}, physics \citep{gebhard2025flow, brehmer2022simulation, dax2024real}, and neuroscience \citep{lueckmann2017flexible, sohn2021neural}. We use the notion of \emph{full Bayesian inference} for methods yielding potentially complex and high-dimensional posterior distributions—in contrast to, for instance, methods that yield only the posterior predictive or point estimates of the posterior as, for example \citet{hollmann2022tabpfn}. However, performing full Bayesian inference can be challenging, even for relatively simple models such as generalized linear models \citep[GLMs;][]{nelder1972generalized}. Two common issues when performing full Bayesian inference include (a) slow inference time, particularly when using sampling-based methods \citep{sommer2025mile, sommer2024connecting}, and (b) model misspecification. Although potentially restrictive modeling assumptions are often necessary to make Bayesian inference efficient or even feasible, they can lead to suboptimal predictive performance \citep{wang2019variational, walker2013bayesian}. 

In this paper, we address the following question: \textit{Can we leverage in-context learning to effectively perform full Bayesian inference?}
In doing so, we aim to obtain an in-context learner that can perform the mapping \smash{$\vx \mapsto P^{\vz|\vx}$} for a specific probabilistic model, and, analogous to LLMs, (a) allows for the rapid generation of samples from a posterior of interest during deployment and (b) can flexibly adapt to a broad range of inputs, thereby overcoming issues arising from model misspecification. More specifically, our approach combines a TabPFN encoder \citep{hollmann2022tabpfn} and a diffusion transformer-decoder \citep{peebles2023scalable} that is trained via flow matching \citep{lipman2022flow}.

We present the results of our in-context learning approach on extensive real-world and synthetic datasets in \Cref{sec:results} and discuss the challenges and the transformative potential of in-context learning for full Bayesian inference in \Cref{sec:discussion}.

% The remainder of this paper is structured as follows: \Cref{sec:related_work} discusses how ideas from the in-context learning literature, amortized inference and simulation-based inference relate to our work. \Cref{sec:methodology} provides the background for our methodology and explains how the generative nature of many probabilistic models allows training the in-context learner via normalizing flows. Then, we present the results of our in-context learning approach on extensive real-world and synthetic datasets in \Cref{sec:results}. 
% Finally, we discuss the challenges and the transformative potential of in-context learning for full Bayesian inference in \Cref{sec:discussion}.

%\subsection{Our contribution}

\pagebreak

% \paragraph{Contributions.}
{To summarize, our main contributions are as follows:}\vspace*{-8pt}
\begin{enumerate}[leftmargin=15pt]
\setlength\itemsep{0pt}
    \item
    We develop, train, and examine a model that yields samples from the posterior distribution $P^{\vz|\vx}$ 
    given data $\vx$ as context without any (explicit) parameter updates or parametric assumptions about the posterior. 
    \item
    To achieve this, we propose to use synthetic samples from the joint distribution $P^{\vx, \vz}$ in order to train a large transformer model that performs ICL regarding the posterior $P^{\vz|\vx}$, and provide a general framework to analyze the circumstances that enable learning $P^{\vz|\vx}$ purely through samples from $P^{\vx, \vz}$. %Furthermore, to facilitate ICL of the posteriors of commonly used statistical models, we develop an architecture that combines a powerful transformer encoder and a decoder.
%that allows to approximate arbitrary posterior distributions. 
\item We then analyze the efficacy of our approach for GLMs and latent factor models, namely Gaussian mixture models (GMMs) and factor analysis (FA).
For these applications, we show that including the ``prior'' used for TabPFNs results in reliably inferring posterior distributions on real-world data.
\item
In a variety of experiments, we demonstrate that this approach yields posterior samples that are very similar to those from a Hamiltonian Monte Carlo sampler. Furthermore, we find that the quality of the samples from our ICL approach is preferable, when compared to various popular VI techniques that do not operate in context.
\item 
Finally, we conduct ablation studies of our approach, examining, for instance, alternative diffusion objectives and Gaussian approximations in place of flow matching, the model's performance on out-of-distribution data, and the impact of problem dimensionality.
%\footnote{The source code for this paper is available at \small \url{https://github.com/ArikReuter/ICL_for_Full_Bayesian_Inference}.}
\end{enumerate}

\clearpage

% \vspace{-0.3cm}
\section{Related Work}

\label{sec:related_work}

Beyond the perspective of prior-data fitted networks, the contribution of this work can be summarized from the viewpoints of recent work on in-context learning, amortized Bayesian inference, and, simulation-based inference.  

\paragraph{In-Context Learning.}
ICL is a special case of meta-learning \citep{hospedales2021meta} characterized by using a large pre-trained model in order to learn from a context dataset without explicitly updating task-specific parameters. Several recent lines of work investigate the in-context learning capabilities of transformers \citep{garg2022can, ahuja2023context, wang2024large, chan2022transformers}. 

\citet{garg2022can} show that a model similar to GPT-2 can implicitly implement various interesting function classes in context. More specifically, the model learns to reproduce the predictions of different statistical models such as (sparse) linear functions, decision trees, and even two-layer neural networks. This approach can be extended to multiple families of functions and even mixtures of tasks \citep{ahuja2023context}. \citet{kirsch2022general} investigate ICL as a general principle for meta-learning. However, the results by \citet{garg2022can} and  \citet{ahuja2023context} are restricted to relatively small problem scales and scalar-valued predictions instead of multivariate posterior distributions. Additionally, the experiments are conducted exclusively on simulated data. In contrast, our results show that (large) transformer models can effectively learn multivariate posterior distributions over latent variables in context on real-world datasets. Furthermore, the focus on latent-variable models naturally steers our investigation toward unsupervised in-context learning, where the primary objective is to uncover the underlying structure of the data rather than to make predictions based on input-target paris presented in the context.

Concurrently, \citet{mittal2025context} conduct a comparative analysis of amortized in-context Bayesian posterior estimation methods, ablating over different optimization objectives and architectural choices, and also reporting results on out-of-distribution performance and flow-matching methods. They focus on evaluating the posterior mean and downstream predictive performance, whereas we evaluate full posterior distributions. In an analogous setup, \citet{mittal2025amortized} assess the effectiveness of learning point estimates versus learning entire distributions in context for the goal of predictive performance.

\paragraph{Amortized Inference.}

Amortized inference is a central paradigm in the field of variational inference \citep{kingma2013auto, zhai2018autoencoder, kim2018semi, margossian2023amortized}. A commonly used idea here is to model the posterior distribution \smash{$P^{\vz|\vx}$} of latent variables $\vz$ given a dataset $\vx$ via a factorized density \smash{$p(\vz|\vx) \approx \prod_{j=1}^K q_\theta(\vz_j|h_{\bm{\phi}}(\vx_j))$}. In contrast to our more general assumption, each datapoint $\vx_j$ in $\vx$ is assumed to have a corresponding latent variable $\vz_j$. While the parameter $\theta$ determines global aspects of the variational distribution, the function $h_{\bm{\phi}}$ is shared for all $\vx_j$ and thus amortized across data $\vx$. Variational autoencoders \citep{kingma2013auto, rezende2014stochastic} and neural processes \citep{garnelo2018conditional, garnelo2018neural, rudner2018connection} are important model classes based on amortized inference.

In comparison, our ICL approach amortizes its parameters on the level of datasets, such that a single functional relationship is learned for a set \smash{$\mathcal{D} \subset (\mathcal{X} \times \mathcal{Z})^N$} of datasets. From this point of view, \smash{$\mathcal{D} = \left \{\left ( \vx_i, \vz_i \right) \right \}_{i=1}^N$} comprising $N$ datasets $\vx_i \in \mathcal{X}$ and the corresponding latent variables $\vz_i \in \mathcal{Z}$ can be seen as a ``meta-dataset'' for which we perform amortized inference. This is similar in nature to the setup by \citet{le2017inference}, who use recurrent neural networks to ``compile'' inference based on execution traces of probabilistic programs by training on simulated data. 

Unlike in amortized variational inference, we do not use the notion of an evidence lower bound \citep{blei2017variational} or even the Kullback-Leibler divergence to learn the posterior, but utilize ideas that also appear in the context of simulation-based inference.
%Note that assuming that $\mathcal{D}$ contains i.i.d. samples of datasets directly implies that a sufficiently expressive in-context-learner can completely close the amortization gap \citep{}.

\paragraph{Simulation-Based Inference.}

Analogously to latent variable models, some scientific simulations, for instance in neuroscience or astrophysics \citep{fan2019brief, schmit2018emulation}, allow to draw samples from the joint distribution \smash{$P^{\vx, \vz}$} of data and latent variable of interest. Amortized posterior inference in this context is referred to as simulation-based inference \citep[SBI;][]{cranmer2020frontier}. Several recent approaches focus on using neural networks to directly infer aspects of the likelihood $p(\vx | \vz)$, the posterior \smash{$P^{\vz | \vx}$} or the joint distribution \smash{$P^{\vx, \vz}$}. More specifically, techniques based on %variational inference \citep{glockler2022variational} 
discrete normalizing flows \citep{dax2021real} or flow-matching \citep{wildberger2024flow} are used to approximate the posterior \smash{$P^{\vz | \vx}$}, while \cite{gloeckler2024all} propose to use a transformer-based diffusion model in order to approximate the joint distribution \smash{$P^{\vx, \vz}$}. In recent work, \citet{vetter2025effortless} directly a pre-trained TabPFN to auto-regressively sample \smash{$P^{\vz | \vx}$} leveraging ICL.

From a simulation-based inference viewpoint, we demonstrate that sample-based posterior estimation can be used for full Bayesian inference in complex scenarios arising in commonly used latent variable models, and demonstrate the effectiveness of this approach on real-world datasets.

\section{In-Context Learning for Full Bayesian Inference} \label{sec:methodology}

Bayesian inference is a tool of central importance for countless applications. However, exact posterior inference can become computationally expensive when using sampling-based methods \citep{hastings1970monte, hoffman2014no, betancourt2017conceptual} and even impossible when relying on fully factorized VI methods, which can incur substantial approximation errors \citep{bishop2002vibes, blei2012probabilistic,margossian2023amortized}. 
Amortized variational inference can alleviate those issues but typically requires the development of specialized and complex modeling frameworks \citep{kingma2013auto, srivastava2017autoencoding, garnelo2018neural, lin2021task}. Another issue with variational inference arises from having to choose a variational distribution. While insufficient flexibility in this respect can lead to overly simplistic posteriors, a too flexible variational distribution might overfit the given data \citep{cremer2018inference}.

We propose a simple and effective solution based on ideas from ICL, which can be seen as conducting amortized inference on a dataset level. %Concretely, investing substantial computational resources for training 
Training a model on a potentially unlimited amount of synthetic datasets yields %an in-context learner that 
an in-context learner that can not only approximate a vast, almost arbitrarily large, class of distributions but is also highly efficient when used for sampling. Furthermore, this does not incur the same issues with overly or insufficiently flexible distribution assumptions that are present in VI. More specifically, empirical results show that a major strength of TabPFN, for instance, is its ability to adapt flexibly to the complexity of the problem at hand, thus removing the need for extensive hyperparameter tuning \citep{hollmann2022tabpfn}.

In the following, we describe a sufficient general condition, as well as a specific framework that allows to train probabilistic in-context learners on simulated data. 

The idea underlying the proposed approach is founded on two observations relating to full Bayesian inference and the working principle of PFNs:  First, many Bayesian models have a generative formulation that allows the simulation of arbitrarily large amounts of training samples from the joint distribution $P^{\vx, \vz}$. We assume that samples from $P^{\vx, \vz}$ comprise a dataset $\smash{\vx = \left \{ \vx_j \right  \}_{j=1}^K}$ containing $K$ samples $\vx_j \in \mathcal{X}$ and a corresponding (latent) variable $\vz \in \mathcal{Z}$.\footnote{We do not assume any specific form of $\vz$. That is, there can be a single $\vz_j$ associated with each data point $\vx_j$ in $\vx$, but the case where a single ``global'' $\vz$ governs the behavior of each $\vx_j$ in $\vx$ is equally included in this notation.} This joint distribution $P^{\vx, \vz}$ corresponds to the ``prior'' in PFNs and allows the training of a large neural network that implicitly learns to perform Bayesian inference. Second, Bayesian inference is especially useful for smaller datasets $\vx$ that can be processed in a single forward pass. This makes an entire dataset a viable context for Bayesian ICL.

More specifically, the central goal is to develop a method allowing to infer the posterior distribution $P^{\vz|\vx}$ of latent variables $\vz \in \mathcal{Z}$, given observations $\vx \in \mathcal{X}$ using ICL. 
%As always in Bayesian inference, the posterior is defined via Bayes' rule: $p(\vz|\vx) = \frac{p(\vx| \vz) p(\vz)}{p(\vx)}$.
From a supervised-learning perspective, we thus aim to directly learn the mapping $f_0: \mathcal{X} \rightarrow \mathcal{M}(\mathcal{Z}), \vx \mapsto P^{\vz|\vx}$, where $\mathcal{M}(\mathcal{Z})$ is the space of all probability measures. Therefore, we want a model $f_{\theta}(\vx) = Q^{\vz|\vx}_{\theta}$ for the posterior %, which can also be seen as a function $f_{\theta}(\vx) = Q^{\vz|\vx}_{\theta}$, 
to be as close as possible to the true posterior $P^{\vz|\vx} = f_0(\vx)$. We measure ``closeness'' w.r.t.\ some divergence $d:\mathcal{M}(\mathcal{Z}) \times \mathcal{M}(\mathcal{Z}) \to [0,\infty)$. When considering the expected divergence over data samples $\vx \sim P^{\vx}$, this gives rise to the following objective: $\mathcal{R}_{\theta} \defeq \mathbb{E}_{\vx \sim p(\vx)} \left [ d \left (f_{\theta}(\vx), f_0(\vx) \right )\right ]$, which can also be directly expressed as
\begin{equation}
    \label{eq:general_objective}
    \mathcal{R}_{\theta} = \mathbb{E}_{\vx \sim p(\vx)} \left [ d \left (Q^{\vz|\vx}_{\theta},  P^{\vz|\vx} \right )\right ].
\end{equation}
Note that we use the notion of a divergence $d$ loosely to refer to any measure of similarity of two distributions. 
Although $\mathcal{R}_{\theta}$  itself is usually intractable, specific choices of $d$ and the use of the joint distribution $P^{\vx, \vz}$  make \Cref{eq:general_objective} accessible via
\newcommand\Rthetatilde{\stackrel{\sim}{\smash{\mathcal{R}_{\theta}}\rule{0pt}{1.1ex}}}
\begin{equation}
     \Rthetatilde\defeq \mathbb{E}_{\vx, \vz \sim p(\vx, \vz)} \left [ \mathcal{L}_d(\vx, \vz, \theta) \right],
\end{equation}
where the loss function $\mathcal{L}_d$ depends on $d$ and the structure of $Q_{\theta}^{\vz|\vx}$ (discussed in detail later). Performing empirical risk minimization for $\Rthetatilde$ with samples from the joint distribution $P^{\vx, \vz}$ then corresponds to learning to approximate $ P^{\vz|\vx}$. The model for the posterior $P^{\vz|\vx}$ is thereby only implicitly defined by the joint distribution $P^{\vx, \vz}$. While this requires the ability to sample from $P^{\vx, \vz}$, drawing samples from the joint distribution is often a weak requirement in terms of model specification that immediately follows from specifying the generative process of a model.
Furthermore, a simple sufficient condition that follows directly from the law of total expectation implies the equivalence of $\mathcal{R}_{\theta}$ and $\Rthetatilde$:
\begin{prop}
\label{th:suffcond}
    Let $d(Q_{\theta}^{\vz|\vx},  P^{\vz|\vx}) = \displaystyle \int \gamma \left (Q_{\theta}^{\vz|\vx} \right ) dP^{\vz|\vx}$ for some measurable functional $\gamma: \mathcal{M}(\mathcal{Z}) \rightarrow \mathbb{R}$. \\ 
    Then $\mathcal{R}_{\theta} = \Rthetatilde$ with $\mathcal{L}_d(\vx, \vz, \theta) = \gamma \left (Q_{\theta}^{\vz|\vx} \right )$. 
\end{prop}
For instance, choosing $d$ to be the forward Kullback-Leibler divergence $d_{\textrm{KL}}(Q_{\theta}^{\vz|\vx},  P^{\vz|\vx}) = \mathbb{D}_{\textrm{KL}} \left [ p(\cdot| \vx) \lvert \rvert q_{\theta}(\cdot| \vx)  \right ]$ implies that $\mathcal{L}_{d_{\textrm{KL}}}(\vx, \vz, \theta) = - \log q_{\theta}(\vz|\vx) + const.$ \citep{muller2021transformers}. In this case, minimizing $ \Rthetatilde$ thus directly corresponds to performing maximum likelihood inference on samples from $P^{\vx, \vz}$. 

\subsection{Defining the Form of the Posterior} \label{sec:posterior}

\newcommand{\PG}{P^{\mathcal{G}} }  %ground-truth distribution
\newcommand{\PBase}{P_{\mathcal{B}} }  %Simple noise base distributin

\newcommand{\pg}{p_{\mathcal{G}}}  %ground-truth density
\newcommand{\pbase}{p_{\mathcal{B}}}  %base noise density
\DeclarePairedDelimiter\psw{[}{]_\sharp}
\newcommand{\Id}{I}

%Conditional Flow Matching (CFM), originating in the domain of image generation \citep{lipman2022flow}, is a technique to model arbitrary distributions based on normalizing flows \citep{papamakarios2021normalizing}. Applications of this framework range from state-of-the-art image synthesis \citep{esser2024scaling}, over speech LLMs \citep{zhang2024speechgpt} to protein ensembles \citep{jing2024alphafold}. In the area of SBI, in particular an application to gravitational waves, \cite{wildberger2024flow} explore the potential of CFM.

To learn the posterior distribution $P^{\vz|\vx}$ in context, we use the framework of flow matching \citep{lipman2022flow}. More specifically, we utilize continuous normalizing flows (CNFs) to specify and ultimately sample from $P^{\vz|\vx}$. CNFs, currently excelling in the field of image synthesis \citep{esser2024scaling}, do not only allow to flexibly learn almost arbitrary distributions, but are also found to be more sample-efficient in training than for instance diffusion objectives \citep{lipman2022flow, wildberger2024flow}.
Furthermore, unlike discrete normalizing flows \citep{JMLR:v22:19-1028}, CNF objectives do not limit the architecture of the used neural network, allowing to incorporate complex conditioning on the data $\vx$ in addition to flexibly modeling the posterior, which is a crucial aspect of our ICL framework. Refer to \Cref{app:background_cfm} for more information on CNFs.

\subsubsection{Normalizing flows}
The key idea of modeling a distribution $P^{\vz|\vx}$ with normalizing flows \citep[see, e.g.,][]{papamakarios2021normalizing}, which are the basis of CNFs, is to assume that $P^{\vz|\vx}$ is the result of ``pushing forward'' a simple base distribution $\PBase$ into $P^{\vz|\vx}$ using a conditional flow $\psi_{\theta}(\cdot| \vx)$:
\begin{equation}
    P^{\vz|\vx} \approx \psw{\psi_{\theta}(\cdot| \vx)} \PBase.
\end{equation}
Therefore, one assumes that samples from $P^{\vz|\vx}$ are generated by first drawing $\vz^{(0)} \sim \PBase$, and then applying $\psi_{\theta}(\cdot |
\vx)$, such that $\psi_{\theta}(\vz^{(0)} | \vx) \sim P^{\vz|\vx}$. The base distribution $\PBase$ is commonly set to be a standard normal distribution, i.e., $\PBase = \mathcal{N}(0, \Id)$. The conditional flow $\psi_{\theta}(\cdot| \vx)$ is the object to be learned, such that our model of $P^{\vz|\vx}$ is defined as $Q^{\vz|\vx}_{\theta} \defeq  \psw{\psi_{\theta}(\cdot| \vx)} \PBase$.

\subsubsection{Continuous Normalizing Flows}

In flow matching \citep{lipman2022flow}, which we will use to obtain an in-context learner for full Bayesian inference, the normalizing flow $\psi_{\theta}(\cdot| \vx)$ is implicitly defined via a (conditional) vector field $v^{\theta}_{t, \vx}$ of an ordinary differential equation (ODE):
\begin{equation}
    \label{eq:ode_flows}
    \frac{d}{dt} \psi_{\theta, t}(\vz| \vx) = v^{\theta}_{t, \vx}(\psi_{\theta, t}(\vz| \vx)), \ \psi_{\theta, 0}(\vz| \vx) = \vz, 
\end{equation}
where $0\leq t \leq 1$. The first condition $\frac{d}{dt} \psi_{\theta, t}(\vz| \vx) = v^{\theta}_{t, \vx}(\psi_{\theta, t}(\vz| \vx))$ means that $v^{\theta}_{t, \vx}$ describes the change in $\psi_{\theta, t}(\vz| \vx)$ at time $t$, and the second condition $\psi_{\theta, 0}(\vz| \vx) = \vz$ implies that initially the flow is just the identity. The family of vector fields $v^{\theta}_{t, \vx}$ is parameterized by a neural network whose parameters $\theta$ will be learned. In order to ultimately compute the flow $v^{\theta}_{1, \vx}$, that yields $Q^{\vz|\vx}_{\theta} = \psw{\psi_{\theta, 1}(\cdot| \vx)} \PBase$, a numerical ODE solver can be used to forward-solve the ODE, which ultimately corresponds to evaluating $\psi_{1, \vx}$ at a data point $\vz^{(0)} \sim \PBase$. This construction  implies very generic assumptions regarding the structure of $Q^{\vz|\vx}$, which include the existence of a density of the target distribution wrt. the Lebesgue measure, and the assumption that $P^{\vz|\vx}$ can be represented by a mixture distribution over the marginal probability paths at time point $t=1$ \citep{lipman2022flow}. Please note that the mathematical understanding of Flow Matching, its properties and assumptions, are still actively researched \citep{wildberger2024flow}.

Assuming Gaussian conditional probability paths with an optimal-transport mean- and variance-function \citep{lipman2022flow}, one obtains the following discrepancy measure $d_{\textrm{CFM}}$ between $Q^{\vz|\vx}_{\theta} \defeq  \psw{\psi_{\theta, 1}(\cdot| \vx)} \PBase$ and $P^{\vz|\vx}$:
\begin{multline}
        \label{eq:CFMdist}
     d_{\textrm{CFM}} \left (Q^{\vz|\vx}_{\theta},  P^{\vz|\vx} \right ) \defeq \\
    \mathbb{E} \left [ \left \lvert \left \rvert v_{t,\vx}^{\theta}(\gamma_t(\vz^{(1)} |\vz^{(0)} )) - {(\vz^{(1)}  - \omega \vz^{(0)}) } \right \lvert \right \rvert_2 ^2  \right ],
\end{multline}

where the expectation is taken w.r.t.\ to three random variables: a uniform time-step $t \sim \mathcal{U}([0,1])$, samples from the base distribution $\vz^{(0)} \sim \PBase$, and samples from the ground-truth conditional distribution $\vz^{(1)}  \sim P^{\vz|\vx}$. We define $\gamma_t(\vz^{(1)} |\vz^{(0)} ) \defeq (1 - {\omega}t)\vz^{(0)} + t \vz^{(1)} $.

 We refer to \cite{wildberger2024flow} for mathematical results on the relationship of $d_{\textrm{CFM}}$ and the (forward) Kullback-Leibler divergence. The hyperparameter $\omega = 1 - \sigma_{\textrm{min}}$, where $\sigma_{\textrm{min}}$ is the variance at time $t=1$ in the Gaussian conditional probability paths, appears to have negligible influence when set to a value sufficiently close to one \citep{lipman2022flow}.\footnote{In our experiments, we follow \cite{wildberger2024flow} and set $\omega \defeq 1 - 10^{-4}$ for all experiments.} 

In order to make optimizing 
\begin{align}
    \mathbb{E}_{\vx \sim p(\vx)} \left [ d_{\textrm{CFM}} \left (Q^{\vz|\vx}_{\theta},  P^{\vz|\vx} \right )\right ]
\end{align}
tractable, and thus train our in-context learner, we make use of the sufficient condition in \Cref{th:suffcond}. Thus, the divergence $d_{\textrm{CFM}}$ admits the re-formulation as an objective $\Rthetatilde$ using samples from the joint distribution $P^{\vx, \vz}$. 
We can therefore optimize $\Rthetatilde$ using $N$ independent and identically distributed (i.i.d.) samples $t_i \sim \mathcal{U}([0,1])$ from the time-distribution,  $\vz^{(0)}_i \sim \PBase$ from the base distribution, and $(\vz^{(1)}_i, \vx_i) \sim P^{\vx, \vz}$ from the joint distribution. With this, we obtain the following objective function used for the training of the ICL models: 
\begin{equation}
\label{eq:emprisk}
    \hat{\mathcal{R}}_{\theta} = \sum_{i=1}^N \left \lvert \left \rvert v_{t_i,\vx_i}^{\theta}(\gamma_{t_i}(\vz_{i}^{(1)}|\vz_{i}^{(0)})) + {\vz_{i}^{(1)} - \omega \vz_{i}^{(0)}} \right \lvert \right \rvert_2 ^2
\end{equation}

\subsection{Sampling from the Joint Distribution}
\label{sec:sampling_from _the_joint}

In order to learn a model that can perform posterior inference according to \Cref{sec:posterior}, we require to sample $(\vx, \vz) \sim P^{\vx, \vz}$. Given $ p(\vx, \vz) = p(\vx| \vz) p(\vz)$, this is always possible as long as one can draw samples from $P^{\vz}$ and then from $P^{\vx| \vz}$. Hence, this is a relatively weak requirement allowing for a broad variety of priors and observation models. More specifically, for ICL, we generate a training dataset $\mathcal{D}$ which comprises i.i.d. samples $\left \{\left ( \vx_i, \vz_i \right) \right \}_{i=1}^N$ resulting from sampling $\vz_i \sim P^{\vz}$ and then $\vx_i \sim P^{\vx | \vz_i}$. We use this simple yet fundamental and very general template to generate samples from the joint $P^{\vx, \vz}$ for GLMs, factor analysis (FA), and Gaussian mixture models (GMMs) in our later applications.
Please refer to Appendix \ref{app:datageneration} for more details on the data generating processes.

\subsection{The Architecture}

\label{sec:architecture}

\begin{figure}[t]
\centering
\scalebox{0.6}{
\begin{tikzpicture}

\node (Encoder) [rectangle,rounded corners = 2pt, text centered, draw=black, fill=white, line width = 1.5pt, minimum width = 3cm, minimum height = 2cm, fill = laqua, xshift = -5cm, yshift = -0.5cm] {\Large Encoder};

\node (x) [rectangle,rounded corners = 2pt, text centered, draw=black, fill=white, line width = 1.5pt, minimum width = 1cm, minimum height = 1cm, fill = white, below of = Encoder, yshift = -1 cm] {\Large $\vx$};

\node (TimeMLP) [rectangle,rounded corners = 2pt, text centered, draw=black, fill=white, line width = 1.5pt, minimum width = 3cm, minimum height = 1cm, fill = lblue, xshift = 5cm, yshift = -1cm] {\Large MLP};

\node (t) [rectangle,rounded corners = 2pt, text centered, draw=black, fill=white, line width = 1.5pt, minimum width =1cm, minimum height = 1cm, fill = white, below of = TimeMLP, yshift = -0.5cm] {\Large $\vt$};

\node (Decoder) [rectangle,rounded corners = 2pt, text centered, draw=black, line width = 1.5pt, minimum width = 5cm, minimum height = 9.75cm, fill = lgrey, yshift = 3.4cm] {};

\node (z) [rectangle,rounded corners = 2pt, text centered, draw=black, fill=white, line width = 1.5pt, minimum width = 1cm, minimum height = 1cm, fill = white, left of = t, yshift = 0 cm, xshift = -4cm
] {\Large $(1 - {\omega}t)\vz^{(0)} + t \vz^{(1)} $};

\node (N1) [rectangle,rounded corners = 2pt, text centered, draw=black, fill=white, line width = 1.5pt, minimum width = 3.4cm, minimum height = 0.5cm, fill = lpurple, above of = z, yshift = 1cm] {\Large Norm};

\node (ScaleShift1) [rectangle,rounded corners = 2pt, text centered, draw=black, fill=white, line width = 1.5pt, minimum width = 3.4cm, minimum height = 0.5cm, fill = lorange, above of = N1, yshift = -0.25cm] {\Large Scale and Shift};

\node (Xatt) [rectangle,rounded corners = 2pt, text centered, draw=black, fill=white, line width = 1.5pt, minimum width =3.4cm, minimum height = 1cm, fill = lgreen, above of = ScaleShift1, yshift = 0 cm] {\Large Cross Attention};

\node (Scale1) [rectangle,rounded corners = 2pt, text centered, draw=black, fill=white, line width = 1.5pt, minimum width = 3.4cm, minimum height = 0.5cm, fill = lred, above of = Xatt, yshift = 0cm] {\Large Scale};

\node (Add1) [circle,rounded corners = 2pt, text centered, draw=black, fill=white, line width = 1.5pt, minimum width = 0.5cm, minimum height = 0.5cm, fill = white, above of = Scale1, yshift = -0.25cm] {$+$};

\node (N2) [rectangle,rounded corners = 2pt, text centered, draw=black, fill=white, line width = 1.5pt, minimum width = 3.4cm, minimum height = 0.5cm, fill = lpurple, above of = Add1, yshift = 0.25cm] {\Large Norm};

\node (ScaleShift2) [rectangle,rounded corners = 2pt, text centered, draw=black, fill=white, line width = 1.5pt, minimum width = 3.4cm, minimum height = 0.5cm, fill = lorange, above of = N2, yshift = -0.25cm] {\Large Scale and Shift};

\node (FF) [rectangle,rounded corners = 2pt, text centered, draw=black, fill=white, line width = 1.5pt, minimum width = 3.4cm, minimum height = 1cm, fill = lblue, above of = ScaleShift2, yshift = 0cm] {\Large Feed Forward};

\node (Scale2) [rectangle,rounded corners = 2pt, text centered, draw=black, fill=white, line width = 1.5pt, minimum width = 3.4cm, minimum height = 0.5cm, fill = lred, above of = FF, yshift = 0cm] {\Large Scale};

\node (Add2) [circle,rounded corners = 2pt, text centered, draw=black, fill=white, line width = 1.5pt, minimum width = 0.5cm, minimum height = 0.5cm, fill = white, above of = Scale2, yshift = -0.25cm] {$+$};

\node (Connection) [circle,rounded corners = 2pt, text centered, draw=none, fill=white, line width = 1.5pt, minimum width = 0.5cm, minimum height = 0.5cm, fill = white, left of = Xatt, xshift = -1.75cm] {};

\node (StartSkip1) [circle,rounded corners = 2pt, text centered, draw=none, line width = 1.5pt, minimum width = 0.5cm, minimum height = 0.5cm, fill = none, below of = N1, xshift =0cm, yshift = 0.4cm] {};

\node (ConnectionSkip1) [circle,rounded corners = 2pt, text centered, draw=none, line width = 1.5pt, minimum width = 0.5cm, minimum height = 0.5cm, fill = none, left of = Xatt, xshift = -1cm] {};

\node (StartSkip2) [circle,rounded corners = 2pt, text centered, draw=none, line width = 1.5pt, minimum width = 0.5cm, minimum height = 0.5cm, fill = none, below of = N2, xshift =0cm, yshift = 0.4cm] {};

\node (ConnectionSkip2) [circle,rounded corners = 2pt, text centered, draw=none, fill=none, line width = 1.5pt, minimum width = 0.5cm, minimum height = 0.5cm, left of = FF, xshift = -1cm] {};

%\node (ConnectionScaleShift1) [circle,rounded corners = 2pt, text centered, draw=black, fill=white, line width = 1.5pt, minimum width = 0.5cm, minimum height = 0.5cm, fill = white, right of = ScaleShift1, xshift = 1cm] {};

%\node (ConnectionScale1) [circle,rounded corners = 2pt, text centered, draw=black, fill=white, line width = 1.5pt, minimum width = 0.5cm, minimum height = 0.5cm, fill = white, right of = Scale1, xshift = 1cm] {};

%\node (ConnectionScaleShift2) [circle,rounded corners = 2pt, text centered, draw=black, fill=white, line width = 1.5pt, minimum width = 0.5cm, minimum height = 0.5cm, fill = white, right of = ScaleShift2, xshift = 1cm] {};

%\node (ConnectionScale2) [circle,rounded corners = 2pt, text centered, draw=black, fill=white, line width = 1.5pt, minimum width = 0.5cm, minimum height = 0.5cm, fill = white, right of = Scale2, xshift = 1cm] {};

\node (FinalMLP) [rectangle,rounded corners = 2pt, text centered, draw=black, fill=white, line width = 1.5pt, minimum width = 5cm, minimum height = 1cm, fill = lblue, above of = Decoder, yshift = 5cm] {\Large MLP with Conditioning};

\node (Extranode) [rectangle,rounded corners = 2pt, text centered, draw=black, line width = 1.5pt, minimum width = 4cm, minimum height = 1cm, fill = none, above of = FinalMLP, xshift =0cm, yshift = 0.75cm] {\Large $v_{t,\vx}^{\theta}((1 - {\omega}t)\vz^{(0)} + t \vz^{(1)} )$};

\node (Nlayers) [circle,rounded corners = 2pt, text centered, draw=none, line width = 1.5pt, minimum width = 0.5cm, minimum height = 0.5cm, fill = none, left of = Decoder, xshift = -2.5cm, yshift = 0cm] {\Large $N_{layers} \times$};

\tikzstyle{arrow} = [->,>=stealth, line width = 0.5mm]
\tikzstyle{arrow_back} = [<-,>=stealth, line width = 0.5mm]
\tikzstyle{arrow_double} = [<->,>=stealth, line width = 0.5mm]
\tikzstyle{line} = [-,>=stealth, line width = 0.5mm]

\draw [arrow] (x) -- (Encoder);
\draw [arrow] (t) -- (TimeMLP);
\draw [arrow] (z) -- (Decoder);

\draw [line] (N1) -- (ScaleShift1);
\draw [line] (ScaleShift1) -- (Xatt);
\draw [line] (Xatt) -- (Scale1);
\draw [line] (Scale1) -- (Add1);
\draw [line] (Add1) -- (N2);

\draw [line] (N2) -- (ScaleShift2);
\draw [line] (ScaleShift2) -- (FF);
\draw [line] (FF) -- (Scale2);
\draw [line] (Scale2) -- (Add2);

\draw [line] (Decoder.south) -- (N1);

\draw [line] (Encoder) |- (Connection.center);
\draw [arrow] (Connection.center) |- (Xatt);

\draw [line] (StartSkip1.center) -| (ConnectionSkip1.center);
\draw [arrow] (ConnectionSkip1.center) |- (Add1);

\draw [line] (StartSkip2.center) -| (ConnectionSkip2.center);
\draw [arrow] (ConnectionSkip2.center) |- (Add2);

\draw [line] (Add2) -- (Decoder.north);

\draw [line] (StartSkip2.center) -| (ConnectionSkip2.center);
\draw [arrow] (ConnectionSkip2.center) |- (Add2);

\draw [arrow] (TimeMLP) |- (ScaleShift1);
\draw [arrow] (TimeMLP) |- (Scale1);
\draw [arrow] (TimeMLP) |- (ScaleShift2);
\draw [arrow] (TimeMLP) |- (Scale2);
\draw [arrow] (TimeMLP) |- (FinalMLP);
\draw [arrow] (Decoder.north) -- (FinalMLP.south);
\draw [arrow] (FinalMLP.north) -- (Extranode.south);

\end{tikzpicture}
}

\caption{Architecture to perform ICL for full Bayesian inference.
A relatively large transformer encoder, similar to that in TabPFN \citep{hollmann2022tabpfn} processes a dataset $\vx$ and yields a representation used in the decoder. The decoder outputs a vector field defining a flow for a given input vector conditioned on the encoder output and the time. We condition on the time in each cross-attention and each feed-forward block. Please note that the size of the parts of the architecture does not correspond to the number of allocated parameters in this figure.} 
%The three core components of the model are a transformer encoder to process the context $\vx$, an MLP to obtain an embedding of the time $t$, and a decoder yielding the vector field %$\v_{t,x}^{\theta}$ conditioned on $\vx$ and $t$.
\label{fig:architecture}
% \vspace{-0.6cm}
\end{figure}
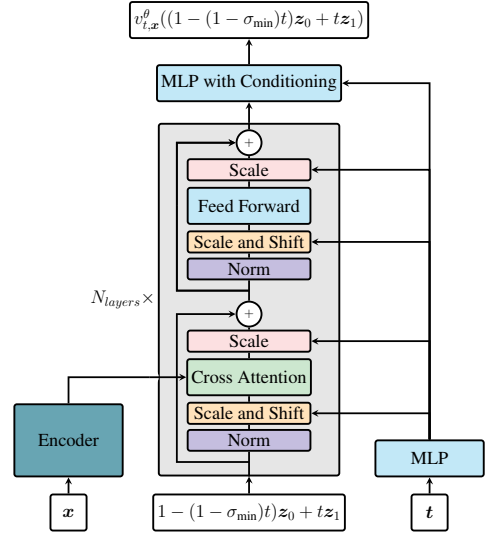

In order to implement the idea of learning full Bayesian inference in context, we extend ideas of diffusion transformers \citep{peebles2023scalable}, where the conditioning on the time $t$ is implemented via adaptive layer norm (adaLN) blocks initialized as the identity function. As we potentially require complex conditioning on the data $\vx$, an additional transformer encoder is added. The input to the decoder is a vector in the form $(1 - {\omega}t)\vz^{(0)} + t \vz^{(1)} $, which is treated as a sequence with length one and processed by a transformer decoder without self-attention, but the adaLN blocks. Therefore, the decoder has an equivalent interpretation as a multi-layer perceptron with skip-connections, cross-attention, and adaptive layer normalization. For the final processing in the decoder, only conditional feedforward layers with adaptive layer normalization are used. This corresponds exactly to the architecture of the decoder before, albeit without cross attention. We call this part an ``MLP with Conditioning". Samples for the time $t \in \left [0,1 \right ]$ are mapped onto a conditioning vector using several fully connected layers, which yields a richer representation of $t$ that is well-suited as an input to the adaLN blocks. Figure \ref{fig:architecture} depicts of the resulting architecture. 

\subsection{\textcolor{rev}{Implementing Flow Matching}}

\textcolor{rev}{
During the training phase, a tuple $(\vz^{(1)} , \vx^{})$ is drawn from the distribution $P^{\vz, \vx}$. Additionally, a time step $t \sim \mathcal{U}[0,1]$ and a sample $\vz^{(0)} $ is drawn from the base distribution $P_{\mathcal{B}}$, which is a standard Gaussian for all our applications. Subsequently, the ground-truth conditional flow $\psi(\vz^{(0)} | \vx) = (1 - {\omega}t)\vz^{(0)} + t \vz^{(1)} $ is computed, pushing forward $P_{\mathcal{B}}$ into $P^{\vz| \vx}$ up to time-point $t$. The transformer encoder processes $\vx$ and the decoder takes the representation of the encoder into account in order to output $v_{t,\vx}^{\theta}(\psi(\vz^{(0)} | \vx))$. This output should match the vector field that describes how the ground-truth flow $\psi(\vz^{(0)} | \vx)$ continues at time $t$. The discrepancy to the ground-truth vector field is measured with the MSE-loss in \Cref{eq:emprisk}.} %Note that all operations are of course batched during training.}

\textcolor{rev}{
In the sampling phase, we are given $\vx$ and the goal is to sample from $P^{\vz|\vx}$. To do so, first a vector $\vz^{(0)} \sim P_{\mathcal{B}}$ is drawn. The data $\vx$ is passed through the encoder. The decoder defines a function that maps a time-point $t$ and a vector $\bm\nu$ onto a vector field: $(t, \bm{\nu}) \mapsto v_{t,\vx}^{\theta}(\bm{\nu})$ taking $\vx$ into account. This function is given to an ODE-solver in order to forward-solve the corresponding ODE with boundary conditions $0 \leq t \leq 1$.}

\begin{table}[t]
\captionsetup{font=normal}
\centering
\small
\caption{Summarized results for GLMs. Average performance of VI methods and our ICL approach on 50 synthetic and 17 real-world datasets across 7 different GLM scenarios. Comparison to the analytical solution when available and HMC otherwise. Lower is better for all metrics. The best average result is marked in \textbf{bold}.}
% \vspace*{-5pt}
\label{Tab:GLM_summary}
\resizebox{0.47\textwidth}{!}{
\begin{tabular}{p{1.8cm}m{0.5cm}m{0.5cm}m{0.6cm}m{0.6cm}m{0.5cm}m{0.5cm}} % No vertica
\toprule
\multirow{2}{0pt}{\raisebox{-0.5ex}{\textbf{Model}}} & \multicolumn{3}{c}{\textbf{Synthetic}} & \multicolumn{3}{c}{\textbf{Real-World}} \\
\cmidrule(lr){2-4} \cmidrule(lr){5-7}
& C2ST & MMD & $\hspace{0.1cm} \mathcal{W}_2$ & C2ST & MMD & $\mathcal{W}_2$ \\
\midrule
LA & 1.000 & 2.770 & 2.049 & 1.000 & 2.091 & 0.849 \\
VI: Diagonal & 0.869 & 1.586 & 1.742 & 0.819 & 0.583 & 0.529 \\
VI: Full & 0.714 & 1.016 & 1.601 & 0.668 & 0.116 & 0.374 \\
VI: Structured & 0.711 & 0.929 & 1.580 & 0.664 & 0.109 & \textbf{0.370} \\
VI: IAF & 0.784 & 1.648 & 2.349 & 0.732 & 0.516 & 0.680 \\
ICL (ours) & \textbf{0.657} & \textbf{0.183} & \textbf{0.556} & \textbf{0.648} & \textbf{0.090} & 0.387 \\
\bottomrule
\end{tabular}
}
\label{tab:summarized_GLM}
\end{table}

\section{Experiments}

To show that the proposed methodology is not just an abstract concept, we derive exemplary use cases that demonstrate how well ICL is able to keep up with MCMC and VI approaches in practice. 

%\subsection{Experimental setup}

For this, we will use two prominent statistical modeling classes, namely generalized linear models (GLMs) and latent factor models. For the latent factor models, we consider factor analysis (FA) and Gaussian mixture models (GMMs).

\paragraph{Modeling Scenarios.} We use seven different scenarios for the GLMs, where we vary the prior distribution on the parameters, the conditional distribution of the response, and whether an intercept is included. For FA, we vary the form of the priors and dimensionalities of variables leading to four different scenarios. For the GMMs, we investigate different dimensionalities as well as prior configurations also in four different scenarios. We refer to  \Cref{app:datageneration} for details on the model structure and scenarios. 

For our experiments, we train a separate model from scratch for each GLM, GMM, and FA scenario using synthetic samples from the joint distribution $P^{\vx, \vx}$; i.e. we train seven separate models to cover the GLM scenarios, six separate models for the FA scenarios and four separate models for the GMM scenarios. 
Please refer to \Cref{sec:generating_real_data} for more details.

\paragraph{Datasets.} We evaluate the methods on 50 synthetic datasets and 17 real-world datasets from a benchmark suite for tabular regression problems proposed by \citet{grinsztajn2022tree}. We refer to \Cref{app:preprocessing} for more details on the preprocessing of the datasets.

\paragraph{Methods.}
Apart from a comparison with a gold standard, we compare our ICL approach to a Laplace approximation \citep[LA;][]{daxberger2021laplace} and different established VI methods based on automatic differentiation VI \citep{kucukelbir2017automatic}. For the variational distribution, we use a normal distribution with 1) a diagonal and 2) a full covariance matrix, as well as 3) a structured normal distribution with linear dependencies between the latent variables, and 4) an approach based on inverse autoregressive flows \citep[IAF;][]{kingma2016improved}. \Cref{app:hyper} includes a discussion regarding the hyperparameters of all considered methods. %More specifically, we evaluate the critical learning rate parameter for all VI methods. 

\paragraph{Evaluation Process.}
For every synthetic and real-world dataset, 1000 posterior samples from each method are compared against samples from the analytical solution, if available, or from a Hamiltonian Monte Carlo (HMC) sampler with a NUTS kernel \citep{hoffman2014no} as the gold standard. If posteriors are unimodal, we run a single chain. In the multimodal case, we use three times the number of modes as the number of Markov chains. 

\paragraph{Evaluation Metrics.}
Three metrics are employed to compare samples from different approximations of the posterior distribution. The first metric is a classifier 2-sample test \citep[C2ST;][]{lueckmann2021benchmarking, lopez2016revisiting}, where the ROC-AUC score of a random forest classifier, trained to distinguish between samples from the gold standard and the method in question, is utilized. For random forest, we use default hyperparameters, as defined in Scikit-learn \citep{pedregosa2011scikit} and 10-fold cross-validation. We use a random forest with the given hyperparameters as a highly performative classifier in order to detect small deviations in distributions, even though this incurs the risk that the C2ST quickly saturates at a value of one, especially in high-dimensional cases. 
The second metric is the maximum mean discrepancy (MMD) between the two distributions (gold-standard and each tested method) with an exponential kernel \citep{gretton2012kernel}. The third metric is the empirical Wasserstein-2 distance \citep[$\mathcal{W}_2$;][]{givens1984class} of the two distributions, as implemented in the POT library \citep{flamary2021pot}.

%\begin{figure}[t]
%    \centering
%    \includegraphics[width=1\linewidth]{Images/PosteriorMarginalsGammaPrior.png}
%    \caption{Density plots for first three the marginals of the posterior in a GLM with a gamma prior on the coefficients $\beta$, and an inverse gamma prior on the variance $\sigma^2$ of the responses. The data is part of the Miami housing 2016 dataset.}
 %   \label{fig:posteriormarginalgamma}
%\end{figure}

\label{sec:results}

\begin{table}[t] 
\captionsetup{font=normal}
\centering
\caption{Results for GLMs. Real-world Evaluation on 17 datasets: Linear regression with a gamma prior on the coefficients $\beta$, and an inverse gamma prior on the variance $\sigma^2$ of the responses (scenario 5). Comparison to HMC samples. All results within two standard errors of the best average result are marked in \textbf{bold}.}
% \vspace*{-5pt}
\label{tab:GLM_example}
\resizebox{0.47\textwidth}{!}{
\begin{tabular}{p{2.0cm}m{2.2cm}m{2.3cm}m{2.2cm}}
\toprule
\textbf{Model} & {\hspace{6.2mm} C2ST} & \hspace{6mm} MMD & \hspace{6.8mm} $\mathcal{W}_2$ \\
\midrule
LA & 1.000 ($\pm$ 0.000) & 1.982 $(\pm$ 0.126) & 0.623 ($\pm$ 0.084) \\
VI: Diagonal & 0.810 ($\pm$ 0.036) & 0.441 $(\pm$ 0.252) & 0.384 ($\pm$ 0.089) \\
VI: Full & 0.711 ($\pm$ 0.038) & 0.148 $(\pm$ 0.093) & \textbf{0.279} ($\pm$ 0.056) \\
VI: Structured & 0.705 ($\pm$ 0.032) & 0.140 $(\pm$ 0.081) & \textbf{0.269} ($\pm$ 0.045) \\
VI: IAF & 0.777 ($\pm$ 0.106) & 0.684 $(\pm$ 0.939) & 0.625 ($\pm$ 0.525) \\ICL (ours) & \textbf{0.610} ($\pm$ 0.045) & \textbf{0.046} $(\pm$ 0.020) & \textbf{0.242} ($\pm$ 0.038) \\
\bottomrule
\end{tabular}
}
\label{tab:GLM_scen5}
\vspace{-10pt}
\end{table}

\subsection{Generalized Linear Models}

Across seven different variants of GLMs, we find that ICL yields samples that have overall the highest agreement with the gold standard (see Table \ref{Tab:GLM_summary}). Specifically on the synthetic datasets, the C2ST, MMD and $\mathcal{W}_2$ metrics indicate that the posterior distribution can be approximated more accurately with ICL than via variational inference. 

Particularly in cases where the posterior has a shape deviating from a normal distribution, ICL and HMC agree more closely than VI. For instance, in the case where a gamma prior, i.e.\ a skewed distribution, is used on the coefficients of a regression model, we find that ICL substantially outperforms VI both on synthetic and real-world data (see Table \ref{tab:GLM_example}). %This phenomenon can also be observed in the tails of the marginals of the posterior distribution (\Cref{fig:posteriormarginalgamma}).
On the real-world data, ICL still matches the performance of VI methods and has the best (or not significantly worse than the best) performance in terms of C2ST in four out of seven cases (see Table \ref{tab:GLM_example}). Please refer to \Cref{sec:detailed_results} for the detailed experimental results summarized in Table \ref{tab:summarized_GLM}.

\begin{table}[t]
\captionsetup{font=normal}
\centering
\small
\caption{Summarized results for FA. Average performance of VI methods and our ICL approach on 50 synthetic and 17 real-world datasets across 6 different FA scenarios. Comparison to HMC samples. Lower is better for all metrics. The best average result is marked in \textbf{bold}.}
% \vspace*{-5pt}
\label{tab:FA_summary}
\resizebox{0.47\textwidth}{!}{
\begin{tabular}{p{1.8cm}m{0.5cm}m{0.5cm}m{0.6cm}m{0.6cm}m{0.5cm}m{0.5cm}} % No vertica

\toprule
\multirow{2}{*}{\raisebox{-0.5ex}{\textbf{Model}}} & \multicolumn{3}{c}{\textbf{Synthetic}} & \multicolumn{3}{c}{\textbf{Real-World}} \\
\cmidrule(lr){2-4} \cmidrule(lr){5-7}
& C2ST & MMD & $\mathcal{W}_2$ & C2ST & MMD & $\mathcal{W}_2$\\
\midrule
LA & 1.000 & 4.115 & 2.543 & 1.000 & 4.127 & 0.597 \\
VI: Diagonal & 0.999 & 3.321 & 1.998 & 0.960 & 1.220 & 0.288 \\
VI: Full & 0.993 & 3.222 & 1.955 & 0.950 & 1.173 & 0.281 \\
VI: Structured & 0.995 & 3.404 & 2.079 & 0.955 & 1.189 & 0.283 \\
VI: IAF & 0.987 & 3.226 & 1.973 & 0.902 & 0.969 & \textbf{0.251} \\
ICL (ours) & \textbf{0.568} & \textbf{0.057} & \textbf{0.409} & \textbf{0.751} & \textbf{0.673} & 0.583 \\
\bottomrule
\end{tabular}
}
\label{tab:FA_Summarized}
\vspace{-5pt}
\end{table}

\subsection{Factor analysis} 

On the factor analysis tasks, ICL has notably lower dissimilarity scores compared to the gold standard than all other considered methods in the synthetic evaluation (Table \ref{tab:FA_summary}). Notably, an average C2ST score of 0.568 is remarkably close to the theoretical lower bound of 0.5. Regarding the real world datasets, C2ST and MMD indicate that our ICL approach yields samples most similar to the reference, while the average $\mathcal{W}_2$ score is substantially higher. We hypothesize that this discrepancy in the metrics might be caused by numerical issues when computing the empirical $\mathcal{W}_2$ distance.
Furthermore, the relatively high number of latent variables in comparison to the limited number of data-points can yield overly flexible assumptions on the variational posterior causing the VI methods to overfit. %While the ICL approach is well suited for cases with little data, the small number of data points is likely the cause for the poor performance of the VI methods on the FA tasks.
See \Cref{sec:detailed_results} for the detailed experimental results summarized in Table \ref{tab:FA_Summarized}.

\subsection{Gaussian Mixture Models} 

Full Bayesian inference for GMMs is more challenging than for GLMs or FA. First, the generative process of GMMs involves discrete assignments to clusters, which poses a challenge not only for NUTS, but especially for VI methods. Second, the dimensionality of the posterior samples can be relatively large since for diagonal normal distributions, each component of the mixture has a mean and a variance parameter per dimension. Finally, the considered GMMs are not identifiable leading to multi-modal posterior distributions, which are impossible to perfectly approximate with commonly used VI methods based on Gaussian approximations. 

\begin{figure}[t]
\captionsetup{font=normal}
    \centering
    \includegraphics[width=1.0\linewidth]{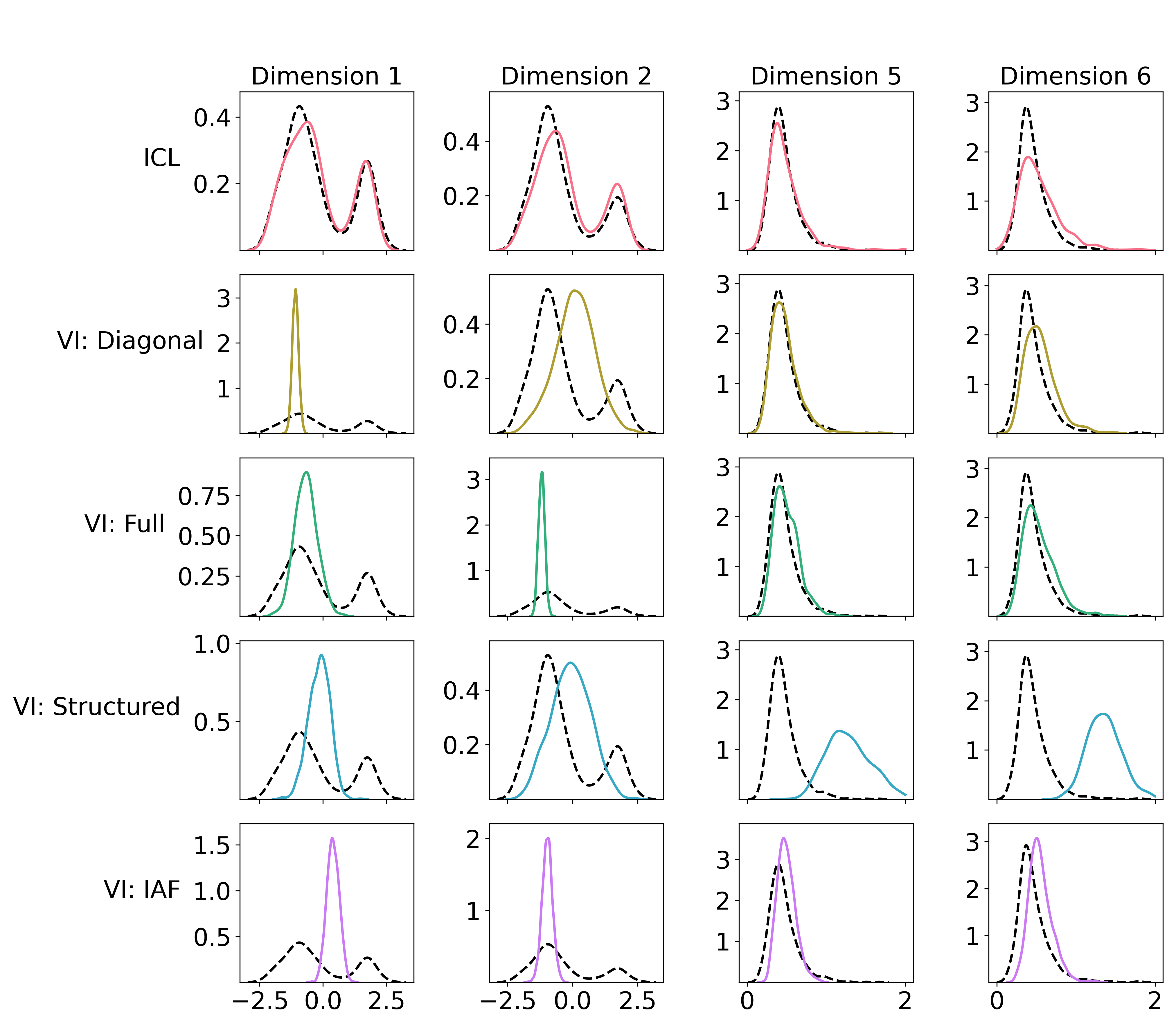}
    \caption{Density plots for the marginals of the posterior for GMM scenario 1. Comparison to HMC samples on a synthetic dataset whose density is depicted as a dotted line. Only the marginals of the first two components of the mean and the variance are shown. The density of the posterior obtained via HMC is depicted as a dotted line. While the ICL method aligns with the gold-standard HMC, the VI methods have a lower level of agreement and exhibit mode-seeking behaviour.}
    \label{fig:gmm_marginals}
\end{figure}

%\paragraph{Results}
Due to this inherent difficulty of the GMM scenarios, we find the overall performances of all models to be worse than in the GLM and FA cases. In particular, the C2ST metric is almost saturated for the VI approaches and has a value of around 83 percent for ICL (Table \ref{tab:gmm_summary}). The MMD and $\mathcal{W}_2$ metrics also indicate that ICL yields samples with higher agreement with the reference than the other approaches on synthetic data. A plot of the marginals of the posterior shows high agreement between the posterior distributions of both HMC and ICL while VI is incapable of perfectly approximating a bimodal distribution and exhibits typical mode-seeking behavior (Figure \ref{fig:gmm_marginals}). Note that also the VI approach based on inverse autoregressive flows, which in theory allows flexible modeling of a wide range of posterior shapes, fails to learn the bi-modality accurately from the limited number of 50 data points in this GMM scenario. This demonstrates the strength of our ICL approach in flexibly learning distributions agnostic of the provided sample size. Please refer to \Cref{sec:detailed_results} for the detailed experimental results summarized in Table \ref{tab:Summarzied_GMM}. 
%
%On the real-world evaluation, the differences are similar, albeit slightly less pronounced. While C2ST and MMD are better for ICL than for VI, the $\mathcal{W}_2$ metric is not substantially different. 

\begin{table}[t]
\captionsetup{font=normal}
\centering
\small
\caption{Summarized Results for GMMs. Average performance of VI methods and our ICL approach on 50 synthetic and 17 real-world datasets across 4 different GMM scenarios. Comparison to HMC samples in all cases. The best average result is marked in \textbf{bold}.}
% \vspace*{-5pt}
\label{tab:gmm_summary}
\resizebox{0.49\textwidth}{!}{
\begin{tabular}{p{1.8cm}m{0.5cm}m{0.5cm}m{0.6cm}m{0.6cm}m{0.5cm}m{0.7cm}} % No vertica

\toprule
\multirow{2}{*}{\raisebox{-0.5ex}{\textbf{Model}}} & \multicolumn{3}{c}{\textbf{Synthetic}} & \multicolumn{3}{c}{\textbf{Real-World}} \\
\cmidrule(lr){2-4} \cmidrule(lr){5-7}
& C2ST & MMD & $\hspace{0.1cm} \mathcal{W}_2$ & C2ST & MMD & $ \hspace{0.2cm} \mathcal{W}_2$\\
\midrule
LA & 1.000 & 3.916 & 8.324 & 1.000 & 3.385 & 12.740 \\
VI: Diagonal & 0.994 & 2.676 & 7.938 & 0.992 & 2.182 & 11.633 \\
VI: Full & 0.995 & 2.556 & 7.947 & 0.987 & 2.143 & 11.696 \\
VI: Structured & 0.994 & 2.595 & 7.929 & 0.988 & 2.129 & 11.521 \\
VI: IAF & 0.985 & 2.308 & 7.489 & 0.957 & 1.845 & 11.541 \\
ICL (ours) & \textbf{0.825} & \textbf{0.706} & \textbf{4.348} & \textbf{0.881} & \textbf{1.051} & \textbf{10.691} \\
\bottomrule
\end{tabular}
}
\label{tab:Summarzied_GMM}
\end{table}

\subsection{Ablations and Further Experimental Results}

In this subsection, we present various ablations concerning our ICL approach to full Bayesian inference. Due to limited space, most of the results are deferred to the appendix.

%\paragraph{Detailed results.} Please refer to \Cref{sec:detailed_results} for the detailed experimental results summarized in Table \ref{tab:summarized_GLM}, Table \ref{tab:FA_Summarized}, and Table \ref{tab:Summarzied_GMM}. 
%

\paragraph{Alternatives to Flow Matching.}

Appendix \ref{sec:diffusion_abl} contains results from an ablation study using diffusion objectives instead of flow matching, while Appendix \ref{sec:Gaussian_Approximation} investigates the use of a multivariate Gaussian to parametrize $Q_\theta^{\vz|\vx}$. The empirical results from these ablations strongly indicate that flow matching is essential for achieving a close approximation of the gold-standard posterior in the scenarios we consider. 

\paragraph{Dimensionality.}

In addition, we investigate the effect of the dimensionality $K$ of the latent variable $\vz \in \mathbb{R}^K$ for all seven different GLM scenarios. The key takeaway from our results is that for $K=20$ and $K=50$, the ICL approach performs comparably to the other methods in terms of sample similarity to HMC, but does not outperform them. Please refer to Appendix \ref{sec:ablation_dimensionality} for more details. We hypothesize that a key reason for the failure to detect meaningful differences between the methods in high dimensions is due to the curse of dimensionality affecting our metrics. 

\paragraph{Out-of-distribution Performance.}

We further investigate the robustness of our method under mild distribution shifts in Appendix \ref{sec:ood_abl}. Our results indicate that the performance of our ICL method remains relatively stable for small distribution shifts, but increasingly degrades for a larger gap between the training and testing distribution.

\paragraph{Predictive Performance.} Additionally, we evaluate our ICL method and all variational approaches with respect to the predictive performance of the considered GLM setups in Appendix \ref{app:eval_predictions}. Results confirm the strong performance of variational inference methods in terms of point prediction, especially for high dimensionalities, while the ICL method is generally competitive.

\paragraph{Architecture.} Appendix \ref{sec:abl_mlp} discusses results regarding the effect of using an MLP-based architecture. Our experimental findings confirm that the transformer-based encoder performs significantly better than an equally sized MLP encoder.

\paragraph{The Classifier in the C2ST Metric.}

Finally, we validate the choice of a random forest classifier for the C2ST metric (Appendix \ref{sec:c2st_eval}). We find that employing a nonlinear neural network and utilizing a random forest yields an overall analogous picture in terms of the performance of all methods.

\section{Discussion}

\label{sec:discussion}

This paper explores in-context learning for the purpose of full Bayesian inference in latent variable models. 
We propose to use conditional flow matching as a generic and flexible framework to approximate posterior distributions and an architecture that utilizes a transformer encoder for potentially complex conditioning on the data. %Our extensive experimental results show that this approach allows to accurately approximate posterior distributions for generalized linear models, in the case of factor analysis and for Gaussian mixture models. 
We find that our ICL approach yields a closer approximation of the posterior than several state-of-the-art variational inference methods across different datasets and model setups. This does not only hold for synthetic, but also real-world tabular datasets.
%emphasizing the flexibility of ICL and its applicability for full Bayesian inference. 

\paragraph{Limitations.}
While our experiments indicate the effectiveness of ICL as a Bayesian inference method, it requires an extensive up-front training routine on modern GPU hardware. %even on real-world datasets, this performance incurs a substantial computational costs for training the in-context learner in the first place. 
%Despite being faster at inference time than the considered HMC and VI methods, the overall computational burden to train our approach is much higher. 
Despite ICL being consistently faster at inference time than the considered HMC methods, the overall computational burden to train our approach is much higher. 

Furthermore, the goal of this work is to show that ICL can effectively learn full Bayesian inference. Our experiments therefore focus on relatively simple posterior distributions where we can compare against established methods, such as HMC. Additionally, increased dimensionality of the problems considered poses a challenge to both the ICL method and the metrics we employ. 
%for Bayesian inference can take more time than our ICL approach once the training is done,
%they do not require any of the extensive up-front training of the ICL approach. 
%they still require at most a few minutes on a single CPU core instead of several hours on a modern L4 or A100 graphics card. %
%
%Furthermore, we train one individual model for each scenario, i.e. for each latent factor model, from scratch. From a more practical perspective fine-tuning an in-context learner trained on a sufficiently similar task could substantially alleviate this issue.
%
%Finally, all considered scenarios only involve small or moderate numbers of data-points and features. While it can be a challenge to effectively model small data, for which ICL is arguably well-suited, 
Further, as with many other ICL approaches, large datasets as a context can become computationally very expensive. 

\paragraph{Outlook and Future Work.}

Despite its vast up-front computational cost, ICL has not only proven fundamentally transformative in the field of NLP \citep{brown2020language, touvron2023llama}, but has recently started to transform the field of tabular machine learning \citep{hollmann2022tabpfn}. Exploring the frontiers of ICL in terms of full Bayesian inference, starting from the feasibility results of this work, might therefore lead to similarly fertile territories. 

Although ICL performs well even when trained on data that may differ from real-world distributions, its flexibility is limited by the structure of the training data. If the synthetic data is highly unrealistic, ICL may fail--- much like any model with a misspecified hypothesis space that imposes an unsuitable inductive bias.

While flexible state-of-the-art sampling-based methods, such as HMC, serve as an efficient and highly effective reference in terms of inference for standard and statistical methods discussed in this paper, the proposed ICL approach is fundamentally more general in nature. In particular, any probabilistic model for which a generative process is conceivable can be fitted using our ICL approach---the potential for fitting models beyond the horizon of standard Bayesian methods is therefore manifold. 
%This includes the pursuit of probabilistic foundation models where broad applicability amortizes the pre-training cost. 
%More specifically, our ICL approach would, for instance, allow to perform inference for a GLM with priors on every aspect of the modeling such as the distribution of the response, the coefficients of the model or even basis expansions, making model choice naturally part of posterior inference.

\clearpage

\section*{Impact Statement}
This paper presents work whose goal is to advance the field of 
machine learning. There are many potential societal consequences 
of our work, none of which we feel must be specifically highlighted here.

\section*{Acknowledgements}
We thank Beste Aydemir and Svea Reuter for their valuable support and insightful comments. VF was supported by the Branco Weiss Fellowship. DR’s research is funded by the Deutsche Forschungsgemeinschaft (DFG, German Research Foundation) – 548823575. We also gratefully acknowledge funding provided to AR by the Munich Center for Machine Learning (MCML).


\bibliography{example_paper}

\begin{thebibliography}{102}
\providecommand{\natexlab}[1]{#1}
\providecommand{\url}[1]{\texttt{#1}}
\expandafter\ifx\csname urlstyle\endcsname\relax
  \providecommand{\doi}[1]{doi: #1}\else
  \providecommand{\doi}{doi: \begingroup \urlstyle{rm}\Url}\fi

\bibitem[Abdullah et~al.(2022)Abdullah, Hassan, and Mustafa]{abdullah2022review}
Abdullah, A.~A., Hassan, M.~M., and Mustafa, Y.~T.
\newblock A review on bayesian deep learning in healthcare: Applications and challenges.
\newblock \emph{IEEE Access}, 10:\penalty0 36538--36562, 2022.

\bibitem[Ahuja et~al.(2023)Ahuja, Panwar, and Goyal]{ahuja2023context}
Ahuja, K., Panwar, M., and Goyal, N.
\newblock In-context learning through the bayesian prism.
\newblock \emph{arXiv preprint arXiv:2306.04891}, 2023.

\bibitem[Anil et~al.(2023)Anil, Borgeaud, Alayrac, Yu, Soricut, Schalkwyk, Dai, Hauth, Millican, et~al.]{team2023gemini}
Anil, R., Borgeaud, S., Alayrac, J.-B., Yu, J., Soricut, R., Schalkwyk, J., Dai, A.~M., Hauth, A., Millican, K., et~al.
\newblock Gemini: a family of highly capable multimodal models.
\newblock \emph{arXiv preprint arXiv:2312.11805}, 2023.

\bibitem[Betancourt(2017)]{betancourt2017conceptual}
Betancourt, M.
\newblock A conceptual introduction to hamiltonian monte carlo.
\newblock \emph{arXiv preprint arXiv:1701.02434}, 2017.

\bibitem[Bingham et~al.(2019)Bingham, Chen, Jankowiak, Obermeyer, Pradhan, Karaletsos, Singh, Szerlip, Horsfall, and Goodman]{bingham2019pyro}
Bingham, E., Chen, J.~P., Jankowiak, M., Obermeyer, F., Pradhan, N., Karaletsos, T., Singh, R., Szerlip, P., Horsfall, P., and Goodman, N.~D.
\newblock Pyro: Deep universal probabilistic programming.
\newblock \emph{Journal of machine learning research}, 20\penalty0 (28):\penalty0 1--6, 2019.

\bibitem[Bishop et~al.(2002)Bishop, Spiegelhalter, and Winn]{bishop2002vibes}
Bishop, C., Spiegelhalter, D., and Winn, J.
\newblock Vibes: A variational inference engine for bayesian networks.
\newblock \emph{Advances in neural information processing systems}, 15, 2002.

\bibitem[Blei(2012)]{blei2012probabilistic}
Blei, D.~M.
\newblock Probabilistic topic models.
\newblock \emph{Communications of the ACM}, 55\penalty0 (4):\penalty0 77--84, 2012.

\bibitem[Blei et~al.(2017)Blei, Kucukelbir, and McAuliffe]{blei2017variational}
Blei, D.~M., Kucukelbir, A., and McAuliffe, J.~D.
\newblock Variational inference: A review for statisticians.
\newblock \emph{Journal of the American statistical Association}, 112\penalty0 (518):\penalty0 859--877, 2017.

\bibitem[Brehmer \& Cranmer(2022)Brehmer and Cranmer]{brehmer2022simulation}
Brehmer, J. and Cranmer, K.
\newblock Simulation-based inference methods for particle physics.
\newblock In \emph{Artificial Intelligence for High Energy Physics}, pp.\  579--611. World Scientific, 2022.

\bibitem[Brosse et~al.(2018)Brosse, Durmus, and Moulines]{brosse2018promises}
Brosse, N., Durmus, A., and Moulines, E.
\newblock The promises and pitfalls of stochastic gradient langevin dynamics.
\newblock \emph{Advances in Neural Information Processing Systems}, 31, 2018.

\bibitem[Brown et~al.(2020)Brown, Mann, Ryder, Subbiah, Kaplan, Dhariwal, Neelakantan, Shyam, Sastry, Askell, et~al.]{brown2020language}
Brown, T., Mann, B., Ryder, N., Subbiah, M., Kaplan, J.~D., Dhariwal, P., Neelakantan, A., Shyam, P., Sastry, G., Askell, A., et~al.
\newblock Language models are few-shot learners.
\newblock \emph{Advances in neural information processing systems}, 33:\penalty0 1877--1901, 2020.

\bibitem[Chan et~al.(2022)Chan, Dasgupta, Kim, Kumaran, Lampinen, and Hill]{chan2022transformers}
Chan, S.~C., Dasgupta, I., Kim, J., Kumaran, D., Lampinen, A.~K., and Hill, F.
\newblock Transformers generalize differently from information stored in context vs in weights.
\newblock \emph{arXiv preprint arXiv:2210.05675}, 2022.

\bibitem[Chen(2018)]{torchdiffeq}
Chen, R. T.~Q.
\newblock torchdiffeq, 2018.
\newblock URL \url{https://github.com/rtqichen/torchdiffeq}.

\bibitem[Chen et~al.(2014)Chen, Fox, and Guestrin]{chen2014stochastic}
Chen, T., Fox, E., and Guestrin, C.
\newblock Stochastic gradient hamiltonian monte carlo.
\newblock In \emph{International conference on machine learning}, pp.\  1683--1691. PMLR, 2014.

\bibitem[Cranmer et~al.(2020)Cranmer, Brehmer, and Louppe]{cranmer2020frontier}
Cranmer, K., Brehmer, J., and Louppe, G.
\newblock The frontier of simulation-based inference.
\newblock \emph{Proceedings of the National Academy of Sciences}, 117\penalty0 (48):\penalty0 30055--30062, 2020.

\bibitem[Cremer et~al.(2018)Cremer, Li, and Duvenaud]{cremer2018inference}
Cremer, C., Li, X., and Duvenaud, D.
\newblock Inference suboptimality in variational autoencoders.
\newblock In \emph{International conference on machine learning}, pp.\  1078--1086. PMLR, 2018.

\bibitem[Dao et~al.(2023)Dao, Phung, Nguyen, and Tran]{dao2023flow}
Dao, Q., Phung, H., Nguyen, B., and Tran, A.
\newblock Flow matching in latent space.
\newblock \emph{arXiv preprint arXiv:2307.08698}, 2023.

\bibitem[Dax et~al.(2021)Dax, Green, Gair, Macke, Buonanno, and Sch{\"o}lkopf]{dax2021real}
Dax, M., Green, S.~R., Gair, J., Macke, J.~H., Buonanno, A., and Sch{\"o}lkopf, B.
\newblock Real-time gravitational wave science with neural posterior estimation.
\newblock \emph{Physical review letters}, 127\penalty0 (24):\penalty0 241103, 2021.

\bibitem[Dax et~al.(2024)Dax, Green, Gair, Gupte, P{\"u}rrer, Raymond, Wildberger, Macke, Buonanno, and Sch{\"o}lkopf]{dax2024real}
Dax, M., Green, S.~R., Gair, J., Gupte, N., P{\"u}rrer, M., Raymond, V., Wildberger, J., Macke, J.~H., Buonanno, A., and Sch{\"o}lkopf, B.
\newblock Real-time gravitational-wave inference for binary neutron stars using machine learning.
\newblock \emph{arXiv preprint arXiv:2407.09602}, 2024.

\bibitem[Daxberger et~al.(2021)Daxberger, Kristiadi, Immer, Eschenhagen, Bauer, and Hennig]{daxberger2021laplace}
Daxberger, E., Kristiadi, A., Immer, A., Eschenhagen, R., Bauer, M., and Hennig, P.
\newblock Laplace redux-effortless bayesian deep learning.
\newblock \emph{Advances in Neural Information Processing Systems}, 34:\penalty0 20089--20103, 2021.

\bibitem[Dong et~al.(2022)Dong, Li, Dai, Zheng, Wu, Chang, Sun, Xu, and Sui]{dong2022survey}
Dong, Q., Li, L., Dai, D., Zheng, C., Wu, Z., Chang, B., Sun, X., Xu, J., and Sui, Z.
\newblock A survey on in-context learning.
\newblock \emph{arXiv preprint arXiv:2301.00234}, 2022.

\bibitem[Dormand \& Prince(1980)Dormand and Prince]{dormand1980family}
Dormand, J.~R. and Prince, P.~J.
\newblock A family of embedded runge-kutta formulae.
\newblock \emph{Journal of computational and applied mathematics}, 6\penalty0 (1):\penalty0 19--26, 1980.

\bibitem[Eloundou et~al.(2023)Eloundou, Manning, Mishkin, and Rock]{eloundou2023gpts}
Eloundou, T., Manning, S., Mishkin, P., and Rock, D.
\newblock Gpts are gpts: An early look at the labor market impact potential of large language models.
\newblock \emph{arXiv preprint arXiv:2303.10130}, 2023.

\bibitem[Esser et~al.(2024)Esser, Kulal, Blattmann, Entezari, M{\"u}ller, Saini, Levi, Lorenz, Sauer, Boesel, et~al.]{esser2024scaling}
Esser, P., Kulal, S., Blattmann, A., Entezari, R., M{\"u}ller, J., Saini, H., Levi, Y., Lorenz, D., Sauer, A., Boesel, F., et~al.
\newblock Scaling rectified flow transformers for high-resolution image synthesis.
\newblock In \emph{Forty-first International Conference on Machine Learning}, 2024.

\bibitem[Etzioni \& Kadane(1995)Etzioni and Kadane]{etzioni1995bayesian}
Etzioni, R.~D. and Kadane, J.~B.
\newblock Bayesian statistical methods in public health and medicine.
\newblock \emph{Annual review of public health}, 16\penalty0 (1):\penalty0 23--41, 1995.

\bibitem[Fahrmeir et~al.(2013)Fahrmeir, Kneib, Lang, Marx, Fahrmeir, Kneib, Lang, and Marx]{fahrmeir2013regression}
Fahrmeir, L., Kneib, T., Lang, S., Marx, B., Fahrmeir, L., Kneib, T., Lang, S., and Marx, B.
\newblock \emph{Regression models}.
\newblock Springer, 2013.

\bibitem[Fan \& Markram(2019)Fan and Markram]{fan2019brief}
Fan, X. and Markram, H.
\newblock A brief history of simulation neuroscience.
\newblock \emph{Frontiers in neuroinformatics}, 13:\penalty0 32, 2019.

\bibitem[Flamary et~al.(2021)Flamary, Courty, Gramfort, Alaya, Boisbunon, Chambon, Chapel, Corenflos, Fatras, Fournier, et~al.]{flamary2021pot}
Flamary, R., Courty, N., Gramfort, A., Alaya, M.~Z., Boisbunon, A., Chambon, S., Chapel, L., Corenflos, A., Fatras, K., Fournier, N., et~al.
\newblock Pot: Python optimal transport.
\newblock \emph{Journal of Machine Learning Research}, 22\penalty0 (78):\penalty0 1--8, 2021.

\bibitem[Garg et~al.(2022)Garg, Tsipras, Liang, and Valiant]{garg2022can}
Garg, S., Tsipras, D., Liang, P.~S., and Valiant, G.
\newblock What can transformers learn in-context? a case study of simple function classes.
\newblock \emph{Advances in Neural Information Processing Systems}, 35:\penalty0 30583--30598, 2022.

\bibitem[Garnelo et~al.(2018{\natexlab{a}})Garnelo, Rosenbaum, Maddison, Ramalho, Saxton, Shanahan, Teh, Rezende, and Eslami]{garnelo2018conditional}
Garnelo, M., Rosenbaum, D., Maddison, C., Ramalho, T., Saxton, D., Shanahan, M., Teh, Y.~W., Rezende, D., and Eslami, S.~A.
\newblock Conditional neural processes.
\newblock In \emph{International conference on machine learning}, pp.\  1704--1713. PMLR, 2018{\natexlab{a}}.

\bibitem[Garnelo et~al.(2018{\natexlab{b}})Garnelo, Schwarz, Rosenbaum, Viola, Rezende, Eslami, and Teh]{garnelo2018neural}
Garnelo, M., Schwarz, J., Rosenbaum, D., Viola, F., Rezende, D.~J., Eslami, S., and Teh, Y.~W.
\newblock Neural processes.
\newblock \emph{arXiv preprint arXiv:1807.01622}, 2018{\natexlab{b}}.

\bibitem[Gebhard et~al.(2025)Gebhard, Wildberger, Dax, Kofler, Angerhausen, Quanz, and Sch{\"o}lkopf]{gebhard2025flow}
Gebhard, T.~D., Wildberger, J., Dax, M., Kofler, A., Angerhausen, D., Quanz, S.~P., and Sch{\"o}lkopf, B.
\newblock Flow matching for atmospheric retrieval of exoplanets: Where reliability meets adaptive noise levels.
\newblock \emph{Astronomy \& Astrophysics}, 693:\penalty0 A42, 2025.

\bibitem[Givens \& Shortt(1984)Givens and Shortt]{givens1984class}
Givens, C.~R. and Shortt, R.~M.
\newblock A class of wasserstein metrics for probability distributions.
\newblock \emph{Michigan Mathematical Journal}, 31\penalty0 (2):\penalty0 231--240, 1984.

\bibitem[Gloeckler et~al.(2024)Gloeckler, Deistler, Weilbach, Wood, and Macke]{gloeckler2024all}
Gloeckler, M., Deistler, M., Weilbach, C., Wood, F., and Macke, J.~H.
\newblock All-in-one simulation-based inference.
\newblock \emph{arXiv preprint arXiv:2404.09636}, 2024.

\bibitem[Gretton et~al.(2012)Gretton, Borgwardt, Rasch, Sch{\"o}lkopf, and Smola]{gretton2012kernel}
Gretton, A., Borgwardt, K.~M., Rasch, M.~J., Sch{\"o}lkopf, B., and Smola, A.
\newblock A kernel two-sample test.
\newblock \emph{The Journal of Machine Learning Research}, 13\penalty0 (1):\penalty0 723--773, 2012.

\bibitem[Grinsztajn et~al.(2022)Grinsztajn, Oyallon, and Varoquaux]{grinsztajn2022tree}
Grinsztajn, L., Oyallon, E., and Varoquaux, G.
\newblock Why do tree-based models still outperform deep learning on typical tabular data?
\newblock \emph{Advances in neural information processing systems}, 35:\penalty0 507--520, 2022.

\bibitem[Gr{\"u}nwald \& van Ommen(2017)Gr{\"u}nwald and van Ommen]{grunwald2017inconsistency}
Gr{\"u}nwald, P. and van Ommen, T.
\newblock Inconsistency of bayesian inference for misspecified linear models, and a proposal for repairing it.
\newblock \emph{Bayesian Analysis}, 12\penalty0 (4):\penalty0 1069--1103, 2017.

\bibitem[Hastings(1970)]{hastings1970monte}
Hastings, W.
\newblock Monte carlo sampling methods using markov chains and their applications.
\newblock \emph{Biometrika}, 57\penalty0 (1):\penalty0 97--109, 1970.

\bibitem[Hoffman et~al.(2014)Hoffman, Gelman, et~al.]{hoffman2014no}
Hoffman, M.~D., Gelman, A., et~al.
\newblock The no-u-turn sampler: adaptively setting path lengths in hamiltonian monte carlo.
\newblock \emph{J. Mach. Learn. Res.}, 15\penalty0 (1):\penalty0 1593--1623, 2014.

\bibitem[Hollmann et~al.(2022)Hollmann, M{\"u}ller, Eggensperger, and Hutter]{hollmann2022tabpfn}
Hollmann, N., M{\"u}ller, S., Eggensperger, K., and Hutter, F.
\newblock Tabpfn: A transformer that solves small tabular classification problems in a second.
\newblock \emph{arXiv preprint arXiv:2207.01848}, 2022.

\bibitem[Hollmann et~al.(2025)Hollmann, M{\"u}ller, Purucker, Krishnakumar, K{\"o}rfer, Hoo, Schirrmeister, and Hutter]{hollmann2025accurate}
Hollmann, N., M{\"u}ller, S., Purucker, L., Krishnakumar, A., K{\"o}rfer, M., Hoo, S.~B., Schirrmeister, R.~T., and Hutter, F.
\newblock Accurate predictions on small data with a tabular foundation model.
\newblock \emph{Nature}, 637\penalty0 (8045):\penalty0 319--326, 2025.

\bibitem[Hoo et~al.(2024)Hoo, M{\"u}ller, Salinas, and Hutter]{hootabular}
Hoo, S.~B., M{\"u}ller, S., Salinas, D., and Hutter, F.
\newblock The tabular foundation model tabpfn outperforms specialized time series forecasting models based on simple features.
\newblock In \emph{NeurIPS Workshop on Time Series in the Age of Large Models}, 2024.

\bibitem[Hospedales et~al.(2021)Hospedales, Antoniou, Micaelli, and Storkey]{hospedales2021meta}
Hospedales, T., Antoniou, A., Micaelli, P., and Storkey, A.
\newblock Meta-learning in neural networks: A survey.
\newblock \emph{IEEE transactions on pattern analysis and machine intelligence}, 44\penalty0 (9):\penalty0 5149--5169, 2021.

\bibitem[Ioffe(2015)]{ioffe2015batch}
Ioffe, S.
\newblock Batch normalization: Accelerating deep network training by reducing internal covariate shift.
\newblock \emph{arXiv preprint arXiv:1502.03167}, 2015.

\bibitem[Izmailov et~al.(2021)Izmailov, Vikram, Hoffman, and Wilson]{izmailov2021bayesian}
Izmailov, P., Vikram, S., Hoffman, M.~D., and Wilson, A. G.~G.
\newblock What are bayesian neural network posteriors really like?
\newblock In \emph{International conference on machine learning}, pp.\  4629--4640. PMLR, 2021.

\bibitem[Kim et~al.(2018)Kim, Wiseman, Miller, Sontag, and Rush]{kim2018semi}
Kim, Y., Wiseman, S., Miller, A., Sontag, D., and Rush, A.
\newblock Semi-amortized variational autoencoders.
\newblock In \emph{International Conference on Machine Learning}, pp.\  2678--2687. PMLR, 2018.

\bibitem[Kingma(2013)]{kingma2013auto}
Kingma, D.~P.
\newblock Auto-encoding variational bayes.
\newblock \emph{arXiv preprint arXiv:1312.6114}, 2013.

\bibitem[Kingma(2014)]{kingma2014adam}
Kingma, D.~P.
\newblock Adam: A method for stochastic optimization.
\newblock \emph{arXiv preprint arXiv:1412.6980}, 2014.

\bibitem[Kingma et~al.(2016)Kingma, Salimans, Jozefowicz, Chen, Sutskever, and Welling]{kingma2016improved}
Kingma, D.~P., Salimans, T., Jozefowicz, R., Chen, X., Sutskever, I., and Welling, M.
\newblock Improved variational inference with inverse autoregressive flow.
\newblock \emph{Advances in neural information processing systems}, 29, 2016.

\bibitem[Kirsch et~al.(2022)Kirsch, Harrison, Sohl-Dickstein, and Metz]{kirsch2022general}
Kirsch, L., Harrison, J., Sohl-Dickstein, J., and Metz, L.
\newblock General-purpose in-context learning by meta-learning transformers.
\newblock \emph{arXiv preprint arXiv:2212.04458}, 2022.

\bibitem[Kucukelbir et~al.(2017)Kucukelbir, Tran, Ranganath, Gelman, and Blei]{kucukelbir2017automatic}
Kucukelbir, A., Tran, D., Ranganath, R., Gelman, A., and Blei, D.~M.
\newblock Automatic differentiation variational inference.
\newblock \emph{Journal of machine learning research}, 18\penalty0 (14):\penalty0 1--45, 2017.

\bibitem[Kyrimi et~al.(2021)Kyrimi, McLachlan, Dube, Neves, Fahmi, and Fenton]{kyrimi2021comprehensive}
Kyrimi, E., McLachlan, S., Dube, K., Neves, M.~R., Fahmi, A., and Fenton, N.
\newblock A comprehensive scoping review of bayesian networks in healthcare: Past, present and future.
\newblock \emph{Artificial Intelligence in Medicine}, 117:\penalty0 102108, 2021.

\bibitem[Lawley \& Maxwell(1962)Lawley and Maxwell]{lawley1962factor}
Lawley, D.~N. and Maxwell, A.~E.
\newblock Factor analysis as a statistical method.
\newblock \emph{Journal of the Royal Statistical Society. Series D (The Statistician)}, 12\penalty0 (3):\penalty0 209--229, 1962.

\bibitem[Le et~al.(2017)Le, Baydin, and Wood]{le2017inference}
Le, T.~A., Baydin, A.~G., and Wood, F.
\newblock Inference compilation and universal probabilistic programming.
\newblock In \emph{Artificial Intelligence and Statistics}, pp.\  1338--1348. PMLR, 2017.

\bibitem[Li et~al.(2016)Li, Chen, Carlson, and Carin]{li2016preconditioned}
Li, C., Chen, C., Carlson, D., and Carin, L.
\newblock Preconditioned stochastic gradient langevin dynamics for deep neural networks.
\newblock In \emph{Proceedings of the Thirtieth AAAI Conference on Artificial Intelligence}, pp.\  1788--1794, 2016.

\bibitem[Lienen et~al.(2025)Lienen, Kollovieh, and G{\"u}nnemann]{lienen2025generative}
Lienen, M., Kollovieh, M., and G{\"u}nnemann, S.
\newblock Generative modeling with bayesian sample inference.
\newblock \emph{arXiv preprint arXiv:2502.07580}, 2025.

\bibitem[Lin et~al.(2021)Lin, Wu, Zhou, Pan, Cao, and Wang]{lin2021task}
Lin, X., Wu, J., Zhou, C., Pan, S., Cao, Y., and Wang, B.
\newblock Task-adaptive neural process for user cold-start recommendation.
\newblock In \emph{Proceedings of the Web Conference 2021}, pp.\  1306--1316, 2021.

\bibitem[Lipman et~al.(2022)Lipman, Chen, Ben-Hamu, Nickel, and Le]{lipman2022flow}
Lipman, Y., Chen, R.~T., Ben-Hamu, H., Nickel, M., and Le, M.
\newblock Flow matching for generative modeling.
\newblock \emph{arXiv preprint arXiv:2210.02747}, 2022.

\bibitem[Lopes \& West(2004)Lopes and West]{lopes2004bayesian}
Lopes, H.~F. and West, M.
\newblock Bayesian model assessment in factor analysis.
\newblock \emph{Statistica Sinica}, pp.\  41--67, 2004.

\bibitem[Lopez-Paz \& Oquab(2016)Lopez-Paz and Oquab]{lopez2016revisiting}
Lopez-Paz, D. and Oquab, M.
\newblock Revisiting classifier two-sample tests.
\newblock \emph{arXiv preprint arXiv:1610.06545}, 2016.

\bibitem[Loshchilov \& Hutter(2016)Loshchilov and Hutter]{loshchilov2016sgdr}
Loshchilov, I. and Hutter, F.
\newblock Sgdr: Stochastic gradient descent with warm restarts.
\newblock \emph{arXiv preprint arXiv:1608.03983}, 2016.

\bibitem[Lueckmann et~al.(2017)Lueckmann, Goncalves, Bassetto, {\"O}cal, Nonnenmacher, and Macke]{lueckmann2017flexible}
Lueckmann, J.-M., Goncalves, P.~J., Bassetto, G., {\"O}cal, K., Nonnenmacher, M., and Macke, J.~H.
\newblock Flexible statistical inference for mechanistic models of neural dynamics.
\newblock \emph{Advances in neural information processing systems}, 30, 2017.

\bibitem[Lueckmann et~al.(2021)Lueckmann, Boelts, Greenberg, Goncalves, and Macke]{lueckmann2021benchmarking}
Lueckmann, J.-M., Boelts, J., Greenberg, D., Goncalves, P., and Macke, J.
\newblock Benchmarking simulation-based inference.
\newblock In \emph{International conference on artificial intelligence and statistics}, pp.\  343--351. PMLR, 2021.

\bibitem[Mangoubi \& Vishnoi(2019)Mangoubi and Vishnoi]{mangoubi2019nonconvex}
Mangoubi, O. and Vishnoi, N.~K.
\newblock Nonconvex sampling with the metropolis-adjusted langevin algorithm.
\newblock In \emph{Conference on learning theory}, pp.\  2259--2293. PMLR, 2019.

\bibitem[Margossian \& Blei(2023)Margossian and Blei]{margossian2023amortized}
Margossian, C.~C. and Blei, D.~M.
\newblock Amortized variational inference: When and why?
\newblock \emph{arXiv preprint arXiv:2307.11018}, 2023.

\bibitem[Mittal et~al.(2025{\natexlab{a}})Mittal, Bengio, Malkin, and Lajoie]{mittal2025context}
Mittal, S., Bengio, Y., Malkin, N., and Lajoie, G.
\newblock In-context parametric inference: Point or distribution estimators?
\newblock \emph{arXiv preprint arXiv:2502.11617}, 2025{\natexlab{a}}.

\bibitem[Mittal et~al.(2025{\natexlab{b}})Mittal, Bracher, Lajoie, Jaini, and Brubaker]{mittal2025amortized}
Mittal, S., Bracher, N.~L., Lajoie, G., Jaini, P., and Brubaker, M.
\newblock Amortized in-context bayesian posterior estimation.
\newblock \emph{arXiv preprint arXiv:2502.06601}, 2025{\natexlab{b}}.

\bibitem[M{\"u}ller et~al.(2021)M{\"u}ller, Hollmann, Arango, Grabocka, and Hutter]{muller2021transformers}
M{\"u}ller, S., Hollmann, N., Arango, S.~P., Grabocka, J., and Hutter, F.
\newblock Transformers can do bayesian-inference by meta-learning on prior-data.
\newblock In \emph{Fifth Workshop on Meta-Learning at the Conference on Neural Information Processing Systems}, 2021.

\bibitem[Murphy(2023)]{murphy2023probabilistic}
Murphy, K.~P.
\newblock \emph{Probabilistic machine learning: Advanced topics}.
\newblock MIT press, 2023.

\bibitem[Nelder \& Wedderburn(1972)Nelder and Wedderburn]{nelder1972generalized}
Nelder, J.~A. and Wedderburn, R.~W.
\newblock Generalized linear models.
\newblock \emph{Journal of the Royal Statistical Society Series A: Statistics in Society}, 135\penalty0 (3):\penalty0 370--384, 1972.

\bibitem[OpenAI(2023)]{achiam2023gpt}
OpenAI.
\newblock Gpt-4 technical report.
\newblock \emph{arXiv preprint arXiv:2303.08774}, 2023.

\bibitem[Papamakarios et~al.(2021{\natexlab{a}})Papamakarios, Nalisnick, Rezende, Mohamed, and Lakshminarayanan]{JMLR:v22:19-1028}
Papamakarios, G., Nalisnick, E., Rezende, D.~J., Mohamed, S., and Lakshminarayanan, B.
\newblock Normalizing flows for probabilistic modeling and inference.
\newblock \emph{Journal of Machine Learning Research}, 22\penalty0 (57):\penalty0 1--64, 2021{\natexlab{a}}.

\bibitem[Papamakarios et~al.(2021{\natexlab{b}})Papamakarios, Nalisnick, Rezende, Mohamed, and Lakshminarayanan]{papamakarios2021normalizing}
Papamakarios, G., Nalisnick, E., Rezende, D.~J., Mohamed, S., and Lakshminarayanan, B.
\newblock Normalizing flows for probabilistic modeling and inference.
\newblock \emph{Journal of Machine Learning Research}, 22\penalty0 (57):\penalty0 1--64, 2021{\natexlab{b}}.

\bibitem[Pedregosa et~al.(2011)Pedregosa, Varoquaux, Gramfort, Michel, Thirion, Grisel, Blondel, Prettenhofer, Weiss, Dubourg, et~al.]{pedregosa2011scikit}
Pedregosa, F., Varoquaux, G., Gramfort, A., Michel, V., Thirion, B., Grisel, O., Blondel, M., Prettenhofer, P., Weiss, R., Dubourg, V., et~al.
\newblock Scikit-learn: Machine learning in python.
\newblock \emph{the Journal of machine Learning research}, 12:\penalty0 2825--2830, 2011.

\bibitem[Peebles \& Xie(2023)Peebles and Xie]{peebles2023scalable}
Peebles, W. and Xie, S.
\newblock Scalable diffusion models with transformers.
\newblock In \emph{Proceedings of the IEEE/CVF International Conference on Computer Vision}, pp.\  4195--4205, 2023.

\bibitem[Phan et~al.(2019)Phan, Pradhan, and Jankowiak]{phan2019composable}
Phan, D., Pradhan, N., and Jankowiak, M.
\newblock Composable effects for flexible and accelerated probabilistic programming in numpyro.
\newblock \emph{arXiv preprint arXiv:1912.11554}, 2019.

\bibitem[Rezende et~al.(2014)Rezende, Mohamed, and Wierstra]{rezende2014stochastic}
Rezende, D.~J., Mohamed, S., and Wierstra, D.
\newblock Stochastic backpropagation and approximate inference in deep generative models.
\newblock In \emph{International conference on machine learning}, pp.\  1278--1286. PMLR, 2014.

\bibitem[Robertson et~al.(2024)Robertson, Hollmann, Awad, and Hutter]{robertson2024fairpfn}
Robertson, J., Hollmann, N., Awad, N., and Hutter, F.
\newblock Fairpfn: Transformers can do counterfactual fairness.
\newblock In \emph{ICML 2024 Next Generation of AI Safety Workshop}, 2024.

\bibitem[Rudner et~al.(2018)Rudner, Fortuin, Teh, and Gal]{rudner2018connection}
Rudner, T.~G., Fortuin, V., Teh, Y.~W., and Gal, Y.
\newblock On the connection between neural processes and gaussian processes with deep kernels.
\newblock In \emph{Workshop on Bayesian Deep Learning, NeurIPS}, pp.\ ~14, 2018.

\bibitem[Rummel(1988)]{rummel1988applied}
Rummel, R.~J.
\newblock \emph{Applied factor analysis}.
\newblock Northwestern University Press, 1988.

\bibitem[Sahoo et~al.(2024)Sahoo, Gokaslan, De~Sa, and Kuleshov]{sahoo2024diffusion}
Sahoo, S., Gokaslan, A., De~Sa, C.~M., and Kuleshov, V.
\newblock Diffusion models with learned adaptive noise.
\newblock \emph{Advances in Neural Information Processing Systems}, 37:\penalty0 105730--105779, 2024.

\bibitem[Salazar(2023)]{salazar2023vart}
Salazar, S.
\newblock Vart: variational regression trees.
\newblock \emph{Advances in Neural Information Processing Systems}, 36:\penalty0 45681--45693, 2023.

\bibitem[Schmit \& Pritchard(2018)Schmit and Pritchard]{schmit2018emulation}
Schmit, C.~J. and Pritchard, J.~R.
\newblock Emulation of reionization simulations for bayesian inference of astrophysics parameters using neural networks.
\newblock \emph{Monthly Notices of the Royal Astronomical Society}, 475\penalty0 (1):\penalty0 1213--1223, 2018.

\bibitem[Sohn \& Narain(2021)Sohn and Narain]{sohn2021neural}
Sohn, H. and Narain, D.
\newblock Neural implementations of bayesian inference.
\newblock \emph{Current Opinion in Neurobiology}, 70:\penalty0 121--129, 2021.

\bibitem[Sommer et~al.(2024)Sommer, Wimmer, Papamarkou, Bothmann, Bischl, and R{\"u}gamer]{sommer2024connecting}
Sommer, E., Wimmer, L., Papamarkou, T., Bothmann, L., Bischl, B., and R{\"u}gamer, D.
\newblock Connecting the dots: Is mode-connectedness the key to feasible sample-based inference in bayesian neural networks?
\newblock In \emph{Proceedings of the 41st International Conference on Machine Learning}. PMLR, 2024.

\bibitem[Sommer et~al.(2025)Sommer, Robnik, Nozadze, Seljak, and R\"ugamer]{sommer2025mile}
Sommer, E., Robnik, J., Nozadze, G., Seljak, U., and R\"ugamer, D.
\newblock {Microcanonical Langevin Ensembles: Advancing the Sampling of Bayesian Neural Networks}.
\newblock In \emph{The Thirteenth International Conference on Learning Representations}, 2025.

\bibitem[Song \& Ermon(2019)Song and Ermon]{song2019generative}
Song, Y. and Ermon, S.
\newblock Generative modeling by estimating gradients of the data distribution.
\newblock \emph{Advances in neural information processing systems}, 32, 2019.

\bibitem[Song et~al.(2020)Song, Sohl-Dickstein, Kingma, Kumar, Ermon, and Poole]{song2020score}
Song, Y., Sohl-Dickstein, J., Kingma, D.~P., Kumar, A., Ermon, S., and Poole, B.
\newblock Score-based generative modeling through stochastic differential equations.
\newblock \emph{arXiv preprint arXiv:2011.13456}, 2020.

\bibitem[Srivastava \& Sutton(2017)Srivastava and Sutton]{srivastava2017autoencoding}
Srivastava, A. and Sutton, C.
\newblock Autoencoding variational inference for topic models.
\newblock \emph{arXiv preprint arXiv:1703.01488}, 2017.

\bibitem[Touvron et~al.(2023)Touvron, Martin, Stone, Albert, Almahairi, Babaei, Bashlykov, Batra, Bhargava, Bhosale, et~al.]{touvron2023llama}
Touvron, H., Martin, L., Stone, K., Albert, P., Almahairi, A., Babaei, Y., Bashlykov, N., Batra, S., Bhargava, P., Bhosale, S., et~al.
\newblock Llama 2: Open foundation and fine-tuned chat models.
\newblock \emph{arXiv preprint arXiv:2307.09288}, 2023.

\bibitem[Vetter et~al.(2025)Vetter, Gloeckler, Gedon, and Macke]{vetter2025effortless}
Vetter, J., Gloeckler, M., Gedon, D., and Macke, J.~H.
\newblock Effortless, simulation-efficient bayesian inference using tabular foundation models.
\newblock \emph{arXiv preprint arXiv:2504.17660}, 2025.

\bibitem[Walker(2013)]{walker2013bayesian}
Walker, S.~G.
\newblock Bayesian inference with misspecified models.
\newblock \emph{Journal of statistical planning and inference}, 143\penalty0 (10):\penalty0 1621--1633, 2013.

\bibitem[Wang et~al.(2024)Wang, Zhu, Saxon, Steyvers, and Wang]{wang2024large}
Wang, X., Zhu, W., Saxon, M., Steyvers, M., and Wang, W.~Y.
\newblock Large language models are latent variable models: Explaining and finding good demonstrations for in-context learning.
\newblock \emph{Advances in Neural Information Processing Systems}, 36, 2024.

\bibitem[Wang \& Blei(2019)Wang and Blei]{wang2019variational}
Wang, Y. and Blei, D.
\newblock Variational bayes under model misspecification.
\newblock \emph{Advances in Neural Information Processing Systems}, 32, 2019.

\bibitem[Welling \& Teh(2011)Welling and Teh]{welling2011bayesian}
Welling, M. and Teh, Y.~W.
\newblock Bayesian learning via stochastic gradient langevin dynamics.
\newblock In \emph{Proceedings of the 28th international conference on machine learning (ICML-11)}, pp.\  681--688. Citeseer, 2011.

\bibitem[Wildberger et~al.(2024)Wildberger, Dax, Buchholz, Green, Macke, and Sch{\"o}lkopf]{wildberger2024flow}
Wildberger, J., Dax, M., Buchholz, S., Green, S., Macke, J.~H., and Sch{\"o}lkopf, B.
\newblock Flow matching for scalable simulation-based inference.
\newblock \emph{Advances in Neural Information Processing Systems}, 36, 2024.

\bibitem[Yeo \& Johnson(2000)Yeo and Johnson]{yeo2000new}
Yeo, I.-K. and Johnson, R.~A.
\newblock A new family of power transformations to improve normality or symmetry.
\newblock \emph{Biometrika}, 87\penalty0 (4):\penalty0 954--959, 2000.

\bibitem[Yim et~al.(2023)Yim, Campbell, Foong, Gastegger, Jim{\'e}nez-Luna, Lewis, Satorras, Veeling, Barzilay, Jaakkola, et~al.]{yim2023fast}
Yim, J., Campbell, A., Foong, A.~Y., Gastegger, M., Jim{\'e}nez-Luna, J., Lewis, S., Satorras, V.~G., Veeling, B.~S., Barzilay, R., Jaakkola, T., et~al.
\newblock Fast protein backbone generation with se (3) flow matching.
\newblock \emph{arXiv preprint arXiv:2310.05297}, 2023.

\bibitem[Yim et~al.(2024)Yim, Campbell, Mathieu, Foong, Gastegger, Jim{\'e}nez-Luna, Lewis, Satorras, Veeling, No{\'e}, et~al.]{yim2024improved}
Yim, J., Campbell, A., Mathieu, E., Foong, A.~Y., Gastegger, M., Jim{\'e}nez-Luna, J., Lewis, S., Satorras, V.~G., Veeling, B.~S., No{\'e}, F., et~al.
\newblock Improved motif-scaffolding with se (3) flow matching.
\newblock \emph{arXiv preprint arXiv:2401.04082}, 2024.

\bibitem[Zhai et~al.(2018)Zhai, Zhang, Chen, and He]{zhai2018autoencoder}
Zhai, J., Zhang, S., Chen, J., and He, Q.
\newblock Autoencoder and its various variants.
\newblock In \emph{2018 IEEE international conference on systems, man, and cybernetics (SMC)}, pp.\  415--419. IEEE, 2018.

\bibitem[Zhao et~al.(2024)Zhao, Shi, Yu, Zhou, and Lu]{zhao2024flowturbo}
Zhao, W., Shi, M., Yu, X., Zhou, J., and Lu, J.
\newblock Flowturbo: Towards real-time flow-based image generation with velocity refiner.
\newblock \emph{arXiv preprint arXiv:2409.18128}, 2024.

\bibitem[Zheng et~al.(2023)Zheng, Lu, Chen, and Zhu]{zheng2023improved}
Zheng, K., Lu, C., Chen, J., and Zhu, J.
\newblock Improved techniques for maximum likelihood estimation for diffusion odes.
\newblock In \emph{International Conference on Machine Learning}, pp.\  42363--42389. PMLR, 2023.

\end{thebibliography}
\bibliographystyle{icml2025}

%%%%%%%%%%%%%%%%%%%%%%%%%%%%%%%%%%%%%%%%%%%%%%%%%%%%%%%%%%%%%%%%%%%%%%%%%%%%%%%
%%%%%%%%%%%%%%%%%%%%%%%%%%%%%%%%%%%%%%%%%%%%%%%%%%%%%%%%%%%%%%%%%%%%%%%%%%%%%%%
% APPENDIX
%%%%%%%%%%%%%%%%%%%%%%%%%%%%%%%%%%%%%%%%%%%%%%%%%%%%%%%%%%%%%%%%%%%%%%%%%%%%%%%
%%%%%%%%%%%%%%%%%%%%%%%%%%%%%%%%%%%%%%%%%%%%%%%%%%%%%%%%%%%%%%%%%%%%%%%%%%%%%%%
\newpage
\appendix
\onecolumn

\newpage 
\section*{Appendix}

\section{Data-generating Processses}
\label{app:datageneration}

This section contains more details on the data generating processes of the latent variable models we fit via ICL.

\subsection{Generalized Linear Models}

\label{app:datageneration_glm}

In this section we expand the description and explanation regarding GLMs from \Cref{sec:sampling_from _the_joint}. GLMs are among the most commonly used statistical models with myriads of applications \citep{nelder1972generalized, fahrmeir2013regression}. In the context of GLMs, we assume that the response $y$ follows a distribution $P^{y|\vu}$ depending on the linear predictor $\eta \defeq \vu^\top \bm{\beta}$ and an additional parameter $\sigma^2$. We denote the covariates as $\vu$, the regression coefficients as $\bm{\beta}$, and use $\sigma^2$ for the variance of the response. The mean of $P^{y|\vu}$ depends on the linear predictor via a link function $g$, such that $g\left (\mathbb{E}[y|\vu] \right) = \vu^\top \bm{\beta}$. Ultimately, the  density of distribution of the response $y$ depending on the linear predictor and the additional parameter is denoted by $p(y|g\left (\bm{u}^\top  \bm{\beta}\right ), \sigma^2)$. To showcase the flexibility of our framework, we experiment with different priors $P^{\beta}$ on the regression coefficients, $P^{\sigma^2}$ on the parameter $\sigma^2$, and also different parametric distributions of the response. Additionally, to include covariates $\vu$ that resemble practically relevant tabular data in the generative process, allowing for meaningful inference on real-world datasets, we utilize samples from the Tab-PFN ``prior'' for $P^{\vu}$.

GLMs belong to the framework of latent variable models defined by data $\vx$ and (latent) variables $\vz$, where the data comprises covariates and response $\vx \defeq (\vu, y)$. The variables of interest are the coefficients $\vz \defeq \bm{\beta}$. This yields the following generative process for a set of synthetic samples $\mathcal{D} \defeq \left \{ (\vx_i, \vz_i)\right \}_{i=1}^N$ from $P^{\vx, \vz}$:

We consider seven different GLM scenarios by varying the structure of the prior distributions and the conditional distribution of the response (\Cref{tab:glm_variables}). In particular, we consider a normal $\mathcal{N}(0,1)$ prior, a $\text{Laplace}(0, 1)$ and a gamma $\text{Ga}(1, 1)$ prior that factorizes over the coefficients $\beta_j$ contained in $\bm{\beta} = (\beta_1,\ldots,\beta_p)$. In two cases we include an intercept in the model using a normal prior $\mathcal{N}(0,9)$ with a relatively large variance. We consider regression cases with a normally distributed response $\mathcal{N}(\bm{u}^\top \bm{\beta}, \sigma^2)$, a Bernoulli distributed response $\text{Bin}(1, \operatorname{sigmoid}(\bm{u}^\top \bm{\beta}))$, i.e. logistic regression, and a response following a gamma distribution $\text{Ga}({\sigma^{-2}}\exp{(\bm{u}^\top \bm{\beta})}, {\sigma^{-2}}\exp(2 \bm{u}^\top \bm{\beta}))$. In the last case, we set $\exp(\bm{u}^\top \bm{\beta})$ to be the mean and $\sigma^2$ to be the conditional variance of the response. An inverse gamma prior $\text{IG}(5,2)$ is used on the variance $\sigma^2$ for each scenario except the logistic regression. We fix the number of covariates and thus also the dimensionality of $\bm{\beta}$ at $p = 5$ and set the number of data points per dataset to $K = 50$.

\begin{algorithm}[h]
\caption{Data Generation Algorithm}
\begin{algorithmic}[1]
    \REQUIRE Number of datasets $N$, number of samples per dataset $K$, distributions $P^{\bm{\beta}}, P^{\bm{\sigma^2}}, P^{\bm{u}}$, 
    \ENSURE A dataset $\mathcal{D}$ of input-output pairs $\left(\vx_i, \vz_i\right)$ for $i = 1, \dots, N$.
    \STATE Initialize $\mathcal{D} \leftarrow \varnothing$
    \FOR{$i = 1 \to N$}
        \STATE Draw $\bm{\beta}_i \sim P^{\bm{\beta}}$
        \STATE Draw $\sigma^2_i \sim P^{\bm{\sigma^2}}$
        \FOR{$j = 1 \to K$}
            \STATE Draw $\bm{u}_{i,j} \sim P^{\bm{u}}$
            \STATE Draw $y_{i,j} \sim p\!\Bigl(y \;\big|\; g^{-1}\bigl(\bm{u}_{i,j}^\top \bm{\beta}_i\bigr),\, \sigma^2_i\Bigr)$
        \ENDFOR
        \STATE Set $\vx_i \coloneqq \bigl(\,(\bm{u}_{i,j},\, y_{i,j})\bigr)_{j=1}^K$
        \STATE Set $\vz_i \coloneqq \bm{\beta}_i$
        \STATE Update $\mathcal{D} \leftarrow \mathcal{D} \cup \{\bigl(\vx_i, \vz_i\bigr)\}$
    \ENDFOR
\end{algorithmic}

\caption{Generation of synthetic data for GLMs}
\label{Algo:glmAppendix}

\end{algorithm}

\begin{table}[t]
    \centering
    \caption{Distribution of variables for the considered GLM scenarios.}
    \begin{tabular}{l|c|c|c|c}  % Removed the extra '|' after the last column
        \hline
        \textbf{Scenario} & $\beta_{i,j}$ & $\beta_{i,0}$ & $\sigma^2_i$ & $y_{i,j}|(\bm{u}_{i,j}, \bm{\beta}_i, \beta_{0,i}, \sigma^2_i)$ \\  % Added header row
        \hline
        Scenario 1 & $\mathcal{N}(0, 1)$ & - &  $\text{IG}(5,2)$ & $\mathcal{N}(\bm{u}_{i,j}^\top \bm{\beta}_i, \sigma^2_i)$ \\    
        Scenario 2 & $\mathcal{N}(0, 1)$ & $\mathcal{N}(0, 9)$&  $\text{IG}(5,2)$ & $\mathcal{N}(\bm{u}_{i,j}^\top \bm{\beta}_i, \sigma^2_i)$ \\ 
        Scenario 3 & $\text{Laplace}(0, 1)$ & - &  $\text{IG}(5,2)$ & $\mathcal{N}(\bm{u}_{i,j}^\top \bm{\beta}_i, \sigma^2_i)$ \\ 
        Scenario 4 & $\text{Laplace}(0, 1)$ & $\mathcal{N}(0, 9)$ &  $\text{IG}(5,2)$ & $\mathcal{N}(\bm{u}_{i,j}^\top \bm{\beta}_i, \sigma^2_i)$ \\  
        Scenario 5 & $\text{Ga}(1, 1)$ & - & $\text{IG}(5,2)$& $\mathcal{N}(\bm{u}_{i,j}^\top \bm{\beta}_i, \sigma^2_i)$\\   
        Scenario 6 & $\mathcal{N}(0, 1)$ & - &  - & $\text{Bin}(1, \operatorname{sigmoid}(\bm{u}_{i,j}^\top \bm{\beta}_i))$ \\ 
        Scenario 7 & $\mathcal{N}(0, 1)$ & - &  $\text{IG}(5,2)$ & $\text{Ga}(\sigma^{-2}_i\exp{(\bm{u}_{i,j}^\top \bm{\beta}_i)}, \sigma^{-2}_i\exp(2 \bm{u}_{i,j}^\top \bm{\beta}_i))$ \\ 
        
        \hline
    \end{tabular}
    \label{tab:glm_variables}
\end{table}

\newpage 
\subsection{Factor Analysis}

\label{app:datageneration_fa}

The goal of factor analysis is to explain data $\vx$ in terms of latent, typically lower-dimensional, factors $\vz$ \citep{lawley1962factor, rummel1988applied}. In the Bayesian setting, one assumes a prior $P^{\vz}$ on the latent variable $\vz$, a prior $P^{\bm{W}}$ on the factor loading matrix $\bm{W}$ and additional priors $P^{\bm{\Psi}}$ and $P^{\bm{\mu}}$ on the covariance matrix and the mean vector. The conditional distribution $P^{\vz|\vx}$ of the data given $\vz$ has mean $ \mathbb{E}[\vz|\vx] = \bm{W} \vz + \bm{\mu}$ and covariance matrix $\text{Cov}[\vz|\vx] = \bm{\Psi}$. In the case where $P^{\vz}$ and $P^{\vz|\vx}$ are Gaussian, one can set $P^{\vz} = \mathcal{N}(\bm{0}, \Id)$ and assume a diagonal covariance matrix $\bm{\Psi}$ without loosing expressiveness of the model \citep{murphy2023probabilistic}. We make the assumption that $\bm{W}$ is lower triangular with positive entries on the diagonal in order to ensure identifiability of the model \citep{lopes2004bayesian}. Additionally, we assume that the distributions $\bm{\mu}$, $\bm{\Psi}$ and $P^{\bm{W}}$ fully factorize. In order to ensure that the diagonal of $\bm{W}$ is positive, we consider absolute values in the generative process. 
\Cref{Algo:fa} details the data generating process.

Table \ref{tab:fa_variables} summarizes the different configurations for FA. We assume a Gaussian prior on the mean components, and an inverse gamma prior on the elements of the diagonal covariance matrix $\bm{\Psi}$. For the factor loading matrix $\bm{W}$, independent normal and Laplace priors are investigated. Furthermore, we use a normal prior on the latent factors $\vz_i$ in five cases and a Laplace prior in one case. We vary the number of samples $K$ per dataset $\vx$, the dimensionality $P$ of each data point, as well as the dimensionality $\bm{z}_{dim}$. 

\begin{table}[h]
    \centering
    \caption{Distribution and dimensionalitites of variables for the considered FA scenarios.}
    \begin{tabular}{l|c|c|c|c|c|c|c}  % Removed the extra '|' after the last column
        \hline
        \textbf{Scenario} & $K$ & $P$ & $\mu_{i,j}$ & $\Psi_{i,j,j}$ & $W_{i,j,k}$ & $z_{i,j}$ & $\bm{z}_{dim}$ \\  % Added header row
        \hline
        Scenario 1 & $50$ & $3$ & $\mathcal{N}(0,1)$ & $\text{IG}(5,1)$ & $\mathcal{N}(0,1)$ & $\mathcal{N}(0,1)$ & $3$ \\  
        Scenario 2 & $50$ & $3$ & $\mathcal{N}(0,0.1)$ & $\text{IG}(5,1)$ & $\text{Laplace}(0,10)$ & $\mathcal{N}(0,1)$ & $3$ \\  
        Scenario 3 & $25$ & $5$ & $\mathcal{N}(0,0.1)$ & $\text{IG}(5,2)$ & $\mathcal{N}(0,3)$ & $\mathcal{N}(0,1)$ & $3$ \\  
        Scenario 4 & $25$ & $15$ & $\mathcal{N}(0,0.1)$ & $\text{IG}(5,2)$ & $\mathcal{N}(0,3)$ & $\mathcal{N}(0,1)$ & $5$ \\  
        Scenario 5 & $25$ & $5$ & $\mathcal{N}(0,0.1)$ & $\text{IG}(5,2)$ & $\text{Laplace}(0,3)$ & $\mathcal{N}(0,1)$ & $3$ \\
        Scenario 6 & $25$ & $5$ & $\mathcal{N}(0,0.1)$ & $\text{IG}(5,2)$ & $\mathcal{N}(0,3)$ & $\text{Laplace}(0,1)$ & $3$ \\ 
        \hline
    \end{tabular}
    \label{tab:fa_variables}
\end{table}

\begin{algorithm}[h]
\caption{Generation of synthetic data for FA}
\label{Algo:fa}
\begin{algorithmic}[1]
    \REQUIRE Number of datasets $N$, number of samples $K$, and distributions $P^{\bm{\mu}}, P^{\bm{\Psi}}, P^{\bm{W}}, P^{\vz}$.
    \ENSURE A dataset $\mathcal{D}$ containing $(\vx_i, \vz_i)$ for $i=1,\dots,N$.

    \STATE Initialize $\mathcal{D} \leftarrow \varnothing$
    \FOR{$i = 1 \to N$}
        \STATE Draw $\bm{\mu}_i \sim P^{\bm{\mu}}$
        \STATE Draw $\bm{\Psi}_i \sim P^{\bm{\Psi}}$
        \STATE Draw $\bm{W}_i \sim P^{\bm{W}}$
        \STATE Draw $\vz_i \sim P^{\vz}$
        \FOR{$j = 1 \to K$}
            \STATE Draw $\bm{x}_{i,j} \sim \mathcal{N}\!\bigl(\bm{W}_i\,\vz_i + \bm{\mu}_i,\;\bm{\Psi}_i\bigr)$
        \ENDFOR
        \STATE Update $\mathcal{D} \leftarrow \mathcal{D} \cup \{\bigl(\vx_i, \vz_i\bigr)\}$
    \ENDFOR
\end{algorithmic}
\end{algorithm}

\subsection{Gaussian Mixture Models}

\label{app:datageneration_gmm}

\begin{table}[h]
    \centering
    \caption{Distribution and dimensionalitites of variables for the considered GMM scenarios.}    \label{tab:gmm_variables}
    \begin{tabular}{l|c|c|c|c|c|c}  % Removed the extra '|' after the last column
        \hline
        \textbf{Scenario} & $K$  & $M$ & $L$ & $\bm{\phi}_i$ & $\sigma^2_{i,m,l}$ & $\mu_{i,m,l}|\sigma^2_{i,m,l}$ \\
        \hline
          % Added header row
        Scenario 1 & $50$ & $5$ & $1$ & $\text{Dir}(1)$ & $\text{IG}(5,2)$ & $\mathcal{N}(0, 3 \sigma^2_{i,m,l}) $ \\
        Scenario 2& $25$ & $3$ & $3$ & $\text{Dir}(1)$ & $\text{IG}(5,2)$ & $\mathcal{N}(0, 3 \sigma^2_{i,m,l}) $ \\  % Added header row
        Scenario 3 & $50$  & $3$ & $5$ & $\text{Dir}(0.5)$ & $\text{IG}(5,2)$ & $\mathcal{N}(0, 5 \sigma^2_{i,m,l})$ \\  % Added header row
        Scenario 4 & $50$  & $3$ & $3$ & $\text{Dir}(1)$ & $\text{IG}(5,2)$ & $\mathcal{N}(0, 3 \sigma^2_{i,m,l})$ \\  % Added header row
        \hline
    \end{tabular}
\end{table}

In GMMs one assumes that the data of interest is generated by a convex combination of $M$ (multivariate) normal distributions, such that $p(\vx|\vz) = \sum_{m=1}^M \bm{\phi}_m p_m(\vx)$, where the probability vector $\bm{\phi} = (\phi_1,\ldots,\phi_M)$ comprises the mixture weights and $p_m$ denotes the $m$-th mixture component. We consider $p_m$ to take the form of a diagonal Gaussian with mean vector $\bm{\mu}_m$ and covariance matrix with diagonal elements $\bm{\sigma}_m^2$. We assume a prior $P^{\bm{\phi}}$ on $\bm{\phi}$, a prior $P^{\bm{\sigma}^2}$ on the variances of each component and a prior $P^{\bm{\mu}|\bm{\sigma}^2}$ for the means that depends on the variance of the respective component. More specifically, we assume a symmetric Dirichlet prior on $\bm{\phi}$ such that $P^{\bm{\phi}} = \text{Dir}(\mathcal{\alpha}_{Dir})$ and an independent inverse gamma distribution as prior on each component $\sigma^2_m$ of $\bm{\sigma}_m^2$. The prior on each component of $\bm{\mu}_{i,m} \in \mathbb{R}^L$ is then given by an independent normal distribution $ P^{\bm{\mu}|\bm{\sigma}^2_{i,m,l}} = \mathcal{N}(0, \lambda \sigma^2_{i,m,l})$. We use $\omega_{i,j}$ to denote the assignment of datapoint $j$ a component. \Cref{Algo:gmm} details the data generating process and \Cref{tab:gmm_variables} summarizes the different setups regarding the prior distributions.

\begin{algorithm}[h]
\caption{Generation of synthetic data for a GMM.}
\label{Algo:gmm}
\begin{algorithmic}[1]
    \REQUIRE Number of datasets $N$, mixture dimension parameters $M$, $L$, number of samples $K$, and distributions $P^{\bm{\phi}},\,P^{\bm{\sigma}^2},\,P^{\bm{\mu}\mid\bm{\sigma}^2}$.
    \ENSURE A dataset $\mathcal{D}$ containing $(\vx_i, \vz_i)$ for $i=1,\dots,N$.

    \STATE Initialize $\mathcal{D} \leftarrow \varnothing$
    \FOR{$i = 1 \to N$}
        \STATE Draw $\bm{\phi}_i \sim P^{\bm{\phi}}$
        \FOR{$m = 1 \to M$}
            \FOR{$l = 1 \to L$}
                \STATE Draw $\sigma^2_{i,m,l} \sim P^{\bm{\sigma}^2}$
                \STATE Draw $\mu_{i,m,l} \sim P^{\bm{\mu}\mid\bm{\sigma}^2_{i,m,l}}$
            \ENDFOR
        \ENDFOR
        \FOR{$j = 1 \to K$}
            \STATE Draw $\omega_{i,j} \sim \mathrm{Cat}\bigl(\bm{\phi}_i\bigr)$
            \STATE Draw $\bm{x}_{i,j} \sim \mathcal{N}\!\Bigl(\bm{\mu}_{i,\omega_{i,j}}, \bm{\sigma}^2_{i,\omega_{i,j}}\Bigr)$
        \ENDFOR
        \STATE Set $\vz_i \coloneqq \bigl(\,\bigl(\sigma^2_{i,m,l},\,\mu_{i,m,l}\bigr)\bigr)_{\substack{m=1,\dots,M\\l=1,\dots,L}}$
        \STATE Update $\mathcal{D} \leftarrow \mathcal{D} \cup \bigl\{(\vx_i, \vz_i)\bigr\}$
    \ENDFOR
\end{algorithmic}
\end{algorithm}

\newpage 

\section{Generating Realistic Data}
\label{sec:generating_real_data}
While we assume a data-generating process such as the one in \Cref{Algo:glmAppendix}, this is not necessarily the data-generating process that produces the data in the model's application as an in-context learner. Even when the generative process $P^{\vx, \vz}$ underlying a statistical model is sophisticated and complex in nature, model misspecification is inevitable in almost every practical application. While mismatches between the real data-generating processes and model assumptions can lead to various problems in traditional Bayesian modeling \citep{grunwald2017inconsistency}, the question of model misspecification plays a somewhat different and yet an especially central role for our ICL approach. 

More specifically, the ICL model learns the relationship between $P^{\vz|\vx}$ and a datapoint $\vx$ exclusively based on synthetic samples from the marginal $P^{\vx}$ implied by the statistical model with generative process $P^{\vx, \vz}$.
Given a real-world dataset $\vx^{\ast}\sim P^{\vx^{\ast}}$, model misspecification in terms of $P^{\vx^{\ast}}$ implies that the in-context learner needs to infer the posterior based on out-of-distribution data, where the problem is aggravated the more unrealistic $P^{\vx}$ is. 

\textcolor{rev}{To be able to access a reference or ground truth distribution, the data generating processes in our experiments need to match the structure of the GLM, FA and GMM approaches. While the generative processes of FA and GMMs directly prescribe how all parts of the data are generated, this can potentially cause a discrepancy between synthetically generated and real-world datasets. However, our empirical results (Section 4.1) demonstrate that the in-context learner can generalize to real-world data despite the discrepancy to the simulated datasets.} 

In the aforementioned GLM case, the distribution of the covariates $P^{\vu}$ does not affect the structure of $P^{\vz|\vx}$ in the data generating process (cf.~\Cref{Algo:glmAppendix}). We can therefore use a flexible prior $P^{\vu}$ such as the TabPFN-``prior'' \citep{hollmann2022tabpfn} to generate covariates $\vu$ and thereby effectively tackle the issue of model specification.

\section{Preprocessing of the Real-world Datasets}
\label{app:preprocessing}

The real-world datasets considered for the evaluation of all methods are proposed in a benchmark study by \citet{grinsztajn2022tree}. We standardize all features, scale and shift the target such that it has the mean and variance implied by the prior structure of the respective generative model. Furthermore, for the GLM scenarios, we apply a Yeo-Johnson transform on the target variable \citep{yeo2000new} before applying the scaling. 
In cases where the number of features in the real-world dataset exceeds that of our scenario, we select those features with the most distinct values in the original dataset and randomly sub-sample the appropriate number of samples from the real-world datasets for our experiments.

\section{Background on Conditional Flow-matching}

\label{app:background_cfm}

Flow matching, initially used in image synthesis leverages normalizing flows \citep{papamakarios2021normalizing} to model arbitrary distributions. Continuous normalizing flows \citep{lipman2022flow} have emerged as a potent tool for modeling complex distributions. For example, recent advancements have shown its effectiveness in state-of-the-art image generation, outperforming diffusion-based methods in likelihood and sample quality on ImageNet \citep{lipman2022flow}. Techniques like FlowTurbo have accelerated class-conditional and text-to-image generation, setting new benchmarks \citep{zhao2024flowturbo}. Additionally, applying flow matching in latent spaces of pretrained autoencoders has enhanced computational efficiency and scalability for high-resolution image synthesis \citep{dao2023flow}. Similarly, flow-based models have been successfully applied to protein structure prediction, improving accuracy and efficiency in modeling complex protein conformations \citep{yim2024improved, yim2023fast}.

In the area of simulation-based inference, \citet{wildberger2024flow} introduce the idea of using continuous normalizing flows in order to efficiently approximate complex posterior distributions. In particular, they apply the framework to the field of gravitational-wave inference, substantially outperforming approaches based on discrete flows. Furthermore, they demonstrate good performance on the existing SBI-Benchmark \citep{lueckmann2021benchmarking} using a simple MLP-based architecture. 

\section{Relationship of our approach to density estimation methods}

\label{sec:density_estimation}

An alternative to flow matching for parameterizing our model $Q^{\vz|\vx}$ would be explicit (conditional) density estimation. While knowledge of the explicit density of a distribution can be useful for downstream tasks, we would like to reemphasize that we consider the problem of full Bayesian inference via \textit{sampling} from the posterior in this paper. 

Popular explicit density estimation methods include i-DODE \citep{zheng2023improved}, which proposes several techniques for improving maximum likelihood estimation from diffusion ordinary differential equations (ODEs), including velocity parameterization and techniques for variance reduction, leading to faster convergence. Additionally,in \citep{sahoo2024diffusion} log-likelihood estimation is improved by casting the learned diffusion process as a variational posterior that yields a tighter evidence lower bound on the actual likelihood. \citep{lienen2025generative} propose a novel generative model using iterative Gaussian posterior inference and empirically demonstrate that it yields strong results in log-likelihood estimation. Furthermore, \citet{salazar2023vart} use variational inference to learn Bayesian regression trees, which could be used for multivariate density estimation. 

Note that it is, in principle, also possible to recover the explicit density of $Q^{\vz|\vx}$, which is parametrized in  the Flow Matching Framework that is used for our approach \citep{wildberger2024flow}.

\section{Hyperparameters, Software and Computational Setup}

In this section, we detail the hyperparameters, used software and computational setups for all our experiments. 

\label{app:hyper}

\subsection{ICL}

To ensure maximum comparability across different experiments, we fix the hyperparameters for all ICL experiments:
For the architecture of the model introduced in \Cref{sec:architecture}, we use the following configuration: The dimensionality of encoder representations is set to 512 and is expanded to 1024 in the feed-forward blocks. We use 8 heads and 8 encoder layers with a dropout rate of 0.1. For the decoder part we also use 512 as the dimensionality of the representations and 1024 as the intermediate representation in the feed-forward layers and a dropout rate of 0.1. Furthermore, 3 simple fully connected layers with adaLN conditioning are used for final processing in the decoder. For the time conditioning, we use 3 simple fully connected layers to map the scalar-valued time $t$ onto a 512 dimensional conditioning vector that is used for the adaLN blocks in the decoder. This yields a model of around 43.1 million parameters. We use no tokenization for either the encoder or the decoder and simple embedding layers to map the encoder- and decoder-input onto the feed-forward dimensions. 
%The encoder-setup is thus analogous to that in Tab-PFN \citep{muller2021transformers}. 

We use an Adam optimizer \citep{kingma2014adam} with a cosine learning rate schedule \citep{loshchilov2016sgdr}, where the maximum learning rate is $5 \cdot 10^{-4}$, the final division factor is $10^4$ and 10 percent of the epochs are used for warm-up. We use a weight decay parameter of $10^{-5}$ and a batch size of 1024 and gradient clipping with a maximum gradient norm of one. We use in total 75 million synthetic samples for all scenarios. Of the total number, half, i.e.\ 37.5 million, are used for training and 10 percent for validation and the remaining 40 percent for testing. Note that we observe convergence of the loss usually much earlier than after this training duration, but fix the number of samples for consistency across experiments. A single L4 GPU is used for the GLM scenarios and a single A100 GPU for the FA and GMM cases. 

To solve the ODE for the sample generation, dopri5 \citep{dormand1980family} as implemented in Torchdiffeq \citep{torchdiffeq} is used in the adjoint version. We set the relative and absolute tolerance to $10^{-7}$. The $\sigma_{\textrm{min}}$ parameter in the CNF-loss is set to $10^{-4}$.

\begin{figure}[!htb]
\minipage{0.32\textwidth}
  \includegraphics[width=\linewidth]{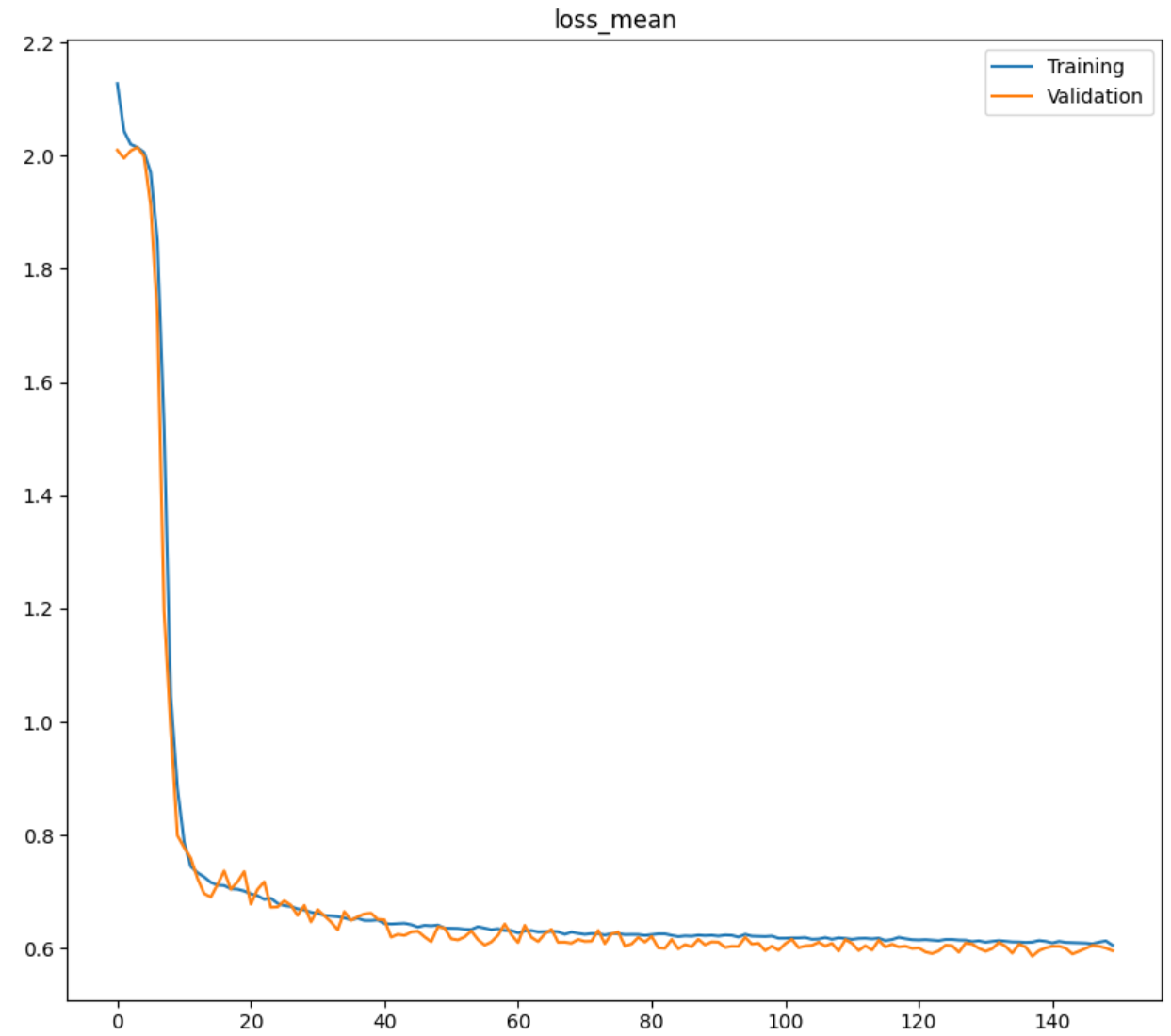}
  \caption{Learning curves for GLM scenario 1 with a Normal Prior on the coefficients $\bm{\beta}$ and
an Inverse Gamma prior on $\sigma_2$.}\label{fig:curves_glm}
\endminipage\hfill
\minipage{0.32\textwidth}
  \includegraphics[width=\linewidth]{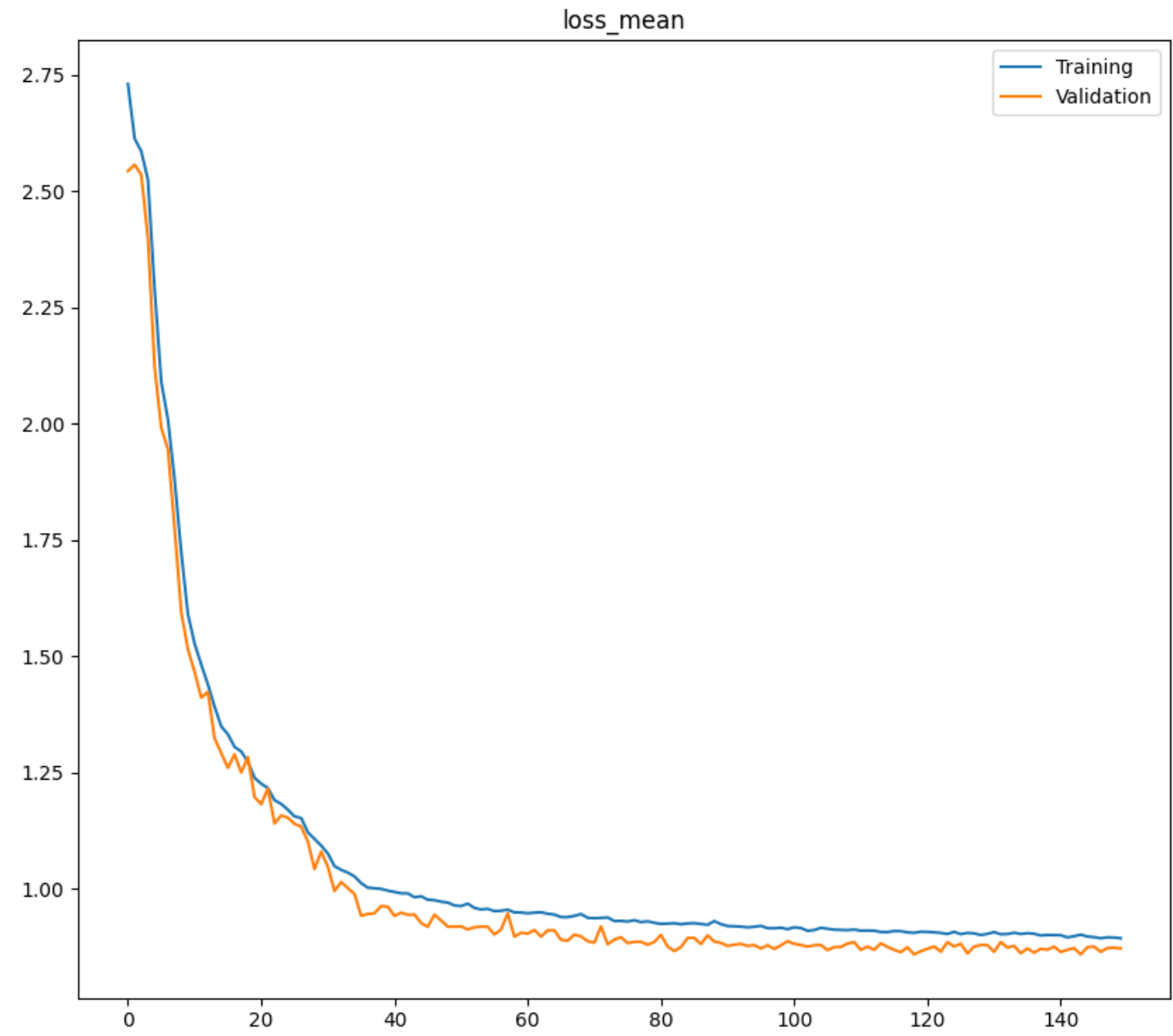}
  \caption{Learning curves for GMM scenario 1 with $M=5$ components, $K=50$ datapoints and $L=1$ dimensions.}\label{fig:curves_univariate_gmm}
\endminipage\hfill
\minipage{0.32\textwidth}%
  \includegraphics[width=\linewidth]{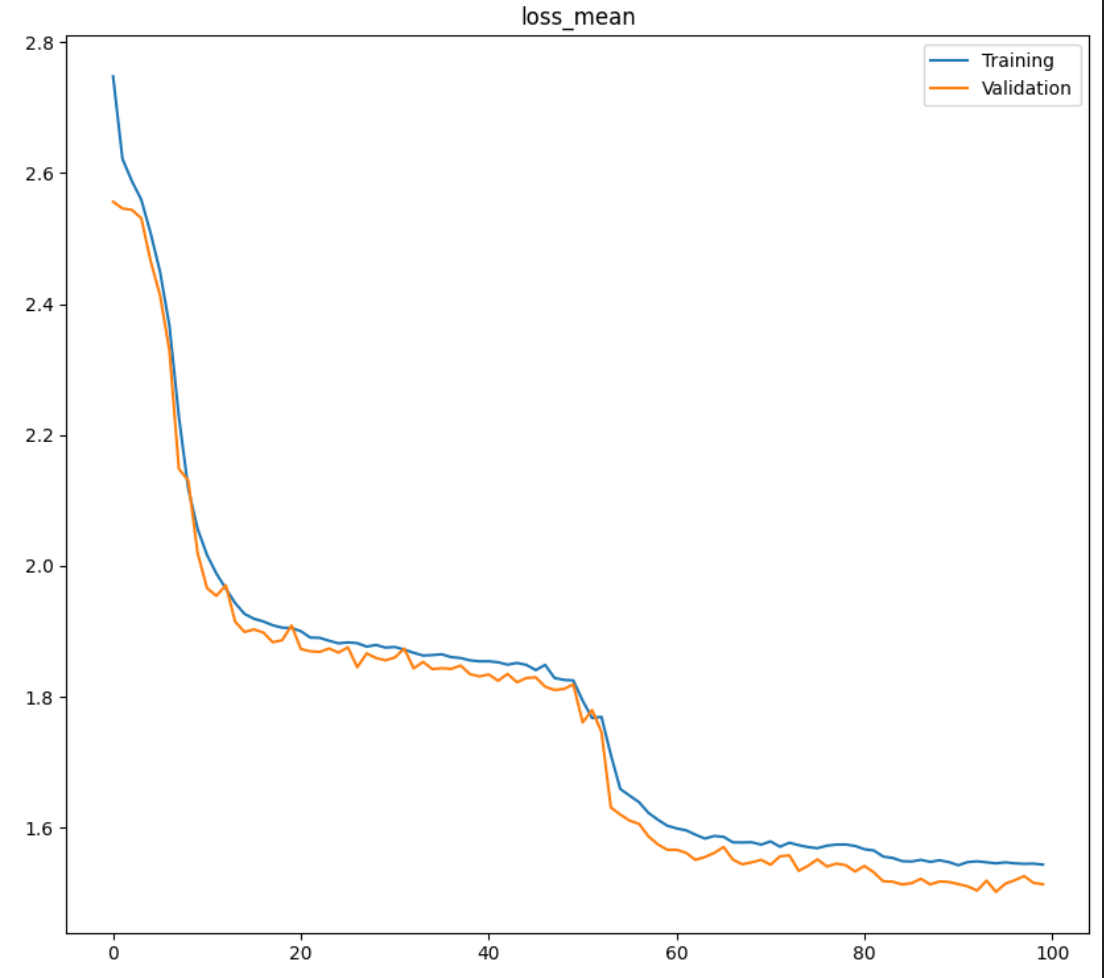}
  \caption{Learning curves for GMM scenario 3 with $M=3$ components, $K=50$ datapoints and $L=5$ dimensions.}\label{fig:curves_multivariate_gmm}
\endminipage
\end{figure}

\subsection{HMC}

We use HMC with a NUTS kernel \citep{hoffman2014no} as a reference for all experiments where no analytical solution is available. We set the number of burn-in samples to 500 and use one chain for all uni-modal problems and three times the number of potential modes in all other cases. More specifically, we use $M \times 3$ chains for all GMM scenarios. The Pyro implementation of NUTS is used for the GLM scenarios \citep{bingham2019pyro} and the conceptually identical, albeit computationally faster implementation in Numpyro for the FA and GMM cases \citep{phan2019composable}.

\subsection{VI}

For the variational inference methods, we utilize automatic guide generation based on the ground-truth data-generating processes \citep{kucukelbir2017automatic}. Pyro is used for the implementation of the probabilistic programs, which we also use to sample the synthetic training data, for the automatic guide generation, and for the implementation of the actual VI methods \citep{bingham2019pyro}. Default hyperparameters, as well as an Adam optimizer \citep{kingma2014adam} with a learning rate of $10^{-2}$ is used for all methods except for AutoIAF where a learning rate of $10^{-3}$ is used. We perform 2000 full-batch gradient update steps for each method.

\section{\textcolor{rev}{Runtimes}}

\textcolor{rev}{
We use a single L4 GPU for generating samples based on our ICL approach and HMC in the GLM scenarios, a single A100 for our ICL approach and HMC in the FA and GMM scenarios, and an Intel(R) Xeon(R) CPU @ 2.20GHz CPU with two virtual cores and 40 gigabytes of RAM for the VI methods. 
Across all considered GLM scenarios, pre-training takes on average $14.89$ hours with a standard error of $18.01$ minutes. For the FA scenarios, on average $3.95$ hours with a standard error of $11.38$ minutes is used for pretraining and for the GMM scenarios $10.63$ with a standard error of   $72.88$ minutes.}

\textcolor{rev}{
When applied in order to generate samples for a new dataset, the benchmarked VI methods have, as expected the lowest runtime. The Laplace approximation is the fastest of all methods, while our ICL appraoch has consistently a lower runtime compared to HMC. Overall, the ICL method takes around 2 minutes on the GLM tasks, around 30 seconds in the FA scenarios and less than 2 minutes for the inference regarding the GMM tasks.}

\textcolor{rev}{
This difference is especially pronounced in the FA and GMM scenarios. Please note that the runtime of the ICL method also fundamentally depends on the used precision for solving the underlying differential equation where we use a relatively high relative and absolute precision of $10^{-7}$. Decreasing this value might lead to significantly faster inference time while maintaining sample quality.}

\begin{table}[h!]
    \centering
    \textcolor{rev}{
    \caption{\textcolor{rev}{Runtime Metrics for all GLM, FA, and GMM Scenarios}}
    \begin{tabular}{l|l|c}
        \textbf{Scenario} & \textbf{Method}              & \textbf{Mean Runtime (s)} \\ \hline
        \multirow{7}{*}{GLM} 
                         & Laplace Approximation        & $10.48 \, (\pm 0.25)$          \\ 
                         & VI: DiagonalNormal           & $12.02 \, (\pm 0.26)$          \\ 
                         & VI: MultivariateNormal       & $13.70 \, (\pm 0.29)$          \\ 
                         & VI: Structured Normal        & $19.81 \, (\pm 0.98)$          \\ 
                         & VI:IAF                       & $15.44 \, (\pm 0.30)$          \\ 
                         & HMC                          & $120.24 \, (\pm 13.94)$        \\ 
                         & \textbf{ICL (ours)}                 & $107.79 \, (\pm 17.36)$        \\ \hline
        \multirow{7}{*}{FA} 
                         & Laplace Approximation        & $17.85 \, (\pm 0.21)$          \\ 
                         & VI: DiagonalNormal           & $20.94 \, (\pm 0.66)$          \\ 
                         & VI: MultivariateNormal       & $20.84 \, (\pm 0.28)$          \\ 
                         & VI: Structured Normal        & $36.17 \, (\pm 0.61)$          \\ 
                         & VI:IAF                       & $23.75 \, (\pm 0.38)$          \\ 
                         & HMC                          & $248.26 \, (\pm 57.88)$        \\ 
                         & \textbf{ICL (ours)}                 & $31.49 \, (\pm 4.97)$          \\ \hline
        \multirow{7}{*}{GMM} 
                         & Laplace Approximation        & $27.52 \, (\pm 0.40)$          \\ 
                         & VI: DiagonalNormal           & $29.74 \, (\pm 0.57)$          \\ 
                         & VI: MultivariateNormal       & $30.50 \, (\pm 0.41)$          \\ 
                         & VI: Structured Normal        & $42.44 \, (\pm 0.44)$          \\ 
                         & VI:IAF                       & $33.39 \, (\pm 0.49)$          \\ 
                         & HMC                          & $239.67 \, (\pm 32.71)$        \\ 
                         & \textbf{ICL (ours)}                 & $93.88 \, (\pm 10.47)$         \\ \hline
    \end{tabular}
    \label{tab:combined_runtime_scenarios}
    }
\end{table}

\newpage

\section{Ablation: Different Learning Rates for VI}

To investigate the role of the learning rate parameter for the benchmarked VI methods, we record the performance for learning-rate values of $10^{-2}$, $10^{-3}$ and $10^{-4}$ across a prototypical GLM, a FA and a GMM scenario, where we use 10 synthetic and 10 real-world datasets. In summary, while we find the VI methods to often be quite robust to the choice of the learning rate, those results also confirm our choice of setting the learning rate to $10^{-2}$ for the Laplace approximation, variational inference with a diagonal normal distribution, a multivariate normal distribution and a structured normal distribution, and to a value of $10^{-3}$ for the VI approach with inverse autoregressive flows. 

For the GLM-scenario, we find in terms of the C2ST metric that VI with an ordinary multivariate normal distribution and VI with a structured normal distribution and a learning rate of $10^{-2}$ are the best models on the synthetic data. While MMD also indicates that this learning rate yields ideal results for those models, VI with inverse auoregressive flows has good values across the different learning rates with the minimum for $10^{-3}$. The $\mathcal{W}_2$ metric indicates a similar tendency. 

\begin{table}[h]
\centering
\caption{Results of VI methods with different learning rates on 10 synthetic and 10 real-world
datasets: Linear regression with a normal prior on the coefficients $\bm{\beta}$ and an inverse gamma prior on the variance $\sigma^2$ (scenario 1). Comparison to HMC samples. All results within two standard errors of the best average result are marked in \textbf{bold}.}\label{Tab:Vi_ablation_GLM}
\resizebox{1.0\textwidth}{!}{
\begin{tabular}{p{3.4cm}m{0.7cm} m{2.3cm}m{2.3cm}m{2.3cm}| m{2.3cm}m{2.3cm}m{2.3cm}} % No vertical lines for header row
\toprule
\textbf{Model} & \textbf{LR} & \multicolumn{3}{c}{\textbf{Synthetic Evaluation}} & \multicolumn{3}{c}{\textbf{Real-World Evaluation}} \\
\cmidrule(lr){3-5} \cmidrule(lr){6-8}
& & C2ST ($\downarrow$) & MMD ($\downarrow$) & $\mathcal{W}_2$ ($\downarrow$) & C2ST ($\downarrow$) & MMD ($\downarrow$) & $\mathcal{W}_2$ ($\downarrow$) \\
\midrule
Laplace Approximation & 1e-2 & 1.000 ($\pm$ 0.000) & 2.342 $(\pm$ 0.390) & 2.121 ($\pm$ 0.100) & 1.000 ($\pm$ 0.000) & 2.134 $(\pm$ 0.107) & 2.095 ($\pm$ 0.062) \\
Laplace Approximation & 1e-3 & 1.000 ($\pm$ 0.000) & 2.341 $(\pm$ 0.389) & 2.121 ($\pm$ 0.100) & 1.000 ($\pm$ 0.000) & 2.133 $(\pm$ 0.108) & 2.095 ($\pm$ 0.062) \\
Laplace Approximation & 1e-4 & 1.000 ($\pm$ 0.000) & 2.341 $(\pm$ 0.389) & 2.121 ($\pm$ 0.100) & 1.000 ($\pm$ 0.000) & 2.133 $(\pm$ 0.108) & 2.095 ($\pm$ 0.062) \\
\midrule
VI: DiagonalNormal & 1e-2 & 0.892 ($\pm$ 0.074) & 0.921 $(\pm$ 0.374) & \textbf{1.411} ($\pm$ 0.174) & 0.889 ($\pm$ 0.062) & 0.819 $(\pm$ 0.343) & 1.339 ($\pm$ 0.190) \\
VI: DiagonalNormal & 1e-3 & 0.966 ($\pm$ 0.024) & 1.588 $(\pm$ 0.540) &  \textbf{1.672} ($\pm$ 0.203) & 0.981 ($\pm$ 0.017) & 1.685 $(\pm$ 0.331) & 1.739 ($\pm$ 0.139) \\
VI: DiagonalNormal & 1e-4 & 0.971 ($\pm$ 0.010) & 1.572 $(\pm$ 0.300) & 1.666 ($\pm$ 0.081) & 0.849 ($\pm$ 0.030) & 0.575 $(\pm$ 0.127) & \textbf{1.221} ($\pm$ 0.098) \\
\midrule
VI: MultivariateNormal & 1e-2 & \textbf{0.725} ($\pm$ 0.064) & \textbf{0.523} $(\pm$ 0.242) & \textbf{1.114} ($\pm$ 0.261) & \textbf{0.625} ($\pm$ 0.051) & \textbf{0.470} $(\pm$ 0.066) & \textbf{0.918} ($\pm$ 0.119) \\
VI: MultivariateNormal & 1e-3 & 0.964 ($\pm$ 0.008) & 1.455 $(\pm$ 0.327) & 1.617 ($\pm$ 0.100) & 0.853 ($\pm$ 0.052) & 0.634 $(\pm$ 0.266) & \textbf{1.238} ($\pm$ 0.151) \\
VI: MultivariateNormal & 1e-4 & 0.984 ($\pm$ 0.005) & 1.848 $(\pm$ 0.324) & 1.773 ($\pm$ 0.079) & 0.899 ($\pm$ 0.020) & 0.807 $(\pm$ 0.094) & \textbf{1.345} ($\pm$ 0.079) \\
\midrule
VI: Structured Normal & 1e-2 & \textbf{0.734} ($\pm$ 0.063) & \textbf{0.541} $(\pm$ 0.254) & \textbf{1.119} ($\pm$ 0.264) & \textbf{0.670} ($\pm$ 0.047) & \textbf{0.467} $(\pm$ 0.086) & \textbf{1.060} ($\pm$ 0.130) \\
VI: Structured Normal & 1e-3 & 0.882 ($\pm$ 0.042) & 0.719 $(\pm$ 0.315) & 1.335 ($\pm$ 0.149) & 0.776 ($\pm$ 0.045) & \textbf{0.473} $(\pm$ 0.081) & \textbf{1.064} ($\pm$ 0.131) \\
VI: Structured Normal & 1e-4 & 0.890 ($\pm$ 0.027) & 0.710 $(\pm$ 0.290) & 1.347 ($\pm$ 0.138) & 0.771 ($\pm$ 0.049) & \textbf{0.468} $(\pm$ 0.078) & \textbf{1.062} ($\pm$ 0.128) \\
\hline
VI: IAF & 1e-2 & 0.840 ($\pm$ 0.036) & \textbf{0.502} $(\pm$ 0.262) & \textbf{1.272} ($\pm$ 0.170) & \textbf{0.614} ($\pm$ 0.045) & \textbf{0.455} $(\pm$ 0.048) & \textbf{0.957} ($\pm$ 0.105) \\
VI: IAF & 1e-3 & 0.797 ($\pm$ 0.065) & \textbf{0.485} $(\pm$ 0.556) & \textbf{1.169} ($\pm$ 0.313) & \textbf{0.619} ($\pm$ 0.036) & \textbf{0.469} $(\pm$ 0.064) & \textbf{0.989} ($\pm$ 0.124) \\
VI: IAF & 1e-4 & 0.803 ($\pm$ 0.068) & \textbf{0.475} $(\pm$ 0.535) & \textbf{1.162} ($\pm$ 0.291) & \textbf{0.612} ($\pm$ 0.034) & \textbf{0.457} $(\pm$ 0.055) & \textbf{0.977} ($\pm$ 0.113) \\
\bottomrule
\end{tabular}
}
\end{table}

Regarding the learning rate for the FA scenario, one can first see that no single learning rate seems to dominate substantially given the variance of the results. However, on the synthetic data for the Laplace approximation, as well as VI with a diagonal normal distribution, a multivariate normal and a structured normal distribution, the lowest average result is obtained for a learning rate of $10^{-2}$, while for VI with inverse autoregressive flows the best performance is obtained when the learning rate equals $10^{-3}$. The real-world results are the best for VI with a structured normal distribution and a learning rate of $10^{-2}$.

\begin{table}[h]
\centering
\caption{Results of VI methods with different learning rates on 10 synthetic and 10 real-world
datasets: Factor analysis with Gaussian priors on the weights and the latents and $K = 25$ datapoints, $P = 5$ features, and dimensionality of the latents $\textbf{z}_{dim} = 3$ (scenario 3). Comparison to HMC samples. All results within two standard errors of the best average result are marked in \textbf{bold}.}\label{Tab:Vi_ablation_FA}
\resizebox{1.0\textwidth}{!}{
\begin{tabular}{p{3.4cm}m{0.7cm} m{2.3cm}m{2.3cm}m{2.3cm}| m{2.3cm}m{2.3cm}m{2.3cm}} % No vertical lines for header row
\toprule
\textbf{Model} & \textbf{LR} & \multicolumn{3}{c}{\textbf{Synthetic Evaluation}} & \multicolumn{3}{c}{\textbf{Real-World Evaluation}} \\
\cmidrule(lr){3-5} \cmidrule(lr){6-8}
& & C2ST ($\downarrow$) & MMD ($\downarrow$) & $\mathcal{W}_2$ ($\downarrow$) & C2ST ($\downarrow$) & MMD ($\downarrow$) & $\mathcal{W}_2$ ($\downarrow$) \\
\midrule
Laplace Approximation & 1e-2 & 1.000 ($\pm$ 0.000) & 3.449 $(\pm$ 0.821) & \textbf{1.773} ($\pm$ 0.539)  & 1.000 ($\pm$ 0.000) & 2.703 $(\pm$ 0.312) & \textbf{0.362} ($\pm$ 0.017) \\
Laplace Approximation & 1e-3 & 1.000 ($\pm$ 0.000) & 4.288 $(\pm$ 0.853) & \textbf{2.263} ($\pm$ 0.732)  & 1.000 ($\pm$ 0.000) & 2.896 $(\pm$ 0.238) & \textbf{0.376} ($\pm$ 0.022) \\
Laplace Approximation & 1e-4 & 1.000 ($\pm$ 0.000) & 4.252 $(\pm$ 0.611) & \textbf{2.122} ($\pm$ 0.430)  & 1.000 ($\pm$ 0.000) & 2.805 $(\pm$ 0.181) & \textbf{0.368} ($\pm$ 0.017) \\
\midrule
VI: DiagonalNormal & 1e-2 & 0.998 ($\pm$ 0.002) & 2.880 $(\pm$ 1.046) & \textbf{1.457} ($\pm$ 0.559) & 0.944 ($\pm$ 0.008) & 1.022 $(\pm$ 0.067) & \textbf{0.230} ($\pm$ 0.010) \\
VI: DiagonalNormal & 1e-3 & 0.998 ($\pm$ 0.002) & 2.973 $(\pm$ 0.834) & \textbf{1.465} ($\pm$ 0.540) & 0.941 ($\pm$ 0.006) & 0.997 $(\pm$ 0.056) & \textbf{0.229} ($\pm$ 0.010) \\
VI: DiagonalNormal & 1e-4 & 1.000 ($\pm$ 0.001) & 3.416 $(\pm$ 0.761) & \textbf{1.602} ($\pm$ 0.437) & 0.943 ($\pm$ 0.009) & 0.997 $(\pm$ 0.057) & \textbf{0.229} ($\pm$ 0.010) \\
\midrule
VI: MultivariateNormal & 1e-2 & 0.993 ($\pm$ 0.007) & 2.969 $(\pm$ 1.089) & \textbf{1.506} ($\pm$ 0.659) & \textbf{0.929} ($\pm$ 0.007) & \textbf{0.957} $(\pm$ 0.048) & \textbf{0.224} ($\pm$ 0.010) \\
VI: MultivariateNormal & 1e-3 & 0.996 ($\pm$ 0.004) & 3.140 $(\pm$ 0.910) & \textbf{1.570} ($\pm$ 0.625) & 0.934 ($\pm$ 0.009) & \textbf{0.971} $(\pm$ 0.054) & \textbf{0.225} ($\pm$ 0.010) \\
VI: MultivariateNormal & 1e-4 & 0.997 ($\pm$ 0.007) & 3.464 $(\pm$ 0.791) & \textbf{1.639} ($\pm$ 0.426) & 0.934 ($\pm$ 0.005) & \textbf{0.962} $(\pm$ 0.049) & \textbf{0.225} ($\pm$ 0.010) \\
\midrule
VI: Structured Normal & 1e-2 & 0.998 ($\pm$ 0.002) & 3.005 $(\pm$ 0.871) & \textbf{1.481} ($\pm$ 0.504)  & 0.947 ($\pm$ 0.005) & 1.003 $(\pm$ 0.066) & \textbf{0.230} ($\pm$ 0.009) \\
VI: Structured Normal & 1e-3 & 0.999 ($\pm$ 0.001) & 3.244 $(\pm$ 0.665) & \textbf{1.619} ($\pm$ 0.559)  & 0.948 ($\pm$ 0.007) & 1.033 $(\pm$ 0.078) & \textbf{0.232} ($\pm$ 0.009) \\
VI: Structured Normal & 1e-4 & 0.999 ($\pm$ 0.001) & 3.119 $(\pm$ 0.612) & \textbf{1.487} ($\pm$ 0.400)  & 0.943 ($\pm$ 0.007) & 0.998 $(\pm$ 0.056) & \textbf{0.229} ($\pm$ 0.010) \\
\midrule
VI: IAF & 1e-2 & \textbf{0.939} ($\pm$ 0.040) & \textbf{2.836} $(\pm$ 0.293) & \textbf{1.247} ($\pm$ 0.297) & 0.944 ($\pm$ 0.008) & 1.518 $(\pm$ 0.048) & 1.332 ($\pm$ 0.027) \\
VI: IAF & 1e-3 & \textbf{0.927} ($\pm$ 0.047) & \textbf{2.758} $(\pm$ 0.342) & \textbf{1.195} ($\pm$ 0.331) & 0.949 ($\pm$ 0.009) & 1.560 $(\pm$ 0.031) & \textbf{1.392} ($\pm$ 0.024) \\
VI: IAF & 1e-4 & \textbf{0.842} ($\pm$ 0.038) & \textbf{2.862} $(\pm$ 0.296) & \textbf{1.281} ($\pm$ 0.292) & 0.943 ($\pm$ 0.008) & 1.493 $(\pm$ 0.039) & 1.302 ($\pm$ 0.039) \\
\bottomrule
\end{tabular}
}
\end{table}

For the GMM scenario, we find that VI with a diagonal, structured and ordinary normal distribution obtain the best results, namely for learning rates of $10^{-2}$ and $10^{-3}$, taking the variance into account. Just considering the averages leads to the conclusion that $10^{-2}$ is the best choice here. 
The results on the real-world data confirm that $10^{-2}$ is the optimal choice for VI with a diagonal normal and ordinary multivariate normal, while VI with inverse autoregressive flows has good results across all choices regarding the learning rate.

\begin{table}[h]
\centering
\caption{
Results of VI methods with different learning rates on 10 synthetic and 10 real-world datasets: Gaussian Mixture Model with $K=50$ datapoints, $L=1$ features (univariate case), $M=5$ components, $\lambda = 3$, and $\alpha_{dir} = 1$ (scenario 1). Comparison to HMC samples.All results within two standard errors of the best average result are marked in \textbf{bold}.}\label{Tab:Vi_ablation_GMM}
\resizebox{1.0\textwidth}{!}{
\begin{tabular}{p{3.4cm}m{0.7cm} m{2.3cm}m{2.3cm}m{2.3cm}| m{2.3cm}m{2.3cm}m{2.3cm}} % No vertical lines for header row
\toprule
\textbf{Model} & \textbf{LR} & \multicolumn{3}{c}{\textbf{Synthetic Evaluation}} & \multicolumn{3}{c}{\textbf{Real-World Evaluation}} \\
\cmidrule(lr){3-5} \cmidrule(lr){6-8}
& & C2ST ($\downarrow$) & MMD ($\downarrow$) & $\mathcal{W}_2$ ($\downarrow$) & C2ST ($\downarrow$) & MMD ($\downarrow$) & $\mathcal{W}_2$ ($\downarrow$) \\
\midrule
Laplace Approximation & 1e-2 & 1.000 ($\pm$ 0.000) & 4.380 $(\pm$ 1.386) & \textbf{4.838} ($\pm$ 1.521) & 1.000 ($\pm$ 0.000) & 4.588 $(\pm$ 1.229) & \textbf{6.813} ($\pm$ 1.697) \\
Laplace Approximation & 1e-3 & 1.000 ($\pm$ 0.000) & 3.893 $(\pm$ 1.433) & \textbf{4.010} ($\pm$ 1.233) & 1.000 ($\pm$ 0.000) & 4.699 $(\pm$ 1.193) & \textbf{6.986} ($\pm$ 0.981) \\
Laplace Approximation & 1e-4 & 1.000 ($\pm$ 0.000) & 4.463 $(\pm$ 1.117) & \textbf{4.610} ($\pm$ 1.027) & 1.000 ($\pm$ 0.000) & 4.710 $(\pm$ 1.205) & \textbf{6.995} ($\pm$ 0.869) \\
\midrule
VI: DiagonalNormal & 1e-2 & \textbf{0.979} ($\pm$ 0.138) & \textbf{1.370} $(\pm$ 1.394) & \textbf{3.522} ($\pm$ 1.634) & \textbf{0.985} ($\pm$ 0.030) & 2.384 $(\pm$ 1.318) & \textbf{6.202} ($\pm$ 1.747) \\
VI: DiagonalNormal & 1e-3 & \textbf{0.990} ($\pm$ 0.096) & \textbf{1.454} $(\pm$ 1.454) & \textbf{3.650} ($\pm$ 1.743) & 0.999 ($\pm$ 0.002) & 3.026 $(\pm$ 0.977) & \textbf{6.959} ($\pm$ 0.890) \\
VI: DiagonalNormal & 1e-4 & 1.000 ($\pm$ 0.001) & 2.390 $(\pm$ 1.177) & \textbf{4.903} ($\pm$ 1.278) & 0.998 ($\pm$ 0.007) & 2.830 $(\pm$ 1.001) & \textbf{7.007} ($\pm$ 0.987) \\
\midrule
VI: MultivariateNormal & 1e-2 & \textbf{0.978} ($\pm$ 0.119) & \textbf{1.351} $(\pm$ 1.410) & \textbf{3.474} ($\pm$ 1.604) & \textbf{0.987} ($\pm$ 0.024) & 2.375 $(\pm$ 1.304) & \textbf{6.189} ($\pm$ 1.761) \\
VI: MultivariateNormal & 1e-3 & \textbf{0.980} ($\pm$ 0.089) & \textbf{1.476} $(\pm$ 1.480) & \textbf{3.681} ($\pm$ 1.734) & 0.997 ($\pm$ 0.008) & 2.808 $(\pm$ 1.014) & \textbf{6.964} ($\pm$ 0.944) \\
VI: MultivariateNormal & 1e-4 & 1.000 ($\pm$ 0.001) & 2.114 $(\pm$ 1.140) & \textbf{4.532} ($\pm$ 1.187) & 0.997 ($\pm$ 0.007) & 2.799 $(\pm$ 1.012) & \textbf{6.963} ($\pm$ 0.950) \\
\midrule
VI: Structured Normal & 1e-2 & \textbf{0.958} ($\pm$ 0.129) & \textbf{1.246} $(\pm$ 1.615) & \textbf{3.225} ($\pm$ 1.701) & 1.000 ($\pm$ 0.001) & 2.911 $(\pm$ 0.753) & \textbf{6.675} ($\pm$ 1.403) \\
VI: Structured Normal & 1e-3 & \textbf{0.979} ($\pm$ 0.092) & \textbf{1.593} $(\pm$ 1.561) & \textbf{3.395} ($\pm$ 1.440) & 0.998 ($\pm$ 0.007) & 2.882 $(\pm$ 1.070) & \textbf{6.968} ($\pm$ 0.941) \\
VI: Structured Normal & 1e-4 & 1.000 ($\pm$ 0.001) & 2.270 $(\pm$ 1.133) & \textbf{4.733} ($\pm$ 1.162) & 0.997 ($\pm$ 0.009) & 2.802 $(\pm$ 1.012) & \textbf{6.953} ($\pm$ 0.948) \\
\midrule
VI: IAF & 1e-2 & 0.998 ($\pm$ 0.003) & \textbf{1.539} $(\pm$ 0.691) & \textbf{8.371} ($\pm$ 0.750) & \textbf{0.987} ($\pm$ 0.022) & \textbf{1.376} $(\pm$ 0.799) & \textbf{8.082} ($\pm$ 1.352) \\
VI: IAF & 1e-3 & 0.997 ($\pm$ 0.004) & \textbf{1.443} $(\pm$ 0.564) & \textbf{8.517} ($\pm$ 0.820) & \textbf{0.988} ($\pm$ 0.020) & \textbf{1.304} $(\pm$ 0.855) & \textbf{8.425} ($\pm$ 1.281) \\
VI: IAF & 1e-4 & 0.997 ($\pm$ 0.004) & \textbf{1.602} $(\pm$ 0.628) & \textbf{7.888} ($\pm$ 0.783) & \textbf{0.987} ($\pm$ 0.020) & \textbf{1.380} $(\pm$ 0.848) & \textbf{7.729} ($\pm$ 1.322) \\
\bottomrule
\end{tabular}
}
\end{table}

\clearpage

\section{Detailed Experimental Results}
\label{sec:detailed_results}

In this section, we describe our experimental results in detail, discussing how different scenarios for GLMs, FA and GMMs affect the performance of different approaches. 

\subsection{Generalized Linear Models}
\label{app:addres_glm}

\begin{figure}[h]
   \centering
    \includegraphics[width=1\linewidth]{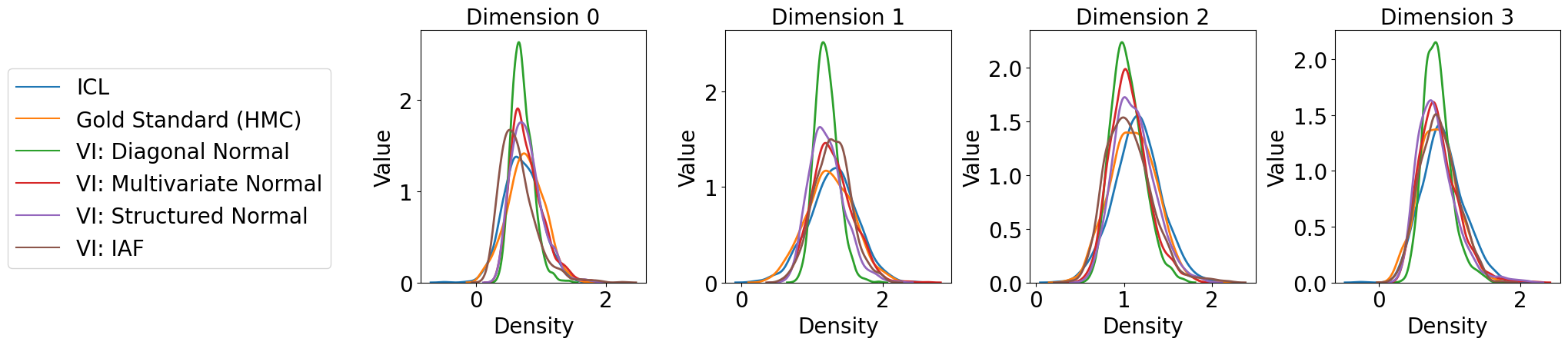}
    \caption{Density plots for first three the marginals of the posterior in a GLM with a gamma prior on the coefficients $\beta$, and an inverse gamma prior on the variance $\sigma^2$ of the responses. The data is part of the Miami housing 2016 dataset.}
   \label{fig:posteriormarginalgamma}
\end{figure}

\Cref{tab:glm_res_detail} contains detailed results regarding the performance of the proposed ICL and the reference VI approaches. In summary, we find that on the synthetic data, our ICL method has the overall best performance, or a performance not significantly worse than that of the best model, with respect to the C2ST metric.\footnote{We refer to a difference that is larger than two standard deviations as ``significant''.}
More specifically, ICL significantly outperforms all other models in 5 out of seven cases w.r.t.\ the C2ST and also the MMD metric. While the $\mathcal{W}_2$ metric exhibits a larger variance, it also indicates that on the synthetic data, ICL yields the significantly best result in those 5 cases. 

On the real-world data, the differences between ICL and VI are less pronounced, and ICL attains the best average result without any other model within two standard errors in three scenarios in terms of the C2ST metric. ICL is among those models not significantly worse than the best in four cases with respect to the C2ST metric, in six cases in terms of the MMD metric, and also in six cases in terms of $\mathcal{W}_2$.

In scenario 1, which is a linear regression scenario with a normal prior on the coefficients $\bm{\beta}$ and an inverse gamma prior on the variance $\sigma^2$, ICL and HMC show a similarly large agreement with the analytical solution. Furthermore, the VI approaches with an ordinary multivariate normal distribution, a structured normal distribution as well as the approach based on inverse autoregressive flows also show a large agreement with the analytical solution, which is to be expected since scenario 1 is has a conjugate prior structure yielding a multivariate t-distribution for the posterior of the coefficients \citep{murphy2023probabilistic}.

Scenario 2 and scenario 4 are those where an intercept is included in the generative structure of the GLM. The notably superior performance of the ICL approach in those two cases might be explained by its ability to model distributions with substantially different variances in different dimensions better than VI. Similarly, the posterior in scenario 5 is determined by the gamma prior on the coefficients leading to a (slightly) skewed posterior distribution, which might explain the good relative performance of ICL. See \Cref{fig:posteriormarginalgamma} for a plot of the marginals of the posterior in this scenario on the Miami housing 2016 dataset. 

Finally, scenarios 6 and 7 demonstrate the versatility of the ICL method in terms of posterior inference for logistic regression and regression with a gamma response.

\begin{table}[htp]
\centering
\caption{Generalized Linear Models: Evaluation on 50 synthetic and 17 real-world datasets for seven different scenarios.  All results within two standard errors of the best average result for each scenario are marked in \textbf{bold}.}
\label{tab:glm_res_detail}
\resizebox{1\textwidth}{!}{
\begin{tabular}{p{1.5cm} p{3.5cm} m{2.4cm}m{2.4cm}m{2.4cm}| m{2.4cm}m{2.4cm}m{2.4cm}}
\toprule
\multicolumn{1}{c}{\multirow[c]{2}{*}[-0.5ex]{\textbf{Scenario}}} & \multicolumn{1}{c}{\multirow{2}{*}[-0.5ex]{\textbf{Model}}} & \multicolumn{3}{c}{\textbf{Synthetic Evaluation}} & \multicolumn{3}{c}{\textbf{Real-World Evaluation}} \\
\cmidrule(lr){3-5} \cmidrule(lr){6-8}
& & C2ST ($\downarrow$) & MMD ($\downarrow$) & $\mathcal{W}_2$ ($\downarrow$) & C2ST ($\downarrow$) & MMD ($\downarrow$) & $\mathcal{W}_2$ ($\downarrow$) \\
\midrule
\multirow{7}{*}{Scenario 1} & Laplace Approximation & 1.000 ($\pm$ 0.000) & 2.738 $(\pm$ 0.721) & \textbf{0.825} ($\pm$ 0.279) & 1.000 ($\pm$ 0.000) & 2.150 $(\pm$ 0.323) & \textbf{0.642} ($\pm$ 0.124) \\
& VI: DiagonalNormal & 0.904 ($\pm$ 0.076) & 1.452 $(\pm$ 0.984) & \textbf{0.669} ($\pm$ 0.301) & 0.797 ($\pm$ 0.083) & 0.612 $(\pm$ 0.511) & \textbf{0.414} ($\pm$ 0.152) \\
& VI: MultivariateNormal & \textbf{0.750} ($\pm$ 0.128) & \textbf{0.735} $(\pm$ 0.733) & \textbf{0.565} ($\pm$ 0.292) & \textbf{0.607} ($\pm$ 0.070) & \textbf{0.167} $(\pm$ 0.196) & \textbf{0.301} ($\pm$ 0.123) \\
& VI: Structured Normal & \textbf{0.753} ($\pm$ 0.126) & \textbf{0.736} $(\pm$ 0.737) & \textbf{0.570} ($\pm$ 0.310) & \textbf{0.600} ($\pm$ 0.070) & \textbf{0.169} $(\pm$ 0.214) & \textbf{0.306} ($\pm$ 0.131) \\
& VI: IAF & \textbf{0.777} ($\pm$ 0.122) & \textbf{0.864} $(\pm$ 0.844) & 0.725 ($\pm$ 0.523) & 0.683 ($\pm$ 0.132) & 0.440 $(\pm$ 0.559) & 0.503 ($\pm$ 0.383) \\
& HMC & \textbf{0.745} ($\pm$ 0.130) & \textbf{0.722} $(\pm$ 0.732) & \textbf{0.569} ($\pm$ 0.301) & \textbf{0.595} ($\pm$ 0.075) & \textbf{0.173} $(\pm$ 0.213) & \textbf{0.321} ($\pm$ 0.140) \\
& \textbf{ICL (ours)} & \textbf{0.765} ($\pm$ 0.123) & \textbf{0.767} $(\pm$ 0.727) & \textbf{0.585} ($\pm$ 0.301) & \textbf{0.614} ($\pm$ 0.074) & \textbf{0.175} $(\pm$ 0.219) & \textbf{0.310} ($\pm$ 0.138) \\
\midrule
\multirow{6}{*}{Scenario 2} & Laplace Approximation & 1.000 ($\pm$ 0.000) & 4.853 $(\pm$ 2.333) & 5.770 ($\pm$ 5.946) & 1.000 ($\pm$ 0.000) & 2.572 $(\pm$ 0.206) & 0.809 ($\pm$ 0.149) \\
& VI: DiagonalNormal & 0.957 ($\pm$ 0.091) & 3.906 $(\pm$ 2.679) & 5.628 ($\pm$ 6.092) & 0.892 ($\pm$ 0.044) & 0.847 $(\pm$ 0.389) & \textbf{0.530} ($\pm$ 0.175) \\
& VI: MultivariateNormal & 0.910 ($\pm$ 0.131) & 3.407 $(\pm$ 2.781) & 5.584 ($\pm$ 6.104) & 0.820 ($\pm$ 0.031) & 0.243 $(\pm$ 0.148) & \textbf{0.408} ($\pm$ 0.118) \\
& VI: Structured Normal & 0.908 ($\pm$ 0.119) & 3.139 $(\pm$ 2.763) & 5.480 ($\pm$ 6.164) & 0.824 ($\pm$ 0.023) & 0.215 $(\pm$ 0.110) & \textbf{0.392} ($\pm$ 0.109) \\
& VI: IAF & 0.968 ($\pm$ 0.063) & 4.416 $(\pm$ 2.473) & 7.474 ($\pm$ 6.235) & 0.888 ($\pm$ 0.067) & 0.921 $(\pm$ 0.860) & 0.942 ($\pm$ 0.733) \\
& \textbf{ICL (ours)} & \textbf{0.839} ($\pm$ 0.072) & \textbf{0.707} $(\pm$ 0.658) & \textbf{1.111} ($\pm$ 0.300) & \textbf{0.768} ($\pm$ 0.033) & \textbf{0.143} $(\pm$ 0.089) & \textbf{0.411} ($\pm$ 0.094) \\
\midrule
\multirow{6}{*}{Scenario 3} & Laplace Approximation & 1.000 ($\pm$ 0.000) & 2.203 $(\pm$ 0.997) & 1.170 ($\pm$ 0.949) & 1.000 ($\pm$ 0.000) & 1.841 $(\pm$ 0.185) & 0.729 ($\pm$ 0.175) \\
& VI: DiagonalNormal & 0.866 ($\pm$ 0.101) & 1.069 $(\pm$ 1.150) & 0.846 ($\pm$ 0.747) & 0.797 ($\pm$ 0.083) & 0.526 $(\pm$ 0.361) & 0.480 ($\pm$ 0.207) \\
& VI: MultivariateNormal & 0.656 ($\pm$ 0.131) & 0.445 $(\pm$ 1.061) & 0.660 ($\pm$ 0.737) & \textbf{0.560} ($\pm$ 0.035) & \textbf{0.032} $(\pm$ 0.028) & \textbf{0.249} ($\pm$ 0.069) \\
& VI: Structured Normal & 0.653 ($\pm$ 0.125) & 0.421 $(\pm$ 0.993) & 0.659 ($\pm$ 0.736) & \textbf{0.552} ($\pm$ 0.028) & \textbf{0.027} $(\pm$ 0.015) & \textbf{0.239} ($\pm$ 0.055) \\
& VI: IAF & 0.751 ($\pm$ 0.148) & 0.939 $(\pm$ 1.349) & 0.964 ($\pm$ 0.924) & 0.673 ($\pm$ 0.141) & 0.399 $(\pm$ 0.543) & 0.563 ($\pm$ 0.433) \\
& \textbf{ICL (ours)} & \textbf{0.611} ($\pm$ 0.070) & \textbf{0.089} $(\pm$ 0.114) & \textbf{0.423} ($\pm$ 0.348) & 0.576 ($\pm$ 0.027) & \textbf{0.037} $(\pm$ 0.026) & \textbf{0.257} ($\pm$ 0.044) \\
\midrule
\multirow{6}{*}{Scenario 4} & Laplace Approximation & 1.000 ($\pm$ 0.000) & 3.511 $(\pm$ 2.025) & 2.166 ($\pm$ 1.722) & 1.000 ($\pm$ 0.000) & 2.011 $(\pm$ 0.058) & 0.993 ($\pm$ 0.144) \\
& VI: DiagonalNormal & 0.968 ($\pm$ 0.036) & 2.798 $(\pm$ 2.255) & 2.065 ($\pm$ 1.745) & 0.916 ($\pm$ 0.040) & 0.928 $(\pm$ 0.339) & 0.732 ($\pm$ 0.181) \\
& VI: MultivariateNormal & 0.855 ($\pm$ 0.123) & 1.648 $(\pm$ 2.052) & 1.853 ($\pm$ 1.745) & 0.771 ($\pm$ 0.017) & \textbf{0.087} $(\pm$ 0.030) & \textbf{0.539} ($\pm$ 0.070) \\
& VI: Structured Normal & 0.847 ($\pm$ 0.116) & 1.505 $(\pm$ 1.978) & 1.889 ($\pm$ 1.883) & 0.769 ($\pm$ 0.012) & \textbf{0.083} $(\pm$ 0.018) & \textbf{0.543} ($\pm$ 0.070) \\
& VI: IAF & 0.942 ($\pm$ 0.077) & 3.029 $(\pm$ 2.210) & 3.554 ($\pm$ 2.715) & 0.833 ($\pm$ 0.069) & 0.636 $(\pm$ 0.756) & 0.978 ($\pm$ 0.600) \\
& \textbf{ICL (ours)} & \textbf{0.753} ($\pm$ 0.049) & \textbf{0.171} $(\pm$ 0.153) & \textbf{0.631} ($\pm$ 0.294) & \textbf{0.762} ($\pm$ 0.015) & \textbf{0.105} $(\pm$ 0.046) & \textbf{0.597} ($\pm$ 0.104) \\
\midrule
\multirow{6}{*}{Scenario 5} & Laplace Approximation & 1.000 ($\pm$ 0.000) & 2.060 $(\pm$ 0.472) & 0.797 ($\pm$ 0.577) & 1.000 ($\pm$ 0.000) & 1.982 $(\pm$ 0.126) & 0.623 ($\pm$ 0.084) \\
& VI: DiagonalNormal & 0.866 ($\pm$ 0.085) & 0.954 $(\pm$ 1.022) & 0.651 ($\pm$ 0.549) & 0.810 ($\pm$ 0.036) & 0.441 $(\pm$ 0.252) & 0.384 ($\pm$ 0.089) \\
& VI: MultivariateNormal & 0.765 ($\pm$ 0.100) & 0.537 $(\pm$ 1.019) & 0.633 ($\pm$ 1.067) & 0.711 ($\pm$ 0.038) & 0.148 $(\pm$ 0.093) & \textbf{0.279} ($\pm$ 0.056) \\
& VI: Structured Normal & 0.758 ($\pm$ 0.098) & 0.447 $(\pm$ 0.818) & 0.572 ($\pm$ 0.816) & 0.705 ($\pm$ 0.032) & 0.140 $(\pm$ 0.081) & \textbf{0.269} ($\pm$ 0.045) \\
& VI: IAF & 0.814 ($\pm$ 0.105) & 0.953 $(\pm$ 1.165) & 0.881 ($\pm$ 1.067) & 0.777 ($\pm$ 0.106) & 0.684 $(\pm$ 0.939) & 0.625 ($\pm$ 0.525) \\
& \textbf{ICL (ours)} & \textbf{0.621} ($\pm$ 0.063) & \textbf{0.067} $(\pm$ 0.080) & \textbf{0.299} ($\pm$ 0.195) & \textbf{0.610} ($\pm$ 0.045) & \textbf{0.046} $(\pm$ 0.020) & \textbf{0.242} ($\pm$ 0.038) \\
\midrule
\multirow{6}{*}{Scenario 6} & Laplace Approximation & 1.000 ($\pm$ 0.000) & 2.026 $(\pm$ 0.027) & 1.612 ($\pm$ 0.162) & 1.000 ($\pm$ 0.000) & 1.993 $(\pm$ 0.032) & 1.299 ($\pm$ 0.106) \\
& VI: DiagonalNormal & 0.724 ($\pm$ 0.060) & 0.185 $(\pm$ 0.082) & \textbf{0.787} ($\pm$ 0.078) & 0.703 ($\pm$ 0.039) & 0.147 $(\pm$ 0.063) & 0.637 ($\pm$ 0.089) \\
& VI: MultivariateNormal & \textbf{0.534} ($\pm$ 0.018) & \textbf{0.014} $(\pm$ 0.006) & \textbf{0.581} ($\pm$ 0.074) & \textbf{0.538} ($\pm$ 0.019) & \textbf{0.016} $(\pm$ 0.007) & \textbf{0.466} ($\pm$ 0.029) \\
& VI: Structured Normal & \textbf{0.536} ($\pm$ 0.016) & \textbf{0.014} $(\pm$ 0.005) & \textbf{0.583} ($\pm$ 0.071) & \textbf{0.536} ($\pm$ 0.019) & \textbf{0.017} $(\pm$ 0.009) & \textbf{0.469} ($\pm$ 0.033) \\
& VI: IAF & 0.542 ($\pm$ 0.026) & 0.031 $(\pm$ 0.031) & 0.613 ($\pm$ 0.092) & \textbf{0.535} ($\pm$ 0.015) & \textbf{0.015} $(\pm$ 0.006) & \textbf{0.467} ($\pm$ 0.031) \\
& \textbf{ICL (ours)} & \textbf{0.532} ($\pm$ 0.019) & 0.016 $(\pm$ 0.008) & \textbf{0.590} ($\pm$ 0.066) & 0.556 ($\pm$ 0.017) & 0.035 $(\pm$ 0.015) & \textbf{0.504} ($\pm$ 0.038) \\

\midrule
\multirow{6}{*}{Scenario 7} & Laplace Approximation & 1.000 ($\pm$ 0.000) & 3.559 $(\pm$ 1.933) & 1.347 ($\pm$ 1.067) & 1.000 ($\pm$ 0.000) & 2.016 $(\pm$ 0.080) & 0.763 ($\pm$ 0.174) \\
& VI: DiagonalNormal & 0.938 ($\pm$ 0.074) & 2.536 $(\pm$ 2.097) & 1.142 ($\pm$ 0.993) & 0.936 ($\pm$ 0.024) & 1.029 $(\pm$ 0.255) & 0.579 ($\pm$ 0.181) \\
& VI: MultivariateNormal & 0.814 ($\pm$ 0.181) & 1.999 $(\pm$ 2.283) & 1.033 ($\pm$ 0.969) & \textbf{0.741} ($\pm$ 0.020) & 0.093 $(\pm$ 0.025) & \textbf{0.391} ($\pm$ 0.074) \\
& VI: Structured Normal & 0.824 ($\pm$ 0.177) & 1.891 $(\pm$ 2.127) & 1.041 ($\pm$ 0.934) & \textbf{0.734} ($\pm$ 0.025) & \textbf{0.072} $(\pm$ 0.019) & \textbf{0.385} ($\pm$ 0.065) \\
& VI: IAF & 0.939 ($\pm$ 0.091) & 2.707 $(\pm$ 1.712) & 1.590 ($\pm$ 0.820) & 0.864 ($\pm$ 0.093) & 0.830 $(\pm$ 0.697) & 1.064 ($\pm$ 0.616) \\
& \textbf{ICL (ours)} & \textbf{0.700} ($\pm$ 0.116) & \textbf{0.317} $(\pm$ 0.355) & \textbf{0.400} ($\pm$ 0.286) & 0.773 ($\pm$ 0.048) & \textbf{0.294} $(\pm$ 0.457) & 0.559 ($\pm$ 0.256) \\
\bottomrule
\end{tabular}

}
\end{table}

\clearpage
\subsection{Factor Analysis}
\label{app:addres_fa}

\Cref{tab:fa_res_detail} contains detailed results regarding FA for 50 synthetic and 17 real-world datasets across 6 different scenarios. We find that overall the ICL method has a very high agreement with the gold standard HMC reference with scores of more than than 56 percent in five scenarios on the synthetic data. In comparison, the C2ST metric is almost saturated for all considered VI methods. For MMD and $\mathcal{W}_2$ the ICL method is again the best. 

The real-world datasets show a similar picture except for scenario 4 where C2ST and MMD indicate that VI with inverse autoregressive flows performs best. The $\mathcal{W}_2$ metric, however exhibits a relatively large variance in those cases and does not yield significant results regarding the best performance. 

\begin{table}[htp]
\centering
\caption{Factor Analysis: Evaluation on 50 synthetic and 17 real-world datasets for six different scenarios. All results within two standard errors of the best average result for each scenario are marked in \textbf{bold}.}
\label{tab:fa_res_detail}
\resizebox{1\textwidth}{!}{
\begin{tabular}{p{1.5cm} p{3.5cm} m{2.4cm}m{2.4cm}m{2.4cm}| m{2.4cm}m{2.4cm}m{2.4cm}}
\toprule
\multicolumn{1}{c}{\multirow[c]{2}{*}[-0.5ex]{\textbf{Scenario}}} & \multicolumn{1}{c}{\multirow{2}{*}[-0.5ex]{\textbf{Model}}} & \multicolumn{3}{c}{\textbf{Synthetic Evaluation}} & \multicolumn{3}{c}{\textbf{Real-World Evaluation}} \\
\cmidrule(lr){3-5} \cmidrule(lr){6-8}
& & C2ST ($\downarrow$) & MMD ($\downarrow$) & $\mathcal{W}_2$ ($\downarrow$) & C2ST ($\downarrow$) & MMD ($\downarrow$) & $\mathcal{W}_2$ ($\downarrow$) \\
\midrule
\multirow{7}{*}{Scenario 1} & Laplace Approximation & 1.000 ($\pm$ 0.000) & 3.459 $(\pm$ 1.553) & 1.987 ($\pm$ 1.363) & 1.000 ($\pm$ 0.000) & 2.487 $(\pm$ 0.454) & \textbf{0.875} ($\pm$ 0.036) \\
& VI: DiagonalNormal & 1.000 ($\pm$ 0.001) & 4.695 $(\pm$ 1.488) & 2.865 ($\pm$ 1.681) & 0.979 ($\pm$ 0.008) & 1.283 $(\pm$ 0.225) & \textbf{0.625} ($\pm$ 0.058) \\
& VI: MultivariateNormal & 0.998 ($\pm$ 0.003) & 4.163 $(\pm$ 1.473) & 2.603 ($\pm$ 1.959) & 0.966 ($\pm$ 0.010) & 1.213 $(\pm$ 0.260) & \textbf{0.608} ($\pm$ 0.047) \\
& VI: Structured Normal & 0.997 ($\pm$ 0.004) & 4.655 $(\pm$ 1.189) & 2.700 ($\pm$ 1.333) & 0.979 ($\pm$ 0.010) & 1.231 $(\pm$ 0.132) & \textbf{0.611} ($\pm$ 0.041) \\
& VI: IAF & 0.953 ($\pm$ 0.104) & 3.992 $(\pm$ 2.089) & 2.750 ($\pm$ 1.838) & 0.849 ($\pm$ 0.075) & 0.772 $(\pm$ 0.335) & \textbf{0.503} ($\pm$ 0.063) \\
& \textbf{ICL (ours)} & \textbf{0.552} ($\pm$ 0.028) & \textbf{0.034} $(\pm$ 0.034) & \textbf{0.289} ($\pm$ 0.083) & \textbf{0.606} ($\pm$ 0.038) & \textbf{0.068} $(\pm$ 0.069) & 0.265 ($\pm$ 0.078) \\
\midrule
\multirow{7}{*}{Scenario 2} & Laplace Approximation & 1.000 ($\pm$ 0.000) & 3.687 $(\pm$ 1.661) & 1.954 ($\pm$ 1.129) & 1.000 ($\pm$ 0.000) & 1.690 $(\pm$ 0.182) & \textbf{0.598} ($\pm$ 0.058) \\
& VI: DiagonalNormal & 0.998 ($\pm$ 0.002) & 3.135 $(\pm$ 1.482) & 1.629 ($\pm$ 0.938) & 0.975 ($\pm$ 0.010) & 1.156 $(\pm$ 0.068) & \textbf{0.496} ($\pm$ 0.052) \\
& VI: MultivariateNormal & 0.989 ($\pm$ 0.009) & 2.945 $(\pm$ 1.019) & 1.482 ($\pm$ 0.683) & 0.951 ($\pm$ 0.025) & 0.764 $(\pm$ 0.053) & \textbf{0.421} ($\pm$ 0.052) \\
& VI: Structured Normal & 0.984 ($\pm$ 0.031) & 3.790 $(\pm$ 1.572) & 2.106 ($\pm$ 1.429) & 0.958 ($\pm$ 0.025) & 1.001 $(\pm$ 0.126) & \textbf{0.465} ($\pm$ 0.056) \\
& VI: IAF & 0.966 ($\pm$ 0.066) & 3.523 $(\pm$ 1.340) & 2.153 ($\pm$ 0.968) & 0.799 ($\pm$ 0.058) & 0.462 $(\pm$ 0.226) & \textbf{0.342} ($\pm$ 0.070) \\
& \textbf{ICL (ours)} & \textbf{0.542} ($\pm$ 0.006) & \textbf{0.017} $(\pm$ 0.006) & \textbf{0.244} ($\pm$ 0.033) & \textbf{0.622} ($\pm$ 0.032) & \textbf{0.098} $(\pm$ 0.039) & \textbf{0.287} ($\pm$ 0.046) \\
\midrule
\multirow{7}{*}{Scenario 3} & Laplace Approximation & 1.000 ($\pm$ 0.000) & 4.137 $(\pm$ 0.932) & 2.188 ($\pm$ 1.011) & 1.000 ($\pm$ 0.000) & 3.653 $(\pm$ 0.183) & \textbf{0.473} ($\pm$ 0.026) \\
&VI: DiagonalNormal & 0.999 ($\pm$ 0.002) & 3.339 $(\pm$ 0.985) & 1.722 ($\pm$ 0.870) & 0.951 ($\pm$ 0.007) & 1.114 $(\pm$ 0.080) & \textbf{0.245} ($\pm$ 0.016) \\
&VI: MultivariateNormal & 0.994 ($\pm$ 0.007) & 3.189 $(\pm$ 0.960) & 1.644 ($\pm$ 0.859) & 0.945 ($\pm$ 0.007) & 1.085 $(\pm$ 0.082) & \textbf{0.242} ($\pm$ 0.015) \\
&VI: Structured Normal & 0.997 ($\pm$ 0.003) & 3.159 $(\pm$ 0.968) & 1.614 ($\pm$ 0.793) & 0.942 ($\pm$ 0.009) & 1.084 $(\pm$ 0.071) & \textbf{0.242} ($\pm$ 0.018) \\
&VI: IAF & 0.990 ($\pm$ 0.011) & 3.145 $(\pm$ 1.203) & 1.705 ($\pm$ 0.990) & 0.928 ($\pm$ 0.015) & 1.022 $(\pm$ 0.093) & \textbf{0.235} ($\pm$ 0.018) \\
& \textbf{ICL (ours)} & \textbf{0.537} ($\pm$ 0.023) & \textbf{0.024} $(\pm$ 0.021) & \textbf{0.259} ($\pm$ 0.088) & \textbf{0.609} ($\pm$ 0.019) & \textbf{0.124} $(\pm$ 0.037) & \textbf{0.179} ($\pm$ 0.018) \\
\midrule
\multirow{7}{*}{Scenario 4} & Laplace Approximation & 1.000 ($\pm$ 0.000) & 4.354 $(\pm$ 0.572) & 3.339 ($\pm$ 0.932) & 1.000 ($\pm$ 0.000) & 6.617 $(\pm$ 0.259) & 0.598 ($\pm$ 0.135) \\

& VI: DiagonalNormal & 1.000 ($\pm$ 0.000) & 3.396 $(\pm$ 0.591) & 2.420 ($\pm$ 0.720) & 0.977 ($\pm$ 0.003) & 1.499 $(\pm$ 0.066) & \textbf{0.096} ($\pm$ 0.003) \\
& VI: MultivariateNormal & 0.999 ($\pm$ 0.001) & 3.447 $(\pm$ 0.567) & 2.479 ($\pm$ 0.848) & 0.973 ($\pm$ 0.008) & 1.484 $(\pm$ 0.097) & 0.096 ($\pm$ 0.005) \\
& VI: Structured Normal & 1.000 ($\pm$ 0.000) & 3.421 $(\pm$ 0.610) & 2.481 ($\pm$ 0.884) & 0.973 ($\pm$ 0.007) & 1.474 $(\pm$ 0.078) & \textbf{0.095} ($\pm$ 0.004) \\
& VI: IAF & 0.999 ($\pm$ 0.001) & 3.269 $(\pm$ 0.552) & 2.307 ($\pm$ 0.779) & \textbf{0.961} ($\pm$ 0.018) & \textbf{1.337} $(\pm$ 0.142) & \textbf{0.092} ($\pm$ 0.005) \\
& \textbf{ICL (ours)} & \textbf{0.684} ($\pm$ 0.060) & \textbf{0.198} $(\pm$ 0.141) & \textbf{0.918} ($\pm$ 0.246) & 0.988 ($\pm$ 0.003) & 1.764 $(\pm$ 0.026) & 1.248 ($\pm$ 0.008) \\
\midrule
\multirow{7}{*}{Scenario 5} & Laplace Approximation & 1.000 ($\pm$ 0.000) & 4.456 $(\pm$ 0.785) & 2.608 ($\pm$ 0.946) & 1.000 ($\pm$ 0.000) & 4.559 $(\pm$ 0.494) & 0.663 ($\pm$ 0.127) \\
& VI: DiagonalNormal & 0.999 ($\pm$ 0.002) & 3.520 $(\pm$ 1.073) & 2.012 ($\pm$ 0.886) & 0.944 ($\pm$ 0.010) & 1.007 $(\pm$ 0.129) & \textbf{0.261} ($\pm$ 0.036) \\
& VI: MultivariateNormal & 0.995 ($\pm$ 0.007) & 3.472 $(\pm$ 1.021) & 1.982 ($\pm$ 0.814) & 0.930 ($\pm$ 0.017) & 0.964 $(\pm$ 0.111) & \textbf{0.255} ($\pm$ 0.038) \\
& VI: Structured Normal & 0.998 ($\pm$ 0.005) & 3.369 $(\pm$ 1.044) & 1.916 ($\pm$ 0.852) & 0.934 ($\pm$ 0.011) & 0.996 $(\pm$ 0.133) & \textbf{0.259} ($\pm$ 0.035) \\
& VI: IAF & 0.992 ($\pm$ 0.012) & 3.166 $(\pm$ 0.967) & 1.761 ($\pm$ 0.671) & 0.910 ($\pm$ 0.011) & \textbf{0.892} $(\pm$ 0.094) & \textbf{0.247} ($\pm$ 0.037) \\
& \textbf{ICL (ours)} & \textbf{0.535} ($\pm$ 0.016) & \textbf{0.021} $(\pm$ 0.011) & \textbf{0.279} ($\pm$ 0.060) & \textbf{0.886} ($\pm$ 0.017) & 1.207 $(\pm$ 0.101) & \textbf{1.002} ($\pm$ 0.042) \\
\midrule
\multirow{7}{*}{Scenario 6} & Laplace Approximation & 1.000 ($\pm$ 0.000) & 3.942 $(\pm$ 0.971) & 2.624 ($\pm$ 1.682) & 1.000 ($\pm$ 0.000) & 3.319 $(\pm$ 0.196) & \textbf{0.377} ($\pm$ 0.020) \\
& VI: DiagonalNormal & 0.998 ($\pm$ 0.002) & 3.214 $(\pm$ 1.072) & 2.209 ($\pm$ 1.543) & 0.949 ($\pm$ 0.008) & 1.196 $(\pm$ 0.093) & \textbf{0.210} ($\pm$ 0.011) \\
& VI: MultivariateNormal & 0.991 ($\pm$ 0.013) & 3.056 $(\pm$ 1.237) & 2.189 ($\pm$ 1.698) & 0.938 ($\pm$ 0.009) & 1.121 $(\pm$ 0.075) & \textbf{0.205} ($\pm$ 0.012) \\
& VI: Structured Normal & 0.997 ($\pm$ 0.005) & 3.279 $(\pm$ 1.071) & 2.276 ($\pm$ 1.787) & 0.944 ($\pm$ 0.006) & 1.161 $(\pm$ 0.066) & \textbf{0.208} ($\pm$ 0.012) \\
& VI: IAF & 0.989 ($\pm$ 0.029) & 3.027 $(\pm$ 0.910) & 1.936 ($\pm$ 1.060) & 0.865 ($\pm$ 0.027) & 0.822 $(\pm$ 0.106) & \textbf{0.179} ($\pm$ 0.015) \\
& \textbf{ICL (ours)} & \textbf{0.543} ($\pm$ 0.021) & \textbf{0.023} $(\pm$ 0.015)& \textbf{0.345} ($\pm$ 0.173) & \textbf{0.666} ($\pm$ 0.020) & \textbf{0.200} $(\pm$ 0.034) & \textbf{0.224} ($\pm$ 0.014) \\
\bottomrule
\end{tabular}

}
\end{table}

\newpage

\subsection{Gaussian Mixture Models}

We summarize the results of the ICL approach and the different VI methods regarding the GMM scenarios in \Cref{tab:gmm_res_detail}. First, one can note that on the synthetic data, the ICL approach has a much lower C2ST score for scenario 1 and scenario 2 than the other methods. However, for scenarios 3 and 4, C2ST saturates, or at least almost saturates for all approaches. The MMD metric, however, shows that ICL not only has a high agreement with HMC in scenarios 1 and 2, but that it attains the significantly best result in scenarios 3 and 4 as well. This is supported by the $\mathcal{W}_2$ metric, which has the significantly lowest values for ICL in scenarios 2,3 and 4. 

Analogously, on the real-world data, MMD shows that ICL is the best approach in all four scenarios without any other model coming into the two standard-deviation range. While the C2ST score is the lowest in scenario 1 and scenario 2 for ICL, it saturates for cases 3 and 4.

\begin{table}[htp]
\centering
\caption{Gaussian Mixture Models: Evaluation on 50 synthetic and 17 real-world datasets for six different scenarios. All results within two standard errors of the best average result for each scenario are marked in \textbf{bold}.}
\label{tab:gmm_res_detail}
\resizebox{1\textwidth}{!}{
\begin{tabular}{p{1.5cm} p{3.5cm} m{2.4cm}m{2.5cm}m{2.7cm}| m{2.4cm}m{2.4cm}m{2.4cm}}
\toprule
\multicolumn{1}{c}{\multirow[c]{2}{*}[-0.5ex]{\textbf{Scenario}}} & \multicolumn{1}{c}{\multirow{2}{*}[-0.5ex]{\textbf{Model}}} & \multicolumn{3}{c}{\textbf{Synthetic Evaluation}} & \multicolumn{3}{c}{\textbf{Real-World Evaluation}} \\
\cmidrule(lr){3-5} \cmidrule(lr){6-8}
& & C2ST ($\downarrow$) & MMD ($\downarrow$) & $\mathcal{W}_2$ ($\downarrow$) & C2ST ($\downarrow$) & MMD ($\downarrow$) & $\mathcal{W}_2$ ($\downarrow$) \\
\midrule
\multirow{7}{*}{Scenario 1} & Laplace Approximation & 1.000 ($\pm$ 0.000) & 3.367 $(\pm$ 1.030) & \textbf{4.341} ($\pm$ 2.018) & 1.000 ($\pm$ 0.000) & 3.374 $(\pm$ 0.941) & \textbf{6.440} ($\pm$ 1.994) \\
& VI: DiagonalNormal & 0.988 ($\pm$ 0.013) & 1.175 $(\pm$ 1.189) & \textbf{2.961} ($\pm$ 1.669) & 0.995 ($\pm$ 0.006) & 1.919 $(\pm$ 1.217) & \textbf{5.145} ($\pm$ 2.489) \\
& VI: MultivariateNormal & 0.988 ($\pm$ 0.013) & 1.135 $(\pm$ 1.149) & \textbf{2.926} ($\pm$ 1.651) & 0.994 ($\pm$ 0.007) & 2.007 $(\pm$ 1.367) & \textbf{5.379} ($\pm$ 2.845)\\
& VI: Structured Normal & 0.987 ($\pm$ 0.015) & 1.126 $(\pm$ 1.145) & \textbf{2.944} ($\pm$ 1.663) & 0.993 ($\pm$ 0.009) & 1.943 $(\pm$ 1.359) & \textbf{5.313} ($\pm$ 2.737) \\
& VI: IAF & 0.989 ($\pm$ 0.013) & 1.017 $(\pm$ 1.036) & \textbf{3.104} ($\pm$ 1.523) & 0.995 ($\pm$ 0.010) & 1.888 $(\pm$ 1.051) & \textbf{5.402} ($\pm$ 2.310) \\
& \textbf{ICL (ours)} & \textbf{0.760 ($\pm$ 0.092)} & \textbf{0.303} $(\pm$ 0.548) & \textbf{2.095} ($\pm$ 1.692) & \textbf{0.847} ($\pm$ 0.082) & \textbf{0.486} $(\pm$ 0.623) & \textbf{4.054} ($\pm$ 2.782) \\

\midrule
\multirow{7}{*}{Scenario 2} & Laplace Approximation & 1.000 ($\pm$ 0.000) & 2.864 $(\pm$ 0.607) & 5.407 ($\pm$ 2.320) & 1.000 ($\pm$ 0.000) & 2.928 $(\pm$ 0.438) & \textbf{7.228} ($\pm$ 1.323) \\
& VI: DiagonalNormal & 0.989 ($\pm$ 0.024) & 1.425 $(\pm$ 0.829) & 4.933 ($\pm$ 2.379) & 0.998 ($\pm$ 0.003) & 1.525 $(\pm$ 0.356) & \textbf{6.091} ($\pm$ 0.931) \\
& VI: MultivariateNormal & 0.991 ($\pm$ 0.021) & 1.532 $(\pm$ 0.940) & 5.119 ($\pm$ 2.521) & 0.999 ($\pm$ 0.002) & 1.619 $(\pm$ 0.269) & \textbf{6.258} ($\pm$ 0.872) \\
& VI: Structured Normal & 0.992 ($\pm$ 0.017) & 1.487 $(\pm$ 0.899) & 5.085 ($\pm$ 2.530) & 0.999 ($\pm$ 0.002) & 1.580 $(\pm$ 0.337) & \textbf{6.241} ($\pm$ 0.960) \\
& VI: IAF & 0.992 ($\pm$ 0.021) & 1.319 $(\pm$ 0.854) & 5.265 ($\pm$ 2.534) & 0.998 ($\pm$ 0.004) & 1.256 $(\pm$ 0.320) & \textbf{6.201} ($\pm$ 0.892) \\
& \textbf{ICL (ours)} & \textbf{0.812 ($\pm$ 0.061)} & \textbf{0.159} $(\pm$ 0.154) & \textbf{2.314} ($\pm$ 0.926) & \textbf{0.937} ($\pm$ 0.041) & \textbf{0.282} $(\pm$ 0.131) & \textbf{3.947} ($\pm$ 1.055) \\
\midrule
\multirow{7}{*}{Scenario 3} & Laplace Approximation & 1.000 ($\pm$ 0.000) & 3.631 $(\pm$ 1.362) & 16.387 ($\pm$ 19.604) & 1.000 ($\pm$ 0.000) & 3.009 $(\pm$ 0.768) & \textbf{37.034} ($\pm$ 7.178) \\
& VI: DiagonalNormal & \textbf{0.996} ($\pm$ 0.011) & 2.127 $(\pm$ 1.479) & 16.864 ($\pm$ 19.301) & \textbf{0.992} ($\pm$ 0.018) & 2.429 $(\pm$ 0.516) & \textbf{35.355} ($\pm$ 6.608) \\
& VI: MultivariateNormal & 0.997 ($\pm$ 0.009) & 2.076 $(\pm$ 1.388) & 16.938 ($\pm$ 19.636) & \textbf{0.993} ($\pm$ 0.016) & 2.427 $(\pm$ 0.510) & \textbf{35.312} ($\pm$ 6.655) \\
& VI: Structured Normal & \textbf{0.995} ($\pm$ 0.017) & 2.049 $(\pm$ 1.462) & 16.723 ($\pm$ 19.093) & \textbf{0.993} ($\pm$ 0.016) & 2.301 $(\pm$ 0.549) & \textbf{34.217} ($\pm$ 5.461) \\
& VI: IAF & \textbf{0.994} ($\pm$ 0.018) & 1.675 $(\pm$ 1.049) & 14.311 ($\pm$ 9.266) & \textbf{0.993} ($\pm$ 0.017) & 2.148 $(\pm$ 0.528) & \textbf{34.336} ($\pm$ 5.398) \\
& \textbf{ICL (ours)} & 1.000 ($\pm$ 0.000) & \textbf{0.582} $(\pm$ 0.280) & \textbf{8.708} ($\pm$ 4.945) & 1.000 ($\pm$ 0.000) & \textbf{1.869} $(\pm$ 0.342) & \textbf{33.230} ($\pm$ 8.095) \\
\midrule
\multirow{7}{*}{Scenario 4} & Laplace Approximation & 1.000 ($\pm$ 0.000) & 6.260 $(\pm$ 1.427) & 13.497 ($\pm$ 29.702) & 1.000 ($\pm$ 0.000) & 5.924 $(\pm$ 1.145) & \textbf{12.400} ($\pm$ 4.313) \\
& VI: DiagonalNormal & \textbf{1.000} ($\pm$ 0.002) & 3.958 $(\pm$ 1.641) & 12.068 ($\pm$ 21.301) & 1.000 ($\pm$ 0.000) & 3.879 $(\pm$ 1.061) & \textbf{11.080} ($\pm$ 3.341)\\
& VI: MultivariateNormal & \textbf{1.000} ($\pm$ 0.002) & 3.875 $(\pm$ 1.691) & 12.150 ($\pm$ 22.198) & 1.000 ($\pm$ 0.000) & 3.896 $(\pm$ 1.057) & \textbf{11.112} ($\pm$ 3.321)\\
& VI: Structured Normal & \textbf{1.000} ($\pm$ 0.001) & 3.661 $(\pm$ 1.717) & 12.195 ($\pm$ 22.874) & \textbf{0.996} ($\pm$ 0.016) & 3.822 $(\pm$ 1.302) & \textbf{11.368} ($\pm$ 4.216) \\
& VI: IAF & \textbf{1.000} ($\pm$ 0.002) & 3.536 $(\pm$ 1.597) & 12.015 ($\pm$ 20.884) & 1.000 ($\pm$ 0.000) & 3.471 $(\pm$ 1.036) & \textbf{11.421} ($\pm$ 3.233) \\
& \textbf{ICL (ours)} & 1.000 ($\pm$ 0.000) & \textbf{2.451} $(\pm$ 0.868) & \textbf{8.333} ($\pm$ 4.202) & \textbf{1.000} ($\pm$ 0.000) & \textbf{2.518 $(\pm$ 0.694)} & \textbf{11.938} ($\pm$ 2.956) \\

\bottomrule
\end{tabular}
}
\end{table}

\label{app:addres_gmm}

\clearpage

\section{Evaluating Predictive Performance}
\label{app:eval_predictions}

In this section, we discuss the results of our ICL approach in terms of predictive performance. In scenarios one to four, TabPFN gives the best overall performance, which is expected since it is not limited to the GLM structure. Besides that, the MAP approach obtains consistently the best results, while our ICL method performs on par (scenarios 1,2,3) or better than (scenario 4) compared to the fully Bayesian methods on the real-world data. On the synthetic data there is no significant difference to the other fully Bayesian methods, except for the real-world data in scenario where HMC is clearly the best method. In scenario 5 (gamma prior on the regression coefficients), the in-context learner performs significantly worse than all other methods, while this difference is less pronounced in scenario 7. In scenario 6, TabPFN also has a substantially better performance than all other methods. The MAP approach performs on average better than all fully Bayesian methods, which themselves do not differ significantly.

\begin{table}[htp]
\centering
\caption{Evaluating the predictive performance across 50 synthetic and 17 real-world datasets for scenarios 1-4 in terms of Root Mean Squared Error (RMSE). The best result among all fully Bayesian methods is marked in \textbf{bold}. For the fully Bayesian approaches, we use the posterior mean as a point estimate for the response. MAP denotes the predictive performance of the model with the maximum a posteriori estimate for the latents.}
\label{tab:tabular_results}
\resizebox{0.8\textwidth}{!}{
\begin{tabular}{p{2.5cm} p{4.5cm} m{3.4cm}m{3.4cm}}
\toprule
\textbf{Scenario} & \textbf{Model} & \textbf{RMSE Real-World} ($\downarrow$) & \textbf{RMSE Synthetic}  ($\downarrow$) \\
\midrule

% Scenario 1
\multirow{9}{*}{Scenario 1} 
& {HMC} & \textbf{0.591} ($\pm$ 0.023) & \textbf{0.510} ($\pm$ 0.040) \\
& {Laplace Approximation} & \textbf{0.594} ($\pm$ 0.023) & \textbf{0.510} ($\pm$ 0.040) \\
& {VI: DiagonalNormal} & \textbf{0.591} ($\pm$ 0.023) & \textbf{0.509} ($\pm$ 0.040) \\
& {VI: MultivariateNormal} & \textbf{0.591} ($\pm$ 0.023) & \textbf{0.510} ($\pm$ 0.040) \\
& {VI: Structured Normal} & \textbf{0.629} ($\pm$ 0.017) & 0.555 ($\pm$ 0.039) \\
& {VI: IAF} & \textbf{0.593} ($\pm$ 0.023) & \textbf{0.510} ($\pm$ 0.040) \\
& {ICL (ours)} & \textbf{0.593} ($\pm$ 0.020) & \textbf{0.524} ($\pm$ 0.038) \\
\cmidrule{2-4}
& MAP & 0.555 ($\pm$ 0.024) & 0.491 ($\pm$ 0.038) \\
%& Linear Model & 0.543 ($\pm$ 0.024) & 0.466 ($\pm$ 0.036) \\
%& Random Forest & 0.232 ($\pm$ 0.011) & 0.210 ($\pm$ 0.016) \\
& TabPFN & 0.483 ($\pm$ 0.036) & 0.453 ($\pm$ 0.036) \\

\midrule

\multirow{9}{*}{Scenario 2}
& {HMC} & \textbf{0.559} ($\pm$ 0.023) & \textbf{0.556} ($\pm$ 0.049) \\
& {Laplace Approximation} & \textbf{0.561} ($\pm$ 0.022) & \textbf{0.557} ($\pm$ 0.049) \\
& {VI: DiagonalNormal} & \textbf{0.560} ($\pm$ 0.023) & \textbf{0.557} ($\pm$ 0.049) \\
& {VI: MultivariateNormal} & \textbf{0.559} ($\pm$ 0.023) & \textbf{0.556} ($\pm$ 0.049) \\
& {VI: Structured Normal} & \textbf{0.604} ($\pm$ 0.016) & 0.685 ($\pm$ 0.054) \\
& {VI: IAF} & \textbf{0.563} ($\pm$ 0.023) & \textbf{0.557} ($\pm$ 0.049) \\
& {ICL (ours)} & \textbf{0.561} ($\pm$ 0.019) & \textbf{0.653} ($\pm$ 0.049) \\
\cmidrule{2-4}
& MAP & 0.513 ($\pm$ 0.023) & 0.522 ($\pm$ 0.048) \\
%& Linear Model & 0.510 ($\pm$ 0.023) & 0.508 ($\pm$ 0.046) \\
%& Random Forest & 0.219 ($\pm$ 0.011) & 0.224 ($\pm$ 0.019) \\
& TabPFN & 0.449 ($\pm$ 0.034) & 0.498 ($\pm$ 0.047) \\

\midrule
% Scenario 3
\multirow{9}{*}{Scenario 3}
& {HMC} & \textbf{0.684} ($\pm$ 0.027) & \textbf{0.512} ($\pm$ 0.040) \\
& {Laplace Approximation} & \textbf{0.688} ($\pm$ 0.026) & 0.516 ($\pm$ 0.040) \\
& {VI: DiagonalNormal} & \textbf{0.686} ($\pm$ 0.027) & \textbf{0.513} ($\pm$ 0.040) \\
& {VI: MultivariateNormal} & \textbf{0.685} ($\pm$ 0.027) & \textbf{0.512} ($\pm$ 0.040) \\
& {VI: Structured Normal} & \textbf{0.733} ($\pm$ 0.016) & 0.607 ($\pm$ 0.043) \\
& {VI: IAF} & \textbf{0.686} ($\pm$ 0.027) & \textbf{0.512} ($\pm$ 0.040) \\
& {ICL (ours)} & \textbf{0.690} ($\pm$ 0.023) & \textbf{0.588} ($\pm$ 0.045) \\
\cmidrule{2-4}
& MAP & 0.646 ($\pm$ 0.028) & 0.495 ($\pm$ 0.039) \\
%& Linear Model & 0.630 ($\pm$ 0.028) & 0.468 ($\pm$ 0.037) \\
%& Random Forest & 0.269 ($\pm$ 0.013) & 0.214 ($\pm$ 0.016) \\
& TabPFN & 0.556 ($\pm$ 0.041) & 0.462 ($\pm$ 0.037) \\

\midrule
% Scenario 4
\multirow{9}{*}{Scenario 4}
& {HMC} & \textbf{0.642} ($\pm$ 0.027) & \textbf{0.559} ($\pm$ 0.051) \\
& Laplace Approximation & 0.737 ($\pm$ 0.048) & 2.457 ($\pm$ 0.493) \\
& VI: DiagonalNormal & 0.751 ($\pm$ 0.038) & 2.046 ($\pm$ 0.399) \\
& {VI: MultivariateNormal} & \textbf{0.690} ($\pm$ 0.037) & 2.155 ($\pm$ 0.454) \\
& {VI: Structured Normal} & \textbf{0.686} ($\pm$ 0.015) & 3.019 ($\pm$ 0.545) \\
& {VI: IAF} & \textbf{0.643} ($\pm$ 0.027) & 1.751 ($\pm$ 0.422) \\
& {ICL (ours)} & \textbf{0.649} ($\pm$ 0.023) & 1.464 ($\pm$ 0.151) \\
\cmidrule{2-4}
& MAP & 0.626 ($\pm$ 0.038) & 2.377 ($\pm$ 0.529) \\
%& Linear Model & 0.588 ($\pm$ 0.026) & 0.506 ($\pm$ 0.046) \\
%& Random Forest & 0.244 ($\pm$ 0.012) & 0.235 ($\pm$ 0.019) \\
& TabPFN & 0.522 ($\pm$ 0.037) & 0.496 ($\pm$ 0.047) \\

\bottomrule
\end{tabular}
}
\end{table}

\begin{table}[h]
\centering
\caption{Evaluating the predictive performance across 50 synthetic and 17 real-world datasets for scenarios 5 and 7 in terms of Root Mean Squared Error (RMSE). The best result among all fully Bayesian methods is marked in \textbf{bold}. For the fully Bayesian approaches, we use the posterior mean as a point estimate for the response. MAP denotes the predictive performance of the model with the maximum a posteriori estimate for the latents.}
\label{tab:scenario5_7_results}
\resizebox{0.8\textwidth}{!}{
\begin{tabular}{p{2.5cm} p{4.5cm} m{3.4cm}m{3.4cm}}
\toprule
\textbf{Scenario} & \textbf{Model} & \textbf{RMSE Real-World} ($\downarrow$) & \textbf{RMSE Synthetic}  ($\downarrow$) \\
\midrule

% Scenario 5
\multirow{9}{*}{Scenario 5}
& {HMC} & \textbf{0.699} ($\pm$ 0.022) & \textbf{0.490} ($\pm$ 0.036) \\
& {Laplace Approximation} & \textbf{0.699} ($\pm$ 0.022) & \textbf{0.491} ($\pm$ 0.036) \\
& {VI: DiagonalNormal} & \textbf{0.702} ($\pm$ 0.022) & \textbf{0.491} ($\pm$ 0.036) \\
& {VI: MultivariateNormal} & \textbf{0.698} ($\pm$ 0.021) & \textbf{0.491} ($\pm$ 0.036) \\
& VI: Structured Normal & 1.507 ($\pm$ 0.089) & 0.741 ($\pm$ 0.053) \\
& {VI: IAF} & \textbf{0.699} ($\pm$ 0.022) & \textbf{0.490} ($\pm$ 0.036) \\
& {ICL (ours)} & 0.769 ($\pm$ 0.020) & 0.701 ($\pm$ 0.049) \\
\cmidrule{2-4}
& MAP & 0.658 ($\pm$ 0.022) & 0.471 ($\pm$ 0.035) \\
%& Linear Model & 0.604 ($\pm$ 0.027) & 0.451 ($\pm$ 0.034) \\
%& Random Forest & 0.256 ($\pm$ 0.012) & 0.207 ($\pm$ 0.014) \\
& TabPFN & 0.534 ($\pm$ 0.040) & 0.442 ($\pm$ 0.035) \\

\midrule

% Scenario 7
\multirow{9}{*}{Scenario 7}
& {HMC} & \textbf{0.953} ($\pm$ 0.015) & \textbf{0.719} ($\pm$ 0.041) \\
& {Laplace Approximation} & \textbf{0.950} ($\pm$ 0.016) & \textbf{0.719} ($\pm$ 0.041) \\
& {VI: DiagonalNormal} & \textbf{0.954} ($\pm$ 0.015) & \textbf{0.718} ($\pm$ 0.041) \\
& {VI: MultivariateNormal} & \textbf{0.953} ($\pm$ 0.015) & \textbf{0.718} ($\pm$ 0.041) \\
& VI: Structured Normal & 1.082 ($\pm$ 0.026) & 1.028 ($\pm$ 0.118) \\
& {VI: IAF} & \textbf{0.954} ($\pm$ 0.014) & \textbf{0.720} ($\pm$ 0.041) \\
& {ICL (ours)} & 1.019 ($\pm$ 0.017) & \textbf{0.765} ($\pm$ 0.041) \\
\cmidrule{2-4}
& MAP & 0.945 ($\pm$ 0.017) & 0.686 ($\pm$ 0.048) \\
%& Linear Model & 0.874 ($\pm$ 0.018) & 0.695 ($\pm$ 0.045) \\
%& Random Forest & 0.390 ($\pm$ 0.015) & 0.320 ($\pm$ 0.023) \\
& TabPFN & 0.817 ($\pm$ 0.040) & 0.654 ($\pm$ 0.039) \\

\bottomrule
\end{tabular}
}
\end{table}

\begin{table}[h]
\centering
\caption{Evaluating the predictive performance across 50 synthetic and 17 real-world datasets for scenarios 5 and 7 in terms of accuracy (Acc.). The best result among all fully Bayesian methods is marked in \textbf{bold}. For the fully Bayesian approaches, we use the posterior mean as a point estimate for the response. MAP denotes the predictive performance of the model with the maximum a posteriori estimate for the latents. }
\label{tab:scenario6_results}
\resizebox{0.8\textwidth}{!}{
\begin{tabular}{p{2.5cm} p{4.5cm} m{3.4cm}m{3.4cm}}
\toprule
\textbf{Scenario} & \textbf{Model} & \textbf{Acc. Real-World} ($\uparrow$) & \textbf{Acc. Synthetic}  ($\uparrow$) \\
\midrule
\multirow{9}{*}{Scenario 6}
& {HMC} & \textbf{0.694} ($\pm$ 0.028) & \textbf{0.546} ($\pm$ 0.015) \\
& {Laplace Approximation} & \textbf{0.692} ($\pm$ 0.027) & \textbf{0.547} ($\pm$ 0.015) \\
& {VI: DiagonalNormal} & \textbf{0.700} ($\pm$ 0.028) & \textbf{0.546} ($\pm$ 0.015) \\
& {VI: MultivariateNormal} & \textbf{0.691} ($\pm$ 0.029) & \textbf{0.546} ($\pm$ 0.015) \\
& {VI: Structured Normal} & \textbf{0.686} ($\pm$ 0.028) & \textbf{0.546} ($\pm$ 0.015) \\
& VI: IAF & \textbf{0.689} ($\pm$ 0.029) & \textbf{0.545} ($\pm$ 0.015) \\
& ICL (ours) & \textbf{ 0.688} ($\pm$ 0.027) & \textbf{0.545} ($\pm$ 0.015) \\
\cmidrule{2-4}
& MAP & 0.723 ($\pm$ 0.025) & 0.610 ($\pm$ 0.016) \\
%& Linear Model & 0.728 ($\pm$ 0.026) & 0.600 ($\pm$ 0.011) \\
%& Random Forest & 1.000 ($\pm$ 0.000) & 1.000 ($\pm$ 0.000) \\
& TabPFN & 0.862 ($\pm$ 0.021) & 0.673 ($\pm$ 0.011) \\

\bottomrule
\end{tabular}
}
\end{table}

\clearpage

\section{Ablation: Using a Gaussian Approximation}
\label{sec:Gaussian_Approximation}

In this section, we present results on using a Gaussian approximation instead of Flow Matching to parameterize the approximation of the posterior.

The key takeaway from these results is that the in-context learning approach performs substantially better with Flow Matching \citep{lipman2022flow} than when a Gaussian approximation of the posterior is employed. 

\begin{table}[htp]
\centering
\caption{Generalized Linear Models: Comparing the in-context learner with a Gaussian approximation, fitted via the forward KL-divergence, to the proposed flow matching method. Evaluation on 50 synthetic and 17 real-world datasets for seven different scenarios. If one method is by more than two standard errors better than the other, it is marked in \textbf{bold}. Overall, the ICL + Flow Matching method clearly outperforms the Gaussian approximation, fitted via the forward KL-divergence,: it yields significantly better results (according to the two-standard-error criterion) in 6 out of 7 scenarios on synthetic datasets and in all 7 scenarios on real-world datasets, across at least two of the three considered metrics (C2ST, MMD, or $\mathcal{W}_2$). In addition, the flow matching method consistently achieves lower or comparable standard errors, indicating more stable and reliable performance across datasets.}

\label{tab:glm_res_detail}
\resizebox{1\textwidth}{!}{
\begin{tabular}{p{1.5cm} p{3.5cm} m{2.4cm}m{2.4cm}m{2.4cm}| m{2.4cm}m{2.4cm}m{2.4cm}}
\toprule
\multicolumn{1}{c}{\multirow[c]{2}{*}[-0.5ex]{\textbf{Scenario}}} & \multicolumn{1}{c}{\multirow{2}{*}[-0.5ex]{\textbf{Model}}} & \multicolumn{3}{c}{\textbf{Synthetic Evaluation}} & \multicolumn{3}{c}{\textbf{Real-World Evaluation}} \\
\cmidrule(lr){3-5} \cmidrule(lr){6-8}
& & C2ST ($\downarrow$) & MMD ($\downarrow$) & $\mathcal{W}_2$ ($\downarrow$) & C2ST ($\downarrow$) & MMD ($\downarrow$) & $\mathcal{W}_2$ ($\downarrow$) \\
\midrule

\multirow{2}{*}{Scenario 1} 
& ICL + Gaussian & \textbf{0.845} ($\pm$ 0.213) & \textbf{1.601} $(\pm$ 1.213) & 2.024 ($\pm$ 0.874) & 0.980 ($\pm$ 0.007) & 1.715 $(\pm$ 0.295) & 1.976 ($\pm$ 0.238) \\
& \textbf{ICL + Flow Matching} & \textbf{0.765} ($\pm$ 0.123) & \textbf{0.767} $(\pm$ 0.727) & \textbf{0.585} ($\pm$ 0.301) & \textbf{0.614} ($\pm$ 0.074) & \textbf{0.175} $(\pm$ 0.219) & \textbf{0.310} ($\pm$ 0.138) \\

\midrule
\multirow{2}{*}{Scenario 2} 
& ICL + Gaussian & \textbf{0.941} ($\pm$ 0.056) & \textbf{1.000} $(\pm$ 0.953) & 1.943 ($\pm$ 0.657) & 0.969 ($\pm$ 0.013) & 1.490 $(\pm$ 0.310) & 2.068 ($\pm$ 0.259) \\
& \textbf{ICL + Flow Matching} & \textbf{0.839} ($\pm$ 0.072) & \textbf{0.707} $(\pm$ 0.658) & \textbf{1.111} ($\pm$ 0.300) & \textbf{0.768} ($\pm$ 0.033) & \textbf{0.143} $(\pm$ 0.089) & \textbf{0.411} ($\pm$ 0.094) \\

\midrule
\multirow{2}{*}{Scenario 3} 
& ICL + Gaussian & 0.907 ($\pm$ 0.138) & 1.779 $(\pm$ 1.363) & 4.713 ($\pm$ 1.560) & 0.985 ($\pm$ 0.006) & 1.526 $(\pm$ 0.198) & 4.144 ($\pm$ 0.438) \\
& \textbf{ICL + Flow Matching} & \textbf{0.611} ($\pm$ 0.070) & \textbf{0.089} $(\pm$ 0.114) & \textbf{0.423} ($\pm$ 0.348) & \textbf{0.576} ($\pm$ 0.027) & \textbf{0.037} $(\pm$ 0.026) & \textbf{0.257} ($\pm$ 0.044) \\

\midrule
\multirow{2}{*}{Scenario 4} 
& ICL + Gaussian & 0.989 ($\pm$ 0.011) & 3.544 $(\pm$ 0.343) & 23.035 ($\pm$ 6.549) & 0.990 ($\pm$ 0.003) & 3.858 $(\pm$ 0.061) & 13.601 ($\pm$ 0.427)\\
& \textbf{ICL + Flow Matching} & \textbf{0.753} ($\pm$ 0.049) & \textbf{0.171} $(\pm$ 0.153) &\textbf{ 0.631} ($\pm$ 0.294) &\textbf{0.762} ($\pm$ 0.015) & \textbf{0.105} $(\pm$ 0.046) & \textbf{0.597} ($\pm$ 0.104) \\
\midrule
\multirow{2}{*}{Scenario 5} 
& ICL + Gaussian & 0.962 ($\pm$ 0.037) & 1.444 $(\pm$ 1.640) & 3.299 ($\pm$ 1.614) & 0.991 ($\pm$ 0.005) & 1.666 $(\pm$ 0.387) & 2.963 ($\pm$ 0.239) \\
& \textbf{ICL + Flow Matching} & \textbf{0.621} ($\pm$ 0.063) & \textbf{0.067} $(\pm$ 0.080) & \textbf{0.299} ($\pm$ 0.195) & \textbf{0.610} ($\pm$ 0.045) & \textbf{0.046} $(\pm$ 0.020) & \textbf{0.242} ($\pm$ 0.038) \\

\midrule
\multirow{2}{*}{Scenario 6} 
& ICL + Gaussian & 0.909 ($\pm$ 0.048) & 1.020 $(\pm$ 0.505) & 1.515 ($\pm$ 0.358) & 0.939 ($\pm$ 0.047) & 1.799 $(\pm$ 0.751) & 1.904 ($\pm$ 0.541) \\
& \textbf{ICL + Flow Matching} & \textbf{0.532} ($\pm$ 0.019) & \textbf{0.016} $(\pm$ 0.008) & \textbf{0.590} ($\pm$ 0.066) & \textbf{0.556} ($\pm$ 0.017) & \textbf{0.035} $(\pm$ 0.015) & \textbf{0.504} ($\pm$ 0.038) \\

\midrule
\multirow{2}{*}{Scenario 7} 
& ICL + Gaussian & 0.970 ($\pm$ 0.030) & 2.169 $(\pm$ 1.473) & 1.707 ($\pm$ 0.480) & 0.993 ($\pm$ 0.006) & 2.390 $(\pm$ 0.414) & 1.362 ($\pm$ 0.152) \\
& \textbf{ICL + Flow Matching} & \textbf{0.700} ($\pm$ 0.116) & \textbf{0.317} $(\pm$ 0.355) & \textbf{0.400} ($\pm$ 0.286) & \textbf{0.773} ($\pm$ 0.048) & \textbf{0.294} $(\pm$ 0.457) & \textbf{0.559} ($\pm$ 0.256) \\

\bottomrule
\end{tabular}
}
\end{table}

\begin{table}[htp]
\centering
\caption{Factor Analysis: Comparing the in-context learner with a Gaussian approximation, fitted via the forward KL-divergence, to the proposed flow matching method. Evaluation on 50 synthetic and 17 real-world datasets for seven different scenarios. If one method is by more than two standard errors better than the other, it is marked in \textbf{bold}. The flow matching approach shows favorable performance in the majority of cases. Specifically, it achieves statistically significant improvements in all 6 scenarios on synthetic data and in 5 out of 6 scenarios on real-world data. Notably, it often reduces discrepancy measures such as MMD and Wasserstein-2 distance by a large margin. In addition, the variability of the flow matching estimates is generally lower, leading to more reliable and consistent results across different datasets. In scenario 4, the Gaussian in-context learner learned a singular covariance matrix.}

\label{tab:fa_res_detail}
\resizebox{1\textwidth}{!}{
\begin{tabular}{p{1.5cm} p{3.5cm} m{2.4cm}m{2.4cm}m{2.4cm}| m{2.4cm}m{2.4cm}m{2.4cm}}
\toprule
\multicolumn{1}{c}{\multirow[c]{2}{*}[-0.5ex]{\textbf{Scenario}}} & \multicolumn{1}{c}{\multirow{2}{*}[-0.5ex]{\textbf{Model}}} & \multicolumn{3}{c}{\textbf{Synthetic Evaluation}} & \multicolumn{3}{c}{\textbf{Real-World Evaluation}} \\
\cmidrule(lr){3-5} \cmidrule(lr){6-8}
& & C2ST ($\downarrow$) & MMD ($\downarrow$) & $\mathcal{W}_2$ ($\downarrow$) & C2ST ($\downarrow$) & MMD ($\downarrow$) & $\mathcal{W}_2$ ($\downarrow$) \\
\midrule

\multirow{2}{*}{Scenario 1} 
& ICL + Gaussian & 0.974 ($\pm$ 0.028) & 1.838 $(\pm$ 0.778) & 1.450 ($\pm$ 0.607) & \textbf{0.589} ($\pm$ 0.015) & \textbf{0.080} $(\pm$ 0.010) & 0.459 ($\pm$ 0.017) \\
& \textbf{ICL + Flow Matching} & \textbf{0.552} ($\pm$ 0.028) & \textbf{0.034} $(\pm$ 0.034) & \textbf{0.289} ($\pm$ 0.083) & \textbf{0.606} ($\pm$ 0.038) & \textbf{0.068} $(\pm$ 0.069) & \textbf{0.265} ($\pm$ 0.078) \\

\midrule
\multirow{2}{*}{Scenario 2} 
& ICL + Gaussian & 0.835 ($\pm$ 0.040) & 0.813 $(\pm$ 0.276) & 1.250 ($\pm$ 0.316) & 0.889 ($\pm$ 0.027) & 0.778 $(\pm$ 0.109) & 1.074 ($\pm$ 0.073) \\
& \textbf{ICL + Flow Matching} & \textbf{0.542} ($\pm$ 0.006) & \textbf{0.017} $(\pm$ 0.006) & \textbf{0.244} ($\pm$ 0.033) & \textbf{0.622} ($\pm$ 0.032) & \textbf{0.098} $(\pm$ 0.039) & \textbf{0.287} ($\pm$ 0.046) \\

\midrule
\multirow{2}{*}{Scenario 3} 
& ICL + Gaussian & 0.826 ($\pm$ 0.035) & 0.826 $(\pm$ 0.226) & 1.210 ($\pm$ 0.239) & 0.942 ($\pm$ 0.008) & 1.466 $(\pm$ 0.078) & 1.317 ($\pm$ 0.038) \\
& \textbf{ICL + Flow Matching} & \textbf{0.537} ($\pm$ 0.023) & \textbf{0.024} $(\pm$ 0.021) & \textbf{0.259} ($\pm$ 0.088) & \textbf{0.609} ($\pm$ 0.019) & \textbf{0.124} $(\pm$ 0.037) & \textbf{0.179} ($\pm$ 0.018) \\

\midrule
\multirow{2}{*}{Scenario 4} 
& ICL + Gaussian & 0.870 ($\pm$ 0.043) & 0.706 $(\pm$ 0.218) & 1.635 ($\pm$ 0.297) & 0.999 ($\pm$ 0.001) & 2.025 $(\pm$ 0.017) & 2.013 ($\pm$ 0.019) \\
& \textbf{ICL + Flow Matching} & \textbf{0.684} ($\pm$ 0.060) & \textbf{0.198} $(\pm$ 0.141) & \textbf{0.918} ($\pm$ 0.246) & \textbf{0.988} ($\pm$ 0.003) & \textbf{1.764} $(\pm$ 0.026) & \textbf{1.248} ($\pm$ 0.008) \\

\midrule
\multirow{2}{*}{Scenario 5} 
& ICL + Gaussian & 0.838 ($\pm$ 0.029) & 0.831 $(\pm$ 0.219) & 1.248 ($\pm$ 0.249) & 0.944 ($\pm$ 0.009) & 1.477 $(\pm$ 0.073) & 1.316 ($\pm$ 0.031) \\
& \textbf{ICL + Flow Matching} & \textbf{0.535} ($\pm$ 0.016) & \textbf{0.021} $(\pm$ 0.011) & \textbf{0.279} ($\pm$ 0.060) & \textbf{0.886} ($\pm$ 0.017) & \textbf{1.207} $(\pm$ 0.101) & \textbf{1.002} ($\pm$ 0.042) \\

\midrule
\multirow{2}{*}{Scenario 6} 
& ICL + Gaussian & 0.837 ($\pm$ 0.030) & 0.831 $(\pm$ 0.219) & 1.248 ($\pm$ 0.249) & 0.944 ($\pm$ 0.008) & 1.477 $(\pm$ 0.073) & 1.316 ($\pm$ 0.031) \\
& \textbf{ICL + Flow Matching} & \textbf{0.543} ($\pm$ 0.021) & \textbf{0.023} $(\pm$ 0.015) & \textbf{0.345} ($\pm$ 0.173) & \textbf{0.666} ($\pm$ 0.020) & \textbf{0.200} $(\pm$ 0.034) & \textbf{0.224} ($\pm$ 0.014) \\

\bottomrule
\end{tabular}
}
\end{table}

\begin{table}[htp]
\centering
\caption{Gaussian Mixture Models: Comparing the in-context learner with a Gaussian approximation, fitted via the forward KL-divergence, to the proposed flow matching method. Evaluation on 50 synthetic and 17 real-world datasets for seven different scenarios. If one method is by more than two standard errors better than the other, it is marked in \textbf{bold}. While the differences are less clear-cut than in the previous models, ICL + Flow Matching demonstrates favorable performance in several scenarios, particularly for the Wasserstein-2 distance and MMD. Notably, its advantage is most visible in lower-dimensional settings (Scenario 1 and 2), where it consistently improves upon the Gaussian approximation, fitted via the forward KL-divergence, across most metrics. However, as the dimensionality increases, the performance gap tends to narrow, and in some cases, the inherent variability of the datasets, especially for the Gaussian approximation, fitted via the forward KL-divergence,, makes it difficult to conclusively determine a clear winner. Nonetheless, the flow matching approach often achieves smaller standard errors and lower discrepancy measures, underlining its potential for more stable modeling.}

\label{tab:gmm_res_detail}
\resizebox{1\textwidth}{!}{
\begin{tabular}{p{1.5cm} p{3.5cm} m{2.4cm}m{2.5cm}m{2.7cm}| m{2.4cm}m{2.4cm}m{2.4cm}}
\toprule
\multicolumn{1}{c}{\multirow[c]{2}{*}[-0.5ex]{\textbf{Scenario}}} & \multicolumn{1}{c}{\multirow{2}{*}[-0.5ex]{\textbf{Model}}} & \multicolumn{3}{c}{\textbf{Synthetic Evaluation}} & \multicolumn{3}{c}{\textbf{Real-World Evaluation}} \\
\cmidrule(lr){3-5} \cmidrule(lr){6-8}
& & C2ST ($\downarrow$) & MMD ($\downarrow$) & $\mathcal{W}_2$ ($\downarrow$) & C2ST ($\downarrow$) & MMD ($\downarrow$) & $\mathcal{W}_2$ ($\downarrow$) \\
\midrule

\multirow{2}{*}{Scenario 1} 
& ICL + Gaussian & \textbf{0.926} ($\pm$ 0.029) & \textbf{0.555} $(\pm$ 0.452) & \textbf{2.586} ($\pm$ 0.560) & \textbf{0.957} ($\pm$ 0.034) & \textbf{0.765} $(\pm$ 0.958) & \textbf{3.717} ($\pm$ 1.709) \\
& \textbf{ICL + Flow Matching} & \textbf{0.760} ($\pm$ 0.092) & \textbf{0.303} $(\pm$ 0.548) & \textbf{2.095} ($\pm$ 1.692) & \textbf{0.847} ($\pm$ 0.082) & \textbf{0.486} $(\pm$ 0.623) & \textbf{4.054} ($\pm$ 2.782) \\

\midrule
\multirow{2}{*}{Scenario 2} 
& ICL + Gaussian & 0.985 ($\pm$ 0.010) & 0.761 $(\pm$ 0.227) & 5.022 ($\pm$ 0.945) & \textbf{0.999} ($\pm$ 0.001) & 0.801 $(\pm$ 0.256) & 7.525 ($\pm$ 1.513) \\
& \textbf{ICL + Flow Matching} & \textbf{0.812} ($\pm$ 0.061) & \textbf{0.159} $(\pm$ 0.154) & \textbf{2.314} ($\pm$ 0.926) & \textbf{0.937} ($\pm$ 0.041) & \textbf{0.282} $(\pm$ 0.131) & \textbf{3.947} ($\pm$ 1.055) \\

\midrule
\multirow{2}{*}{Scenario 3} 
& ICL + Gaussian & \textbf{0.998} ($\pm$ 0.002) & \textbf{0.829} $(\pm$ 0.241) & \textbf{11.536} ($\pm$ 2.365) & \textbf{1.000} ($\pm$ 0.000) & \textbf{1.500} $(\pm$ 0.251) & \textbf{26.242} ($\pm$ 4.171) \\
& \textbf{ICL + Flow Matching} & \textbf{1.000} ($\pm$ 0.000) & \textbf{0.582} $(\pm$ 0.280) & \textbf{8.708} ($\pm$ 4.945) & \textbf{1.000} ($\pm$ 0.000) & \textbf{1.869} $(\pm$ 0.342) & \textbf{33.230} ($\pm$ 8.095) \\

\midrule
\multirow{2}{*}{Scenario 4} 
& ICL + Gaussian & \textbf{0.998} ($\pm$ 0.001) & \textbf{6.314} $(\pm$ 0.449) & \textbf{13.404} ($\pm$ 0.609) & \textbf{0.997} ($\pm$ 0.001) & \textbf{2.770} $(\pm$ 1.201) & 22.596 ($\pm$ 5.717) \\
& \textbf{ICL + Flow Matching} & 1.000 ($\pm$ 0.000) & \textbf{2.451} $(\pm$ 0.868) & \textbf{8.333} ($\pm$ 4.202) & 1.000 ($\pm$ 0.000) & \textbf{2.518} $(\pm$ 0.694) & \textbf{11.938} ($\pm$ 2.956) \\

\bottomrule
\end{tabular}
}
\end{table}

\clearpage

\clearpage

\section{Ablation: Using a Diffusion Objective}

\label{sec:diffusion_abl}

\textcolor{rev}{To validate choosing the flow matching objective with optimal transport (OT) paths resulting in the objective in equation \Cref{eq:emprisk}, we also conduct experiments using a diffusion-objective with variance preserving paths introduced by \citet{song2020score}. We choose three selected GLM, FA and GMM scenarios with the same 50 synthetic and 17 real-world datasets for each scenario as in the other benchmarks.}

\subsection{Diffusion with Flow-Matching}

First, we use the diffusion objective learned via flow matching, as described in \citep{lipman2022flow}, where we choose the same hyperparameters as \citep{lipman2022flow}.

\begin{table}[htp]
\centering
\caption{\textcolor{rev}{GLMs: Comparison of the OT flow matching and the VP diffusion objective on 50 synthetic and 17 real-world datasets for three different scenarios. All results within two standard errors of the best average result for each scenario are marked in \textbf{bold}.}}
\label{tab:gml_diff_abl}
\resizebox{1\textwidth}{!}{
\textcolor{rev}{
\begin{tabular}{p{1.5cm} p{3.5cm} m{2.4cm}m{2.5cm}m{2.7cm}| m{2.4cm}m{2.4cm}m{2.4cm}}
\toprule
\multicolumn{1}{c}{\multirow[c]{2}{*}[-0.5ex]{\textbf{Scenario}}} & \multicolumn{1}{c}{\multirow{2}{*}[-0.5ex]{\textbf{Model}}} & \multicolumn{3}{c}{\textbf{Synthetic Evaluation}} & \multicolumn{3}{c}{\textbf{Real-World Evaluation}} \\
\cmidrule(lr){3-5} \cmidrule(lr){6-8}
& & C2ST ($\downarrow$) & MMD ($\downarrow$) & $\mathcal{W}_2$ ($\downarrow$) & C2ST ($\downarrow$) & MMD ($\downarrow$) & $\mathcal{W}_2$ ($\downarrow$) \\
\midrule
\multirow{2}{*}{Scenario 2} 
& Diffusion paths + FM & 0.961 ($\pm$ 0.040) & \textbf{1.525} $(\pm$ 0.777) & 3.354 ($\pm$ 1.333) & 0.961 ($\pm$ 0.016) & 1.347 $(\pm$ 0.365) & 2.025 ($\pm$ 0.270) \\
& \textbf{OT paths} & \textbf{0.839} ($\pm$ 0.072) & \textbf{0.707} $(\pm$ 0.658) & \textbf{1.111} ($\pm$ 0.300) & \textbf{0.768} ($\pm$ 0.033) & \textbf{0.143} $(\pm$ 0.089) & \textbf{0.411} ($\pm$ 0.094) \\
\midrule
\multirow{2}{*}{Scenario 3} 
& Diffusion paths + FM & 0.903 ($\pm$ 0.111) & 1.080 $(\pm$ 0.564) & 1.733 ($\pm$ 0.408) & 0.936 ($\pm$ 0.013) & 1.002 $(\pm$ 0.203) & 1.442 ($\pm$ 0.103) \\
& \textbf{OT paths} & \textbf{0.611} ($\pm$ 0.070) & \textbf{0.089} $(\pm$ 0.114) & \textbf{0.423} ($\pm$ 0.348) & \textbf{0.576} ($\pm$ 0.027) & \textbf{0.037} $(\pm$ 0.026) & \textbf{0.257} ($\pm$ 0.044) \\
\midrule
\multirow{2}{*}{Scenario 5} 
& Diffusion paths + FM & \textbf{0.691} ($\pm$ 0.074) & 0.211 $(\pm$ 0.143) & 0.708 ($\pm$ 0.233) & \textbf{0.681} ($\pm$ 0.038) & 0.182 $(\pm$ 0.093) & \textbf{0.554 ($\pm$ 0.090)} \\
& \textbf{OT paths + FM} & \textbf{0.621} ($\pm$ 0.063) & \textbf{0.067} $(\pm$ 0.080) & \textbf{0.299} ($\pm$ 0.195) & \textbf{0.610} ($\pm$ 0.045) & \textbf{0.046} $(\pm$ 0.020) & \textbf{0.242} ($\pm$ 0.038) \\
\bottomrule
\end{tabular}
}
}
\end{table}

\textcolor{rev}{In summary, the empirical results demonstrate that using the OT paths consistently outperforms the VP diffusion objective across all scenarios for both GLMs and FAs. For GLMs, OT paths achieve significantly lower C2ST values in all scenarios. In Scenario 2, OT paths reduce C2ST from 0.961 to 0.839 on synthetic data and from 0.961 to 0.768 on real-world data. Similarly, in Scenario 3, OT paths achieve substantial improvements, with C2ST dropping from 0.903 to 0.611 on synthetic data and from 0.936 to 0.576 on real-world data. This trend is complemented by consistent improvements in other metrics such as $\mathcal{W}_2$, where OT paths often achieve reductions by over 50\%.}

\begin{table}[htp]
\centering
\caption{\textcolor{rev}{FA: Comparison of the OT flow matching and the VP diffusion objective on 50 synthetic and 17 real-world datasets for three different scenarios. All results within two standard errors of the best average result for each scenario are marked in \textbf{bold}.}}
\label{tab:fa_diff_abl}
\resizebox{1\textwidth}{!}{
\textcolor{rev}{
\begin{tabular}{p{1.5cm} p{3.5cm} m{2.4cm}m{2.5cm}m{2.7cm}| m{2.4cm}m{2.4cm}m{2.4cm}}
\toprule
\multicolumn{1}{c}{\multirow[c]{2}{*}[-0.5ex]{\textbf{Scenario}}} & \multicolumn{1}{c}{\multirow{2}{*}[-0.5ex]{\textbf{Model}}} & \multicolumn{3}{c}{\textbf{Synthetic Evaluation}} & \multicolumn{3}{c}{\textbf{Real-World Evaluation}} \\
\cmidrule(lr){3-5} \cmidrule(lr){6-8}
& & C2ST ($\downarrow$) & MMD ($\downarrow$) & $\mathcal{W}_2$ ($\downarrow$) & C2ST ($\downarrow$) & MMD ($\downarrow$) & $\mathcal{W}_2$ ($\downarrow$) \\
\midrule
\multirow{2}{*}{Scenario 1} 
& Diffusion paths + FM & 0.622 ($\pm$ 0.043) & 0.207 $(\pm$ 0.121) & 0.692 ($\pm$ 0.192) & \textbf{0.595} ($\pm$ 0.012) & 0.089 $(\pm$ 0.011) & 0.475 ($\pm$ 0.019) \\
& \textbf{OT paths + FM} & \textbf{0.552} ($\pm$ 0.028) & \textbf{0.034} $(\pm$ 0.034) & \textbf{0.289} ($\pm$ 0.083) & \textbf{0.606} ($\pm$ 0.038) & \textbf{0.068} $(\pm$ 0.069) & \textbf{0.265} ($\pm$ 0.078) \\
\midrule
\multirow{2}{*}{Scenario 2} 
& Diffusion paths + FM  & 0.826 ($\pm$ 0.036) & 0.768 $(\pm$ 0.238) & 1.219 ($\pm$ 0.276) & 0.878 ($\pm$ 0.028) & 0.793 $(\pm$ 0.154) & 1.056 ($\pm$ 0.084) \\
& \textbf{OT paths + FM} & \textbf{0.542} ($\pm$ 0.006) & \textbf{0.017} $(\pm$ 0.006) & \textbf{0.244} ($\pm$ 0.033) & \textbf{0.622} ($\pm$ 0.032) & \textbf{0.098} $(\pm$ 0.039) & \textbf{0.287} ($\pm$ 0.046) \\
\midrule
\multirow{2}{*}{Scenario 3} 
& Diffusion paths + FM & 0.751 ($\pm$ 0.048) & 0.387 $(\pm$ 0.216) & 0.834 ($\pm$ 0.163) & 0.944 ($\pm$ 0.008) & 1.514 $(\pm$ 0.056) & 1.332 ($\pm$ 0.028)  \\
& \textbf{OT paths + FM} & \textbf{0.537} ($\pm$ 0.023) & \textbf{0.024} $(\pm$ 0.021) & \textbf{0.259} ($\pm$ 0.088) & \textbf{0.609} ($\pm$ 0.019) & \textbf{0.124} $(\pm$ 0.037) & \textbf{0.179} ($\pm$ 0.018)\\
\bottomrule
\end{tabular}
}
}
\end{table}

\textcolor{rev}{For FA, the performance gap in C2ST remains notable. In Scenario 1, OT paths achieve the best results on synthetic data, reducing C2ST from 0.622 to 0.552, while also delivering improvements in $\mathcal{W}_2$ (0.289 compared to 0.692). On real-world datasets, OT paths maintain competitive results, matching or exceeding the performance of diffusion paths. The advantage is even more pronounced in Scenario 2, where OT paths consistently lead across all metrics, with a particularly striking reduction in MMD on synthetic data (0.017 compared to 0.768) and strong results for C2ST on real-world data (0.622 vs. 0.878). Similarly, in Scenario 3, OT paths achieve the lowest C2ST values, with synthetic results improving from 0.751 to 0.537 and real-world results from 0.944 to 0.609.}

\begin{table}[htp]
\centering
\caption{\textcolor{rev}{GMMs: Comparison of the OT flow matching and the VP diffusion objective on 50 synthetic and 17 real-world datasets for three different scenarios. All results within two standard errors of the best average result for each scenario are marked in \textbf{bold}.}}
\label{tab:gmm_diff_abl}
\resizebox{1\textwidth}{!}{
\textcolor{rev}{
\begin{tabular}{p{1.5cm} p{3.5cm} m{2.4cm}m{2.5cm}m{2.7cm}| m{2.4cm}m{2.4cm}m{2.4cm}}
\toprule
\multicolumn{1}{c}{\multirow[c]{2}{*}[-0.5ex]{\textbf{Scenario}}} & \multicolumn{1}{c}{\multirow{2}{*}[-0.5ex]{\textbf{Model}}} & \multicolumn{3}{c}{\textbf{Synthetic Evaluation}} & \multicolumn{3}{c}{\textbf{Real-World Evaluation}} \\
\cmidrule(lr){3-5} \cmidrule(lr){6-8}
& & C2ST ($\downarrow$) & MMD ($\downarrow$) & $\mathcal{W}_2$ ($\downarrow$) & C2ST ($\downarrow$) & MMD ($\downarrow$) & $\mathcal{W}_2$ ($\downarrow$) \\
\midrule
\multirow{2}{*}{Scenario 1} 
& Diffusion paths + FM & 0.924 ($\pm$ 0.024) & \textbf{0.241} $(\pm$ 0.381) & \textbf{2.195} ($\pm$ 1.431) & 0.958 ($\pm$ 0.030) & 0.890 $(\pm$ 0.912) & \textbf{5.328} ($\pm$ 2.544) \\
& \textbf{OT paths + FM} & \textbf{0.760} ($\pm$ 0.092) & \textbf{0.303} $(\pm$ 0.548) & \textbf{2.095} ($\pm$ 1.692) & \textbf{0.847} ($\pm$ 0.082) & \textbf{0.486} $(\pm$ 0.623) & \textbf{4.054} ($\pm$ 2.782)  \\
\midrule
\multirow{2}{*}{Scenario 2} 
& Diffusion paths + FM & 0.942 ($\pm$ 0.020) & \textbf{0.213} $(\pm$ 0.187) & \textbf{2.748} ($\pm$ 0.659) & \textbf{0.984} ($\pm$ 0.012) & \textbf{0.411} $(\pm$ 0.162) & \textbf{5.397} ($\pm$ 1.458) \\
& \textbf{OT paths + FM} & \textbf{0.812} ($\pm$ 0.061) & \textbf{0.159} $(\pm$ 0.154) & \textbf{2.314} ($\pm$ 0.926) & \textbf{0.937} ($\pm$ 0.041) & \textbf{0.282} $(\pm$ 0.131) & \textbf{3.947} ($\pm$ 1.055) \\
\midrule
\multirow{2}{*}{Scenario 3} 
& Diffusion paths + FM & \textbf{1.000} ($\pm$ 0.000) & 0.582 $(\pm$ 0.280) & \textbf{8.708} ($\pm$ 4.945) & \textbf{1.000} ($\pm$ 0.000) & 1.869 $(\pm$ 0.342) & \textbf{33.230} ($\pm$ 8.095) \\
& \textbf{OT paths + FM} & \textbf{0.999} ($\pm$ 0.001) & \textbf{0.267} $(\pm$ 0.154) & \textbf{7.234} ($\pm$ 2.974) & \textbf{1.000} ($\pm$ 0.000) & \textbf{1.155} $(\pm$ 0.258) & \textbf{26.956} ($\pm$ 3.114)  \\
\bottomrule
\end{tabular}
}
}
\end{table}

\textcolor{rev}{In the case of Gaussian Mixture Models (GMMs), the empirical results indicate that the OT paths generally outperform the VP diffusion objective across most scenarios and metrics, though the differences are not always statistically significant in pair-wise comparisons. For example, in Scenario 1, OT paths achieve notably better results for C2ST on both synthetic and real-world datasets, with reductions from 0.924 to 0.760 and from 0.958 to 0.847, respectively. Similarly, for $\mathcal{W}_2$, OT paths exhibit better performance on real-world data (4.054 vs. 5.328). In Scenario 2, OT paths maintain a consistent advantage in metrics such as C2ST and $\mathcal{W}_2$. For instance, synthetic data shows a C2ST improvement from 0.942 to 0.812, while real-world data improves from 0.984 to 0.937. The OT paths also achieve lower MMD on synthetic data (0.159 vs. 0.213), supporting their effectiveness in this scenario. For Scenario 3, OT paths achieve better results for $\mathcal{W}_2$ on both synthetic and real-world data, reducing it from 8.708 to 7.234 and from 33.230 to 26.956, respectively. }

\subsection{Diffusion with Score-Matching}

\begin{figure}
    \centering
    \includegraphics[width=1.0\linewidth]{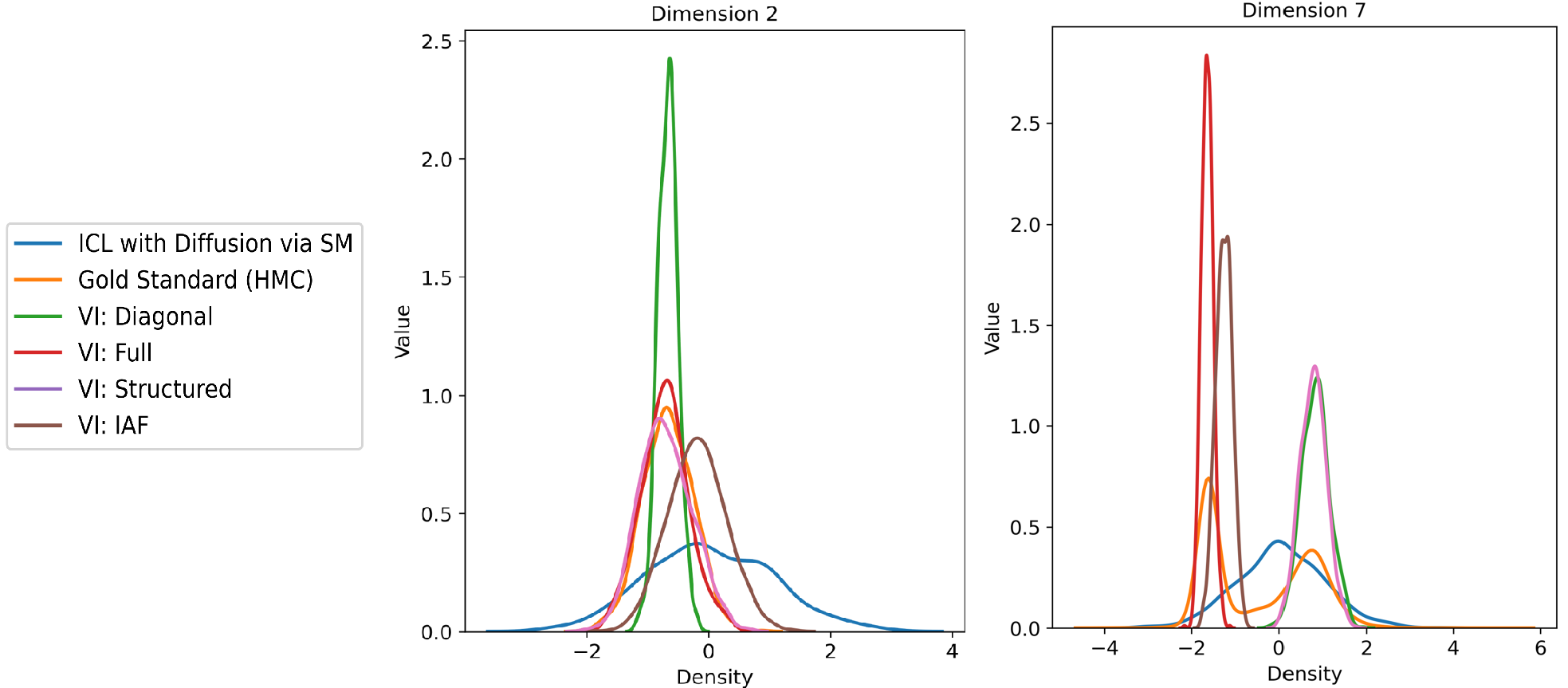}
    \caption{Marginal distribution for GLM scenario 2 (left) and GMM scenario 1 (right). The in-context learner is trained with a diffusion objective using VP paths.}
    \label{fig:diffusion_ablation_marginals}
\end{figure}

Second, we compare the results of using OT paths with flow matching to the results obtained when using VP paths and score matching. We use the score matching objective introduced by \citet{song2019generative} and maintain the VP hyperparameters from \citet{lipman2022flow} that we previously used for the diffusion objective with flow matching. 

We find that, across all three considered GLM scenarios, using OT paths and flow matching yields substantially better results than using Diffusion VP paths and score matching, where the score-matching objective sometimes yields results comparable to those obtained using a Laplace approximation. We observe similar overall results for FA and GMMs, although the effect is less pronounced. Note that the inferiority of score matching compared to flow matching is consistent with findings by \citet{lipman2022flow} and \citet{dax2024real}, who also report that flow matching produces more stable and less noisy training trajectories. 

The large quantity of noise in the diffusion objective might prevent the model from learning complex conditioning on datasets $\vx$, which is arguably the main challenge for performing in-context learning for the posteriors of latent variable models. We find visually that using the diffusion objective leads to a form of collapse where the model only learns a constant posterior distribution $Q_{\theta}^{\vz|\vx}$ that has a relatively large variance and is centered around zero, while largely ignoring the conditioning on $\vx$ (Please refer to figure \ref{fig:diffusion_ablation_marginals}).

\begin{table}[htp]
\centering
\caption{\textcolor{rev}{GLMs: Comparison of the OT flow matching and the VP diffusion objective with score matching on 50 synthetic and 17 real-world datasets for three different scenarios. All results within two standard errors of the best average result for each scenario are marked in \textbf{bold}.}}
\label{tab:gml_diff_abl}
\resizebox{1\textwidth}{!}{
\textcolor{rev}{
\begin{tabular}{p{1.5cm} p{3.5cm} m{2.4cm}m{2.5cm}m{2.7cm}| m{2.4cm}m{2.4cm}m{2.4cm}}
\toprule
\multicolumn{1}{c}{\multirow[c]{2}{*}[-0.5ex]{\textbf{Scenario}}} & \multicolumn{1}{c}{\multirow{2}{*}[-0.5ex]{\textbf{Model}}} & \multicolumn{3}{c}{\textbf{Synthetic Evaluation}} & \multicolumn{3}{c}{\textbf{Real-World Evaluation}} \\
\cmidrule(lr){3-5} \cmidrule(lr){6-8}
& & C2ST ($\downarrow$) & MMD ($\downarrow$) & $\mathcal{W}_2$ ($\downarrow$) & C2ST ($\downarrow$) & MMD ($\downarrow$) & $\mathcal{W}_2$ ($\downarrow$) \\
\midrule
\multirow{2}{*}{Scenario 2} 
& Diffusion paths + SM & 0.996 ($\pm$ 0.011) & 4.121 $(\pm$ 1.625) & 8.761 ($\pm$ 4.415) & 0.998 ($\pm$ 0.002) & 1.574 $(\pm$ 0.906) & 8.483 ($\pm$ 1.580) \\
& \textbf{OT paths + FM} & \textbf{0.839} ($\pm$ 0.072) & \textbf{0.707} $(\pm$ 0.658) & \textbf{1.111} ($\pm$ 0.300) & \textbf{0.768} ($\pm$ 0.033) & \textbf{0.143} $(\pm$ 0.089) & \textbf{0.411} ($\pm$ 0.094) \\
\midrule
\multirow{2}{*}{Scenario 3} 
& Diffusion paths + SM & 0.965 ($\pm$ 0.075) & 2.466 $(\pm$ 1.224) & 3.947 ($\pm$ 1.323) & 0.994 ($\pm$ 0.002) & 2.018 $(\pm$ 0.206) & 3.301 ($\pm$ 0.260) \\
& \textbf{OT paths + FM} & \textbf{0.611} ($\pm$ 0.070) & \textbf{0.089} $(\pm$ 0.114) & \textbf{0.423} ($\pm$ 0.348) & \textbf{0.576} ($\pm$ 0.027) & \textbf{0.037} $(\pm$ 0.026) & \textbf{0.257} ($\pm$ 0.044) \\
\midrule
\multirow{2}{*}{Scenario 5} 
& Diffusion paths + SM & 0.998 ($\pm$ 0.002) & 3.163 $(\pm$ 0.651) & 8.684 ($\pm$ 1.135) & 0.999 ($\pm$ 0.001) & 3.004 $(\pm$ 0.056) & 8.547 ($\pm$ 0.177) \\
& \textbf{OT paths + FM} & \textbf{0.621} ($\pm$ 0.063) & \textbf{0.067} $(\pm$ 0.080) &\textbf{ 0.299} ($\pm$ 0.195) & \textbf{0.610} ($\pm$ 0.045) & \textbf{0.046} $(\pm$ 0.020) & \textbf{0.242} ($\pm$ 0.038) \\
\bottomrule
\end{tabular}
}
}
\end{table}

\begin{table}[htp]
\centering
\caption{\textcolor{rev}{FA: Comparison of the OT flow matching and the VP diffusion objective with score matching on 50 synthetic and 17 real-world datasets for three different scenarios. All results within two standard errors of the best average result for each scenario are marked in \textbf{bold}.}}
\label{tab:fa_diff_abl}
\resizebox{1\textwidth}{!}{
\textcolor{rev}{
\begin{tabular}{p{1.5cm} p{3.5cm} m{2.4cm}m{2.5cm}m{2.7cm}| m{2.4cm}m{2.4cm}m{2.4cm}}
\toprule
\multicolumn{1}{c}{\multirow[c]{2}{*}[-0.5ex]{\textbf{Scenario}}} & \multicolumn{1}{c}{\multirow{2}{*}[-0.5ex]{\textbf{Model}}} & \multicolumn{3}{c}{\textbf{Synthetic Evaluation}} & \multicolumn{3}{c}{\textbf{Real-World Evaluation}} \\
\cmidrule(lr){3-5} \cmidrule(lr){6-8}
& & C2ST ($\downarrow$) & MMD ($\downarrow$) & $\mathcal{W}_2$ ($\downarrow$) & C2ST ($\downarrow$) & MMD ($\downarrow$) & $\mathcal{W}_2$ ($\downarrow$) \\
\midrule
\multirow{2}{*}{Scenario 1} 
& Diffusion paths + SM & 0.880 ($\pm$ 0.024) & 0.875 $(\pm$ 0.134) & 1.787 ($\pm$ 0.155) & 0.906 ($\pm$ 0.007) & 0.845 $(\pm$ 0.026) & 1.723 ($\pm$ 0.029) \\
& \textbf{OT paths + FM}& \textbf{0.552} ($\pm$ 0.028) & \textbf{0.034} $(\pm$ 0.034) & \textbf{0.289} ($\pm$ 0.083) & \textbf{0.606} ($\pm$ 0.038) & \textbf{0.068} $(\pm$ 0.069) & \textbf{0.265} ($\pm$ 0.078) \\
\midrule
\multirow{2}{*}{Scenario 2} 
& Diffusion paths + SM  & 0.932 ($\pm$ 0.022) & 1.459 $(\pm$ 0.128) & 2.798 ($\pm$ 0.141) & 0.980 ($\pm$ 0.008) & 1.772 $(\pm$ 0.065) & 2.927 ($\pm$ 0.085) \\
& \textbf{OT paths + FM}& \textbf{0.542} ($\pm$ 0.006) & \textbf{0.017} $(\pm$ 0.006) & \textbf{0.244} ($\pm$ 0.033) & \textbf{0.622} ($\pm$ 0.032) & \textbf{0.098} $(\pm$ 0.039) & \textbf{0.287} ($\pm$ 0.046) \\
\midrule
\multirow{2}{*}{Scenario 3} 
& Diffusion paths + SM & 0.925 ($\pm$ 0.021) & 1.747 $(\pm$ 0.382) & 3.028 ($\pm$ 0.646) & 0.989 ($\pm$ 0.003) & 2.101 $(\pm$ 0.050) & 2.882 ($\pm$ 0.050)  \\
& \textbf{OT paths + FM}& \textbf{0.537} ($\pm$ 0.023) & \textbf{0.024} $(\pm$ 0.021) & \textbf{0.259} ($\pm$ 0.088) & \textbf{0.609} ($\pm$ 0.019) & \textbf{0.124} $(\pm$ 0.037) & \textbf{0.179} ($\pm$ 0.018)\\
\bottomrule
\end{tabular}
}
}
\end{table}

\begin{table}[htp]
\centering
\caption{\textcolor{rev}{GMMs: Comparison of the OT flow matching and the VP diffusion objective with score matching on 50 synthetic and 17 real-world datasets for three different scenarios. All results within two standard errors of the best average result for each scenario are marked in \textbf{bold}.}}
\label{tab:gmm_diff_abl}
\resizebox{1\textwidth}{!}{
\textcolor{rev}{
\begin{tabular}{p{1.5cm} p{3.5cm} m{2.4cm}m{2.5cm}m{2.7cm}| m{2.4cm}m{2.4cm}m{2.4cm}}
\toprule
\multicolumn{1}{c}{\multirow[c]{2}{*}[-0.5ex]{\textbf{Scenario}}} & \multicolumn{1}{c}{\multirow{2}{*}[-0.5ex]{\textbf{Model}}} & \multicolumn{3}{c}{\textbf{Synthetic Evaluation}} & \multicolumn{3}{c}{\textbf{Real-World Evaluation}} \\
\cmidrule(lr){3-5} \cmidrule(lr){6-8}
& & C2ST ($\downarrow$) & MMD ($\downarrow$) & $\mathcal{W}_2$ ($\downarrow$) & C2ST ($\downarrow$) & MMD ($\downarrow$) & $\mathcal{W}_2$ ($\downarrow$) \\
\midrule
\multirow{2}{*}{Scenario 1} 
& Diffusion paths + SM & 1.000 ($\pm$ 0.001) & 1.412 $(\pm$ 0.365) & 7.038 ($\pm$ 0.655) & 0.998 ($\pm$ 0.002) & 1.574 $(\pm$ 0.906) & 8.483 ($\pm$ 1.580) \\
& \textbf{OT paths + FM}& \textbf{0.760} ($\pm$ 0.092) & \textbf{0.303} $(\pm$ 0.548) & \textbf{2.095} ($\pm$ 1.692) & \textbf{0.847} ($\pm$ 0.082) & \textbf{0.486} $(\pm$ 0.623) & \textbf{4.054} ($\pm$ 2.782)  \\
\midrule
\multirow{2}{*}{Scenario 2} 
& Diffusion paths + SM & 1.000 ($\pm$ 0.000) & 1.275 $(\pm$ 0.240) & 6.621 ($\pm$ 1.091) & 1.000 ($\pm$ 0.000) & 1.032 $(\pm$ 0.163) & 7.931 ($\pm$ 0.748) \\
& \textbf{OT paths + FM} & \textbf{0.812} ($\pm$ 0.061) & \textbf{0.159} $(\pm$ 0.154) & \textbf{2.314} ($\pm$ 0.926) & \textbf{0.937} ($\pm$ 0.041) & \textbf{0.282} $(\pm$ 0.131) & \textbf{3.947} ($\pm$ 1.055) \\
\midrule
\multirow{2}{*}{Scenario 3} 
& Diffusion paths + SM& \textbf{1.000} ($\pm$ 0.000) & 1.337 $(\pm$ 0.476) & \textbf{10.877} ($\pm$ 5.262) & \textbf{1.000} ($\pm$ 0.000) & 2.277 $(\pm$ 0.245) & \textbf{24.269} ($\pm$ 3.841) \\
& \textbf{OT paths + FM}& \textbf{0.999} ($\pm$ 0.001) & \textbf{0.267} $(\pm$ 0.154) & \textbf{7.234} ($\pm$ 2.974) & \textbf{1.000} ($\pm$ 0.000) & \textbf{1.155} $(\pm$ 0.258) & \textbf{26.956} ($\pm$ 3.114)  \\
\bottomrule
\end{tabular}
}
}
\end{table}

\clearpage

\section{\textcolor{rev}{Ablation: Using an MLP-based Encoder}}

\label{sec:abl_mlp}

\textcolor{rev}{To further justify choosing a transformer encoder in our ICL approach, we conduct an ablation study comparing the performance of our original ICL method with the performance obtained when the transformer encoder is replaced by an MLP with batch normalization \citep{ioffe2015batch} and skip-connections. To ensure a fair comparison, we use an MLP encoder with a hidden dimension of 1250 to give the overall model approximately the same number of parameters as in the transformer-based approach. Concretely, our MLP-approach has 43.3 million parameters compared to 43.1 million parameters with the transformer encoder. We choose three selected GLM, FA and GMM scenarios with 50 synthetic and 17 real-world datasets for each scenario.}

\textcolor{rev}{In summary, we find that the transformer encoder yields consistently better, results than the mlp encoder across all scenarios. While the difference is especially pronounced for the GLM scenarios, the difference become smaller for FA and GMMs.} 

%\textcolor{rev}{To validate choosing the flow matching objective with optimal transport (OT) paths resulting in the objective in equation \Cref{eq:emprisk}, we also conduct experiments using a diffusion-objective with variance preserving paths introduced by \citet{song2020score}. }

\begin{table}[htp]
\centering
\caption{\textcolor{rev}{GLMs: Comparison when using an MLP-based encoder and a transformer encoder on 50 synthetic and 17 real-world datasets for three different scenarios.}}
\label{tab:glm_mlp}
\resizebox{1\textwidth}{!}{
\textcolor{rev}{
\begin{tabular}{p{1.5cm} p{3.5cm} m{2.4cm}m{2.5cm}m{2.7cm}| m{2.4cm}m{2.4cm}m{2.4cm}}
\toprule
\multicolumn{1}{c}{\multirow[c]{2}{*}[-0.5ex]{\textbf{Scenario}}} & \multicolumn{1}{c}{\multirow{2}{*}[-0.5ex]{\textbf{Type of Encoder}}} & \multicolumn{3}{c}{\textbf{Synthetic Evaluation}} & \multicolumn{3}{c}{\textbf{Real-World Evaluation}} \\
\cmidrule(lr){3-5} \cmidrule(lr){6-8}
& & C2ST ($\downarrow$) & MMD ($\downarrow$) & $\mathcal{W}_2$ ($\downarrow$) & C2ST ($\downarrow$) & MMD ($\downarrow$) & $\mathcal{W}_2$ ($\downarrow$) \\
\midrule
\multirow{2}{*}{Scenario 2} 
& MLP & 0.942 ($\pm$ 0.093) & 1.783 $(\pm$ 1.048) & 2.503 ($\pm$ 0.814) & 0.968 ($\pm$ 0.012) & 1.528 $(\pm$ 0.394) & 2.271 ($\pm$ 0.315) \\
& \textbf{Transformer} & 0.839 ($\pm$ 0.072) & 0.707 $(\pm$ 0.658) & 1.111 ($\pm$ 0.300) & 0.768 ($\pm$ 0.033) & 0.143 $(\pm$ 0.089) & 0.411 ($\pm$ 0.094) \\
\midrule
\multirow{2}{*}{Scenario 3} 
& MLP & 0.957 ($\pm$ 0.075) & 2.236 $(\pm$ 1.218) & 2.681 ($\pm$ 1.130) & 0.972 ($\pm$ 0.012) & 1.658 $(\pm$ 0.450) & 2.076 ($\pm$ 0.427) \\
& \textbf{Transformer}  & 0.611 ($\pm$ 0.070) & 0.089 $(\pm$ 0.114) & 0.423 ($\pm$ 0.348) & 0.576 ($\pm$ 0.027) & 0.037 $(\pm$ 0.026) & 0.257 ($\pm$ 0.044) \\
\midrule
\multirow{2}{*}{Scenario 5} 
& MLP & 0.845 ($\pm$ 0.115) & 1.066 $(\pm$ 0.859) & 1.166 ($\pm$ 0.996) & 0.890 ($\pm$ 0.055) & 1.223 $(\pm$ 0.791) & 1.102 ($\pm$ 0.383) \\
& \textbf{Transformer}  & 0.621 ($\pm$ 0.063) & 0.067 $(\pm$ 0.080) & 0.299 ($\pm$ 0.195) & 0.610 ($\pm$ 0.045) & 0.046 $(\pm$ 0.020) & 0.242 ($\pm$ 0.038) \\
\bottomrule
\end{tabular}
}
}
\end{table}

\textcolor{rev}{In \Cref{tab:glm_mlp}, the transformer encoder consistently outperforms the MLP encoder across all metrics and scenarios. In Scenario 2, C2ST drops from 0.942 (MLP) to 0.839 (Transformer) on synthetic data and from 0.968 to 0.768 on real-world data. Similarly, $\mathcal{W}_2$ improves significantly, decreasing from 2.503 to 1.111 on synthetic data and from 2.271 to 0.411 on real-world data. In Scenario 3, transformers achieve substantial improvements, reducing C2ST from 0.957 (MLP) to 0.611 on synthetic data and from 0.972 to 0.576 on real-world data. $\mathcal{W}_2$ also sees notable reductions, dropping from 2.681 to 0.423 on synthetic data and from 2.076 to 0.257 on real-world data. Finally, in Scenario 5, transformers maintain their superiority, achieving reductions in C2ST from 0.845 (MLP) to 0.621 on synthetic data and from 0.890 to 0.610 on real-world data. Improvements in $\mathcal{W}_2$ are similarly remarkable, with reductions from 1.166 to 0.299 on synthetic data and from 1.102 to 0.242 on real-world data.
}

\begin{table}[htp]
\centering
\caption{\textcolor{rev}{FA: Comparison when using an MLP-based encoder and a transformer encoder on 50 synthetic and 17 real-world datasets for three different scenarios.}}
\label{tab:fa_mlp}
\resizebox{1\textwidth}{!}{
\textcolor{rev}{
\begin{tabular}{p{1.5cm} p{3.5cm} m{2.4cm}m{2.5cm}m{2.7cm}| m{2.4cm}m{2.4cm}m{2.4cm}}
\toprule
\multicolumn{1}{c}{\multirow[c]{2}{*}[-0.5ex]{\textbf{Scenario}}} & \multicolumn{1}{c}{\multirow{2}{*}[-0.5ex]{\textbf{Type of Encoder}}} & \multicolumn{3}{c}{\textbf{Synthetic Evaluation}} & \multicolumn{3}{c}{\textbf{Real-World Evaluation}} \\
\cmidrule(lr){3-5} \cmidrule(lr){6-8}
& & C2ST ($\downarrow$) & MMD ($\downarrow$) & $\mathcal{W}_2$ ($\downarrow$) & C2ST ($\downarrow$) & MMD ($\downarrow$) & $\mathcal{W}_2$ ($\downarrow$) \\
\midrule
\multirow{2}{*}{Scenario 1} 
& MLP & 0.579 ($\pm$ 0.015) & 0.017 $(\pm$ 0.006) & 0.364 ($\pm$ 0.029) & 0.634 ($\pm$ 0.014) & 0.013 $(\pm$ 0.004) & 0.331 ($\pm$ 0.010) \\
& \textbf{Transformer} & 0.552 ($\pm$ 0.028) & 0.034 $(\pm$ 0.034) & 0.289 ($\pm$ 0.083) & 0.606 ($\pm$ 0.038) & 0.068 $(\pm$ 0.069) & 0.265 ($\pm$ 0.078) \\
\midrule
\multirow{2}{*}{Scenario 2} 
& MLP & 0.562 ($\pm$ 0.038) & 0.037 $(\pm$ 0.042) & 0.308 ($\pm$ 0.097) & 0.632 ($\pm$ 0.068) & 0.182 $(\pm$ 0.407) & 0.339 ($\pm$ 0.174) \\
& \textbf{Transformer} & 0.542 ($\pm$ 0.006) & 0.017 $(\pm$ 0.006) & 0.244 ($\pm$ 0.033) & 0.622 ($\pm$ 0.032) & 0.098 $(\pm$ 0.039) & 0.287 ($\pm$ 0.046) \\
\midrule
\multirow{2}{*}{Scenario 3} 
& MLP & 0.539 ($\pm$ 0.025) & 0.023 $(\pm$ 0.022) & 0.278 ($\pm$ 0.116) & 0.680 ($\pm$ 0.019) & 0.268 $(\pm$ 0.044) & 0.253 ($\pm$ 0.017)  \\
& \textbf{Transformer} & 0.537 ($\pm$ 0.023) & 0.024 $(\pm$ 0.021) & 0.259 ($\pm$ 0.088) & 0.609 ($\pm$ 0.019) & 0.124 $(\pm$ 0.037) & 0.179 ($\pm$ 0.018)\\
\bottomrule
\end{tabular}
}
}
\end{table}

\textcolor{rev}{For the factor analysis cases (\Cref{tab:fa_mlp}), the transformer encoder still has better average performances even though the differences are substantially less pronounced than for the GLMs. In Scenario 1, transformers slightly outperform MLPs, reducing C2ST from 0.579 to 0.552 on synthetic data and from 0.634 to 0.606 on real-world data. $\mathcal{W}_2$ also sees moderate improvements, dropping from 0.364 to 0.289 on synthetic data and from 0.331 to 0.265 on real-world data. In Scenario 2, the advantage of the transformer encoder remains consistent, with C2ST decreasing from 0.562 (MLP) to 0.542 on synthetic data and from 0.632 to 0.622 on real-world data. $\mathcal{W}_2$ also improves slightly, dropping from 0.308 to 0.244 on synthetic data and from 0.339 to 0.287 on real-world data. Scenario 3 shows the smallest differences, where transformers marginally improve C2ST from 0.539 (MLP) to 0.537 on synthetic data and from 0.680 to 0.609 on real-world data. For $\mathcal{W}_2$, the reductions are minor but consistent, dropping from 0.278 to 0.259 on synthetic data and from 0.253 to 0.179 on real-world data.}

\begin{table}[htp]
\centering
\caption{\textcolor{rev}{GMMs: Comparison when using an MLP-based encoder and a transformer encoder on 50 synthetic and 17 real-world datasets for three different scenarios.}}
\label{tab:gmm_mlp}
\resizebox{1\textwidth}{!}{
\textcolor{rev}{
\begin{tabular}{p{1.5cm} p{3.5cm} m{2.4cm}m{2.5cm}m{2.7cm}| m{2.4cm}m{2.4cm}m{2.4cm}}
\toprule
\multicolumn{1}{c}{\multirow[c]{2}{*}[-0.5ex]{\textbf{Scenario}}} & \multicolumn{1}{c}{\multirow{2}{*}[-0.5ex]{\textbf{Type of Encoder}}} & \multicolumn{3}{c}{\textbf{Synthetic Evaluation}} & \multicolumn{3}{c}{\textbf{Real-World Evaluation}} \\
\cmidrule(lr){3-5} \cmidrule(lr){6-8}
& & C2ST ($\downarrow$) & MMD ($\downarrow$) & $\mathcal{W}_2$ ($\downarrow$) & C2ST ($\downarrow$) & MMD ($\downarrow$) & $\mathcal{W}_2$ ($\downarrow$) \\
\midrule
\multirow{2}{*}{Scenario 1} 
& MLP & 0.873 ($\pm$ 0.045) & 0.242 $(\pm$ 0.363) & 2.203 ($\pm$ 1.098) & 0.917 ($\pm$ 0.067) & 0.891 $(\pm$ 1.150) & 4.528 ($\pm$ 2.701) \\
& \textbf{Transformer} & 0.760 ($\pm$ 0.092) & 0.303 $(\pm$ 0.548) & 2.095 ($\pm$ 1.692) & 0.847 ($\pm$ 0.082) & 0.486 $(\pm$ 0.623) & 4.054 ($\pm$ 2.782)  \\
\midrule
\multirow{2}{*}{Scenario 2} 
& MLP & 0.921 ($\pm$ 0.035) & 0.291 $(\pm$ 0.205) & 2.870 ($\pm$ 0.710) & 0.992 ($\pm$ 0.005) & 0.399 $(\pm$ 0.127) & 5.505 ($\pm$ 1.144) \\
& \textbf{Transformer} & 0.812 ($\pm$ 0.061) & 0.159 $(\pm$ 0.154) & 2.314 ($\pm$ 0.926) & 0.937 ($\pm$ 0.041) & 0.282 $(\pm$ 0.131) & 3.947 ($\pm$ 1.055) \\
\midrule
\multirow{2}{*}{Scenario 3} 
& MLP & 0.999 ($\pm$ 0.000) & 0.438 $(\pm$ 0.181) & 11.502 ($\pm$ 9.719) & 1.000 ($\pm$ 0.000) & 1.001 $(\pm$ 0.149) & 26.282 ($\pm$ 3.731) \\
& \textbf{Transformer} & 0.999 ($\pm$ 0.001) & 0.267 $(\pm$ 0.154) & 7.234 ($\pm$ 2.974) & 1.000 ($\pm$ 0.000) & 1.155 $(\pm$ 0.258) & 26.956 ($\pm$ 3.114)  \\
\bottomrule
\end{tabular}
}
}
\end{table}

\textcolor{rev}{For the Gaussian Mixture Models (GMMs), the results indicate a more mixed performance where the transformer still performs slightly better (\Cref{tab:gmm_mlp}): In Scenario 1, transformer encoders slightly outperform MLPs on synthetic data, with C2ST improving from 0.873 (MLP) to 0.760 and $\mathcal{W}_2$ decreasing slightly from 2.203 to 2.095. However, on real-world data, MLPs perform marginally better in terms of MMD, reducing it from 0.486 to 0.242, while transformers show minor improvements in $\mathcal{W}_2$ from 4.528 to 4.054. In Scenario 2, transformers show a more noticeable advantage. On synthetic data, C2ST improves from 0.921 (MLP) to 0.812, and $\mathcal{W}_2$ decreases significantly from 2.870 to 2.314. On real-world data, transformers reduce C2ST from 0.992 to 0.937 and MMD from 0.399 to 0.282, along with a considerable improvement in $\mathcal{W}_2$ from 5.505 to 3.947. In Scenario 3, the differences between the two encoders are relatively small but still favor the transformers on synthetic data, with $\mathcal{W}_2$ decreasing from 11.502 (MLP) to 7.234. For real-world data, the results are nearly identical for C2ST (1.000 for both) but show a slight increase in $\mathcal{W}_2$ for the transformer from 26.282 to 26.956. Overall, for the GMMs, the transformer encoders demonstrate consistent improvements across scenarios for synthetic data, particularly in Scenarios 1 and 2. However, for real-world data, the performance differences are less pronounced.
}

%\textcolor{rev}{In the case of Gaussian Mixture Models (GMMs), the empirical results indicate that the OT paths generally outperform the VP diffusion objective across most scenarios and metrics, though the differences are not always statistically significant in pair-wise comparisons. For example, in Scenario 1, OT paths achieve notably better results for C2ST on both synthetic and real-world datasets, with reductions from 0.924 to 0.760 and from 0.958 to 0.847, respectively. Similarly, for $\mathcal{W}_2$, OT paths exhibit better performance on real-world data (4.054 vs. 5.328). In Scenario 2, OT paths maintain a consistent advantage in metrics such as C2ST and $\mathcal{W}_2$. For instance, synthetic data shows a C2ST improvement from 0.942 to 0.812, while real-world data improves from 0.984 to 0.937. The OT paths also achieve lower MMD on synthetic data (0.159 vs. 0.213), supporting their effectiveness in this scenario. For Scenario 3, the differences in performance between OT paths and diffusion paths are more nuanced. OT paths achieve better results for $\mathcal{W}_2$ on both synthetic and real-world data, reducing it from 8.708 to 7.234 and from 33.230 to 26.956, respectively. }

\newpage 
\section{\textcolor{rev}{Robustness to Out-of-distribution Data}}

\begin{figure}[t]
    \centering
    \includegraphics[width=1.0\linewidth]{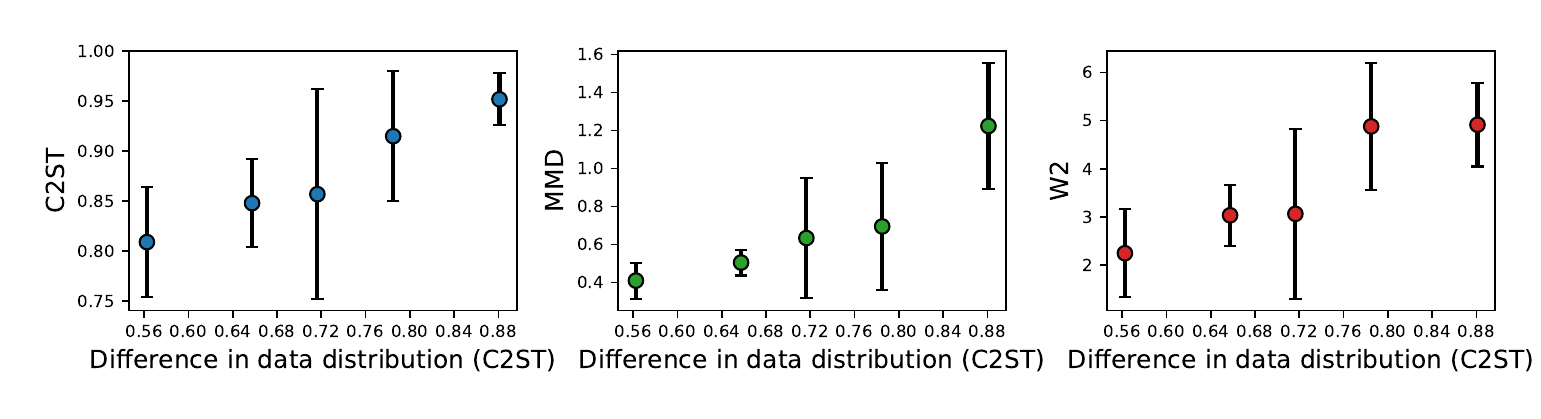}
    \caption{Out-of-distribution (OOD) performance of the ICL method in GLM Scenario 2. The $x$-axis shows the distribution shift between training and test distributions, quantified by C2ST. The $y$-axis displays the performance of the in-context learner, evaluated via C2ST, MMD, and $\mathcal{W}_2$ distances against HMC samples — where higher values indicate worse performance. OOD data is generated by gradually increasing the variance of the prior on the regression coefficients from an initial value of $1.0$ to $\sqrt{2}$, $1.7$, $2$, $2.5$, and $3.5$.}
    \label{fig:enter-label}
\end{figure}

\label{sec:ood_abl}

\textcolor{rev}{To investigate how our ICL approach behaves under mismatches between the distribution of synthetic training data and the data used to infer the posterior, we conduct an ablation study by changing aspects of the distribution of training and testing data.}

\textcolor{rev}{In summary, the results in \Cref{tab:ood_glm_res,tab:ood_fa_res,tab:ood_gmm_res} show that our ICL approach is, in most cases, capable of robustly generalizing beyond its specific pre-training distribution when various aspects of this distribution are changed. While the performance sometimes decreases when a mismatch between training and testing data occurs, the drops in performance are almost always modest and, in many cases, almost negligible.}

\subsection{\textcolor{rev}{GLM Scenarios}}

\textcolor{rev}{For scenario 2, we change the variance of the prior on the covariates from a value of $\mathbb{V}(\bm{\beta}_{i,j})= 1$ to $\mathbb{V}(\bm{\beta}_{i,j}) = 2$ for scenario 2.B and $\mathbb{V}(\bm{\beta}_{i,j}) = 4$ for scenario 2.C. In scenarios 2.D and 2.E we change the scale parameter of the prior on the variance $\sigma^2$ of the noise---thereby changing its mean from $\mathbb{E}[\sigma^2] = 0.5$ to a value of $\mathbb{E}[\sigma^2] \approx 0.7071$ for 2.D and $\mathbb{E}[\sigma^2] =  1 $ for 2.E. The variance is changed from $\mathbb{V}[\sigma^2] \approx 0.0833$ to $\mathbb{V}[\sigma^2] \approx 0.1667$ and $\mathbb{V}[\sigma^2] \approx 0.333$.}

\textcolor{rev}{For scenarios 3.B and 3.C, the variance of the coefficients is doubled from scenario 3 to scenario 3.B and from 3.B to 3.C again, analogously to scenarios 2.B and 2.C.0}

\textcolor{rev}{For scenario 5, the rate parameter of the gamma distribution is changed. This leads to a decrease in the variance from $\mathbb{V}(\bm{\beta}_{i,j})= 1$ to $\mathbb{V}(\bm{\beta}_{i,j}) = 0.5$ for scenario 5.B and $\mathbb{V}(\bm{\beta}_{i,j}) = 0.25$ for scenario 5.C. Notably, we also change the mean in the distribution of the covariates from mean from $\mathbb{E}[\bm{\beta}_{i,j}] = 1$ to a value of $\mathbb{E}[\bm{\beta}_{i,j}] \approx 0.7071$ for 2.D and $\mathbb{E}[\bm{\beta}_{i,j}] =  0.5$ for 2.E.}

\begin{table}[h]
    \centering
    \caption{\textcolor{rev}{Distribution of variables for the OOD analysis on GLM scenarios.}}
    \textcolor{rev}{
    \begin{tabular}{l|c|c|c|c}  % Removed the extra '|' after the last column
        \hline
        \textbf{Scenario} & $\beta_{i,j}$ & $\beta_{i,0}$ & $\sigma^2_i$ & $y_{i,j}|(\bm{u}_{i,j}, \bm{\beta}_i, \beta_{0,i}, \sigma^2_i)$ \\  % Added header row
        \hline 
        Scenario 2 & $\mathcal{N}(0, 1)$ & $\mathcal{N}(0, 9)$&  $\text{IG}(5,2)$ & $\mathcal{N}(\bm{u}_{i,j}^\top \bm{\beta}_i, \sigma^2_i)$ \\ 
         Scenario 2.B & $\mathcal{N}(0, 2)$ & $\mathcal{N}(0, 9)$&  $\text{IG}(5,2)$ & $\mathcal{N}(\bm{u}_{i,j}^\top \bm{\beta}_i, \sigma^2_i)$ \\ 
          Scenario 2.C & $\mathcal{N}(0, 4)$ & $\mathcal{N}(0, 9)$&  $\text{IG}(5,2)$ & $\mathcal{N}(\bm{u}_{i,j}^\top \bm{\beta}_i, \sigma^2_i)$ \\ 
          Scenario 2.D & $\mathcal{N}(0, 1)$ & $\mathcal{N}(0, 9)$&  $\text{IG}(5,2\sqrt{2})$ & $\mathcal{N}(\bm{u}_{i,j}^\top \bm{\beta}_i, \sigma^2_i)$ \\ 
          Scenario 2.E & $\mathcal{N}(0, 1)$ & $\mathcal{N}(0, 9)$&  $\text{IG}(5,4)$ & $\mathcal{N}(\bm{u}_{i,j}^\top \bm{\beta}_i, \sigma^2_i)$ \\ 
          \hline
        Scenario 3 & $\text{Laplace}(0, 1)$ & - &  $\text{IG}(5,2)$ & $\mathcal{N}(\bm{u}_{i,j}^\top \bm{\beta}_i, \sigma^2_i)$ \\ 
        Scenario 3.B & $\text{Laplace}(0, \sqrt{2})$ & - &  $\text{IG}(5,2)$ & $\mathcal{N}(\bm{u}_{i,j}^\top \bm{\beta}_i, \sigma^2_i)$ \\ 
        Scenario 3.C & $\text{Laplace}(0, 2)$ & - &  $\text{IG}(5,2)$ & $\mathcal{N}(\bm{u}_{i,j}^\top \bm{\beta}_i, \sigma^2_i)$ \\ 
        \hline
        Scenario 5 & $\text{Ga}(1, 1)$ & - & $\text{IG}(5,2)$& $\mathcal{N}(\bm{u}_{i,j}^\top \bm{\beta}_i, \sigma^2_i)$\\   
        Scenario 5.B & $\text{Ga}(1, \sqrt{2})$ & - & $\text{IG}(5,2)$& $\mathcal{N}(\bm{u}_{i,j}^\top \bm{\beta}_i, \sigma^2_i)$\\  
        Scenario 5.C & $\text{Ga}(1, 2)$ & - & $\text{IG}(5,2)$& $\mathcal{N}(\bm{u}_{i,j}^\top \bm{\beta}_i, \sigma^2_i)$\\  
        \hline
    \end{tabular}}
    \label{tab:glm_ood_setups}
\end{table}

\begin{table}[htp]
\centering
\caption{\textcolor{rev}{OOD Performance: Evaluation on 50 synthetic datasets for 8 different GLM scenarios.  All results within two standard errors of the non-OOD result for each scenario are marked in \textbf{bold}.}}
\textcolor{rev}{
\begin{tabular}{l l l l }
\toprule
\textbf{Scenario} & C2ST ($\downarrow$) & MMD ($\downarrow$) & $\mathcal{W}_2$ ($\downarrow$)  \\
\midrule
Scenario 2 &  \textbf{0.839} ($\pm$ 0.072) & \textbf{0.707} ($\pm$ 0.658) & \textbf{1.111} ($\pm$ 0.300) \\
Scenario 2.B & \textbf{0.809} ($\pm$ 0.055) & \textbf{0.410} $(\pm$ 0.095) & \textbf{2.250} ($\pm$ 0.916) \\
Scenario 2.C & \textbf{0.857} ($\pm$ 0.105) & \textbf{0.634} $(\pm$ 0.318) & \textbf{3.067} ($\pm$ 1.759) \\
\midrule
Scenario 2 &  \textbf{0.839} ($\pm$ 0.072) & \textbf{0.707} ($\pm$ 0.658) & \textbf{1.111} ($\pm$ 0.300) \\
Scenario 2.D & \textbf{0.840} ($\pm$ 0.109) & \textbf{0.916} $(\pm$ 1.123) & \textbf{4.007} ($\pm$ 3.261) \\
Scenario 2.E & \textbf{0.932} ($\pm$ 0.120) & \textbf{1.556} $(\pm$ 1.127) & \textbf{4.850} ($\pm$ 2.261) \\
\midrule
Scenario 3  & \textbf{0.611} ($\pm$ 0.070) & \textbf{0.089} ($\pm$ 0.114) & \textbf{0.423} ($\pm$ 0.348)  \\
Scenario 3.B & \textbf{0.667} ($\pm$ 0.080) & \textbf{0.210} $(\pm$ 0.117) & 1.172 ($\pm$ 0.258) \\
Scenario 3.C & \textbf{0.720} ($\pm$ 0.108) & \textbf{0.362} $(\pm$ 0.248) & 1.891 ($\pm$ 0.678) \\
\midrule
Scenario 5 & \textbf{0.621} ($\pm$ 0.063) & \textbf{0.067} ($\pm$ 0.080) & \textbf{0.299} ($\pm$ 0.195)  \\
Scenario 5.B & 0.831 ($\pm$ 0.121) & 0.479 $(\pm$ 0.200) & 1.762 ($\pm$ 0.541) \\
Scenario 5.C &  0.920 ($\pm$ 0.064) & 0.753 $(\pm$ 0.424) & 3.159 ($\pm$ 1.254) \\
\midrule
\end{tabular}
\label{tab:ood_glm_res}
}

\end{table}

\textcolor{rev}{\Cref{tab:glm_ood_setups} shows that our ICL approach only exhibits modest degradation in performance when the variance of the coefficients is doubled or quadruple while the mean stays the same (Scenarios 2.B, 2.C and 3.B, 3.C). Increasing the variance of the noise term by a factor of two only has a small effect while multiplying it by four causes a drop in C2ST by 9.3\%. However, decreasing the variance of the gamma prior in scenario 5, combined with decreasing the mean, leads to a notable drop in performance across all metrics.}

%Interestingly, when comparing the performance of the mis-specified ICL appraoch to that of the VI methods, \Cref{tab:glm_res_detail}, the average performance in 2.B, 2.C is still better than the best VI approach in that scenario (VI: Structured Normal).

\subsection{\textcolor{rev}{FA Scenarios}}

\textcolor{rev}{To construct the mismatch between training and test distribution, we vary the variance of the factor loading $W_{i,j,k}$ for scenarios 1, 2 and 3. Concretely, the variance is doubled and quadrupled.}

\begin{table}[h]
    \centering
    \caption{\textcolor{rev}{Distribution of variables for the OOD analysis on the FA scenarios.}}
    \textcolor{rev}{
    \begin{tabular}{l|c|c|c|c|c|c|c}  % Removed the extra '|' after the last column
        \hline
        \textbf{Scenario} & $K$ & $P$ & $\mu_{i,j}$ & $\Psi_{i,j,j}$ & $W_{i,j,k}$ & $z_{i,j}$ & $\bm{z}_{dim}$ \\  % Added header row
        \hline
        Scenario 1 & $50$ & $3$ & $\mathcal{N}(0,1)$ & $\text{IG}(5,1)$ & $\mathcal{N}(0,1)$ & $\mathcal{N}(0,1)$ & $3$ \\  
        Scenario 1.B & $50$ & $3$ & $\mathcal{N}(0,1)$ & $\text{IG}(5,1)$ & $\mathcal{N}(0,2)$ & $\mathcal{N}(0,1)$ & $3$ \\  
        Scenario 1.C & $50$ & $3$ & $\mathcal{N}(0,1)$ & $\text{IG}(5,1)$ & $\mathcal{N}(0,4)$ & $\mathcal{N}(0,1)$ & $3$ \\  
        \midrule
        Scenario 2 & $50$ & $3$ & $\mathcal{N}(0,0.1)$ & $\text{IG}(5,1)$ & $\text{Laplace}(0,10)$ & $\mathcal{N}(0,1)$ & $3$ \\  
        Scenario 2.B & $50$ & $3$ & $\mathcal{N}(0,0.1)$ & $\text{IG}(5,1)$ & $\text{Laplace}(0,10 \cdot \sqrt{2})$ & $\mathcal{N}(0,1)$ & $3$ \\  
        Scenario 2.C & $50$ & $3$ & $\mathcal{N}(0,0.1)$ & $\text{IG}(5,1)$ & $\text{Laplace}(0,20)$ & $\mathcal{N}(0,1)$ & $3$ \\  
        \midrule
        Scenario 3 & $25$ & $5$ & $\mathcal{N}(0,0.1)$ & $\text{IG}(5,2)$ & $\mathcal{N}(0,3)$ & $\mathcal{N}(0,1)$ & $3$ \\
        Scenario 3 & $25$ & $5$ & $\mathcal{N}(0,0.1)$ & $\text{IG}(5,2)$ & $\mathcal{N}(0,3 \cdot \sqrt{2})$ & $\mathcal{N}(0,1)$ & $3$ \\ 
        Scenario 3 & $25$ & $5$ & $\mathcal{N}(0,0.1)$ & $\text{IG}(5,2)$ & $\mathcal{N}(0,6)$ & $\mathcal{N}(0,1)$ & $3$ \\ 
        \hline
    \end{tabular}}
    \label{tab:fa_variables_ood}
\end{table}

\begin{table}[htp]
\centering
\caption{\textcolor{rev}{OOD Performance: Evaluation on 50 synthetic datasets for 6 different FA scenarios.  All results within two standard errors of the non-OOD result for each scenario are marked in \textbf{bold}.}}
\label{tab:fa_conf_ood}

\textcolor{rev}{
\begin{tabular}{l l l l }
\toprule
\textbf{Scenario} & C2ST ($\downarrow$) & MMD ($\downarrow$) & $\mathcal{W}_2$ ($\downarrow$)  \\
\midrule
Scenario 1 &  \textbf{0.552} ($\pm$ 0.028) & \textbf{0.034} $(\pm$ 0.034) & \textbf{0.289} ($\pm$ 0.083)  \\
Scenario 1.B &  0.826 ($\pm$ 0.066) & 0.656 $(\pm$ 0.384) & 0.929 ($\pm$ 0.321) \\
Scenario 1.C & 0.855 ($\pm$ 0.060) & 0.837 $(\pm$ 0.494) & 1.135 ($\pm$ 0.461) \\
\midrule
Scenario 2  & \textbf{0.542} ($\pm$ 0.006) & \textbf{0.017} $(\pm$ 0.006) & \textbf{0.244} ($\pm$ 0.033) \\
Scenario 2.B  & 0.580 ($\pm$ 0.069) & 0.087 $(\pm$ 0.191) & 0.393 ($\pm$ 0.291) \\
Scenario 2.C & 0.589 ($\pm$ 0.076) & 0.089 $(\pm$ 0.113) & 0.446 ($\pm$ 0.233) \\
\midrule
Scenario 3 & \textbf{0.537} ($\pm$ 0.023) & \textbf{0.024} $(\pm$ 0.021) & \textbf{0.259} ($\pm$ 0.088)  \\
Scenario 3.B & \textbf{0.544} ($\pm$ 0.028) & 0.030 $(\pm$ 0.021) & \textbf{0.285} ($\pm$ 0.094) \\
Scenario 3.C & \textbf{0.533} ($\pm$ 0.025) & 0.021 $(\pm$ 0.015) & \textbf{0.347} ($\pm$ 0.152) \\
\midrule
\end{tabular}
\label{tab:ood_fa_res}
}

\end{table}

\textcolor{rev}{For the FA cases (refer to \Cref{tab:ood_fa_res}), there is a notable drop in performance in the first scenario when OOD data is used. Please note that even in the most misspecified scenario (1.C), the performance, as measured in C2ST is still around ten percent better than the best VI method in this scenario (\Cref{tab:fa_res_detail}). While the absolute difference between performance on the training distribution and the test distribution is very small for scenarios 2 and 3, the difference is still not within two standard errors of the non-OOD performance because the standard error itself is quite small. The performance on the OOD data is still better than all other VI methods (see \Cref{tab:FA_summary}).}

\subsection{\textcolor{rev}{GMM Scenarios}}

\textcolor{rev}{To generate several distinct OOD scenarios based on the generative processes of GMMs, we vary scenario 2 in various ways. Note that the structure of the distributions is the same for all GMM scenarios---focusing on this specific scenario thus makes sense when considering OOD generalization. First, in scenario 2.B, we decrease the symmetric parameter of the Dirichlet prior on the assignments from 1 to 0.5 causing larger discrepancy in the number of points per cluster. In scenario 2.C we make the opposite change.}

\textcolor{rev}{In scenarios 2.D and 2.E we first double and then quadruple the variance of the prior on the per-component variances $\sigma_{i,m,l}$. Finally, in scenarios 2.F and 2.G, the prior on the mean is made more dispersed compared to the training data.}

\begin{table}[h]
    \centering
    \caption{\textcolor{rev}{Distribution for the OOD analysis of the GMM scenarios.}}   \label{tab:gmm_ood}
    \textcolor{rev}{
    \begin{tabular}{l|c|c|c|c|c|c}  % Removed the extra '|' after the last column
        \hline
        \textbf{Scenario} & $K$  & $M$ & $L$ & $\bm{\phi}_i$ & $\sigma^2_{i,m,l}$ & $\mu_{i,m,l}|\sigma^2_{i,m,l}$ \\
        \hline
          % Added header row
        Scenario 2& $25$ & $3$ & $3$ & $\text{Dir}(1)$ & $\text{IG}(5,2)$ & $\mathcal{N}(0, 3 \sigma^2_{i,m,l}) $ \\ 
        \midrule
        Scenario 2.B& $25$ & $3$ & $3$ & $\text{Dir}(0.5)$ & $\text{IG}(5,2)$ & $\mathcal{N}(0, 3 \sigma^2_{i,m,l}) $ \\  
        Scenario 2.C& $25$ & $3$ & $3$ & $\text{Dir}(2)$ & $\text{IG}(5,2)$ & $\mathcal{N}(0, 3 \sigma^2_{i,m,l}) $ \\   
        \midrule
        Scenario 2.D& $25$ & $3$ & $3$ & $\text{Dir}(1)$ & $\text{IG}(5,2 \cdot  \sqrt{2})$ & $\mathcal{N}(0, 3 \sigma^2_{i,m,l}) $ \\  
        Scenario 2.E& $25$ & $3$ & $3$ & $\text{Dir}(1)$ & $\text{IG}(5,4)$ & $\mathcal{N}(0, 3 \sigma^2_{i,m,l}) $ \\ 
        \midrule
        Scenario 2.F& $25$ & $3$ & $3$ & $\text{Dir}(1)$ & $\text{IG}(5,2)$ & $\mathcal{N}(0, 4 \sigma^2_{i,m,l}) $ \\ 
        Scenario 2.G& $25$ & $3$ & $3$ & $\text{Dir}(1)$ & $\text{IG}(5,2)$ & $\mathcal{N}(0, 5 \sigma^2_{i,m,l}) $ \\ 
        \hline
    \end{tabular}}
    \label{tab:gmm_ood_conf}
\end{table}

\begin{table}[htp]
\centering
\caption{\textcolor{rev}{OOD Performance: Evaluation on 50 synthetic datasets for 6 different GMM scenarios.  All results within two standard errors of the non-OOD result for each scenario are marked in \textbf{bold}.}}
\label{tab:ood_gmm}

\textcolor{rev}{
\begin{tabular}{l l l l }
\toprule
\textbf{Scenario} & C2ST ($\downarrow$) & MMD ($\downarrow$) & $\mathcal{W}_2$ ($\downarrow$)  \\
\midrule
Scenario 2  & \textbf{0.812} ($\pm$ 0.061) & \textbf{0.159} $(\pm$ 0.154) & \textbf{2.314} ($\pm$ 0.926) \\
Scenario 2.B & \textbf{0.829} ($\pm$ 0.050) & \textbf{0.233} $(\pm$ 0.161) & \textbf{2.595} ($\pm$ 0.998) \\
Scenario 2.C & \textbf{0.816} ($\pm$ 0.057) & 0.149 $(\pm$ 0.135) & 2.272 ($\pm$ 0.654)  \\
\midrule
Scenario 2  & \textbf{0.812} ($\pm$ 0.061) & \textbf{0.159} $(\pm$ 0.154) & \textbf{2.314} ($\pm$ 0.926) \\
Scenario 2.D & \textbf{0.812} ($\pm$ 0.076) & \textbf{0.148} $(\pm$ 0.091) & \textbf{2.557} ($\pm$ 0.837) \\
Scenario 2.E & \textbf{0.880} ($\pm$ 0.057) & \textbf{0.231} $(\pm$ 0.109) & \textbf{3.535} ($\pm$ 1.003) \\
\midrule
Scenario 2  & \textbf{0.812} ($\pm$ 0.061) & \textbf{0.159} $(\pm$ 0.154) & \textbf{2.314} ($\pm$ 0.926) \\
Scenario 2.F &  \textbf{0.821} ($\pm$ 0.076) & \textbf{0.216} $(\pm$ 0.214) & \textbf{2.700} ($\pm$ 1.044 \\
Scenario 2.G  & \textbf{0.844} ($\pm$ 0.046) & \textbf{0.197} $(\pm$ 0.124) & \textbf{2.675} ($\pm$ 0.552)\\
\midrule
\end{tabular}
\label{tab:ood_gmm_res}
}

\end{table}

On the GMM scenarios (\Cref{tab:ood_gmm_res}), the sample quality obtained via ICL is surprisingly stable under various changes to the data-generating process. It is relatively unsurprising that changing the Dirichlet prior, i.e., making the cluster more or less uniform in their number of samples, might lead to cases the ICL method can generalize to relatively easily, as demonstrated in scenarios 2.B and 2.C. The most pronounced drop in performance results from increasing the variance of the prior on the standard deviation of the components of the mixture model (scenario 2.E), while increasing the variance of the mean vector relative to the standard deviation of the components has a less pronounced effect.

\clearpage

\section{Ablation: Dimensionality of the Problem}
\label{sec:ablation_dimensionality}

In this section, the effect of the dimensionality $K$ of the latent variable $\vz \in \mathbb{R}^K$ for the GLM scenarios is investigated. 

In summary, our results show that forward-KL based VI and in particular MAP solutions perform strongly in terms of predictive performance, which is in line with results first presented by \citet{mittal2025amortized, mittal2025context}. 

In table \ref{tab:Dim_v1}, we find that the advantages of the in-context learning approach to deteriorate for higher dimensionalities, with the variational inference methods using a Gaussian approximation performing well for 20 dimensions. This finding is line with work by \citet{mittal2025amortized, mittal2025context}. For 50 dimensions we find that in many cases the used metrics do not allow to significantly discriminate the performance of the different approaches. Note that the randomness in the results, especially for higher dimensionalities, can in rare cases lead to better mean values. This is most likely not significant when taking the standard error into account.

Similar to scenarios 1,2 and 3, we find that in scenarios 4,5 and 6 (Table \ref{tab:Dim_v2}) the advantages of the in-context learning approach to deteriorate for higher dimensionalities, with the variational inference methods using a Gaussian approximation performing well for 20 dimensions. For 50 dimensions we find that in many cases the used metrics do not allow to significantly discriminate the performance of the different approaches. Note that the randomness in the results, especially for higher dimensionalities, can in rare cases lead to better mean values. This is most likely not significant when taking the standard error into account. 

Finally, the results in Table \ref{Tab:Dim_v3} show that also for this scenario 7, the advantages of the in-context learning approach to deteriorate for higher dimensionalities. However, in this specific scenario the in-context learner is not significantly different from the VI methods in terms of C2ST and MMD for 20 dimensions. For 50 dimensions we find that the VI method using IAF performs well, together with the in-context learning approach in terms of MMD while the C2ST score does not indicate a clear winner and $\mathcal{W}_2$ favors the other methods. Note that the randomness in the results, especially for higher dimensionalities, can in rare cases lead to better mean values. This is most likely not significant when taking the standard error into account.

\begin{table}[htp]
\centering
\caption{Generalized Linear Models: Ablation with respect to the dimensionality of the problem on 50 synthetic and 17 real-world datasets for scenarios 1,2 and 3.  All results within two standard errors of the best average result for each scenario are marked in \textbf{bold}. Due to the limitations of the number of features in the real-world data, we can only use 5 datasets for 20 and one dataset for 50 dimensions.  }
\label{tab:glm_res_detail}
\resizebox{1\textwidth}{!}{
\begin{tabular}{p{1.5cm} p{1.5cm} p{3.5cm} m{2.4cm}m{2.4cm}m{2.4cm}| m{2.4cm}m{2.4cm}m{2.4cm}}
\toprule
 \multicolumn{1}{c}{\multirow[c]{2}{*}[-0.5ex]{\textbf{Scenario}}} & \multicolumn{1}{c}{\multirow[c]{2}{*}[-0.5ex]{\textbf{Dim.}}} & \multicolumn{1}{c}{\multirow{2}{*}[-0.5ex]{\textbf{Model}}} & \multicolumn{3}{c}{\textbf{Synthetic Evaluation}} & \multicolumn{3}{c}{\textbf{Real-World Evaluation}} \\
\cmidrule(lr){4-6} \cmidrule(lr){7-9}
& & & C2ST ($\downarrow$) & MMD ($\downarrow$) & $\mathcal{W}_2$ ($\downarrow$) & C2ST ($\downarrow$) & MMD ($\downarrow$) & $\mathcal{W}_2$ ($\downarrow$) \\
\midrule
\multirow{6}{*}{Scenario 1} & \multirow{6}{*}{\hspace{0.3cm} 5} & Laplace Approximation & 1.000 ($\pm$ 0.000) & 2.738 $(\pm$ 0.721) & \textbf{0.825} ($\pm$ 0.279) & 1.000 ($\pm$ 0.000) & 2.150 $(\pm$ 0.323) & \textbf{0.642} ($\pm$ 0.124) \\
& & VI: DiagonalNormal & 0.904 ($\pm$ 0.076) & 1.452 $(\pm$ 0.984) & \textbf{0.669} ($\pm$ 0.301) & 0.797 ($\pm$ 0.083) & 0.612 $(\pm$ 0.511) & \textbf{0.414} ($\pm$ 0.152) \\
& & VI: MultivariateNormal & \textbf{0.750} ($\pm$ 0.128) & \textbf{0.735} $(\pm$ 0.733) & \textbf{0.565} ($\pm$ 0.292) & \textbf{0.607} ($\pm$ 0.070) & \textbf{0.167} $(\pm$ 0.196) & \textbf{0.301} ($\pm$ 0.123) \\
& & VI: Structured Normal & \textbf{0.753} ($\pm$ 0.126) & \textbf{0.736} $(\pm$ 0.737) & \textbf{0.570} ($\pm$ 0.310) & \textbf{0.600} ($\pm$ 0.070) & \textbf{0.169} $(\pm$ 0.214) & \textbf{0.306} ($\pm$ 0.131) \\
& & VI: IAF & \textbf{0.777} ($\pm$ 0.122) & \textbf{0.864} $(\pm$ 0.844) & 0.725 ($\pm$ 0.523) & 0.683 ($\pm$ 0.132) & 0.440 $(\pm$ 0.559) & 0.503 ($\pm$ 0.383) \\
& & HMC & \textbf{0.745} ($\pm$ 0.130) & \textbf{0.722} $(\pm$ 0.732) & \textbf{0.569} ($\pm$ 0.301) & \textbf{0.595} ($\pm$ 0.075) & \textbf{0.173} $(\pm$ 0.213) & \textbf{0.321} ($\pm$ 0.140) \\
& & \textbf{ICL (ours)} & \textbf{0.765} ($\pm$ 0.123) & \textbf{0.767} $(\pm$ 0.727) & \textbf{0.585} ($\pm$ 0.301) & \textbf{0.614} ($\pm$ 0.074) & \textbf{0.175} $(\pm$ 0.219) & \textbf{0.310} ($\pm$ 0.138) \\
\midrule

\multirow{6}{*}{Scenario 1} & \multirow{6}{*}{\hspace{0.3cm} 20} & Laplace Approximation & 1.000 ($\pm$ 0.000) & 2.237 ($\pm$ 0.024) & \textbf{3.252} ($\pm$ 1.172) & 1.000 ($\pm$ 0.000) & 2.206 ($\pm$ 0.021) & 2.792 ($\pm$ 0.339) \\
& & VI: DiagonalNormal & \textbf{0.843} ($\pm$ 0.204) & 1.056 ($\pm$ 0.869) & \textbf{2.976} ($\pm$ 0.927) & 0.983 ($\pm$ 0.019) & 1.217 ($\pm$ 0.463) & 2.406 ($\pm$ 0.348) \\
& & VI: MultivariateNormal & \textbf{0.789} ($\pm$ 0.140) & \textbf{0.714} ($\pm$ 0.351) & \textbf{2.774} ($\pm$ 0.995) & \textbf{0.768} ($\pm$ 0.120) & \textbf{0.180} ($\pm$ 0.159) & \textbf{2.064} ($\pm$ 0.306) \\
& & VI: Structured Normal & \textbf{0.792} ($\pm$ 0.109) & \textbf{0.708} ($\pm$ 0.153) & \textbf{2.703} ($\pm$ 1.069) & \textbf{0.668} ($\pm$ 0.080) & \textbf{0.168} ($\pm$ 0.071) & \textbf{2.052} ($\pm$ 0.275) \\
& & VI: IAF & \textbf{0.832} ($\pm$ 0.196) & \textbf{0.847} ($\pm$ 1.016) & \textbf{4.015} ($\pm$ 0.415) & 0.939 ($\pm$ 0.024) & 0.508 ($\pm$ 0.200) & 3.140 ($\pm$ 0.290) \\
& & {ICL (ours)} & \textbf{0.849} ($\pm$ 0.171) & \textbf{0.844} ($\pm$ 0.905) & 4.564 ($\pm$ 0.622) & 0.970 ($\pm$ 0.020) & 0.724 ($\pm$ 0.287) & 4.250 ($\pm$ 0.312) \\
\midrule

\multirow{6}{*}{Scenario 1} & \multirow{6}{*}{\hspace{0.3cm} 50} & Laplace Approximation & 1.000 ($\pm$ 0.000) & 2.401 ($\pm$ 0.282) & \textbf{5.152} ($\pm$ 2.268) & 1.000 ($\pm$ nan) & 2.339 ($\pm$ nan) & 6.642 ($\pm$ nan) \\
& & VI: DiagonalNormal & \textbf{0.812} ($\pm$ 0.197) & \textbf{0.956} ($\pm$ 1.016) & \textbf{5.541} ($\pm$ 2.356) & 0.915 ($\pm$ nan) & 0.508 ($\pm$ nan) & 6.200 ($\pm$ nan) \\
& & VI: MultivariateNormal & \textbf{0.839} ($\pm$ 0.148) & \textbf{0.926} ($\pm$ 0.682) & \textbf{5.514} ($\pm$ 2.370) & 0.905 ($\pm$ nan) & 0.790 ($\pm$ nan) & 6.258 ($\pm$ nan) \\
& & VI: Structured Normal & \textbf{0.823} ($\pm$ 0.160) & \textbf{0.844} ($\pm$ 0.480) & \textbf{5.752} ($\pm$ 2.098) & 0.910 ($\pm$ nan) & 1.122 ($\pm$ nan) & 6.898 ($\pm$ nan) \\
& & VI: IAF & \textbf{0.820} ($\pm$ 0.182) & \textbf{0.814} ($\pm$ 0.987) & \textbf{6.696} ($\pm$ 1.207) & 0.938 ($\pm$ nan) & 0.256 ($\pm$ nan) & 6.869 ($\pm$ nan) \\
& & {ICL (ours)} & \textbf{0.787} ($\pm$ 0.217) & \textbf{1.015} ($\pm$ 1.255) & \textbf{8.278} ($\pm$ 0.821) & 0.979 ($\pm$ nan) & 0.413 ($\pm$ nan) & 8.368 ($\pm$ nan) \\
\midrule

\multirow{6}{*}{Scenario 2} & \multirow{6}{*}{\hspace{0.3cm} 5} & Laplace Approximation & 1.000 ($\pm$ 0.000) & 4.853 $(\pm$ 2.333) & 5.770 ($\pm$ 5.946) & 1.000 ($\pm$ 0.000) & 2.572 $(\pm$ 0.206) & 0.809 ($\pm$ 0.149) \\
& & VI: DiagonalNormal & 0.957 ($\pm$ 0.091) & 3.906 $(\pm$ 2.679) & 5.628 ($\pm$ 6.092) & 0.892 ($\pm$ 0.044) & 0.847 $(\pm$ 0.389) & \textbf{0.530} ($\pm$ 0.175) \\
& & VI: MultivariateNormal & 0.910 ($\pm$ 0.131) & 3.407 $(\pm$ 2.781) & 5.584 ($\pm$ 6.104) & 0.820 ($\pm$ 0.031) & 0.243 $(\pm$ 0.148) & \textbf{0.408} ($\pm$ 0.118) \\
& & VI: Structured Normal & 0.908 ($\pm$ 0.119) & 3.139 $(\pm$ 2.763) & 5.480 ($\pm$ 6.164) & 0.824 ($\pm$ 0.023) & 0.215 $(\pm$ 0.110) & \textbf{0.392} ($\pm$ 0.109) \\
& & VI: IAF & 0.968 ($\pm$ 0.063) & 4.416 $(\pm$ 2.473) & 7.474 ($\pm$ 6.235) & 0.888 ($\pm$ 0.067) & 0.921 $(\pm$ 0.860) & 0.942 ($\pm$ 0.733) \\
& & {ICL (ours)} & \textbf{0.839} ($\pm$ 0.072) & \textbf{0.707} $(\pm$ 0.658) & \textbf{1.111} ($\pm$ 0.300) & \textbf{0.768} ($\pm$ 0.033) & \textbf{0.143} $(\pm$ 0.089) & \textbf{0.411} ($\pm$ 0.094) \\
\midrule
%\multirow{6}{*}{Scenario 2} & \multirow{6}{*}{\hspace{0.3cm} 10} & Laplace Approximation & 1.000 ($\pm$ 0.000) & 2.182 ($\pm$ 0.074) & 2.089 ($\pm$ 0.913) & 1.000 ($\pm$ 0.000) & 2.088 ($\pm$ 0.039) & 1.430 ($\pm$ 0.159) \\
%& & VI: DiagonalNormal & 0.831 ($\pm$ 0.206) & 0.912 ($\pm$ 0.762) & \textbf{1.574} ($\pm$ 0.597) & 0.973 ($\pm$ 0.010) & 1.209 ($\pm$ 0.376) & 1.195 ($\pm$ 0.213) \\
%& & VI: MultivariateNormal & \textbf{0.733} ($\pm$ 0.131) & \textbf{0.182} ($\pm$ 0.248) & \textbf{1.419} ($\pm$ 0.595) & \textbf{0.709} ($\pm$ 0.055) & \textbf{0.096} ($\pm$ 0.111) & \textbf{0.848} ($\pm$ 0.129) \\
%& & VI: Structured Normal & \textbf{0.683} ($\pm$ 0.039) & \textbf{0.041} ($\pm$ 0.032) & \textbf{1.339} ($\pm$ 0.567) & \textbf{0.676} ($\pm$ 0.021) & \textbf{0.036} ($\pm$ 0.030) & \textbf{0.844} ($\pm$ 0.108) \\
%& & VI: IAF & 0.909 ($\pm$ 0.108) & 1.178 ($\pm$ 1.142) & 2.702 ($\pm$ 0.611) & 0.957 ($\pm$ 0.030) & 1.120 ($\pm$ 0.445) & 2.238 ($\pm$ 0.463) \\
%& & {ICL (ours)} & 0.940 ($\pm$ 0.037) & 1.106 ($\pm$ 0.936) & 2.816 ($\pm$ 0.494) & 0.951 ($\pm$ 0.021) & 1.112 ($\pm$ 0.358) & 2.490 ($\pm$ 0.389) \\
%\midrule

\multirow{6}{*}{Scenario 2} & \multirow{6}{*}{\hspace{0.3cm} 20} & Laplace Approximation & 1.000 ($\pm$ 0.000) & 2.314 ($\pm$ 0.237) & 3.069 ($\pm$ 1.168) & 1.000 ($\pm$ 0.000) & 2.222 ($\pm$ 0.018) & 2.847 ($\pm$ 0.305) \\
& & VI: DiagonalNormal & 0.904 ($\pm$ 0.168) & 1.292 ($\pm$ 0.937) & \textbf{2.863} ($\pm$ 0.919) & 0.990 ($\pm$ 0.009) & 1.277 ($\pm$ 0.452) & 2.483 ($\pm$ 0.318) \\
& & VI: MultivariateNormal & 0.851 ($\pm$ 0.134) & \textbf{0.492} ($\pm$ 0.547) & \textbf{2.694} ($\pm$ 0.916) & 0.843 ($\pm$ 0.069) & 0.243 ($\pm$ 0.170) & \textbf{2.166} ($\pm$ 0.266) \\
& & VI: Structured Normal & \textbf{0.697} ($\pm$ 0.065) & \textbf{0.070} ($\pm$ 0.099) & \textbf{2.497} ($\pm$ 0.993) & \textbf{0.655} ($\pm$ 0.031) & \textbf{0.029} ($\pm$ 0.025) & \textbf{2.191} ($\pm$ 0.271) \\
& & VI: IAF & 0.916 ($\pm$ 0.110) & 1.062 ($\pm$ 1.076) & 4.191 ($\pm$ 0.623) & 0.952 ($\pm$ 0.025) & 0.515 ($\pm$ 0.242) & 3.331 ($\pm$ 0.371) \\
& & {ICL (ours)} & 0.955 ($\pm$ 0.057) & 1.131 ($\pm$ 1.035) & 4.945 ($\pm$ 0.836) & 0.968 ($\pm$ 0.020) & 0.724 ($\pm$ 0.278) & 4.356 ($\pm$ 0.302) \\
\midrule

\multirow{6}{*}{Scenario 2} & \multirow{6}{*}{\hspace{0.3cm} 50} & Laplace Approximation & \textbf{1.000} ($\pm$ 0.000) & 2.437 ($\pm$ 0.271) & \textbf{5.728} ($\pm$ 1.358) & 1.000 ($\pm$ nan) & 2.350 ($\pm$ nan) & 5.620 ($\pm$ nan) \\
& & VI: DiagonalNormal & \textbf{0.853} ($\pm$ 0.182) & 0.787 ($\pm$ 0.687) & \textbf{6.224} ($\pm$ 1.225) & 0.996 ($\pm$ nan) & 1.080 ($\pm$ nan) & 5.426 ($\pm$ nan) \\
& & VI: MultivariateNormal & \textbf{0.878} ($\pm$ 0.150) & \textbf{0.688} ($\pm$ 0.620) & \textbf{6.206} ($\pm$ 1.244) & 0.994 ($\pm$ nan) & 0.791 ($\pm$ nan) & {5.305} ($\pm$ nan) \\
& & VI: Structured Normal & \textbf{0.865} ($\pm$ 0.081) & \textbf{0.186} ($\pm$ 0.169) & {5.874} ($\pm$ 1.233) & {0.819} ($\pm$ nan) & {0.093} ($\pm$ nan) & {5.660} ($\pm$ nan) \\
& & VI: IAF & \textbf{0.909} ($\pm$ 0.130) & 0.649 ($\pm$ 0.650) & 7.465 ($\pm$ 0.335) & 0.985 ($\pm$ nan) & 0.426 ($\pm$ nan) & 6.426 ($\pm$ nan) \\
& & {ICL (ours)} & \textbf{0.972} ($\pm$ 0.039) & 0.741 ($\pm$ 0.713) & 8.313 ($\pm$ 0.608) & 0.971 ($\pm$ nan) & 0.405 ($\pm$ nan) & 7.718 ($\pm$ nan) \\
\midrule

\multirow{6}{*}{Scenario 3} & \multirow{6}{*}{\hspace{0.3cm} 5} & Laplace Approximation & 1.000 ($\pm$ 0.000) & 2.203 ($\pm$ 0.997) & 1.170 ($\pm$ 0.949) & 1.000 ($\pm$ 0.000) & 1.841 ($\pm$ 0.185) & 0.729 ($\pm$ 0.175) \\
& & VI: DiagonalNormal & 0.866 ($\pm$ 0.101) & 1.069 ($\pm$ 1.150) & 0.846 ($\pm$ 0.747) & 0.797 ($\pm$ 0.083) & 0.526 ($\pm$ 0.361) & 0.480 ($\pm$ 0.207) \\
& & VI: MultivariateNormal & \textbf{0.656} ($\pm$ 0.131) & \textbf{0.445} ($\pm$ 1.061) & \textbf{0.660} ($\pm$ 0.737) & \textbf{0.560} ($\pm$ 0.035) & \textbf{0.032} ($\pm$ 0.028) & \textbf{0.249} ($\pm$ 0.069) \\
& & VI: Structured Normal & \textbf{0.653} ($\pm$ 0.125) & \textbf{0.421} ($\pm$ 0.993) & \textbf{0.659} ($\pm$ 0.736) & \textbf{0.552} ($\pm$ 0.028) & \textbf{0.027} ($\pm$ 0.015) & \textbf{0.239} ($\pm$ 0.055) \\
& & VI: IAF & 0.751 ($\pm$ 0.148) & 0.939 ($\pm$ 1.349) & 0.964 ($\pm$ 0.924) & 0.673 ($\pm$ 0.141) & 0.399 ($\pm$ 0.543) & 0.563 ($\pm$ 0.433) \\
& & {ICL (ours)} & \textbf{0.611} ($\pm$ 0.070) & \textbf{0.089} ($\pm$ 0.114) & \textbf{0.423} ($\pm$ 0.348) & \textbf{0.576} ($\pm$ 0.027) & \textbf{0.037} ($\pm$ 0.026) & \textbf{0.257} ($\pm$ 0.044) \\
\midrule

%\multirow{6}{*}{Scenario 3} & \multirow{6}{*}{\hspace{0.3cm} 10} & Laplace Approximation & 1.000 ($\pm$ 0.000) & 2.142 ($\pm$ 0.486) & 2.529 ($\pm$ 1.498) & 1.000 ($\pm$ 0.000) & 2.018 ($\pm$ 0.055) & 1.558 ($\pm$ 0.296) \\
%& & VI: DiagonalNormal & 0.858 ($\pm$ 0.149) & 0.960 ($\pm$ 1.131) & \textbf{1.951} ($\pm$ 1.093) & 0.938 ($\pm$ 0.028) & 1.152 ($\pm$ 0.631) & \textbf{1.376} ($\pm$ 0.558) \\
%& & VI: MultivariateNormal & \textbf{0.691} ($\pm$ 0.108) & \textbf{0.236} ($\pm$ 0.546) & \textbf{1.695} ($\pm$ 1.095) & \textbf{0.632} ($\pm$ 0.077) & \textbf{0.143} ($\pm$ 0.252) & \textbf{0.915} ($\pm$ 0.380) \\
%& & VI: Structured Normal & \textbf{0.639} ($\pm$ 0.081) & \textbf{0.080} ($\pm$ 0.176) & \textbf{1.559} ($\pm$ 1.010) & \textbf{0.607} ($\pm$ 0.058) & \textbf{0.097} ($\pm$ 0.190) & \textbf{0.868} ($\pm$ 0.334) \\
%& & VI: IAF & 0.865 ($\pm$ 0.157) & 1.384 ($\pm$ 1.347) & 3.180 ($\pm$ 1.040) & 0.943 ($\pm$ 0.044) & 1.051 ($\pm$ 0.550) & 2.240 ($\pm$ 0.704) \\
%& & {ICL (ours)} & \textbf{0.787} ($\pm$ 0.154) & \textbf{0.824} ($\pm$ 0.722) & 3.556 ($\pm$ 0.833) & 0.883 ($\pm$ 0.017) & 0.906 ($\pm$ 0.146) & 3.335 ($\pm$ 0.371) \\
%\midrule

\multirow{6}{*}{Scenario 3} & \multirow{6}{*}{\hspace{0.3cm} 20} & Laplace Approximation & 1.000 ($\pm$ 0.000) & 2.726 ($\pm$ 1.116) & 4.127 ($\pm$ 1.927) & 1.000 ($\pm$ 0.000) & 2.234 ($\pm$ 0.092) & 3.589 ($\pm$ 0.519) \\
& & VI: DiagonalNormal & \textbf{0.912} ($\pm$ 0.134) & \textbf{1.704} ($\pm$ 1.467) & \textbf{3.933} ($\pm$ 1.574) & 0.983 ($\pm$ 0.014) & 1.298 ($\pm$ 0.443) & 3.147 ($\pm$ 0.557) \\
& & VI: MultivariateNormal & \textbf{0.863} ($\pm$ 0.113) & \textbf{0.937} ($\pm$ 1.174) & \textbf{3.754} ($\pm$ 1.650) & \textbf{0.796} ($\pm$ 0.099) & \textbf{0.268} ($\pm$ 0.226) & \textbf{2.645} ($\pm$ 0.466) \\
& & VI: Structured Normal & \textbf{0.768} ($\pm$ 0.109) & \textbf{0.302} ($\pm$ 0.518) & \textbf{3.151} ($\pm$ 1.663) & \textbf{0.722} ($\pm$ 0.073) & \textbf{0.131} ($\pm$ 0.141) & \textbf{2.579} ($\pm$ 0.399) \\
& & VI: IAF & \textbf{0.908} ($\pm$ 0.133) & \textbf{1.657} ($\pm$ 1.476) & 5.543 ($\pm$ 1.120) & 0.936 ($\pm$ 0.041) & 0.548 ($\pm$ 0.341) & 3.678 ($\pm$ 0.670) \\
& & {ICL (ours)} & \textbf{0.902} ($\pm$ 0.076) & \textbf{1.053} ($\pm$ 0.782) & 6.206 ($\pm$ 0.783) & 0.932 ($\pm$ 0.019) & 0.635 ($\pm$ 0.183) & 5.281 ($\pm$ 0.317) \\
\midrule

\multirow{6}{*}{Scenario 3} & \multirow{6}{*}{\hspace{0.3cm} 50} & Laplace Approximation & \textbf{1.000} ($\pm$ 0.000) & 2.700 ($\pm$ 0.789) & \textbf{8.841} ($\pm$ 1.691) & 1.000 ($\pm$ nan) & 2.348 ($\pm$ nan) & 7.049 ($\pm$ nan) \\
& & VI: DiagonalNormal & \textbf{0.870} ($\pm$ 0.127) & \textbf{1.154} ($\pm$ 1.321) & \textbf{9.180} ($\pm$ 1.513) & 0.997 ($\pm$ nan) & 1.393 ($\pm$ nan) & 6.791 ($\pm$ nan) \\
& & VI: MultivariateNormal & \textbf{0.896} ($\pm$ 0.101) & \textbf{1.027} ($\pm$ 1.157) & \textbf{9.175} ($\pm$ 1.555) & 0.998 ($\pm$ nan) & 1.092 ($\pm$ nan) & {6.667} ($\pm$ nan) \\
& & VI: Structured Normal & \textbf{0.873} ($\pm$ 0.112) & \textbf{0.539} ($\pm$ 0.667) & \textbf{9.118} ($\pm$ 1.538) & {0.958} ($\pm$ nan) & {0.420} ($\pm$ nan) & {6.665} ($\pm$ nan) \\
& & VI: IAF & \textbf{0.869} ($\pm$ 0.124) & \textbf{0.751} ($\pm$ 0.939) & \textbf{9.917} ($\pm$ 0.870) & 0.971 ($\pm$ nan) & 0.417 ($\pm$ nan) & 7.411 ($\pm$ nan) \\
& & {ICL (ours)} & \textbf{0.931} ($\pm$ 0.062) & \textbf{0.784} ($\pm$ 0.884) & 10.063 ($\pm$ 0.930) & 0.965 ($\pm$ nan) & 0.347 ($\pm$ nan) & 8.482 ($\pm$ nan) \\

%\multirow{6}{*}{Scenario 5} & \multirow{6}{*}{\hspace{0.3cm} 10} & Laplace Approximation & \textbf{1.000} ($\pm$ 0.000) & 2.152 ($\pm$ 0.227) & 1.640 ($\pm$ 0.945) & 1.000 ($\pm$ 0.000) & 2.134 ($\pm$ 0.045) & 1.084 ($\pm$ 0.248) \\
%& & VI: DiagonalNormal & \textbf{0.907} ($\pm$ 0.117) & 0.958 ($\pm$ 0.749) & \textbf{1.258} ($\pm$ 0.704) & 0.913 ($\pm$ 0.021) & 0.498 ($\pm$ 0.170) & \textbf{0.755} ($\pm$ 0.200) \\
%& & VI: MultivariateNormal & \textbf{0.853} ($\pm$ 0.092) & \textbf{0.387} ($\pm$ 0.423) & \textbf{1.129} ($\pm$ 0.692) & \textbf{0.850} ($\pm$ 0.021) & \textbf{0.181} ($\pm$ 0.092) & \textbf{0.629} ($\pm$ 0.161) \\
%& & VI: Structured Normal & \textbf{0.836} ($\pm$ 0.084) & \textbf{0.275} ($\pm$ 0.334) & \textbf{1.066} ($\pm$ 0.691) & \textbf{0.838} ($\pm$ 0.021) & \textbf{0.133} ($\pm$ 0.053) & \textbf{0.621} ($\pm$ 0.140) \\
%& & VI: IAF & \textbf{0.915} ($\pm$ 0.118) & 1.504 ($\pm$ 1.248) & 3.260 ($\pm$ 1.299) & 0.971 ($\pm$ 0.020) & 1.255 ($\pm$ 0.425) & 2.334 ($\pm$ 0.407) \\
%& & {ICL (ours)} & \textbf{0.826} ($\pm$ 0.110) & 1.000 ($\pm$ 0.817) & 2.568 ($\pm$ 0.863) & 0.879 ($\pm$ 0.018) & 0.925 ($\pm$ 0.233) & 2.229 ($\pm$ 0.151) \\
%\midrule

\bottomrule
\end{tabular}
\label{tab:Dim_v1}
}
\end{table}

\begin{table}[htp]
\centering
\caption{Generalized Linear Models: Ablation with respect to the dimensionality of the problem on 50 synthetic and 17 real-world datasets for scenarios 4, 5 and 6.  All results within two standard errors of the best average result for each scenario are marked in \textbf{bold}. Due to the limitations of the number of features in the real-world data, we can only use 5 datasets for 20 and one dataset for 50 dimensions. }
\label{tab:glm_res_detail}
\resizebox{1\textwidth}{!}{
\begin{tabular}{p{1.5cm} p{1.5cm} p{3.5cm} m{2.4cm}m{2.4cm}m{2.4cm}| m{2.4cm}m{2.4cm}m{2.4cm}}
\toprule
 \multicolumn{1}{c}{\multirow[c]{2}{*}[-0.5ex]{\textbf{Scenario}}} & \multicolumn{1}{c}{\multirow[c]{2}{*}[-0.5ex]{\textbf{Dim.}}} & \multicolumn{1}{c}{\multirow{2}{*}[-0.5ex]{\textbf{Model}}} & \multicolumn{3}{c}{\textbf{Synthetic Evaluation}} & \multicolumn{3}{c}{\textbf{Real-World Evaluation}} \\
\cmidrule(lr){4-6} \cmidrule(lr){7-9}
& & & C2ST ($\downarrow$) & MMD ($\downarrow$) & $\mathcal{W}_2$ ($\downarrow$) & C2ST ($\downarrow$) & MMD ($\downarrow$) & $\mathcal{W}_2$ ($\downarrow$) \\
\midrule
\multirow{6}{*}{Scenario 4} & \multirow{6}{*}{\hspace{0.3cm} 5} & Laplace Approximation & 1.000 ($\pm$ 0.000) & 3.511 $(\pm$ 2.025) & 2.166 ($\pm$ 1.722) & 1.000 ($\pm$ 0.000) & 2.011 $(\pm$ 0.058) & 0.993 ($\pm$ 0.144) \\
& & VI: DiagonalNormal & 0.968 ($\pm$ 0.036) & 2.798 $(\pm$ 2.255) & 2.065 ($\pm$ 1.745) & 0.916 ($\pm$ 0.040) & 0.928 $(\pm$ 0.339) & 0.732 ($\pm$ 0.181) \\
& & VI: MultivariateNormal & 0.855 ($\pm$ 0.123) & 1.648 $(\pm$ 2.052) & 1.853 ($\pm$ 1.745) & 0.771 ($\pm$ 0.017) & \textbf{0.087} $(\pm$ 0.030) & \textbf{0.539} ($\pm$ 0.070) \\
& & VI: Structured Normal & 0.847 ($\pm$ 0.116) & 1.505 $(\pm$ 1.978) & 1.889 ($\pm$ 1.883) & 0.769 ($\pm$ 0.012) & \textbf{0.083} $(\pm$ 0.018) & \textbf{0.543} ($\pm$ 0.070) \\
& & VI: IAF & 0.942 ($\pm$ 0.077) & 3.029 $(\pm$ 2.210) & 3.554 ($\pm$ 2.715) & 0.833 ($\pm$ 0.069) & 0.636 $(\pm$ 0.756) & 0.978 ($\pm$ 0.600) \\
& & \textbf{ICL (ours)} & \textbf{0.753} ($\pm$ 0.049) & \textbf{0.171} $(\pm$ 0.153) & \textbf{0.631} ($\pm$ 0.294) & \textbf{0.762} ($\pm$ 0.015) & \textbf{0.105} $(\pm$ 0.046) & \textbf{0.597} ($\pm$ 0.104) \\
\midrule

\multirow{6}{*}{Scenario 4} & \multirow{6}{*}{\hspace{0.3cm} 20} & Laplace Approximation & {1.000} ($\pm$ 0.000) & 4.929 $(\pm$ 1.611) & \textbf{8.863} ($\pm$ 3.796) & \textbf{1.000} ($\pm$ 0.000) & 3.196 $(\pm$ 0.841) & {5.186} ($\pm$ 1.533) \\
& & VI: DiagonalNormal & \textbf{0.988} ($\pm$ 0.060) & 4.418 $(\pm$ 2.013) & \textbf{9.364} ($\pm$ 4.281) & 0.997 ($\pm$ 0.007) & 3.095 $(\pm$ 1.417) & 6.098 ($\pm$ 2.435) \\
& & VI: MultivariateNormal & \textbf{0.986} ($\pm$ 0.054) & 3.388 $(\pm$ 1.907) & \textbf{7.910} ($\pm$ 4.070) & 0.893 ($\pm$ 0.087) & \textbf{0.534} $(\pm$ 0.469) & \textbf{3.175} ($\pm$ 0.751) \\
& & VI: Structured Normal & \textbf{0.954} ($\pm$ 0.076) & \textbf{2.254} $(\pm$ 1.515) & \textbf{7.475} ($\pm$ 4.224) & \textbf{0.727} ($\pm$ 0.034) & \textbf{0.074} $(\pm$ 0.070) & \textbf{2.877} ($\pm$ 0.379) \\
& & VI: IAF & \textbf{0.987} ($\pm$ 0.059) & 3.258 $(\pm$ 1.415) & \textbf{9.865} ($\pm$ 3.515) & 0.955 ($\pm$ 0.030) & 0.629 $(\pm$ 0.308) & 4.098 ($\pm$ 0.341) \\
& & {ICL (ours)} & \textbf{0.978} ($\pm$ 0.038) & \textbf{1.185} $(\pm$ 0.720) & \textbf{11.335} ($\pm$ 1.378) & 0.972 ($\pm$ 0.018) & 0.668 $(\pm$ 0.199) & 9.937 ($\pm$ 0.466) \\
\midrule
\multirow{6}{*}{Scenario 4} & \multirow{6}{*}{\hspace{0.3cm} 50} & {Laplace Approximation} & \textbf{1.000} ($\pm$ 0.000) & 6.695 ($\pm$ 1.329) & \textbf{12.323} ($\pm$ 4.091) & 1.000 ($\pm$ nan) & 5.491 ($\pm$ nan) & 7.518 ($\pm$ nan) \\
& & {VI: DiagonalNormal} & \textbf{0.965} ($\pm$ 0.084) & 2.395 ($\pm$ 1.958) & \textbf{12.022} ($\pm$ 3.673) & 0.996 ($\pm$ nan) & 4.368 ($\pm$ nan) & 6.951 ($\pm$ nan) \\
& & {VI: MultivariateNormal} & \textbf{0.984} ($\pm$ 0.054) & 5.395 ($\pm$ 1.847) & \textbf{12.141} ($\pm$ 3.079) & 1.000 ($\pm$ nan) & 5.146 ($\pm$ nan) & 9.002 ($\pm$ nan) \\
& & {VI: Structured Normal} & \textbf{0.982} ($\pm$ 0.026) & \textbf{4.261} ($\pm$ 1.191) & \textbf{11.126} ($\pm$ 3.396) & 0.869 ($\pm$ nan) & 3.181 ($\pm$ nan) & 7.065 ($\pm$ nan) \\
& & {VI: IAF} & \textbf{0.981} ($\pm$ 0.048) & \textbf{4.609} ($\pm$ 1.412) & \textbf{12.567} ($\pm$ 3.131) & 0.988 ($\pm$ nan) & 3.558 ($\pm$ nan) & 7.849 ($\pm$ nan) \\
& & {ICL (ours)} & \textbf{0.960} ($\pm$ 0.045) & \textbf{3.792} ($\pm$ 0.758) & \textbf{14.071} ($\pm$ 0.894) & 0.974 ($\pm$ nan) & 3.443 ($\pm$ nan) & 12.546 ($\pm$ nan) \\
\midrule

\multirow{6}{*}{Scenario 5} & \multirow{6}{*}{\hspace{0.3cm} 5} & Laplace Approximation & 1.000 ($\pm$ 0.000) & 2.060 $(\pm$ 0.472) & 0.797 ($\pm$ 0.577) & 1.000 ($\pm$ 0.000) & 1.982 $(\pm$ 0.126) & 0.623 ($\pm$ 0.084) \\
& & VI: DiagonalNormal & 0.866 ($\pm$ 0.085) & 0.954 $(\pm$ 1.022) & 0.651 ($\pm$ 0.549) & 0.810 ($\pm$ 0.036) & 0.441 $(\pm$ 0.252) & 0.384 ($\pm$ 0.089) \\
& & VI: MultivariateNormal & 0.765 ($\pm$ 0.100) & 0.537 $(\pm$ 1.019) & 0.633 ($\pm$ 1.067) & 0.711 ($\pm$ 0.038) & 0.148 $(\pm$ 0.093) & \textbf{0.279} ($\pm$ 0.056) \\
& & VI: Structured Normal & 0.758 ($\pm$ 0.098) & 0.447 $(\pm$ 0.818) & 0.572 ($\pm$ 0.816) & 0.705 ($\pm$ 0.032) & 0.140 $(\pm$ 0.081) & \textbf{0.269} ($\pm$ 0.045) \\
& & VI: IAF & 0.814 ($\pm$ 0.105) & 0.953 $(\pm$ 1.165) & 0.881 ($\pm$ 1.067) & 0.777 ($\pm$ 0.106) & 0.684 $(\pm$ 0.939) & 0.625 ($\pm$ 0.525) \\
& & {ICL (ours)} & \textbf{0.621} ($\pm$ 0.063) & \textbf{0.067} $(\pm$ 0.080) & \textbf{0.299} ($\pm$ 0.195) & \textbf{0.610} ($\pm$ 0.045) & \textbf{0.046} $(\pm$ 0.020) & \textbf{0.242} ($\pm$ 0.038) \\
\midrule
%\multirow{6}{*}{Scenario 5} & \multirow{6}{*}{\hspace{0.3cm} 10} & Laplace Approximation & \textbf{1.000} ($\pm$ 0.000) & 2.152 ($\pm$ 0.227) & 1.640 ($\pm$ 0.945) & 1.000 ($\pm$ 0.000) & 2.134 ($\pm$ 0.045) & 1.084 ($\pm$ 0.248) \\
%& & VI: DiagonalNormal & \textbf{0.907} ($\pm$ 0.117) & 0.958 ($\pm$ 0.749) & \textbf{1.258} ($\pm$ 0.704) & 0.913 ($\pm$ 0.021) & 0.498 ($\pm$ 0.170) & \textbf{0.755} ($\pm$ 0.200) \\
%& & VI: MultivariateNormal & \textbf{0.853} ($\pm$ 0.092) & \textbf{0.387} ($\pm$ 0.423) & \textbf{1.129} ($\pm$ 0.692) & \textbf{0.850} ($\pm$ 0.021) & \textbf{0.181} ($\pm$ 0.092) & \textbf{0.629} ($\pm$ 0.161) \\
%& & VI: Structured Normal & \textbf{0.836} ($\pm$ 0.084) & \textbf{0.275} ($\pm$ 0.334) & \textbf{1.066} ($\pm$ 0.691) & \textbf{0.838} ($\pm$ 0.021) & \textbf{0.133} ($\pm$ 0.053) & \textbf{0.621} ($\pm$ 0.140) \\
%& & VI: IAF & \textbf{0.915} ($\pm$ 0.118) & 1.504 ($\pm$ 1.248) & 3.260 ($\pm$ 1.299) & 0.971 ($\pm$ 0.020) & 1.255 ($\pm$ 0.425) & 2.334 ($\pm$ 0.407) \\
%& & {ICL (ours)} & \textbf{0.826} ($\pm$ 0.110) & 1.000 ($\pm$ 0.817) & 2.568 ($\pm$ 0.863) & 0.879 ($\pm$ 0.018) & 0.925 ($\pm$ 0.233) & 2.229 ($\pm$ 0.151) \\
%\midrule

\multirow{6}{*}{Scenario 5} & \multirow{6}{*}{\hspace{0.3cm} 20} & Laplace Approximation & \textbf{1.000} ($\pm$ 0.000) & 2.367 ($\pm$ 0.555) & 2.780 ($\pm$ 1.271) & \textbf{1.000} ($\pm$ 0.000) & 2.200 ($\pm$ 0.041) & 2.444 ($\pm$ 0.619) \\
& & VI: DiagonalNormal & \textbf{0.938} ($\pm$ 0.098) & \textbf{1.153} ($\pm$ 0.954) & \textbf{2.552} ($\pm$ 1.147) & \textbf{0.967} ($\pm$ 0.012) & 0.547 ($\pm$ 0.233) & \textbf{1.973} ($\pm$ 0.452) \\
& & VI: MultivariateNormal & \textbf{0.929} ($\pm$ 0.082) & \textbf{0.710} ($\pm$ 0.768) & \textbf{2.473} ($\pm$ 1.145) & \textbf{0.928} ($\pm$ 0.016) & \textbf{0.250} ($\pm$ 0.079) & \textbf{1.776} ($\pm$ 0.399) \\
& & VI: Structured Normal & \textbf{0.909} ($\pm$ 0.082) & \textbf{0.397} ($\pm$ 0.442) & \textbf{2.246} ($\pm$ 1.244) & \textbf{0.924} ($\pm$ 0.018) & \textbf{0.202} ($\pm$ 0.094) & \textbf{1.775} ($\pm$ 0.430) \\
& & VI: IAF & \textbf{0.934} ($\pm$ 0.092) & 1.325 ($\pm$ 1.161) & 4.899 ($\pm$ 1.320) & \textbf{0.980} ($\pm$ 0.016) & 0.892 ($\pm$ 0.404) & 3.593 ($\pm$ 0.597) \\
& & {ICL (ours)} & \textbf{0.961} ($\pm$ 0.046) & 1.330 ($\pm$ 1.125) & 5.084 ($\pm$ 1.297) & \textbf{0.981} ($\pm$ 0.014) & 1.162 ($\pm$ 0.461) & 4.804 ($\pm$ 0.578) \\
\midrule

\multirow{6}{*}{Scenario 5} & \multirow{6}{*}{\hspace{0.3cm} 50} & Laplace Approximation & \textbf{1.000} ($\pm$ 0.000) & 2.582 ($\pm$ 0.606) & \textbf{5.765} ($\pm$ 1.540) & 1.000 ($\pm$ nan) & 2.322 ($\pm$ nan) & 3.485 ($\pm$ nan) \\
& & VI: DiagonalNormal & \textbf{0.925} ($\pm$ 0.074) & 0.925 ($\pm$ 1.056) & \textbf{6.461} ($\pm$ 1.877) & 0.972 ($\pm$ nan) & 0.186 ($\pm$ nan) & 3.251 ($\pm$ nan) \\
& & VI: MultivariateNormal & \textbf{0.934} ($\pm$ 0.064) & \textbf{0.825} ($\pm$ 0.972) & \textbf{6.404} ($\pm$ 1.882) & 0.969 ($\pm$ nan) & 0.165 ($\pm$ nan) & {3.223} ($\pm$ nan) \\
& & VI: Structured Normal & \textbf{0.927} ($\pm$ 0.068) & \textbf{0.481} ($\pm$ 0.588) & \textbf{6.420} ($\pm$ 1.970) & {0.961} ($\pm$ nan) & {0.072} ($\pm$ nan) & {3.324} ($\pm$ nan) \\
& & VI: IAF & \textbf{0.925} ($\pm$ 0.069) & \textbf{0.792} ($\pm$ 0.975) & 8.458 ($\pm$ 0.864) & 0.996 ($\pm$ nan) & 0.519 ($\pm$ nan) & 4.645 ($\pm$ nan) \\
& & {ICL (ours)} & \textbf{0.998} ($\pm$ 0.002) & \textbf{0.762} ($\pm$ 0.987) & 8.195 ($\pm$ 0.820) & 1.000 ($\pm$ nan) & 0.984 ($\pm$ nan) & 7.288 ($\pm$ nan) \\
\midrule

\multirow{6}{*}{Scenario 6} & \multirow{6}{*}{\hspace{0.3cm} 5} & Laplace Approximation & 1.000 ($\pm$ 0.000) & 2.026 $(\pm$ 0.027) & 1.612 ($\pm$ 0.162) & 1.000 ($\pm$ 0.000) & 1.993 $(\pm$ 0.032) & 1.299 ($\pm$ 0.106) \\
& & VI: DiagonalNormal & 0.724 ($\pm$ 0.060) & 0.185 $(\pm$ 0.082) & \textbf{0.787} ($\pm$ 0.078) & 0.703 ($\pm$ 0.039) & 0.147 $(\pm$ 0.063) & 0.637 ($\pm$ 0.089) \\
& & VI: MultivariateNormal & \textbf{0.534} ($\pm$ 0.018) & \textbf{0.014} $(\pm$ 0.006) & \textbf{0.581} ($\pm$ 0.074) & \textbf{0.538} ($\pm$ 0.019) & \textbf{0.016} $(\pm$ 0.007) & \textbf{0.466} ($\pm$ 0.029) \\
& & VI: Structured Normal & \textbf{0.536} ($\pm$ 0.016) & \textbf{0.014} $(\pm$ 0.005) & \textbf{0.583} ($\pm$ 0.071) & \textbf{0.536} ($\pm$ 0.019) & \textbf{0.017} $(\pm$ 0.009) & \textbf{0.469} ($\pm$ 0.033) \\
& & VI: IAF & 0.542 ($\pm$ 0.026) & 0.031 $(\pm$ 0.031) & 0.613 ($\pm$ 0.092) & \textbf{0.535} ($\pm$ 0.015) & \textbf{0.015} $(\pm$ 0.006) & \textbf{0.467} ($\pm$ 0.031) \\
& & \textbf{ICL (ours)} & \textbf{0.532} ($\pm$ 0.019) & 0.016 $(\pm$ 0.008) & \textbf{0.590} ($\pm$ 0.066) & 0.556 ($\pm$ 0.017) & 0.035 $(\pm$ 0.015) & \textbf{0.504} ($\pm$ 0.038) \\

\midrule

\multirow{6}{*}{Scenario 6} & \multirow{6}{*}{\hspace{0.3cm} 20} & Laplace Approximation & 1.000 ($\pm$ 0.000) & 2.247 ($\pm$ 0.006) & 4.158 ($\pm$ 0.243) & 1.000 ($\pm$ 0.000) & 2.240 ($\pm$ 0.007) & 3.714 ($\pm$ 0.127) \\
& & VI: DiagonalNormal & \textbf{0.747} ($\pm$ 0.138) & \textbf{0.136} ($\pm$ 0.123) & \textbf{3.460} ($\pm$ 0.361) & \textbf{0.836} ($\pm$ 0.053) & \textbf{0.203} ($\pm$ 0.086) & \textbf{2.977} ($\pm$ 0.112) \\
& & VI: MultivariateNormal & \textbf{0.621} ($\pm$ 0.016) & \textbf{0.016} ($\pm$ 0.002) & \textbf{3.564} ($\pm$ 0.290) & \textbf{0.608} ($\pm$ 0.017) & \textbf{0.015} ($\pm$ 0.003) & \textbf{3.101} ($\pm$ 0.115) \\
& & VI: Structured Normal & \textbf{0.599} ($\pm$ 0.015) & \textbf{0.012} ($\pm$ 0.002) & \textbf{3.592} ($\pm$ 0.267) & \textbf{0.584} ($\pm$ 0.028) & \textbf{0.012} ($\pm$ 0.002) & \textbf{3.120} ($\pm$ 0.107) \\
& & VI: IAF & \textbf{0.625} ($\pm$ 0.040) & \textbf{0.019} ($\pm$ 0.009) & \textbf{3.572} ($\pm$ 0.266) & \textbf{0.636} ($\pm$ 0.021) & \textbf{0.020} ($\pm$ 0.005) & \textbf{3.106} ($\pm$ 0.128) \\
& & {ICL (ours)} & 0.747 ($\pm$ 0.148) & 0.163 ($\pm$ 0.144) & 4.063 ($\pm$ 0.184) & 0.928 ($\pm$ 0.030) & 0.463 ($\pm$ 0.162) & 4.425 ($\pm$ 0.314) \\
\midrule

\multirow{6}{*}{Scenario 6} & \multirow{6}{*}{\hspace{0.3cm} 50} & Laplace Approximation & 1.000 ($\pm$ 0.000) & 2.291 ($\pm$ 0.003) & \textbf{6.742} ($\pm$ 0.362) & 1.000 ($\pm$ nan) & 2.293 ($\pm$ nan) & 6.587 ($\pm$ nan) \\
& & VI: DiagonalNormal & \textbf{0.761} ($\pm$ 0.138) & \textbf{0.087} ($\pm$ 0.083) & \textbf{6.909} ($\pm$ 0.743) & 0.905 ($\pm$ nan) & 0.175 ($\pm$ nan) & 6.403 ($\pm$ nan) \\
& & VI: MultivariateNormal & \textbf{0.797} ($\pm$ 0.100) & \textbf{0.069} ($\pm$ 0.055) & \textbf{6.956} ($\pm$ 0.736) & 0.891 ($\pm$ nan) & 0.110 ($\pm$ nan) & 6.473 ($\pm$ nan) \\
& & VI: Structured Normal & \textbf{0.647} ($\pm$ 0.017) & \textbf{0.013} ($\pm$ 0.002) & 7.218 ($\pm$ 0.506) & 0.654 ($\pm$ nan) & 0.013 ($\pm$ nan) & 6.890 ($\pm$ nan) \\
& & VI: IAF & \textbf{0.639} ($\pm$ 0.038) & \textbf{0.014} ($\pm$ 0.006) & 7.204 ($\pm$ 0.463) & 0.692 ($\pm$ nan) & 0.024 ($\pm$ nan) & 6.887 ($\pm$ nan) \\
& & {ICL (ours)} & \textbf{0.742} ($\pm$ 0.178) & 0.115 ($\pm$ 0.124) & 7.713 ($\pm$ 0.120) & 0.935 ($\pm$ nan) & 0.203 ($\pm$ nan) & 7.846 ($\pm$ nan) \\
\bottomrule

\label{tab:Dim_v2}

\end{tabular}

}
\end{table}

\begin{table}[htp]
\centering
\caption{Generalized Linear Models: Ablation with respect to the dimensionality of the problem on 50 synthetic and 17 real-world datasets for scenario 7. All results within two standard errors of the best average result for each scenario are marked in \textbf{bold}. Due to the limitations of the number of features in the real-world data, we can only use 5 datasets for 20 and one dataset for 50 dimensions.}
\resizebox{1\textwidth}{!}{
\begin{tabular}{p{1.5cm} p{1.5cm} p{3.5cm} m{2.4cm}m{2.4cm}m{2.4cm}| m{2.4cm}m{2.4cm}m{2.4cm}}
\toprule
 \multicolumn{1}{c}{\multirow[c]{2}{*}[-0.5ex]{\textbf{Scenario}}} & \multicolumn{1}{c}{\multirow[c]{2}{*}[-0.5ex]{\textbf{Dim.}}} & \multicolumn{1}{c}{\multirow{2}{*}[-0.5ex]{\textbf{Model}}} & \multicolumn{3}{c}{\textbf{Synthetic Evaluation}} & \multicolumn{3}{c}{\textbf{Real-World Evaluation}} \\
\cmidrule(lr){4-6} \cmidrule(lr){7-9}
& & & C2ST ($\downarrow$) & MMD ($\downarrow$) & $\mathcal{W}_2$ ($\downarrow$) & C2ST ($\downarrow$) & MMD ($\downarrow$) & $\mathcal{W}_2$ ($\downarrow$) \\
\midrule
\multirow{6}{*}{Scenario 7} & \multirow{6}{*}{\hspace{0.3cm} 5} & Laplace Approximation & 1.000 ($\pm$ 0.000) & 3.559 $(\pm$ 1.933) & 1.347 ($\pm$ 1.067) & 1.000 ($\pm$ 0.000) & 2.016 $(\pm$ 0.080) & 0.763 ($\pm$ 0.174) \\
& & VI: DiagonalNormal & 0.938 ($\pm$ 0.074) & 2.536 $(\pm$ 2.097) & 1.142 ($\pm$ 0.993) & 0.936 ($\pm$ 0.024) & 1.029 $(\pm$ 0.255) & 0.579 ($\pm$ 0.181) \\
& & VI: MultivariateNormal & 0.814 ($\pm$ 0.181) & 1.999 $(\pm$ 2.283) & 1.033 ($\pm$ 0.969) & \textbf{0.741} ($\pm$ 0.020) & 0.093 $(\pm$ 0.025) & \textbf{0.391} ($\pm$ 0.074) \\
& & VI: Structured Normal & 0.824 ($\pm$ 0.177) & 1.891 $(\pm$ 2.127) & 1.041 ($\pm$ 0.934) & \textbf{0.734} ($\pm$ 0.025) & \textbf{0.072} $(\pm$ 0.019) & \textbf{0.385} ($\pm$ 0.065) \\
& &  VI: IAF & 0.939 ($\pm$ 0.091) & 2.707 $(\pm$ 1.712) & 1.590 ($\pm$ 0.820) & 0.864 ($\pm$ 0.093) & 0.830 $(\pm$ 0.697) & 1.064 ($\pm$ 0.616) \\
& & {ICL (ours)} & \textbf{0.700} ($\pm$ 0.116) & \textbf{0.317} $(\pm$ 0.355) & \textbf{0.400} ($\pm$ 0.286) & 0.773 ($\pm$ 0.048) & \textbf{0.294} $(\pm$ 0.457) & 0.559 ($\pm$ 0.256) \\
\midrule

\multirow{6}{*}{Scenario 7} & \multirow{6}{*}{\hspace{0.3cm} 20} & Laplace Approximation & 1.000 ($\pm$ 0.000) & 3.581 ($\pm$ 2.147) & \textbf{3.365} ($\pm$ 1.583) & 1.000 ($\pm$ 0.000) & 2.213 ($\pm$ 0.024) & 2.539 ($\pm$ 0.378) \\
& & VI: DiagonalNormal & \textbf{0.887} ($\pm$ 0.184) & 2.819 ($\pm$ 2.732) & 3.637 ($\pm$ 1.371) & \textbf{0.996} ($\pm$ 0.005) & 1.734 ($\pm$ 0.314) & \textbf{2.348} ($\pm$ 0.423) \\
& & VI: MultivariateNormal & \textbf{0.881} ($\pm$ 0.164) & 2.265 ($\pm$ 2.573) & \textbf{3.524} ($\pm$ 1.392) & \textbf{0.916} ($\pm$ 0.085) & 0.766 ($\pm$ 0.535) & \textbf{2.043} ($\pm$ 0.516) \\
& & VI: Structured Normal & \textbf{0.850} ($\pm$ 0.162) & \textbf{1.667} ($\pm$ 2.266) & \textbf{3.186} ($\pm$ 1.315) & \textbf{0.849} ($\pm$ 0.105) & \textbf{0.391} ($\pm$ 0.244) & \textbf{1.880} ($\pm$ 0.367) \\
& & VI: IAF & \textbf{0.867} ($\pm$ 0.184) & \textbf{1.629} ($\pm$ 1.584) & 4.875 ($\pm$ 1.239) & 0.986 ($\pm$ 0.007) & 0.895 ($\pm$ 0.361) & 4.096 ($\pm$ 0.319) \\
& & {ICL (ours)} & \textbf{0.867} ($\pm$ 0.185) & \textbf{1.428} ($\pm$ 1.352) & 4.836 ($\pm$ 1.032) & \textbf{0.982} ($\pm$ 0.010) & \textbf{0.820} ($\pm$ 0.324) & 4.177 ($\pm$ 0.368) \\
\midrule

\multirow{6}{*}{Scenario 7} & \multirow{6}{*}{\hspace{0.3cm} 50} & Laplace Approximation & 1.000 ($\pm$ 0.000) & 4.768 ($\pm$ 1.171) & \textbf{6.573} ($\pm$ 1.038) & 1.000 ($\pm$ nan) & 2.312 ($\pm$ nan) & {5.270} ($\pm$ nan) \\
& & VI: DiagonalNormal & \textbf{0.771} ($\pm$ 0.191) & 3.263 ($\pm$ 1.853) & \textbf{6.919} ($\pm$ 1.257) & 1.000 ($\pm$ nan) & 2.237 ($\pm$ nan) & {5.417} ($\pm$ nan) \\
& & VI: MultivariateNormal & \textbf{0.816} ($\pm$ 0.154) & 3.245 ($\pm$ 1.793) & 6.978 ($\pm$ 1.226) & {0.997} ($\pm$ nan) & {2.117} ($\pm$ nan) & 5.781 ($\pm$ nan) \\
& & VI: Structured Normal & \textbf{0.795} ($\pm$ 0.171) & 3.126 ($\pm$ 1.677) & \textbf{6.918} ($\pm$ 1.260) & 1.000 ($\pm$ nan) & {1.879} ($\pm$ nan) & 5.461 ($\pm$ nan) \\
& & VI: IAF & \textbf{0.769} ($\pm$ 0.189) & \textbf{2.534} ($\pm$ 0.894) & 7.895 ($\pm$ 0.843) & {0.994} ($\pm$ nan) & {0.584} ($\pm$ nan) & 7.626 ($\pm$ nan) \\
& & {ICL (ours)} & \textbf{0.732} ($\pm$ 0.216) & \textbf{2.451} ($\pm$ 0.790) & 7.787 ($\pm$ 0.661) & {0.980} ($\pm$ nan) & {0.411} ($\pm$ nan) & 7.461 ($\pm$ nan) \\
\bottomrule

\end{tabular}
\label{Tab:Dim_v3}
}
\end{table}

\begin{table}[htp]
\centering
\caption{Evaluating the predictive performance across 50 synthetic and 17 real-world datasets in GLM scenario 2 for different dimensionalities.  All results within two standard errors of the best average result for each scenario are marked in \textbf{bold}. Due to the limitations of the number of features in the real-world data, we can only use 5 datasets for 20 and one dataset for 50 dimensions. We find that the quality of the samples by the in-context learner, when evaluated based on predictive performance, decreases consistently with an increase in the dimensionality of the problem. Note that the randomness in the results, especially for higher dimensionalities, can in rare cases lead to better mean values. This is most likely not significant when taking the standard error into account.}
\label{tab:tabular_results}
\resizebox{1.0\textwidth}{!}{
\begin{tabular}{p{2.5cm} p{2.5cm} p{4.5cm} m{3.4cm}m{3.4cm}}
\toprule
\textbf{Scenario} & \textbf{Dim.} & \textbf{Model} & \textbf{RMSE Real-World} ($\downarrow$) & \textbf{RMSE Synthetic}  ($\downarrow$) \\
\midrule
% Scenario 2
\multirow{9}{*}{Scenario 2} & \multirow{9}{*}{5}
& {HMC} & \textbf{0.559} ($\pm$ 0.023) & \textbf{0.556} ($\pm$ 0.049) \\
& & {Laplace Approximation} & \textbf{0.561} ($\pm$ 0.022) & \textbf{0.557} ($\pm$ 0.049) \\
& & {VI: DiagonalNormal} & \textbf{0.560} ($\pm$ 0.023) & \textbf{0.557} ($\pm$ 0.049) \\
& & {VI: MultivariateNormal} & \textbf{0.559} ($\pm$ 0.023) & \textbf{0.556} ($\pm$ 0.049) \\
& & {VI: Structured Normal} & \textbf{0.604} ($\pm$ 0.016) & 0.685 ($\pm$ 0.054) \\
& & {VI: IAF} & \textbf{0.563} ($\pm$ 0.023) & \textbf{0.557} ($\pm$ 0.049) \\
& & {ICL (ours)} & \textbf{0.561} ($\pm$ 0.019) & \textbf{0.653} ($\pm$ 0.049) \\
\cmidrule{3-5}
& & MAP & 0.513 ($\pm$ 0.023) & 0.522 ($\pm$ 0.048) \\
%& Linear Model & 0.510 ($\pm$ 0.023) & 0.508 ($\pm$ 0.046) \\
%& Random Forest & 0.219 ($\pm$ 0.011) & 0.224 ($\pm$ 0.019) \\
& & TabPFN & 0.449 ($\pm$ 0.034) & 0.498 ($\pm$ 0.047) \\
\midrule

\multirow{9}{*}{Scenario 2} & \multirow{9}{*}{20}
& {HMC} & \textbf{0.682} ($\pm$ 0.029) & \textbf{0.536} ($\pm$ 0.041) \\
& & {Laplace Approximation} & \textbf{0.682} ($\pm$ 0.030) & \textbf{0.538} ($\pm$ 0.040) \\
& & {VI: DiagonalNormal} & \textbf{0.680} ($\pm$ 0.029) & \textbf{0.539} ($\pm$ 0.041) \\
& & {VI: MultivariateNormal} & \textbf{0.685} ($\pm$ 0.029) & \textbf{0.537} ($\pm$ 0.041) \\
& & VI: Structured Normal & 0.746 ($\pm$ 0.019) & 0.681 ($\pm$ 0.041) \\
& & {VI: IAF} & \textbf{0.683} ($\pm$ 0.029) & \textbf{0.539} ($\pm$ 0.041) \\
& & ICL (ours) & 0.777 ($\pm$ 0.011) & 1.122 ($\pm$ 0.078) \\
\cmidrule{3-5}
& & MAP & 0.578 ($\pm$ 0.025) & 0.472 ($\pm$ 0.039) \\
%& Linear Model & 0.562 ($\pm$ 0.025) & 0.448 ($\pm$ 0.037) \\
%& Random Forest & 0.260 ($\pm$ 0.011) & 0.230 ($\pm$ 0.016) \\
& & TabPFN & 0.470 ($\pm$ 0.044) & 0.446 ($\pm$ 0.038) \\
\midrule

\multirow{9}{*}{Scenario 2} & \multirow{9}{*}{50}
& HMC & 0.669 ($\pm$ nan) & \textbf{0.713} ($\pm$ 0.060) \\
& & Laplace Approximation & 0.594 ($\pm$ nan) & 0.878 ($\pm$ 0.068) \\
& & VI: DiagonalNormal & 0.582 ($\pm$ nan) & 0.870 ($\pm$ 0.065) \\
& & VI: MultivariateNormal & 0.729 ($\pm$ nan) & \textbf{0.764} ($\pm$ 0.066) \\
& & VI: Structured Normal & 0.922 ($\pm$ nan) & 1.116 ($\pm$ 0.074) \\
& & VI: IAF & 0.695 ($\pm$ nan) & \textbf{0.770} ($\pm$ 0.060) \\
& & ICL (ours) & 1.256 ($\pm$ nan) & 2.343 ($\pm$ 0.230) \\
\cmidrule{3-5}
& & MAP & 0.301 ($\pm$ nan) & 0.398 ($\pm$ 0.047) \\
%& Linear Model & 0.000 ($\pm$ nan) & 0.012 ($\pm$ 0.005) \\
%& Random Forest & 0.226 ($\pm$ nan) & 0.381 ($\pm$ 0.030) \\
& & TabPFN & 0.235 ($\pm$ nan) & 0.570 ($\pm$ 0.053) \\
\bottomrule

\end{tabular}
}
\end{table}

\begin{table}[htp]
\centering
\caption{Evaluating the predictive performance across 50 synthetic and 17 real-world datasets in GLM scenario 2 for different dimensionalities.  All results within two standard errors of the best average result for each scenario are marked in \textbf{bold}. Due to the limitations of the number of features in the real-world data, we can only use 5 datasets for 20 and one dataset for 50 dimensions. We find that the quality of the samples by the in-context learner, when evaluated based on predictive performance, decreases consistently with an increase in the dimensionality of the problem.}
\label{tab:tabular_results}
\resizebox{1.0\textwidth}{!}{
\begin{tabular}{p{2.5cm} p{2.5cm} p{4.5cm} m{3.4cm}m{3.4cm}}
\toprule
\textbf{Scenario} & \textbf{Dim.} & \textbf{Model} & \textbf{RMSE Real-World} ($\downarrow$) & \textbf{RMSE Synthetic}  ($\downarrow$) \\
\midrule
% Scenario 2
% Scenario 3
\multirow{9}{*}{Scenario 3} & \multirow{9}{*}{5}
& {HMC} & \textbf{0.684} ($\pm$ 0.027) & \textbf{0.512} ($\pm$ 0.040) \\
& & {Laplace Approximation} & \textbf{0.688} ($\pm$ 0.026) & \textbf{0.516} ($\pm$ 0.040) \\
& & {VI: DiagonalNormal} & \textbf{0.686} ($\pm$ 0.027) & \textbf{0.513} ($\pm$ 0.040) \\
& & {VI: MultivariateNormal} & \textbf{0.685} ($\pm$ 0.027) & \textbf{0.512} ($\pm$ 0.040) \\
& & {VI: Structured Normal} & \textbf{0.733} ($\pm$ 0.016) & 0.607 ($\pm$ 0.043) \\
& & {VI: IAF} & \textbf{0.686} ($\pm$ 0.027) & \textbf{0.512} ($\pm$ 0.040) \\
& & {ICL (ours)} & \textbf{0.690} ($\pm$ 0.023) & \textbf{0.588} ($\pm$ 0.045) \\
\cmidrule{3-5}
& & MAP & 0.646 ($\pm$ 0.028) & 0.495 ($\pm$ 0.039) \\
%& Linear Model & 0.630 ($\pm$ 0.028) & 0.468 ($\pm$ 0.037) \\
%& Random Forest & 0.269 ($\pm$ 0.013) & 0.214 ($\pm$ 0.016) \\
& & TabPFN & 0.556 ($\pm$ 0.041) & 0.462 ($\pm$ 0.037) \\
\midrule

%\multirow{9}{*}{Scenario 3} & \multirow{9}{*}{10}
%& {HMC} & \textbf{0.822} ($\pm$ 0.032) & \textbf{0.575} ($\pm$ 0.037) \\
%%& & {Laplace Approximation} & \textbf{0.814} ($\pm$ 0.032) & \textbf{0.604} ($\pm$ 0.040) \\
%& & {VI: DiagonalNormal} & \textbf{0.813} ($\pm$ 0.033) & \textbf{0.589} ($\pm$ 0.039) \\
%& & {VI: MultivariateNormal} & \textbf{0.815} ($\pm$ 0.032) & \textbf{0.589} ($\pm$ 0.039) \\
%& & {VI: Structured Normal} & \textbf{0.875} ($\pm$ 0.025) & 0.804 ($\pm$ 0.050) \\
%& & {VI: IAF} & \textbf{0.814} ($\pm$ 0.032) & \textbf{0.577} ($\pm$ 0.037) \\
%& & {ICL (ours)} & 0.941 ($\pm$ 0.011) & 1.169 ($\pm$ 0.118) \\
%\cmidrule{3-5}
%& & MAP & 0.715 ($\pm$ 0.030) & 0.524 ($\pm$ 0.035) \\
%& Linear Model & 0.671 ($\pm$ 0.030) & 0.476 ($\pm$ 0.030) \\
%& Random Forest & 0.295 ($\pm$ 0.012) & 0.259 ($\pm$ 0.015) \\
%& & TabPFN & 0.558 ($\pm$ 0.053) & 0.474 ($\pm$ 0.031) \\
%\midrule

\multirow{9}{*}{Scenario 3} & \multirow{9}{*}{20}
& {HMC} & \textbf{1.030} ($\pm$ 0.045) & \textbf{0.621} ($\pm$ 0.046) \\
& & {Laplace Approximation} & \textbf{1.053} ($\pm$ 0.047) & 0.755 ($\pm$ 0.052) \\
& & {VI: DiagonalNormal} & \textbf{1.035} ($\pm$ 0.043) & 0.734 ($\pm$ 0.053) \\
& & {VI: MultivariateNormal} & \textbf{1.033} ($\pm$ 0.039) & \textbf{0.705} ($\pm$ 0.055) \\
& & {VI: Structured Normal} & \textbf{1.095} ($\pm$ 0.045) & 1.033 ($\pm$ 0.063) \\
& & {VI: IAF} & \textbf{1.026} ($\pm$ 0.045) & \textbf{0.653} ($\pm$ 0.047) \\
& & {ICL (ours)} & 1.770 ($\pm$ 0.048) & 2.160 ($\pm$ 0.217) \\
\cmidrule{3-5}
& & MAP & 0.861 ($\pm$ 0.038) & 0.581 ($\pm$ 0.050) \\
%& Linear Model & 0.781 ($\pm$ 0.031) & 0.414 ($\pm$ 0.032) \\
%& Random Forest & 0.420 ($\pm$ 0.019) & 0.343 ($\pm$ 0.019) \\
& & TabPFN & 0.654 ($\pm$ 0.062) & 0.475 ($\pm$ 0.039) \\
\midrule

\multirow{9}{*}{Scenario 3} & \multirow{9}{*}{50}
& HMC & 0.858 ($\pm$ nan) & \textbf{0.645} ($\pm$ 0.051) \\
& & Laplace Approximation & 0.866 ($\pm$ nan) & 0.865 ($\pm$ 0.083) \\
& & VI: DiagonalNormal & 0.788 ($\pm$ nan) & 0.870 ($\pm$ 0.084) \\
& & VI: MultivariateNormal & 0.819 ($\pm$ nan) & 0.778 ($\pm$ 0.066) \\
& & VI: Structured Normal & 0.812 ($\pm$ nan) & 1.040 ($\pm$ 0.103) \\
& & VI: IAF & 0.802 ($\pm$ nan) & 0.846 ($\pm$ 0.078) \\
& & {ICL (ours)} & 1.686 ($\pm$ nan) & 3.477 ($\pm$ 0.604) \\
\cmidrule{3-5}
& & MAP & 0.539 ($\pm$ nan) & 0.618 ($\pm$ 0.054) \\
%& Linear Model & 0.000 ($\pm$ nan) & 0.017 ($\pm$ 0.012) \\
%& Random Forest & 0.286 ($\pm$ nan) & 0.368 ($\pm$ 0.035) \\
& & TabPFN & 0.322 ($\pm$ nan) & 0.534 ($\pm$ 0.038) \\

\bottomrule
\end{tabular}
}
\end{table}

\begin{table}[htp]
\centering
\caption{Evaluating the predictive performance across 50 synthetic and 17 real-world datasets in GLM scenario 2 for different dimensionalities.  All results within two standard errors of the best average result for each scenario are marked in \textbf{bold}. Due to the limitations of the number of features in the real-world data, we can only use 5 datasets for 20 and one dataset for 50 dimensions. We find that the quality of the samples by the in-context learner, when evaluated based on predictive performance, decreases consistently with an increase in the dimensionality of the problem.}
\label{tab:tabular_results}
\resizebox{1.0\textwidth}{!}{
\begin{tabular}{p{2.5cm} p{2.5cm} p{4.5cm} m{3.4cm}m{3.4cm}}
\toprule
\textbf{Scenario} & \textbf{Dim.} & \textbf{Model} & \textbf{RMSE Real-World} ($\downarrow$) & \textbf{RMSE Synthetic}  ($\downarrow$) \\
\midrule
% Scenario 2

\multirow{9}{*}{Scenario 5} & \multirow{9}{*}{5}
& HMC & \textbf{0.699} ($\pm$ 0.022) & \textbf{0.490} ($\pm$ 0.036) \\
& & Laplace Approximation & \textbf{0.699} ($\pm$ 0.022) & \textbf{0.491} ($\pm$ 0.036) \\
& & VI: DiagonalNormal & \textbf{0.702} ($\pm$ 0.022) & \textbf{0.491} ($\pm$ 0.036) \\
& & VI: MultivariateNormal & \textbf{0.698} ($\pm$ 0.021) & \textbf{0.491} ($\pm$ 0.036) \\
& & VI: Structured Normal & 1.507 ($\pm$ 0.089) & 0.741 ($\pm$ 0.053) \\
& & VI: IAF & \textbf{0.699} ($\pm$ 0.022) & \textbf{0.490} ($\pm$ 0.036) \\
& & {ICL (ours)} & 0.769 ($\pm$ 0.020) & 0.701 ($\pm$ 0.049) \\
\cmidrule{3-5}
& & MAP & 0.658 ($\pm$ 0.022) & 0.471 ($\pm$ 0.035) \\
%& Linear Model & 0.604 ($\pm$ 0.027) & 0.451 ($\pm$ 0.034) \\
%& Random Forest & 0.256 ($\pm$ 0.012) & 0.207 ($\pm$ 0.014) \\
& & TabPFN & 0.534 ($\pm$ 0.040) & 0.442 ($\pm$ 0.035) \\
\midrule

%\multirow{9}{*}{Scenario 5} & \multirow{9}{*}{10}
%& HMC & \textbf{0.946} ($\pm$ 0.034) & \textbf{0.521} ($\pm$ 0.041) \\
%& & Laplace Approximation & \textbf{0.941} ($\pm$ 0.036) & \textbf{0.526} ($\pm$ 0.041) \\
%& & VI: DiagonalNormal & \textbf{0.955} ($\pm$ 0.035) & \textbf{0.525} ($\pm$ 0.041) \\
%& & VI: MultivariateNormal & \textbf{0.945} ($\pm$ 0.035) & \textbf{0.522} ($\pm$ 0.041) \\
%& & VI: Structured Normal & 1.329 ($\pm$ 0.049) & 0.868 ($\pm$ 0.062) \\
%& & VI: IAF & \textbf{0.945} ($\pm$ 0.033) & \textbf{0.524} ($\pm$ 0.041) \\
%& & {ICL (ours)} & 1.783 ($\pm$ 0.034) & 1.048 ($\pm$ 0.104) \\
%\cmidrule{3-5}
%& & MAP & 0.848 ($\pm$ 0.036) & 0.463 ($\pm$ 0.038) \\
%& Linear Model & 0.659 ($\pm$ 0.029) & 0.434 ($\pm$ 0.034) \\
%& Random Forest & 0.306 ($\pm$ 0.011) & 0.256 ($\pm$ 0.016) \\
%& & TabPFN & 0.547 ($\pm$ 0.053) & 0.430 ($\pm$ 0.038) \\
%\midrule

\multirow{9}{*}{Scenario 5} & \multirow{9}{*}{20}
& HMC & \textbf{1.527} ($\pm$ 0.055) & \textbf{0.553} ($\pm$ 0.044) \\
& & Laplace Approximation & \textbf{1.585} ($\pm$ 0.065) & \textbf{0.586} ($\pm$ 0.043) \\
& & VI: DiagonalNormal & \textbf{1.554} ($\pm$ 0.058) & \textbf{0.586} ($\pm$ 0.042) \\
& & VI: MultivariateNormal & \textbf{1.530} ($\pm$ 0.058) & \textbf{0.564} ($\pm$ 0.043) \\
& & VI: Structured Normal & 2.109 ($\pm$ 0.156) & 1.054 ($\pm$ 0.067) \\
& & VI: IAF & \textbf{1.548} ($\pm$ 0.057) & \textbf{0.562} ($\pm$ 0.043) \\
& & {ICL (ours)} & 3.545 ($\pm$ 0.288) & 1.626 ($\pm$ 0.140) \\
\cmidrule{3-5}
& & MAP & 1.254 ($\pm$ 0.027) & 0.464 ($\pm$ 0.035) \\
%& Linear Model & 0.790 ($\pm$ 0.032) & 0.383 ($\pm$ 0.030) \\
%& Random Forest & 0.424 ($\pm$ 0.017) & 0.325 ($\pm$ 0.022) \\
& & TabPFN & 0.668 ($\pm$ 0.064) & 0.413 ($\pm$ 0.032) \\
\midrule

\multirow{9}{*}{Scenario 5} & \multirow{9}{*}{50}
& HMC & 1.626 ($\pm$ nan) & \textbf{0.521} ($\pm$ 0.028) \\
& & Laplace Approximation & 1.541 ($\pm$ nan) & 0.655 ($\pm$ 0.040) \\
& & VI: DiagonalNormal & 1.576 ($\pm$ nan) & 0.639 ($\pm$ 0.041) \\
& & VI: MultivariateNormal & 1.659 ($\pm$ nan) & 0.592 ($\pm$ 0.035) \\
& & VI: Structured Normal & 2.076 ($\pm$ nan) & 1.018 ($\pm$ 0.102) \\
& & VI: IAF & 1.706 ($\pm$ nan) & 0.627 ($\pm$ 0.040) \\
& & {ICL (ours)} & 10.319 ($\pm$ nan) & 1.458 ($\pm$ 0.193) \\
\cmidrule{3-5}
& & MAP & 1.318 ($\pm$ nan) & 0.416 ($\pm$ 0.018) \\
%& Linear Model & 0.000 ($\pm$ nan) & 0.010 ($\pm$ 0.005) \\
%& Random Forest & 0.304 ($\pm$ nan) & 0.384 ($\pm$ 0.041) \\
& & TabPFN & 0.330 ($\pm$ nan) & 0.443 ($\pm$ 0.024) \\
\bottomrule
\end{tabular}
}
\end{table}

\newpage

\section{\textcolor{rev}{Comparison to SGLD}}

\textcolor{rev}{
Besides comparing the samples from our ICL approach to samples from various VI methods, we additionally compare it against samples generated via stochastic gradient Langevin dynamics (SGLD) \citep{welling2011bayesian}. We run SGLD with a learning rate of $10^{-3}$ for the GLM and GMM cases and a learning rate of $10^{-4}$ for FA and use 1000 gradient steps for warmup and partition the data into ten minibatches. We implement the preconditioning method introduced by \cite{li2016preconditioned} for more stable sampling behavior. Despite the preconditioning, SGLD consistently fails for GLMs scenario 7 because the sampler diverges causing singular covariance matrices. To facilitate running SGLD for the GMMs, which also include discrete variables, we marginalize over the discrete variables.}

%\begin{figure}
%    \centering
 %   \includegraphics[width=1.0\linewidth]{Images/HMC_vs_SGLD.png}
%    \caption{Caption}
 %   \label{fig:enter-label}
%\end{figure}

\textcolor{rev}{
In summary, we find that ICL yields samples with much higher quality than SGLD compared to the gold standard HMC samples across almost all scenarios on both synthetic and real-world data. The poor sample quality with SGLD is expected given that numerous theoretical and empirical findings confirm that, while SGLD is computationally very cheap, it is substantially outperformed by, for instance, HMC, in terms of sample quality, which is especially pronounced when the posterior distributions are complex and parameters are correlated \citep{chen2014stochastic, mangoubi2019nonconvex, izmailov2021bayesian, brosse2018promises} .}

%On the Convergence of Stochastic Gradient MCMC Algorithms with High-Order Integrators

 \begin{table}[htp]
\centering
\caption{\textcolor{rev}{SGLD vs. ICL: Evaluation on 50 synthetic and 17 real-world datasets for six different GLM scenarios.  All results within two standard errors of the best average result for each scenario are marked in \textbf{bold}.}}
\label{tab:sgld_diff_abl}
\resizebox{1\textwidth}{!}{
\textcolor{rev}{
\begin{tabular}{p{1.5cm} p{1.8cm} m{2.4cm}m{2.5cm}m{2.7cm}| m{2.4cm}m{2.4cm}m{2.4cm}}
\toprule
\multicolumn{1}{c}{\multirow[c]{2}{*}[-0.5ex]{\textbf{Scenario}}} & \multicolumn{1}{c}{\multirow{2}{*}[-0.5ex]{\textbf{Model}}} & \multicolumn{3}{c}{\textbf{Synthetic Evaluation}} & \multicolumn{3}{c}{\textbf{Real-World Evaluation}} \\
\cmidrule(lr){3-5} \cmidrule(lr){6-8}
& & C2ST ($\downarrow$) & MMD ($\downarrow$) & $\mathcal{W}_2$ ($\downarrow$) & C2ST ($\downarrow$) & MMD ($\downarrow$) & $\mathcal{W}_2$ ($\downarrow$) \\
\midrule
\multirow{2}{*}{Scenario 1} 
& SGLD & 0.992 ($\pm$ 0.015) & 2.846 ($\pm$ 1.411) & 1.951 ($\pm$ 0.917) & 0.980 ($\pm$ 0.013) & 2.191 ($\pm$ 1.183) & 0.865 ($\pm$ 0.438) \\
& \textbf{ICL (ours)} & \textbf{0.765} ($\pm$ 0.123) & \textbf{0.767} ($\pm$ 0.727) & \textbf{0.585} ($\pm$ 0.301) & \textbf{0.614} ($\pm$ 0.074) & \textbf{0.175} ($\pm$ 0.219) & \textbf{0.310} ($\pm$ 0.138) \\
\midrule
\multirow{2}{*}{Scenario 2} 
& SGLD & 0.999 ($\pm$ 0.004) & 5.650 ($\pm$ 1.762) & 8.295 ($\pm$ 5.629) & 0.994 ($\pm$ 0.006) & 2.699 ($\pm$ 1.093) & 1.289 ($\pm$ 0.454) \\
& \textbf{ICL (ours)} & \textbf{0.839} ($\pm$ 0.072) & \textbf{0.707} ($\pm$ 0.658) & \textbf{1.111} ($\pm$ 0.300) & \textbf{0.768} ($\pm$ 0.033) & \textbf{0.143} ($\pm$ 0.089) & \textbf{0.411} ($\pm$ 0.094) \\
\midrule
\multirow{2}{*}{Scenario 3} 
& SGLD & 0.997 ($\pm$ 0.008) & 3.320 ($\pm$ 1.595) & 3.011 ($\pm$ 1.036) & 0.983 ($\pm$ 0.013) & 2.152 ($\pm$ 1.194) & 0.935 ($\pm$ 0.523) \\
& \textbf{ICL (ours)} & \textbf{0.611} ($\pm$ 0.070) & \textbf{0.089} ($\pm$ 0.114) & \textbf{0.423} ($\pm$ 0.348) & \textbf{0.576} ($\pm$ 0.027) & \textbf{0.037} ($\pm$ 0.026) & \textbf{0.257} ($\pm$ 0.044) \\
\midrule
\multirow{2}{*}{Scenario 4} 
& SGLD & 1.000 ($\pm$ 0.000) & 6.626 ($\pm$ 1.215) & 15.674 ($\pm$ 8.100) & 0.994 ($\pm$ 0.006) & 2.927 ($\pm$ 1.564) & 1.606 ($\pm$ 1.022) \\
& \textbf{ICL (ours)} & \textbf{0.753} ($\pm$ 0.049) & \textbf{0.171} ($\pm$ 0.153) & \textbf{0.631} ($\pm$ 0.294) & \textbf{0.762} ($\pm$ 0.015) & \textbf{0.105} ($\pm$ 0.046) & \textbf{0.597} ($\pm$ 0.104) \\
\midrule
\multirow{2}{*}{Scenario 5} 
& SGLD & 0.999 ($\pm$ 0.003) & 3.308 ($\pm$ 1.728) & 2.216 ($\pm$ 1.247) & 1.000 ($\pm$ 0.000) & 4.012 ($\pm$ 1.413) & 0.996 ($\pm$ 0.406) \\
& \textbf{ICL (ours)} & \textbf{0.621} ($\pm$ 0.063) & \textbf{0.067} ($\pm$ 0.080) & \textbf{0.299} ($\pm$ 0.195) & \textbf{0.610} ($\pm$ 0.045) & \textbf{0.046} ($\pm$ 0.020) & \textbf{0.242} ($\pm$ 0.038) \\
\midrule
\multirow{2}{*}{Scenario 6} 
& SGLD & 0.998 ($\pm$ 0.001) & 2.681 ($\pm$ 0.565) & 2.419 ($\pm$ 0.510) & 0.998 ($\pm$ 0.002) & 2.845 ($\pm$ 0.590) & 1.851 ($\pm$ 0.319) \\
& \textbf{ICL (ours)} & \textbf{0.532} ($\pm$ 0.019) & \textbf{0.016} ($\pm$ 0.008) & \textbf{0.590} ($\pm$ 0.066) & \textbf{0.556} ($\pm$ 0.017) & \textbf{0.035} ($\pm$ 0.015) & \textbf{0.504} ($\pm$ 0.038) \\
\bottomrule
\end{tabular}
\label{tab:glm_sgld}
}
}
\end{table}

\textcolor{rev}{
For GLMs (\Cref{tab:glm_sgld}), ICL achieves significantly better results, with notable improvements in C2ST. In Scenario 1, synthetic C2ST drops from 0.992 to 0.765 and real-world C2ST from 0.980 to 0.614. Similarly, Scenario 3 shows substantial gains, with synthetic C2ST improving from 0.997 to 0.611 and real-world C2ST from 0.983 to 0.576. These trends extend to metrics like $\mathcal{W}_2$, where ICL yields consistent reductions.
}

\begin{table}[htp]
\centering
\caption{\textcolor{rev}{SGLD vs. ICL: Evaluation on 50 synthetic and 17 real-world datasets for six different FA scenarios. All results within two standard errors of the best average result for each scenario are marked in \textbf{bold}.}}
\label{tab:sgld_res_detail}
\resizebox{1\textwidth}{!}{
\textcolor{rev}{
\begin{tabular}{p{1.5cm} p{1.8cm} m{2.4cm} m{2.4cm} m{2.4cm}| m{2.4cm} m{2.4cm} m{2.4cm}}
\toprule
\multicolumn{1}{c}{\multirow[c]{2}{*}[-0.5ex]{\textbf{Scenario}}} & \multicolumn{1}{c}{\multirow{2}{*}[-0.5ex]{\textbf{Model}}} & \multicolumn{3}{c}{\textbf{Synthetic Evaluation}} & \multicolumn{3}{c}{\textbf{Real-World Evaluation}} \\
\cmidrule(lr){3-5} \cmidrule(lr){6-8}
& & C2ST ($\downarrow$) & MMD ($\downarrow$) & $\mathcal{W}_2$ ($\downarrow$) & C2ST ($\downarrow$) & MMD ($\downarrow$) & $\mathcal{W}_2$ ($\downarrow$) \\
\midrule
\multirow{2}{*}{Scenario 1} 
& SGLD & 0.996 ($\pm$ 0.006) & 2.883 ($\pm$ 1.552) & 1.776 ($\pm$ 0.694) & 0.995 ($\pm$ 0.003) & 2.676 ($\pm$ 0.710) & 1.608 ($\pm$ 0.381) \\
& \textbf{ICL (ours)} & \textbf{0.552} ($\pm$ 0.028) & \textbf{0.034} ($\pm$ 0.034) & \textbf{0.289} ($\pm$ 0.083) & \textbf{0.606} ($\pm$ 0.038) & \textbf{0.068} ($\pm$ 0.069) & \textbf{0.265} ($\pm$ 0.078) \\
\midrule
\multirow{2}{*}{Scenario 2} 
& SGLD & 0.997 ($\pm$ 0.003) & 2.950 ($\pm$ 0.786) & 1.892 ($\pm$ 0.533) & 0.995 ($\pm$ 0.003) & 2.517 ($\pm$ 0.583) & 1.500 ($\pm$ 0.268) \\
& \textbf{ICL (ours)} & \textbf{0.542} ($\pm$ 0.006) & \textbf{0.017} ($\pm$ 0.006) & \textbf{0.244} ($\pm$ 0.033) & \textbf{0.622} ($\pm$ 0.032) & \textbf{0.098} ($\pm$ 0.039) & \textbf{0.287} ($\pm$ 0.046) \\
\midrule
\multirow{2}{*}{Scenario 3} 
& SGLD & 0.998 ($\pm$ 0.005) & 3.662 ($\pm$ 1.099) & 2.086 ($\pm$ 0.919) & 0.956 ($\pm$ 0.025) & 1.580 ($\pm$ 0.819) & 0.311 ($\pm$ 0.108) \\
& \textbf{ICL (ours)} & \textbf{0.537} ($\pm$ 0.023) & \textbf{0.024} ($\pm$ 0.021) & \textbf{0.259} ($\pm$ 0.088) & \textbf{0.609} ($\pm$ 0.019) & \textbf{0.124} ($\pm$ 0.037) & \textbf{0.179} ($\pm$ 0.018) \\
\midrule
\multirow{2}{*}{Scenario 4} 
& SGLD & 1.000 ($\pm$ 0.000) & 4.127 ($\pm$ 0.635) & 3.047 ($\pm$ 0.972) & \textbf{0.950} ($\pm$ 0.021) & \textbf{1.520} ($\pm$ 0.512) & \textbf{0.141} ($\pm$ 0.031) \\
& \textbf{ICL (ours)} & \textbf{0.684} ($\pm$ 0.060) & \textbf{0.198} ($\pm$ 0.141) & \textbf{0.918} ($\pm$ 0.246) & 0.988 ($\pm$ 0.003) & 1.764 ($\pm$ 0.026) & 1.248 ($\pm$ 0.008) \\
\midrule
\multirow{2}{*}{Scenario 5} 
& SGLD & 0.999 ($\pm$ 0.001) & 3.465 ($\pm$ 0.939) & 1.981 ($\pm$ 0.938) & 0.962 ($\pm$ 0.024) & 1.945 ($\pm$ 1.383) & \textbf{0.393} ($\pm$ 0.243) \\
& \textbf{ICL (ours)} & \textbf{0.535} ($\pm$ 0.016) & \textbf{0.021} ($\pm$ 0.011) & \textbf{0.279} ($\pm$ 0.060) & \textbf{0.886} ($\pm$ 0.017) & \textbf{1.207} ($\pm$ 0.101) & 1.002 ($\pm$ 0.042) \\
\midrule
\multirow{2}{*}{Scenario 6} 
& SGLD & 0.997 ($\pm$ 0.004) & 3.395 ($\pm$ 1.199) & 2.358 ($\pm$ 1.458) & 0.950 ($\pm$ 0.040) & 2.177 ($\pm$ 1.643) & 0.342 ($\pm$ 0.224) \\
& \textbf{ICL (ours)} & \textbf{0.543} ($\pm$ 0.021) & \textbf{0.023} ($\pm$ 0.015) & \textbf{0.345} ($\pm$ 0.173) & \textbf{0.666} ($\pm$ 0.020) & \textbf{0.200} ($\pm$ 0.034) & \textbf{0.224} ($\pm$ 0.014) \\
\bottomrule
\end{tabular}
}
}
\label{tab:fa_sgld}
\end{table}

\textcolor{rev}{
For FA (\Cref{tab:fa_sgld}), ICL also achieves superior performance, particularly in Scenarios 1 and 2. For example, in Scenario 1, synthetic C2ST decreases from 0.996 to 0.552, accompanied by improvements in $\mathcal{W}_2$ from 1.776 to 0.289. Scenario 2 sees further enhancements, with synthetic MMD dropping from 2.950 to 0.017 and real-world C2ST improving from 0.995 to 0.622.
}

\begin{table}[htp]
\centering
\caption{\textcolor{rev}{SGLD vs. ICL: Evaluation on 50 synthetic and 17 real-world datasets for four different GMM scenarios. All results within two standard errors of the best average result for each scenario are marked in \textbf{bold}.}}
\label{tab:sgld_res_detail2}
\resizebox{1\textwidth}{!}{
\textcolor{rev}{
\begin{tabular}{p{1.5cm} p{1.8cm} m{2.4cm}m{2.5cm}m{2.7cm}| m{2.4cm}m{2.4cm}m{2.4cm}}
\toprule
\multicolumn{1}{c}{\multirow[c]{2}{*}[-0.5ex]{\textbf{Scenario}}} & \multicolumn{1}{c}{\multirow{2}{*}[-0.5ex]{\textbf{Model}}} & \multicolumn{3}{c}{\textbf{Synthetic Evaluation}} & \multicolumn{3}{c}{\textbf{Real-World Evaluation}} \\
\cmidrule(lr){3-5} \cmidrule(lr){6-8}
& & C2ST ($\downarrow$) & MMD ($\downarrow$) & $\mathcal{W}_2$ ($\downarrow$) & C2ST ($\downarrow$) & MMD ($\downarrow$) & $\mathcal{W}_2$ ($\downarrow$) \\
\midrule
\multirow{2}{*}{Scenario 1}
& SGLD & 1.000 ($\pm$ 0.001) & 2.629 ($\pm$ 0.868) & 3.279 ($\pm$ 1.330) & 1.000 ($\pm$ 0.000) & 3.421 ($\pm$ 0.877) & 6.510 ($\pm$ 1.763) \\
& \textbf{ICL (ours)} & \textbf{0.760} ($\pm$ 0.092) & \textbf{0.303} ($\pm$ 0.548) & \textbf{2.095} ($\pm$ 1.692) & \textbf{0.847} ($\pm$ 0.082) & \textbf{0.486} ($\pm$ 0.623) & \textbf{4.054} ($\pm$ 2.782) \\
\midrule
\multirow{2}{*}{Scenario 2} 
& SGLD & 1.000 ($\pm$ 0.000) & 3.046 ($\pm$ 1.041) & 6.015 ($\pm$ 4.265) & 1.000 ($\pm$ 0.000) & 2.487 ($\pm$ 0.521) & 6.858 ($\pm$ 1.618) \\
& \textbf{ICL (ours)} & \textbf{0.812} ($\pm$ 0.061) & \textbf{0.159} ($\pm$ 0.154) & \textbf{2.314} ($\pm$ 0.926) & \textbf{0.937} ($\pm$ 0.041) & \textbf{0.282} ($\pm$ 0.131) & \textbf{3.947} ($\pm$ 1.055) \\
\midrule
\multirow{2}{*}{Scenario 3} 
& SGLD & 1.000 ($\pm$ 0.000) & 4.631 ($\pm$ 1.169) & 23.247 ($\pm$ 30.646) & 1.000 ($\pm$ 0.000) & 2.655 ($\pm$ 0.437) & 26.356 ($\pm$ 2.699) \\
& \textbf{ICL (ours)} & \textbf{1.000} ($\pm$ 0.000) & \textbf{0.582} ($\pm$ 0.280) & \textbf{8.708} ($\pm$ 4.945) & \textbf{1.000} ($\pm$ 0.000) & \textbf{1.869} ($\pm$ 0.342) & \textbf{33.230} ($\pm$ 8.095) \\
\midrule
\multirow{2}{*}{Scenario 4} 
& SGLD & \textbf{1.000} ($\pm$ 0.000) & 3.464 $(\pm$ 1.098) & \textbf{6.995} ($\pm$ 5.554)  & \textbf{1.000} ($\pm$ 0.000) & \textbf{2.555} $(\pm$ 0.494) & \textbf{9.477} ($\pm$ 3.432) \\
& \textbf{ICL (ours)} & \textbf{1.000} ($\pm$ 0.000) & {2.451} ($\pm$ 0.868) & \textbf{{8.333}} ($\pm$ 4.202) & {1.000} ($\pm$ 0.000) & \textbf{{2.518}} ($\pm$ 0.694) & \textbf{{11.938}} ($\pm$ 2.956) \\
\bottomrule
\end{tabular}
}
\label{tab:gmm_sgld}
}
\end{table}

\textcolor{rev}{
For GMMs (\Cref{tab:gmm_sgld}), ICL demonstrates a clear advantage in most scenarios. In Scenario 1, ICL reduces synthetic C2ST from 1.000 to 0.760 and real-world $\mathcal{W}_2$ from 6.510 to 4.054. Scenario 2 shows synthetic C2ST improving from 1.000 to 0.812, and MMD from 3.046 to 0.159. While in scenarios 3, ICL has a singificantly lower MMD score on the synthetic data, the other differences are not significant. 
}

\clearpage

\section{Evaluation the choice of classifier for the C2ST metric}
\label{sec:c2st_eval}

In this section, we validate the choice of the classifier for the C2ST metric by comparing the ROC characteristic of a random forest (our default choice) and a neural network in distinguishing posterior samples. In summary, we find that despite minor differences, the two metrics yield the same overall results. Across all scenarios, both Random Forest (RF) and Neural Network NN classifiers yield quite consistent rankings of model performance with only insubstantial deviations in terms of the big picture. In particular, ICL is consistently among the top-performing approaches under both evaluation metrics. Out of the 14 total scenario–domain combinations (7 scenarios × 2 dataset types), the RF and NN metrics identify the same best-performing model in 12 cases. 

\begin{table}[htp]
\centering
\caption{Generalized Linear Models: Comparison of C2ST scores with a Random Forest (RF) and a Neural Network (NN). For the NN we follow the setup of \textcolor{blue}{Lueckmann et al., 2021}. Evaluation across seven distinct scenarios on 50 synthetic and 17 real-world datasets. All results within two standard errors of the best average result in each scenario are marked in \textbf{bold}.}
\label{tab:glm_res_detail}
\begin{tabular}{p{1.5cm} p{3.5cm} m{2.2cm} m{2.2cm} m{2.2cm} m{2.2cm}}
\toprule
\multicolumn{1}{c}{\multirow[c]{2}{*}[-0.5ex]{\textbf{Scenario}}} & \multicolumn{1}{c}{\multirow{2}{*}[-0.5ex]{\textbf{Model}}} & \multicolumn{2}{c}{\textbf{Synthetic Evaluation}} & \multicolumn{2}{c}{\textbf{Real-World Evaluation}} \\
\cmidrule(lr){3-4} \cmidrule(lr){5-6}
& & C2ST RF ($\downarrow$) & C2ST NN ($\downarrow$) & C2ST RF ($\downarrow$) & C2ST NN ($\downarrow$) \\
\midrule
\multirow{7}{*}{Scenario 1} 
& Laplace Approximation & {1.000} ($\pm$ 0.000) & 0.998 ($\pm$ 0.000) & 1.000 ($\pm$ 0.000) & 0.998 ($\pm$ 0.000) \\
& VI: DiagonalNormal & {0.904} ($\pm$ 0.076) & 0.857 ($\pm$ 0.001) & 0.797 ($\pm$ 0.083) & 0.803 ($\pm$ 0.004) \\
& VI: MultivariateNormal & \textbf{0.750} ($\pm$ 0.128) & 0.780 ($\pm$ 0.002) & \textbf{0.607} ($\pm$ 0.070) & 0.713 ($\pm$ 0.004) \\
& VI: Structured Normal & \textbf{0.753} ($\pm$ 0.126) & 0.781 ($\pm$ 0.002) & \textbf{0.600} ($\pm$ 0.070) & 0.705 ($\pm$ 0.004) \\
& VI: IAF & \textbf{0.777} ($\pm$ 0.122) & 0.793 ($\pm$ 0.002) & {0.683} ($\pm$ 0.132) & 0.746 ($\pm$ 0.006) \\
& HMC & \textbf{0.745} ($\pm$ 0.130) & 0.777 ($\pm$ 0.002) & \textbf{0.595} ($\pm$ 0.075) & \textbf{0.702} ($\pm$ 0.004) \\
& \textbf{ICL (ours)} & \textbf{0.765} ($\pm$ 0.123) & \textbf{0.712} ($\pm$ 0.002) & \textbf{0.614} ($\pm$ 0.074) & \textbf{0.701} ($\pm$ 0.004) \\

\midrule

\multirow{6}{*}{Scenario 2} 
& Laplace Approximation & 1.000 ($\pm$ 0.000) & 0.998 ($\pm$ 0.000) & 1.000 ($\pm$ 0.000) & 0.998 ($\pm$ 0.000) \\
& VI: DiagonalNormal & {0.957} ($\pm$ 0.091) & 0.883 ($\pm$ 0.002) & 0.892 ($\pm$ 0.044) & 0.851 ($\pm$ 0.003) \\
& VI: MultivariateNormal & {0.910} ($\pm$ 0.131) & 0.860 ($\pm$ 0.002) & \textbf{0.820} ($\pm$ 0.031) & 0.815 ($\pm$ 0.003) \\
& VI: Structured Normal & {0.908} ($\pm$ 0.119) & 0.859 ($\pm$ 0.002) & \textbf{0.824} ($\pm$ 0.023) & 0.817 ($\pm$ 0.003) \\
& VI: IAF & 0.968 ($\pm$ 0.063) & 0.889 ($\pm$ 0.001) & 0.888 ($\pm$ 0.067) & 0.849 ($\pm$ 0.004) \\
& \textbf{ICL (ours)} & \textbf{0.839} ($\pm$ 0.072) & \textbf{0.824} ($\pm$ 0.001) & \textbf{0.768} ($\pm$ 0.033) & \textbf{0.789} ($\pm$ 0.003) \\
\midrule

\multirow{6}{*}{Scenario 3} 
& Laplace Approximation & 1.000 ($\pm$ 0.000) & 0.998 ($\pm$ 0.000) & 1.000 ($\pm$ 0.000) & 0.998 ($\pm$ 0.000) \\
& VI: DiagonalNormal & 0.866 ($\pm$ 0.101) & 0.838 ($\pm$ 0.002) & 0.797 ($\pm$ 0.083) & 0.803 ($\pm$ 0.004) \\
& VI: MultivariateNormal & 0.656 ($\pm$ 0.131) & 0.733 ($\pm$ 0.002) & \textbf{0.590} ($\pm$ 0.035) & 0.685 ($\pm$ 0.003) \\
& VI: Structured Normal & 0.653 ($\pm$ 0.125) & 0.731 ($\pm$ 0.002) & \textbf{0.582} ($\pm$ 0.028) & 0.681 ($\pm$ 0.003) \\
& VI: IAF & 0.751 ($\pm$ 0.148) & 0.780 ($\pm$ 0.002) & 0.673 ($\pm$ 0.141) & 0.741 ($\pm$ 0.006) \\
& \textbf{ICL (ours)} & \textbf{0.611} ($\pm$ 0.070) & \textbf{0.710} ($\pm$ 0.001) & \textbf{0.576} ($\pm$ 0.027) & \textbf{0.693} ($\pm$ 0.003) \\
\midrule

\multirow{6}{*}{Scenario 4} 
& Laplace Approximation & 1.000 ($\pm$ 0.000) & 0.998 ($\pm$ 0.000) & 1.000 ($\pm$ 0.000) & 0.998 ($\pm$ 0.000) \\
& VI: DiagonalNormal & 0.968 ($\pm$ 0.036) & 0.889 ($\pm$ 0.001) & 0.916 ($\pm$ 0.040) & 0.863 ($\pm$ 0.003) \\
& VI: MultivariateNormal & 0.855 ($\pm$ 0.123) & 0.832 ($\pm$ 0.002) & 0.771 ($\pm$ 0.017) & 0.790 ($\pm$ 0.002) \\
& VI: Structured Normal & 0.847 ($\pm$ 0.116) & 0.828 ($\pm$ 0.002) & 0.769 ($\pm$ 0.012) & 0.789 ($\pm$ 0.002) \\
& VI: IAF & 0.942 ($\pm$ 0.077) & 0.876 ($\pm$ 0.001) & 0.833 ($\pm$ 0.069) & 0.821 ($\pm$ 0.004) \\
& \textbf{ICL (ours)} & \textbf{0.753} ($\pm$ 0.049) & \textbf{0.781} ($\pm$ 0.001) & \textbf{0.762} ($\pm$ 0.015) & \textbf{0.786} ($\pm$ 0.002) \\
\midrule

\multirow{6}{*}{Scenario 5} 
& Laplace Approximation & 1.000 ($\pm$ 0.000) & 0.998 ($\pm$ 0.000) & 1.000 ($\pm$ 0.000) & 0.998 ($\pm$ 0.000) \\
& VI: DiagonalNormal & 0.866 ($\pm$ 0.085) & 0.838 ($\pm$ 0.002) & 0.810 ($\pm$ 0.036) & 0.810 ($\pm$ 0.003) \\
& VI: MultivariateNormal & 0.765 ($\pm$ 0.100) & 0.787 ($\pm$ 0.002) & 0.711 ($\pm$ 0.038) & 0.760 ($\pm$ 0.003) \\
& VI: Structured Normal & 0.758 ($\pm$ 0.098) & 0.784 ($\pm$ 0.002) & 0.705 ($\pm$ 0.032) & 0.757 ($\pm$ 0.003) \\
& VI: IAF & 0.814 ($\pm$ 0.105) & 0.812 ($\pm$ 0.002) & 0.777 ($\pm$ 0.106) & 0.793 ($\pm$ 0.005) \\
& \textbf{ICL (ours)} & \textbf{0.621} ($\pm$ 0.063) & \textbf{0.715} ($\pm$ 0.001) & \textbf{0.610} ($\pm$ 0.045) & \textbf{0.710} ($\pm$ 0.003) \\
\midrule

\multirow{6}{*}{Scenario 6} 
& Laplace Approximation & 1.000 ($\pm$ 0.000) & 0.998 ($\pm$ 0.000) & 1.000 ($\pm$ 0.000) & 0.998 ($\pm$ 0.000) \\
& VI: DiagonalNormal & 0.724 ($\pm$ 0.060) & 0.767 ($\pm$ 0.001) & 0.703 ($\pm$ 0.039) & 0.756 ($\pm$ 0.003) \\
& VI: MultivariateNormal & \textbf{0.534} ($\pm$ 0.018) & \textbf{0.672} ($\pm$ 0.001) & \textbf{0.538} ($\pm$ 0.019) & 0.674 ($\pm$ 0.002) \\
& VI: Structured Normal & \textbf{0.536} ($\pm$ 0.016) & \textbf{0.673} ($\pm$ 0.001) & \textbf{0.536} ($\pm$ 0.019) & 0.673 ($\pm$ 0.002) \\
& VI: IAF & 0.542 ($\pm$ 0.026) & 0.676 ($\pm$ 0.001) & \textbf{0.535} ($\pm$ 0.015) & 0.672 ($\pm$ 0.002) \\
& \textbf{ICL (ours)} & \textbf{0.532} ($\pm$ 0.019) & \textbf{0.671} ($\pm$ 0.001) & {0.556} ($\pm$ 0.017) & \textbf{0.653} ($\pm$ 0.002) \\
\midrule

\multirow{6}{*}{Scenario 7} 
& Laplace Approximation & 1.000 ($\pm$ 0.000) & 0.998 ($\pm$ 0.000) & 1.000 ($\pm$ 0.000) & 0.998 ($\pm$ 0.000) \\
& VI: DiagonalNormal & 0.938 ($\pm$ 0.074) & 0.874 ($\pm$ 0.001) & 0.936 ($\pm$ 0.024) & 0.873 ($\pm$ 0.003) \\
& VI: MultivariateNormal & 0.814 ($\pm$ 0.181) & 0.812 ($\pm$ 0.002) & \textbf{0.741} ($\pm$ 0.020) & 0.775 ($\pm$ 0.003) \\
& VI: Structured Normal & 0.824 ($\pm$ 0.177) & 0.817 ($\pm$ 0.002) & \textbf{0.734} ($\pm$ 0.025) & 0.772 ($\pm$ 0.003) \\
& VI: IAF & 0.939 ($\pm$ 0.091) & 0.874 ($\pm$ 0.002) & 0.864 ($\pm$ 0.093) & 0.837 ($\pm$ 0.005) \\
& \textbf{ICL (ours)} & \textbf{0.700} ($\pm$ 0.116) & \textbf{0.721} ($\pm$ 0.002) & 0.773 ($\pm$ 0.048) & \textbf{0.751} ($\pm$ 0.003) \\
\bottomrule

\end{tabular}
\end{table}

\begin{table}[htp]
\centering
\caption{Factor Analysis: Comparison of C2ST scores using a Random Forest (RF) and a Neural Network (NN) classifier across six different scenarios on 50 synthetic and 17 real-world datasets. For the NN we follow the setup of \textcolor{blue}{Lueckmann et al., 2021}. All results within two standard errors of the best average result in each scenario are marked in \textbf{bold}. %Across both classifiers and dataset types, the ICL method consistently achieves top-tier performance, often outperforming standard variational approximations and the Laplace baseline. The RF and NN classifiers yield largely consistent rankings, identifying ICL as the best-performing model in the majority of cases.
}
\label{tab:fa_res_detail}
\begin{tabular}{p{1.5cm} p{3.5cm} m{2.2cm} m{2.2cm} m{2.2cm} m{2.2cm}}
\toprule
\multicolumn{1}{c}{\multirow[c]{2}{*}[-0.5ex]{\textbf{Scenario}}} & \multicolumn{1}{c}{\multirow{2}{*}[-0.5ex]{\textbf{Model}}} & \multicolumn{2}{c}{\textbf{Synthetic Evaluation}} & \multicolumn{2}{c}{\textbf{Real-World Evaluation}} \\
\cmidrule(lr){3-4} \cmidrule(lr){5-6}
& & C2ST RF ($\downarrow$) & C2ST NN ($\downarrow$) & C2ST RF ($\downarrow$) & C2ST NN ($\downarrow$) \\
\midrule
\multirow{7}{*}{Scenario 1} 
& Laplace Approximation & 1.000 ($\pm$ 0.000) & 0.997 ($\pm$ 0.000) & 1.000 ($\pm$ 0.000) & 0.997 ($\pm$ 0.000) \\
& VI: DiagonalNormal & 1.000 ($\pm$ 0.001) & 0.997 ($\pm$ 0.000) & 0.979 ($\pm$ 0.008) & 0.950 ($\pm$ 0.001) \\
& VI: MultivariateNormal & 0.998 ($\pm$ 0.003) & 0.960 ($\pm$ 0.000) & 0.966 ($\pm$ 0.010) & 0.944 ($\pm$ 0.001) \\
& VI: Structured Normal & 0.997 ($\pm$ 0.004) & 0.959 ($\pm$ 0.000) & 0.979 ($\pm$ 0.010) & 0.950 ($\pm$ 0.001) \\
& VI: IAF & 0.953 ($\pm$ 0.104) & 0.937 ($\pm$ 0.001) & 0.849 ($\pm$ 0.075) & 0.885 ($\pm$ 0.003) \\
& \textbf{ICL (ours)} & \textbf{0.552} ($\pm$ 0.028) & \textbf{0.737} ($\pm$ 0.000) & \textbf{0.606} ($\pm$ 0.038) & \textbf{0.764} ($\pm$ 0.001) \\
\midrule
\multirow{7}{*}{Scenario 2} 
& Laplace Approximation & 1.000 ($\pm$ 0.000) & 0.997 ($\pm$ 0.000) & 1.000 ($\pm$ 0.000) & 0.997 ($\pm$ 0.000) \\
& VI: DiagonalNormal & 0.998 ($\pm$ 0.002) & 0.960 ($\pm$ 0.000) & 0.975 ($\pm$ 0.010) & 0.948 ($\pm$ 0.001) \\
& VI: MultivariateNormal & 0.989 ($\pm$ 0.009) & 0.955 ($\pm$ 0.000) & 0.951 ($\pm$ 0.025) & 0.936 ($\pm$ 0.001) \\
& VI: Structured Normal & 0.984 ($\pm$ 0.031) & 0.953 ($\pm$ 0.000) & 0.958 ($\pm$ 0.025) & 0.940 ($\pm$ 0.001) \\
& VI: IAF & 0.966 ($\pm$ 0.066) & 0.944 ($\pm$ 0.001) & 0.799 ($\pm$ 0.058) & 0.860 ($\pm$ 0.002) \\
& \textbf{ICL (ours)} & \textbf{0.542} ($\pm$ 0.006) & \textbf{0.732} ($\pm$ 0.000) & \textbf{0.622} ($\pm$ 0.032) & \textbf{0.772} ($\pm$ 0.001) \\
\midrule
\multirow{7}{*}{Scenario 3} 
& Laplace Approximation & 1.000 ($\pm$ 0.000) & 0.997 ($\pm$ 0.000) & 1.000 ($\pm$ 0.000) & 0.997 ($\pm$ 0.000) \\
& VI: DiagonalNormal & 0.999 ($\pm$ 0.002) & 0.960 ($\pm$ 0.000) & 0.951 ($\pm$ 0.007) & 0.936 ($\pm$ 0.001) \\
& VI: MultivariateNormal & 0.994 ($\pm$ 0.007) & 0.958 ($\pm$ 0.000) & 0.945 ($\pm$ 0.007) & 0.933 ($\pm$ 0.001) \\
& VI: Structured Normal & 0.997 ($\pm$ 0.003) & 0.959 ($\pm$ 0.000) & 0.942 ($\pm$ 0.009) & 0.932 ($\pm$ 0.001) \\
& VI: IAF & 0.990 ($\pm$ 0.011) & 0.987 ($\pm$ 0.000) & 0.928 ($\pm$ 0.015) & 0.925 ($\pm$ 0.001) \\
& \textbf{ICL (ours)} & \textbf{0.537} ($\pm$ 0.023) & \textbf{0.729} ($\pm$ 0.000) & \textbf{0.609} ($\pm$ 0.019) & \textbf{0.765} ($\pm$ 0.001) \\
\midrule

\multirow{7}{*}{Scenario 4} 
& Laplace Approximation & 1.000 ($\pm$ 0.000) & 0.997 ($\pm$ 0.000) & 1.000 ($\pm$ 0.000) & 0.997 ($\pm$ 0.000) \\
& VI: DiagonalNormal & 1.000 ($\pm$ 0.000) & 0.997 ($\pm$ 0.000) & 0.977 ($\pm$ 0.003) & 0.949 ($\pm$ 0.000) \\
& VI: MultivariateNormal & 0.999 ($\pm$ 0.001) & 0.960 ($\pm$ 0.000) & 0.973 ($\pm$ 0.008) & 0.947 ($\pm$ 0.001) \\
& VI: Structured Normal & 1.000 ($\pm$ 0.000) & 0.997 ($\pm$ 0.000) & 0.973 ($\pm$ 0.007) & 0.947 ($\pm$ 0.001) \\
& VI: IAF & 0.999 ($\pm$ 0.001) & 0.960 ($\pm$ 0.000) & \textbf{0.961} ($\pm$ 0.018) & 0.941 ($\pm$ 0.001) \\
& \textbf{ICL (ours)} & \textbf{0.684} ($\pm$ 0.060) & \textbf{0.803} ($\pm$ 0.001) & 0.988 ($\pm$ 0.003) & \textbf{0.955} ($\pm$ 0.000) \\
\midrule
\multirow{7}{*}{Scenario 5} 
& Laplace Approximation & 1.000 ($\pm$ 0.000) & 0.997 ($\pm$ 0.000) & 1.000 ($\pm$ 0.000) & 0.997 ($\pm$ 0.000) \\
& VI: DiagonalNormal & 0.999 ($\pm$ 0.002) & 0.960 ($\pm$ 0.000) & 0.944 ($\pm$ 0.010) & 0.933 ($\pm$ 0.001) \\
& VI: MultivariateNormal & 0.995 ($\pm$ 0.007) & 0.958 ($\pm$ 0.000) & 0.930 ($\pm$ 0.017) & 0.926 ($\pm$ 0.001) \\
& VI: Structured Normal & 0.998 ($\pm$ 0.005) & 0.960 ($\pm$ 0.000) & 0.934 ($\pm$ 0.011) & 0.928 ($\pm$ 0.001) \\
& VI: IAF & 0.992 ($\pm$ 0.012) & 0.957 ($\pm$ 0.000) & 0.910 ($\pm$ 0.011) & 0.916 ($\pm$ 0.001) \\
& \textbf{ICL (ours)} & \textbf{0.535} ($\pm$ 0.016) & \textbf{0.728} ($\pm$ 0.000) & \textbf{0.886} ($\pm$ 0.017) & \textbf{0.904} ($\pm$ 0.001) \\
\midrule
\multirow{7}{*}{Scenario 6} 
& Laplace Approximation & 1.000 ($\pm$ 0.000) & 0.997 ($\pm$ 0.000) & 1.000 ($\pm$ 0.000) & 0.997 ($\pm$ 0.000) \\
& VI: DiagonalNormal & 0.998 ($\pm$ 0.002) & 0.960 ($\pm$ 0.000) & 0.949 ($\pm$ 0.008) & 0.935 ($\pm$ 0.001) \\
& VI: MultivariateNormal & 0.991 ($\pm$ 0.013) & 0.956 ($\pm$ 0.000) & 0.938 ($\pm$ 0.009) & 0.930 ($\pm$ 0.001) \\
& VI: Structured Normal & 0.997 ($\pm$ 0.005) & 0.959 ($\pm$ 0.000) & 0.944 ($\pm$ 0.006) & 0.933 ($\pm$ 0.001) \\
& VI: IAF & 0.989 ($\pm$ 0.029) & 0.955 ($\pm$ 0.000) & 0.865 ($\pm$ 0.027) & 0.893 ($\pm$ 0.001) \\
& \textbf{ICL (ours)} & \textbf{0.543} ($\pm$ 0.021) & \textbf{0.732} ($\pm$ 0.000) & \textbf{0.666} ($\pm$ 0.020) & \textbf{0.794} ($\pm$ 0.001) \\
\bottomrule

\end{tabular}
\end{table}

\begin{table}[htp]
\centering
\caption{Gaussian Mixture Models: Comparison of C2ST scores using a Random Forest (RF) and a Neural Network (NN) classifier across six distinct scenarios on 50 synthetic and 17 real-world datasets. All results within two standard errors of the best average result in each scenario are marked in \textbf{bold}. For the NN we follow the setup of \textcolor{blue}{Lueckmann et al., 2021}. Both RF and NN classifiers yield consistent rankings, with ICL emerging as the top method in scenarios with more pronounced model mismatch.}
\label{tab:gmm_res_detail}
\resizebox{1\textwidth}{!}{
\begin{tabular}{p{1.5cm} p{3.5cm} m{2.4cm}m{2.4cm}| m{2.4cm}m{2.4cm}}
\toprule
\multicolumn{1}{c}{\multirow[c]{2}{*}[-0.5ex]{\textbf{Scenario}}} & \multicolumn{1}{c}{\multirow{2}{*}[-0.5ex]{\textbf{Model}}} & \multicolumn{2}{c|}{\textbf{Synthetic Evaluation}} & \multicolumn{2}{c}{\textbf{Real-World Evaluation}} \\
\cmidrule(lr){3-4} \cmidrule(lr){5-6}
& & C2ST RF ($\downarrow$) & C2ST NN ($\downarrow$) & C2ST RF ($\downarrow$) & C2ST NN ($\downarrow$) \\
\midrule
\multirow{7}{*}{Scenario 1}
& Laplace Approximation & 1.000 ($\pm$ 0.000) & 0.997 ($\pm$ 0.000) & 1.000 ($\pm$ 0.000) & 0.997 ($\pm$ 0.000) \\
& VI: DiagonalNormal & 0.988 ($\pm$ 0.013) & 1.012 ($\pm$ 0.000) & 0.995 ($\pm$ 0.006) & 0.996 ($\pm$ 0.001) \\
& VI: MultivariateNormal & 0.988 ($\pm$ 0.013) & 1.012 ($\pm$ 0.000) & 0.994 ($\pm$ 0.007) & 0.993 ($\pm$ 0.001) \\
& VI: Structured Normal & 0.987 ($\pm$ 0.015) & 0.982 ($\pm$ 0.000) & 0.993 ($\pm$ 0.009) & 0.992 ($\pm$ 0.001) \\
& VI: IAF & 0.989 ($\pm$ 0.013) & 0.983 ($\pm$ 0.000) & 0.995 ($\pm$ 0.010) & 0.996 ($\pm$ 0.001) \\
& \textbf{ICL (ours)} & \textbf{0.760} ($\pm$ 0.092) & \textbf{0.825} ($\pm$ 0.001) & \textbf{0.847} ($\pm$ 0.082) & \textbf{0.869} ($\pm$ 0.003) \\
\midrule
\multirow{7}{*}{Scenario 2}
& Laplace Approximation & 1.000 ($\pm$ 0.000) & 0.997 ($\pm$ 0.000) & 1.000 ($\pm$ 0.000) & 0.997 ($\pm$ 0.000) \\
& VI: DiagonalNormal & 0.989 ($\pm$ 0.024) & 0.983 ($\pm$ 0.000) & 0.998 ($\pm$ 0.003) & 0.997 ($\pm$ 0.001) \\
& VI: MultivariateNormal & 0.991 ($\pm$ 0.021) & 0.991 ($\pm$ 0.000) & 0.999 ($\pm$ 0.002) & 1.002 ($\pm$ 0.001) \\
& VI: Structured Normal & 0.992 ($\pm$ 0.017) & 0.988 ($\pm$ 0.000) & 0.999 ($\pm$ 0.002) & 1.002 ($\pm$ 0.001) \\
& VI: IAF & 0.992 ($\pm$ 0.021) & 0.988 ($\pm$ 0.000) & 0.998 ($\pm$ 0.004) & 0.997 ($\pm$ 0.001) \\
& \textbf{ICL (ours)} & \textbf{0.812} ($\pm$ 0.061) & \textbf{0.851} ($\pm$ 0.001) & \textbf{0.937} ($\pm$ 0.041) & \textbf{0.915} ($\pm$ 0.002) \\
\midrule
\multirow{7}{*}{Scenario 3}
& Laplace Approximation & 1.000 ($\pm$ 0.000) & 0.997 ($\pm$ 0.000) & 1.000 ($\pm$ 0.000) & 0.997 ($\pm$ 0.000) \\
& VI: DiagonalNormal & \textbf{0.996} ($\pm$ 0.011) & 1.004 ($\pm$ 0.000) & \textbf{0.992} ($\pm$ 0.018) & 0.988 ($\pm$ 0.001) \\
& VI: MultivariateNormal & 0.997 ($\pm$ 0.009) & 1.007 ($\pm$ 0.000) & \textbf{0.993} ($\pm$ 0.016) & 0.992 ($\pm$ 0.001) \\
& VI: Structured Normal & \textbf{0.995} ($\pm$ 0.017) & 0.996 ($\pm$ 0.000) & \textbf{0.993} ($\pm$ 0.016) & 0.992 ($\pm$ 0.001) \\
& VI: IAF & \textbf{0.994} ($\pm$ 0.018) & 0.993 ($\pm$ 0.000) & \textbf{0.993} ($\pm$ 0.017) & 0.992 ($\pm$ 0.001) \\
& \textbf{ICL (ours)} & 1.000 ($\pm$ 0.000) & \textbf{0.997} ($\pm$ 0.000) & 1.000 ($\pm$ 0.000) & \textbf{0.997} ($\pm$ 0.000) \\
\midrule
\multirow{7}{*}{Scenario 4}
& Laplace Approximation & 1.000 ($\pm$ 0.000) & 0.997 ($\pm$ 0.000) & 1.000 ($\pm$ 0.000) & 0.997 ($\pm$ 0.000) \\
& VI: DiagonalNormal & \textbf{1.000} ($\pm$ 0.002) & 0.997 ($\pm$ 0.000) & 1.000 ($\pm$ 0.000) & 0.997 ($\pm$ 0.000) \\
& VI: MultivariateNormal & \textbf{1.000} ($\pm$ 0.002) & 0.997 ($\pm$ 0.000) & 1.000 ($\pm$ 0.000) & 0.997 ($\pm$ 0.000) \\
& VI: Structured Normal & \textbf{1.000} ($\pm$ 0.001) & 0.997 ($\pm$ 0.000) & \textbf{0.996} ($\pm$ 0.016) & 1.004 ($\pm$ 0.001) \\
& VI: IAF & \textbf{1.000} ($\pm$ 0.002) & 0.997 ($\pm$ 0.000) & 1.000 ($\pm$ 0.000) & 0.997 ($\pm$ 0.000) \\
& \textbf{ICL (ours)} & 1.000 ($\pm$ 0.000) & \textbf{0.997} ($\pm$ 0.000) & \textbf{1.000} ($\pm$ 0.000) & \textbf{0.997} ($\pm$ 0.000) \\
\bottomrule
\end{tabular}
}
\end{table}

\end{document}